\documentclass{article}

\newcommand{\kh}[1]{{\color{orange!80!black}\textsf{\small KH: #1}}}

\PassOptionsToPackage{numbers,sort&compress}{natbib}
\usepackage[preprint]{neurips_2026}

\usepackage[utf8]{inputenc}
\usepackage[T1]{fontenc}
\usepackage{microtype}
\usepackage{url}
\usepackage{nicefrac}
\usepackage{array} 

\usepackage[table]{xcolor}
\usepackage[most]{tcolorbox}
\usepackage{amsmath,amssymb,amsfonts,amsbsy,mathrsfs}
\usepackage{amsthm}
\usepackage[bookmarks=false]{hyperref}
\usepackage{xr}
\usepackage{cleveref}
\usepackage[ruled,vlined, linesnumbered]{algorithm2e}
\usepackage{enumitem}
\usepackage{dutchcal}
\usepackage{graphicx}
\usepackage{placeins}
\usepackage{xkeyval}
\usepackage{caption}
\usepackage{subcaption}
\usepackage{bm}
\usepackage{empheq}
\usepackage{tikz}
\usetikzlibrary{calc,shadows, arrows.meta,arrows, decorations.pathreplacing, positioning, shapes, fit, backgrounds}
\usepackage{pgfplots}
\pgfplotsset{compat=1.18}
\usepackage{booktabs} 
\usepackage{multirow} 
\usepackage{float}
\usepackage{multicol}
\usepackage{changepage}
\usepackage{cancel} 
\definecolor{pnasred}{RGB}{213, 94, 0}      
\definecolor{pnasblue}{RGB}{0, 114, 178}    
\definecolor{pnasgreen}{RGB}{0, 158, 115}   
\definecolor{pnasgray}{RGB}{100, 100, 100}  
\definecolor{pnasyellow}{RGB}{240, 228, 66} 
\definecolor{pnasorange}{RGB}{235, 110, 30}
\definecolor{darkSlate}{RGB}{47, 79, 79}
\definecolor{headergray}{RGB}{240, 240, 240}
\crefname{section}{Section}{Sections}
\Crefname{section}{Section}{Sections} 
\crefname{subsection}{Subsection}{Subsections} 
\Crefname{subsection}{Subsection}{Subsections} 
\crefname{algocf}{Algorithm}{Algorithms} 
\Crefname{algocf}{Algorithm}{Algorithms}
\crefname{appendix}{Appendix}{Appendices} 
\Crefname{appendix}{Appendix}{Appendices}

\definecolor{mygreen}{rgb}{0,0.5,0}

\SetCommentSty{mycommfont}

\crefname{figure}{Fig.}{Figs.}
\Crefname{figure}{Fig.}{Figs.}

\crefname{table}{Table}{Tables}
\Crefname{table}{Table}{Tables}

\crefname{section}{Sec.}{Secs.}
\Crefname{section}{Sec.}{Secs.}

\crefname{subsection}{Subsec.}{Subsecs.}
\Crefname{subsection}{Subsec.}{Subsecs.}

\crefname{equation}{Eq.}{Eqs.}
\Crefname{equation}{Eq.}{Eqs.}

\crefname{algocf}{Algorithm}{Algorithms}
\Crefname{algocf}{Algorithm}{Algorithms}

\crefname{algorithm}{Algorithm}{Algorithms}
\Crefname{algorithm}{Algorithm}{Algorithms}

\crefname{appendix}{Appendix~}{Appendices~}
\Crefname{appendix}{Appendix~}{Appendices~}

\newcommand{\id}[1]{}
\newcommand{\michael}[1]{}
\providecommand{\kh}[1]{}
\newcommand{\antonio}[1]{}
\newcommand{\michaelA}[1]{}

\newcommand{\Vcal}{\mathcal{V}}

\newcommand{\Lp}{\mathrm{L}^2}
\newcommand{\balpha}{\bm{\lambda}}
\newcommand{\bvarphi}{\bm{\varphi}}
\newcommand{\underbalpha}{{\balpha}}
\newcommand{\underbvarphi}{{\bvarphi}}

\newcommand{\bx}{\bm{x}}

\newcommand{\bS}{\bm{\mathcal{S}}}
\newcommand{\bC}{\bm{\mathcal{C}}}
\newcommand{\phiS}{\Phi_{\bS}}
\newcommand{\phiC}{\Phi_{\bC}}
\newcommand{\phiCNN}{\phiC^{\mathrm{NN}}}

\newcommand{\bigoh}{\mathscr{O}}

\newcommand{\sobolev}{{\mathrm{H}}}
\newcommand{\real}{\mathbb{R}}

\newcommand{\velocity}{\bm{\mathrm{v}}}

\newcommand{\nablax}{\nabla_{\bx}}
\newcommand{\pressuretensor}{\boldsymbol{\mathrm{P}}}
\newcommand{\heatflux}{\boldsymbol{\mathrm{q}}}
\newcommand{\bI}{\boldsymbol{I}}

\newcommand{\F}{\mathbf{f}}
\newcommand{\Feqlattice}{\mathbf{f}^{\mathrm{eq}}}
\newcommand{\Geqlattice}{\mathbf{g}^{\mathrm{eq}}}
\newcommand{\G}{\mathbf{g}}
\newcommand{\Hlattice}{\mathbf{h}}
\newcommand{\TV}{\mathsf{TV}}
\newcommand{\Heqlattice}{\mathbf{h}^{\mathrm{eq}}}
\newcommand{\h}{\Delta t}
\newcommand{\deltax}{\Delta \bx}
\newcommand{\relu}{\mathrm{relu}}
\newcommand{\osym}{\mathbin{\otimes^{\scriptscriptstyle \mathrm{sym}}}}
\newcommand{\latticex}{\mathbb{L}_{\bx}}
\newcommand{\latticet}{\mathbb{L}_{t}}

\newcommand{\MLPSymCons}{\mathrel{\mathrm{MLP}^{\vcenter{\hbox{\scriptsize \text{sym+}}}}_{\vcenter{\hbox{\scriptsize \text{cons}}}}}}

\newcommand{\eq}{\mathrm{eq}}
\newcommand{\M}{\bm{M}}
\newcommand{\phiNN}{ \bm{\phi}^{\mathrm{NN}}} 

\def\LBNN {LB+NeurDE}

\def\NN {NeurDE}

\newcommand{\collision}[1]{\bm{{\mathcal{C}}}(#1) }

\DeclareMathOperator*{\argmin}{arg\,min}

\def \e {\varepsilon}
\def \bra {\langle}
\def \ket {\rangle}

\newcommand{\eqdef}{\ensuremath{\stackrel{\mbox{\upshape\tiny def.}}{=}}}

\newcommand{\Velocity}{\bm{\mathrm{u}}}
\newcommand{\x}{\bm{\mathrm{x}}}
\newcommand{\latticevelocity}{\bm{\mathrm{c}}}
\newcommand{\temperature}{{\mathrm{T}}}
\newcommand{\pressure}{{\mathrm{p}}}
\newcommand{\Feq}{f^{\mathrm{eq}}}

\newcommand{\Cp}{\mathrm{C}_p}
\newcommand{\Cv}{\mathrm{C}_v}

\theoremstyle{plain}
\newtheorem{theorem}{Theorem}[section]
\newtheorem{proposition}[theorem]{Proposition}
\newtheorem{lemma}[theorem]{Lemma}

\newtheorem{corollary}[theorem]{Corollary}
\theoremstyle{definition}
\newtheorem{definition}[theorem]{Definition}
\newtheorem*{definition*}{Definition}
\newtheorem{assumption}[theorem]{Assumption}
\theoremstyle{remark}
\newtheorem{remark}[theorem]{Remark}
\newtheorem*{example*}{Example}

\usepackage{tabularx}
\usepackage{siunitx}
\newcolumntype{L}{>{\raggedright\arraybackslash\hangindent=1em\hangafter=1}X}

\pgfplotsset{
    spectrumAxis/.style={
        width=4.5cm, height=2.8cm,
        xlabel={Frequency $k$},
        ylabel={$|\hat{u}_k|$},
        xlabel style={font=\tiny, at={(0.5,-0.15)}},
        ylabel style={font=\tiny, at={(-0.15,0.5)}},
        xmin=0, xmax=8,
        ymin=0, ymax=1.2,
        xtick={0,2,4,6,8},
        ytick={0,0.5,1},
        tick label style={font=\tiny},
        grid=major,
        grid style={dashed, gray!20},
        axis line style={draw=gray!50},
        line width=0.6pt
    }
}

\title{Neural Equilibria for Long-Term Prediction of Nonlinear Conservation Laws}

\author{%
  Jose Antonio Lara Benitez\\
  Rice University\\
  \texttt{antonio.lara@rice.edu}
  \And
  Kareem Hegazy\\
  ICSI and University of California at Berkeley\\
  \texttt{khegazy@icsi.berkeley.edu}
  \And
  Junyi Guo\\
  ICSI and University of California at Berkeley\\
  \texttt{junyiguo@berkeley.edu}
  \AND
  Ivan Dokmani\'c\\
  University of Basel\\
  \texttt{ivan.dokmanic@unibas.ch}
  \And
  Michael W. Mahoney\\
  ICSI, LBNL, and University of California at Berkeley\\
  \texttt{mmahoney@stat.berkeley.edu}
  \And
  Maarten V. de Hoop\\
  Rice University\\
  \texttt{mdehoop@rice.edu}
}

\begin{document}

\maketitle

\begin{abstract}
Nonlinear conservation laws govern a broad class of important physical systems in science and industry and are central to scientific machine learning (SciML). Large general-purpose models offer speed, but replacing the numerical and physical structure of solvers often compromises stability, accuracy, and physical faithfulness. Here, we aim to balance the general inductive bias of conservation with the flexibility and speed of neural networks through a conservation-aware SciML backbone, which we call \emph{Neural Discrete Equilibrium} (NeurDE). NeurDE places machine learning inside a kinetic solver by learning the local equilibrium closure of a Boltzmann formulation. The kinetic solver still performs transport, relaxation, moment recovery, and conservation; the neural network provides only the nonlinear equilibrium target. We test NeurDE on \(6\) conserved systems, including three very challenging subsonic, transonic, and supersonic shock systems. NeurDE outperforms state-of-the-art SciML methods, including neural operators and pretrained SciML foundation models that are \(10^4\) and \(10^6\) times larger, respectively. Most notably, NeurDE improves upon the numerical method from which it is derived. NeurDE therefore provides a compact target for scientific machine learning in conservative simulation: learn the equilibrium law toward which the system relaxes, not the evolution law itself.
\end{abstract}

\section{Introduction}
Nonlinear conservation laws are central to mathematical physics, governing the transport of conserved quantities such as mass, momentum, energy, charge, and vorticity. They arise across diverse systems, including fluid and plasma dynamics, astrophysics, and geophysical flows.
Canonical examples include the Euler, magnetohydrodynamics, and shallow-water equations.
Although these systems admit a compact formulation, their solutions can exhibit intricate structure: nonlinear fluxes generically produce steep gradients and shocks in finite time, even from smooth initial data. These singular features coexist with smooth phenomena such as rarefactions, contacts, and vortex layers, producing multiscale dynamics that are both mathematically subtle and computationally demanding.

The accurate numerical approximation of such equations remains a long-standing challenge in physically relevant regimes.
High-order methods must carefully balance accuracy and stability, while shock capturing typically requires nonlinear flux evaluations or Riemann solvers~\cite{leveque2002finite, toro2013riemann}. These difficulties intensify in under-resolved and high-dimensional regimes, where nonlinear structures interact across scales.
Machine learning (ML)---more recently, scientific machine learning (SciML)~\cite{benitez2024out, lu2021learning, kovachki2023neural, hansen2023learning, herde2024poseidon, brunton2024promising, vinuesa2022enhancing, chen2024constructing, hegazy2025powerformertransformerweightedcausal}---has emerged as a promising tool to accelerate or augment numerical solvers. However, existing approaches face fundamental limitations.

This suggests a more targeted question: can a learning
problem become data efficient when the solver interface already carries the
relevant physical inductive biases, rather than relying primarily on network
capacity?
We introduce the \emph{Neural Discrete Equilibrium} (\NN{}) framework, which
uses conservation and kinetic relaxation to define a data-efficient learning
target: the local equilibrium closure.
The host kinetic solver retains transport and conservation; the neural network
supplies the equilibrium target.
Nonlinear conservation laws can be generally represented by a system of conservation laws of the form:
\begin{equation}\label{eq:conservation}
    \partial_t \boldsymbol{U}(t,\bx) + \nablax \cdot \boldsymbol{F}(\boldsymbol{U}(t,\bx)) = 0,
\end{equation}
where $\bx$ is the spatial variable, $t$ the time variable, $\boldsymbol{U}$ are the conserved fields, and $\boldsymbol{F}(\boldsymbol{U})$ the flux function.
For example, in the compressible Euler equations, the conserved state can be written as $\boldsymbol{U} = (\rho, \rho \Velocity, \rho E)^\top$, where $E$ denotes the specific total energy. The corresponding flux is $\boldsymbol{F}(\boldsymbol{U}) = (\rho \Velocity, \rho \Velocity\Velocity^\top + p \mathbf{I}, (\rho E+p)\Velocity )^\top$, with $p=\rho R \temperature$, where $R$ is the gas constant and $\temperature$ is determined by an equation of state~\cite{majda2012compressible}. 

With the continuity equation (Eq.~\ref{eq:conservation}) as NeurDE's starting point, we leverage kinetic theory to create a more robust learning environment by isolating components best suited for ML while keeping important physical structures.
By \emph{lifting the macroscopic conservation law} (Eq.~\ref{eq:conservation}) into a kinetic representation, we separate the non-local but linear transport from the local but nonlinear operation.
This transformed representation yields the Boltzmann--BGK formulation~\cite{bhatnagar1954model}, which evolves a single-particle distribution function $f(t,\bx,\velocity)$, where $\bx$ and $\velocity$ are the (independent) microscopic (particle) position and velocity, respectively.
Therefore, NeurDE evolves physical systems in the mesoscopic (particle) scale instead of directly tracking the nonlinear macroscopic observables ($\boldsymbol{U}(\bx, t)$).
At the continuum relaxation level, Bouchut's seminal work shows that hyperbolic conservation laws with convex entropy admit such a BGK surrogate under suitable structural hypotheses~\cite{bouchut1999construction}, so the kinetic lifting underlying \NN{} is broad at the level of representation.


Motivated by conservation as an inductive bias, NeurDE explores kinetic theory and the Boltzmann equation (BE) as an architectural backbone to naturally interface between physics, numerical methods, and ML.
This kinetic formalism separates (linear and non-local) transport from (nonlinear and local) relaxation, which naturally suggests operator splitting~\cite{grad1958principles, holden2010splitting}.
This local relaxation is the solver-internal bottleneck: NeurDE learns the equilibrium closure, while transport remains a numerical operation.
Learning the closure within the BE backbone provides a PDE-free formulation in which the governing equations do not need to be specified explicitly and can be learned.
Thus the LB and finite-volume (FV) DUGKS implementations below are realizations of the same BE closure interface, not the definition of the method.

To distinguish NeurDE from modern SciML methods, we empirically evaluate it on conservation problems, particularly challenging supersonic compressible flows with shocks.
Existing SciML approaches address parts of this problem but often ask the
network to learn too much of the numerical update. Physics-informed neural
networks impose equations through penalties and are difficult to optimize near
sharp structures \cite{cai2021physics,krishnapriyan2021characterizing}.
Operator-learning models, including Fourier neural operators
\cite{li2021fourier,kovachki2023neural,JMLR:v24:21-1524,herde2024poseidon},
learn function-space maps and are powerful for smooth regimes, but shocks expose
phase, conservation, and long-rollout weaknesses
\cite{mcgreivy2024weak,FalsePromizeZeroShot_TR,kratsios2024mixture}.
Conservation-aware predictors and causal weighting help in selected regimes
\cite{hansen2023learning,hegazy2025powerformertransformerweightedcausal}, but a
full learned update must still learn transport, closure, and time stepping at
once. Classical shock-capturing solvers avoid this by building transport,
monotonicity, and conservation into the algorithm; high-speed kinetic solvers
add a different bottleneck, because the local equilibrium closure is either a
cheap low-order approximation that fails beyond low Mach number or an expensive
entropy-based nonlinear solve
\cite{frapolli2015entropic,latt2020efficient,tran2022lattice}.



To summarize, this work introduces \NN{} as a method for learning discrete equilibrium closures under a Bouchut-type kinetic lifting, with a flagship compressible realization in the main text and additional scalar conservation-law tests beyond Euler.
Our main contributions are the following.
\begin{itemize}[leftmargin=*,topsep=2pt,itemsep=1pt,parsep=0pt,partopsep=0pt]
    \item Develop NeurDE as a BE-based architecture that learns the local closure rather than a full time-stepper.
    \item Isolate a mesoscopic closure target that is independent of the host transport discretization and does not require a prescribed closed-form macroscopic flux.
    \item Implement NeurDE in LB and finite-volume DUGKS hosts, using DUGKS to test the same closure interface outside lattice streaming.
    \item Challenging shock-focused empirical evidence on deterministic benchmarks: stable long-horizon rollouts with phase-aware diagnostics; comparison to Neural Operators and large SciML foundation model; late-start and out-of-distribution probes without retraining; and a runtime probe showing amortization of a Newton-solved equilibrium.
\end{itemize}



\section{Related Work}
Scientific machine learning for PDEs has developed along several complementary
lines. Physics-informed neural networks impose the governing equations through
loss penalties~\cite{cai2021physics}, while operator-learning models such as
DeepONet and Fourier Neural Operators learn maps between function spaces
~\cite{lu2021learning,li2021fourier,kovachki2023neural}. These methods are
flexible, but shock-dominated conservation laws remain difficult: small phase
errors, spectral ringing, and weak conservation control can dominate
autoregressive rollouts~\cite{mcgreivy2024weak,FalsePromizeZeroShot_TR}.
NeurDE takes a different route. It does not learn the full solution operator or
enforce physics only through a penalty; it learns a local mesoscopic closure
inside a structured kinetic update.

This places NeurDE among structure-preserving learned simulators, where the
learned component is embedded into a numerical scaffold rather than replacing
the simulator. For conservation laws, this distinction matters because
classical shock-capturing methods owe their robustness to carefully designed
transport, monotonicity, and conservation mechanisms
~\cite{leveque2002finite,toro2013riemann}. NeurDE leaves transport and moment
recovery to the kinetic host and uses learning only for the equilibrium map
whose closed form is inaccurate or expensive after velocity discretization.

The closest numerical line is kinetic and entropic closure. BGK and
lattice-Boltzmann methods replace nonlinear macroscopic fluxes by transport and
local relaxation~\cite{bhatnagar1954model,kruger2017lattice,succi2018lattice},
while entropy-based moment closures characterize equilibria variationally
~\cite{levermore1996moment,mieussens2014survey}. Modern entropic LB solvers use
this structure to improve high-speed stability, but often require a nonlinear
equilibrium solve at every cell and time step
~\cite{frapolli2015entropic,latt2020efficient,tran2022lattice}. NeurDE
amortizes this local solve with an exponential-family neural equilibrium while
retaining the solver structure around it.

Recent neural kinetic models also modify BGK dynamics directly. Some insert
neural correctors into a prescribed kinetic operator~\cite{xiao2021using,miller2022neural},
while others learn a full collision or relaxation map
~\cite{xiao2023relaxnet,corbetta2023toward}. NeurDE differs in the object it
learns: the network represents the local equilibrium itself, not the entire
post-collision update. This separation lets the main theorem distinguish the
maximum-entropy projector, the learned closure defect, and the conservative
moment correction. Among these baselines, Corbetta et
al.~\cite{corbetta2023toward} explicitly enforce lattice symmetry and
conservation algebraically; in our compressible Sod ablation this
full-collision construction loses positivity, motivating an
entropy-minimization-derived equilibrium ansatz
(\Cref{sec:ablation}).

\section{Neural Kinetic Equilibria}
\label{sec:method}
We first provide a theoretical background and an overview of the generic BE backbone that underpins \NN. We describe the closure learning problem and provide a proof of its entropy maximization. Finally, we describe the LB host realization used for the main experiments (\LBNN). As mentioned, we also derive and implement a finite-volume (FV) DUGKS realization of \NN{} (FV+\NN) in \Cref{appx:fvfd_dugks}, but focus on \LBNN{} due to its intuitive nature and ability to be highly parallelized.

\subsection{Discrete kinetic framework}
\label{sec:kinetic_background}

The Boltzmann--BGK equation~\cite{bhatnagar1954model} replaces the
nonlinear flux in \eqref{eq:conservation} by linear transport plus a local
relaxation of a particle distribution \(f(t,\bx,\velocity)\) toward a
local equilibrium \(\Feq\):
\begin{equation}\label{eq:Boltzmann_collision}
(\partial_t + \velocity\!\cdot\!\nabla_{\bx}) f
= \tau^{-1}\bigl(\Feq(\boldsymbol{U})(\velocity) - f\bigr),
\qquad
\boldsymbol{U}(t,\bx)
= \int_{\mathbb{R}^{d}}
\bigl(1,\velocity,\tfrac12\velocity\!\cdot\!\velocity\bigr)^{\!\top}
f(t,\bx,\velocity)\,d\velocity .
\end{equation}
Bouchut~\cite{bouchut1999construction} shows that hyperbolic systems with a
convex entropy--entropy flux pair in the sense of Lax~\cite{lax1957hyperbolic}
admit a BGK surrogate of this form, provided the Maxwellian satisfies suitable
moment, realizability, and stability conditions. Thus the kinetic lifting is
broad at the representation level.
In this paper, the kinetic representation, discrete velocity set, and moment
map are fixed; NeurDE then learns the local equilibrium closure.
Discretizing velocity on a finite set
\(\{\latticevelocity_i\}_{i=1}^{Q}\) with positive quadrature weights
\(W_i\) yields populations
\(\F_i(t,\bx)=f(t,\bx,\latticevelocity_i)\) and discrete equilibria
\(\Feqlattice_i\)~\cite{shan1998discretization}:
\begin{equation}\label{eq:Boltzmann_BGK_discrete}
(\partial_t + \latticevelocity_i\!\cdot\!\nabla_{\bx}) \F_i
= \tau^{-1}\bigl(\Feqlattice_i(\boldsymbol{U}) - \F_i\bigr),
\qquad i=1,\ldots,Q .
\end{equation}

A fundamental difficulty in all discrete-velocity kinetic models arises from the fact that the continuous Maxwellian no longer minimizes the kinetic entropy on a finite velocity set.
Standard formulations of the equilibrium distribution \(\Feqlattice\) are inherently low-Mach: their accuracy relies on low-order Hermite expansions and lattice isotropy, leading to errors that grow rapidly with Mach number \cite{frapolli2015entropic}.
Worse, improving these expansions requires solving a nonlinear constrained optimization problem in every cell at each time step---a prohibitively expensive task for large-scale simulations~\cite{latt2020efficient, tran2022lattice}.
These issues make high-speed and/or turbulent flows particularly demanding, especially when small velocity sets are used (\Cref{appendix:LBM}).
With \NN{}, we target this exact regime by introducing a neural surrogate that learns a discrete equilibrium map from \(\boldsymbol{U}\) to the relevant equilibrium channel.

\subsection{NeurDE architecture}
\label{sec:NN_LBNN}

\NN{} learns the problem-specific closure by modeling the discrete equilibrium map $\Feqlattice$.
Entropy-based moment closures express the local equilibrium as
\(f(\underbalpha,\latticevelocity) = \exp(\underbalpha \cdot
\underbvarphi(\latticevelocity))\), where \(\underbvarphi\) are sufficient
statistics and \(\underbalpha\) are Lagrange
multipliers~\cite{kogan1965derivation,dreyer1987maximisation,levermore1996moment,mieussens2014survey}.
Recovering \(\underbalpha\) classically requires a nonlinear inversion at
every cell. \NN{} amortizes this inversion by parameterizing \emph{both}
the sufficient statistics and the natural parameters with neural networks
(\Cref{fig:neurde_overview}):
\begin{equation}\label{eq:levermore_closure_NN}
\Feqlattice_i(t,\bx) \approx \phiNN_i(t,\bx)
= \exp\!\Bigl(
  \textstyle\sum_{k=1}^{p}
  \balpha_k(\boldsymbol{U}(t,\bx);\theta^\lambda)\,
  \bvarphi_k(\latticevelocity_i;\theta^{\bvarphi})
\Bigr),
\end{equation}
where the learned basis spans the moment space
\(\mathbb{M}=\mathrm{span}\{\bvarphi_k(\latticevelocity_i)\}\)
(\Cref{eq:M_space}, defined in \Cref{sec:moment_space}).
The exponential ansatz inherits positivity and entropy structure; the
learned basis \(\bvarphi\) spans non-polynomial moments (heat flux,
pressure tensor) that lie outside any fixed Taylor expansion. This
reduces aliasing errors that plague low-order polynomial closures while
replacing the inner Newton solve by two forward passes.

\begin{figure}[t]
\centering
\vspace{-0.6em}
\resizebox{\linewidth}{!}{\sffamily
\begin{tikzpicture}[scale=0.95, >=stealth]
    \scriptsize

    \pgfmathsetmacro{\zero}{0.}
    \pgfmathsetmacro{\first}{0.8}
    \pgfmathsetmacro{\second}{-0.8}
    \pgfmathsetmacro{\colfirstB}{-4.5}
    \pgfmathsetmacro{\colsecB}{2.5}

    \node[draw=black!60, rectangle, fill=gray!10, line width=0.8pt, minimum width=1.3cm, minimum height=1.cm, text width=1.cm, text centered, rounded corners=2pt] (circ0) at (\colfirstB-1.3,\first) {$\{\latticevelocity_i\}_{i=1}^Q$};
    \node[font=\small] at (\colfirstB-1.3,\first+0.8) {velocity};

    \node[draw=black!60, rectangle, fill=pnasgreen!10, line width=0.8pt, minimum width=1.3cm, minimum height=1.cm, text width=1.cm, text centered, rounded corners=2pt] (circ1) at (\colfirstB-1.3,\second) {$\boldsymbol{U}$};
    \node[font=\small] at (\colfirstB-1.7,\second-1) {\shortstack{macroscopic \\ observables}};

    \node[draw=black!60, circle, fill=pnasblue!10, line width=0.8pt, minimum width=.75cm, minimum height=.75cm] (F) at (\colfirstB-3.7, \second) {$\F$};

    \node[draw=pnasred, rectangle, fill=pnasred!10, line width=0.8pt, minimum width=1.cm, minimum height=1cm, text width=1.2cm, text centered, rounded corners=2pt] (feature_extr_1) at (\colfirstB/2,\first) {$\begin{pmatrix} \bvarphi_1\\ \vdots \\ \bvarphi_p \end{pmatrix}$};

    \node[draw=pnasred, rectangle, fill=pnasred!10, line width=0.8pt, minimum width=1.cm, minimum height=1cm, text width=1.2cm, text centered, rounded corners=2pt] (feature_extr_2) at (\colfirstB/2,\second) {$\begin{pmatrix} \balpha_1\\ \vdots \\ \balpha_p \end{pmatrix}$};

    \node[draw=black!60, rectangle, fill=gray!10, line width=0.8pt, minimum width=1.5cm, minimum height=1.5cm, text width=1.6cm, text centered, rounded corners=2pt] (velocity) at (\colsecB-2.5,\second+1.4) {$\sum_k \balpha_k \bvarphi_k$};

    \node[draw=pnasblue, rectangle, fill=pnasblue!10, line width=0.8pt, minimum width=1.cm, minimum height=1.cm, text width=1.cm, text centered, rounded corners=2pt] (exp) at (\colsecB,\second+1.4) {$\exp(\cdot)$};
    \node[font=\small] at (\colsecB,\second+0.55) {\shortstack{renormalization \\ map}};

    \node[draw=pnasblue, circle, fill=white, line width=1pt, minimum width=.75cm] (eq) at (\colsecB+2., \second+1.4) {$\Feqlattice$};
    \node[font=\small] at (\colsecB+2.2,\second+0.55) {equilibrium};

    \draw[dashed, line width=1.2pt, rounded corners=2pt, rotate=90, draw=pnasgray]
        ($(feature_extr_1.north west) + (.25, 0.25)$) rectangle ($(velocity.south east) + (-1.5,-0.25)$)
        node[left=1cm, above] {$\mathbb{M}$};

    \draw[->, line width=1.3pt] (F) -- (circ1) node[midway, above, font=\small, fill=white, inner sep=1pt] {$\mathcal{D}$};
    \draw[->, line width=1.3pt] (circ0) -- (feature_extr_1) node[midway, below, font=\scriptsize, fill=white, inner sep=1pt] {$\underbvarphi(\cdot; \theta^{\bvarphi})$};
    \draw[->, line width=1.3pt] (circ1) -- (feature_extr_2) node[midway, above, font=\scriptsize, fill=white, inner sep=1pt] {$\underbalpha(\cdot; \theta^{\balpha})$};
    \draw[->, line width=1.3pt] (feature_extr_1) -- (velocity);
    \draw[->, line width=1.3pt] (feature_extr_2) -- (velocity);
    \draw[->, line width=1.3pt] (velocity) -- (exp);
    \draw[->, line width=1.3pt] (exp) -- (eq);

    \node[draw,circle,fill=gray!20,minimum size=0.12cm,line width=0.6pt] (input1) at (\colfirstB-0.45,\first+.6) {};
    \node[draw,circle,fill=gray!20,minimum size=0.12cm,line width=0.6pt] (input2) at (\colfirstB-0.45,\first+1.) {};
    \node[draw,circle,fill=pnasgreen!30,minimum size=0.12cm,line width=0.6pt] (hidden1) at (\colfirstB+.35,\first+.4) {};
    \node[draw,circle,fill=pnasgreen!30,minimum size=0.12cm,line width=0.6pt] (hidden2) at (\colfirstB+.35,\first+0.8) {};
    \node[draw,circle,fill=pnasgreen!30,minimum size=0.12cm,line width=0.6pt] (hidden3) at (\colfirstB+.35,\first+1.2) {};
    \node[draw,circle,fill=pnasred!30,minimum size=0.12cm,line width=0.6pt] (output1) at (\colfirstB+1.,\first+.6) {};
    \node[draw,circle,fill=pnasred!30,minimum size=0.12cm,line width=0.6pt] (output2) at (\colfirstB+1.,\first+1) {};

    \draw[-,line width=0.3pt, black!40] (input1) -- (hidden1); \draw[-,line width=0.3pt, black!40] (input1) -- (hidden2); \draw[-,line width=0.3pt, black!40] (input1) -- (hidden3);
    \draw[-,line width=0.3pt, black!40] (input2) -- (hidden1); \draw[-,line width=0.3pt, black!40] (input2) -- (hidden2); \draw[-,line width=0.3pt, black!40] (input2) -- (hidden3);
    \draw[-,line width=0.3pt, black!40] (hidden1) -- (output1); \draw[-,line width=0.3pt, black!40] (hidden2) -- (output1); \draw[-,line width=0.3pt, black!40] (hidden3) -- (output1);
    \draw[-,line width=0.3pt, black!40] (hidden1) -- (output2); \draw[-,line width=0.3pt, black!40] (hidden2) -- (output2); \draw[-,line width=0.3pt, black!40] (hidden3) -- (output2);

    \node[draw,circle,fill=gray!20,minimum size=0.12cm,line width=0.6pt] (NN2input1) at (\colfirstB-0.45,\first-2.3) {};
    \node[draw,circle,fill=gray!20,minimum size=0.12cm,line width=0.6pt] (NN2input2) at (\colfirstB-0.45,\first-2.7) {};
    \node[draw,circle,fill=pnasgreen!30,minimum size=0.12cm,line width=0.6pt] (NN2hidden1) at (\colfirstB+.35,\first-2.1) {};
    \node[draw,circle,fill=pnasgreen!30,minimum size=0.12cm,line width=0.6pt] (NN2hidden2) at (\colfirstB+.35,\first-2.5) {};
    \node[draw,circle,fill=pnasgreen!30,minimum size=0.12cm,line width=0.6pt] (NN2hidden3) at (\colfirstB+.35,\first-2.9) {};
    \node[draw,circle,fill=pnasred!30,minimum size=0.12cm,line width=0.6pt] (NN2output) at (\colfirstB+1.,\first-2.2) {};
    \node[draw,circle,fill=pnasred!30,minimum size=0.12cm,line width=0.6pt] (NN2output2) at (\colfirstB+1.,\first-2.7) {};

    \draw[-,line width=0.3pt, black!40] (NN2input1) -- (NN2hidden1); \draw[-,line width=0.3pt, black!40] (NN2input1) -- (NN2hidden2); \draw[-,line width=0.3pt, black!40] (NN2input1) -- (NN2hidden3);
    \draw[-,line width=0.3pt, black!40] (NN2input2) -- (NN2hidden1); \draw[-,line width=0.3pt, black!40] (NN2input2) -- (NN2hidden2); \draw[-,line width=0.3pt, black!40] (NN2input2) -- (NN2hidden3);
    \draw[-,line width=0.3pt, black!40] (NN2hidden1) -- (NN2output); \draw[-,line width=0.3pt, black!40] (NN2hidden2) -- (NN2output); \draw[-,line width=0.3pt, black!40] (NN2hidden3) -- (NN2output);
    \draw[-,line width=0.3pt, black!40] (NN2hidden1) -- (NN2output2); \draw[-,line width=0.3pt, black!40] (NN2hidden2) -- (NN2output2); \draw[-,line width=0.3pt, black!40] (NN2hidden3) -- (NN2output2);

\end{tikzpicture}}
\vspace{-0.6em}
\caption{\textbf{NeurDE as a local equilibrium learner.}
\NN{} replaces the prescribed
discrete equilibrium with an exponential-family neural map: a
\(\bvarphi\)-network learns velocity-dependent sufficient statistics, an
\(\balpha\)-network maps the macroscopic state to natural parameters, and
the exponential maps back to the population space. }
\label{fig:neurde_overview}
\vspace{-0.5em}
\end{figure}

\subsection{Closure-level entropy and conservation}
\label{sec:closure_structure}

The exponential form is not merely a positivity device. It is the output
nonlinearity associated with constrained entropy minimization on the
discrete velocity set. The following informal theorem summarizes three
closure-level facts: NeurDE is an approximate maximum-entropy projector, its
approximation error appears as an explicit local defect, and a separate
local correction enforces the chosen physical moments exactly. Assumptions,
full statements, and proofs are in the appendix.

\begin{theorem}[Local entropy and conservation structure of NeurDE; informal]
\label{thm:main_neurde_structure}
Let \(\{\latticevelocity_i,W_i\}_{i=1}^{Q}\) be a discrete velocity set
with admissible learned basis \(\bvarphi\) (\Cref{ass:discrete_admissible}),
and define \(\mathcal{M}_\varphi[\F]=\sum_i W_i\,\bvarphi(\latticevelocity_i)\F_i\)
and the discrete entropy \(H(\F)=\sum_i W_i(\F_i\log\F_i-\F_i)\), with
spatial integral \(\mathscr{H}[\F](t)=\int_\Omega H(\F(t,\bx))\,d\bx\). Then
\textup{(i)}~every interior moment target \(\bm{Y}\) admits a unique
positive maximum-entropy projector
\(\F^\star_i = \exp(\bm{\alpha}^\star(\bm{Y})\!\cdot\!\bvarphi(\latticevelocity_i))\)
with \(\mathcal{M}_\varphi[\F^\star]=\bm{Y}\)
(\Cref{thm:discrete_maxent}). \textup{(ii)}~If the trained branch
returns \(\widehat{\bm{\alpha}}\), the spatially integrated entropy of
BGK relaxation toward \(\widehat{\phiNN}\) satisfies the exact balance
\begin{equation*}
\frac{d}{dt}\mathscr{H}[\F]
= -\frac{1}{\tau}\!\int_\Omega\!\!\textstyle\sum_i
W_i(\F_i-\widehat{\phiNN}_i)\log\!\tfrac{\F_i}{\widehat{\phiNN}_i}\,d\bx
+\frac{1}{\tau}\!\int_\Omega \widehat{\bm{\alpha}}\!\cdot\! m\,d\bx,
	\end{equation*}
	When the trained branch realizes the exact entropy projector
	(\(m\equiv 0\)), the second integral vanishes and one recovers the
	formal \(H\)-theorem; otherwise the right-hand side decomposes into
	non-positive dissipation and a defect linear in the learned moment
	residual. Here the first integrand is non-negative pointwise and
	\(m=\mathcal{M}_\varphi[\widehat{\phiNN}]-\bm{Y}\) is the local
	learned-moment defect (\Cref{thm:discrete_defect,cor:discrete_approx_H}).
\textup{(iii)}~A differentiable local projection enforces a chosen
physical-moment constraint \(C\Feqlattice_{\rm c}=\boldsymbol{U}\) exactly,
where \(C\) contains the corresponding quadrature-weighted moment rows, and
	preserves positivity under the residual smallness condition
	\(\|\Delta\|_\infty\le \underline w^{-1/2}\mu_0^{-1/2}\delta_c
	<\min_i\tilde{\Feqlattice}_i\)
	(\Cref{prop:two_state_weighted_projection,cor:two_state_projection_positivity});
for the hybrid two-population scaffold below
the corresponding energy-channel statement is
\Cref{thm:hybrid_two_population_neurde,cor:hybrid_thermal_specialization}.
\end{theorem}

The defect term \(\widehat{\bm{\alpha}}\!\cdot\! m\) gives a local
diagnostic: errors in the learned equilibrium appear as an explicit,
measurable departure from the exact maximum-entropy closure rather than
only through end-to-end rollout error. \emph{Caveat:} this is a local
closure statement, not a proof of convergence or long-time stability of
the full discretization. We therefore report transport-level diagnostics
(positivity, shock and contact locations, energy-moment residuals)
separately as empirical stability checks in
\Cref{sec:numerical}.

\subsection{\LBNN: the lattice Boltzmann realization of \NN}

For the main experimental host, a lattice Boltzmann (LB) discretization splits one time step into a local
collision and an exact streaming shift along lattice links:
\begin{subequations}\label{eq:LBM_algorithm_dimensionless}
\begin{align}
\F_i^{\rm c}(t,\bx) &= \F_i(t,\bx) + \tau^{-1}\bigl(\Feqlattice_i-\F_i\bigr),
\label{eq:LBM_collision}\\
\F_i(t+1,\bx+\latticevelocity_i) &= \F_i^{\rm c}(t,\bx),
\label{eq:LBM_streaming}
\end{align}
\end{subequations}
with macroscopic moments
\begin{equation}\label{eq:projection}
\rho=\textstyle\sum_i W_i \F_i,\quad
\rho\Velocity=\textstyle\sum_i W_i \latticevelocity_i \F_i,\quad
\rho E=\tfrac12\textstyle\sum_i W_i (\latticevelocity_i\!\cdot\!\latticevelocity_i)\F_i .
\end{equation}
The weights \(W_i\) are therefore part of the discrete velocity quadrature.
Many LB implementations absorb them into the populations
\(\bar{\F}_i=W_i\F_i\), after which the same moment constraints are written
with bare sums; the appendix states this absorbed convention explicitly for
the hybrid compressible solver (\Cref{rem:hybrid_two_pop_convention}).
The streaming step is an exact memory shift; the central closure choice is the
discrete equilibrium \(\Feqlattice_i\).

In standard LBMs, stability deteriorates as the relaxation time approaches its lower bound (\(\tau \to 1/2\) in lattice units), corresponding to vanishing viscosity.
In \LBNN, stability is improved through a lattice-velocity shift and, for the transonic case, TVD regularization; see \Cref{sec:training,appendix:TVD}.

\begin{figure}[t]
\centering
\vspace{-0.6em}
\resizebox{\linewidth}{!}{\sffamily
\begin{tikzpicture}[scale=0.95, >=stealth]
    \scriptsize

    \pgfmathsetmacro{\colfirst}{-7.}
    \pgfmathsetmacro{\colsec}{-2.8}
    \pgfmathsetmacro{\colthird}{2.8}
    \pgfmathsetmacro{\colfourth}{7.}

    \pgfmathsetmacro{\rowfirst}{2.}
    \pgfmathsetmacro{\rowsecond}{-0.8}
    \pgfmathsetmacro{\rowthird}{-2.3}

    \node (domain) at (\colfirst,\rowfirst-1.4) {$(t, \bx, \velocity) \in \real_{\ge 0}\times \Omega \times \real^d$};
    \node (f) at (\colfirst,\rowfirst-1) {$f^\e(t, \bx, \velocity)$};
    \node (BTW) at (\colsec-1.,\rowfirst-.6) {\textbf{Boltzmann-BGK}};

    \node[text=pnasblue, font=\scriptsize\itshape] (scope) at (\colsec-1.,\rowfirst-1.2) {(Lax entropy)};

    \node[draw=pnasred, rectangle, fill=pnasred!10, minimum width=5.cm, minimum height=1.cm, text width=4.5cm, text centered, rounded corners=2pt, line width=0.9pt] (BGK_Eq) at (0,\rowfirst-1) {$(\partial_t + \velocity \cdot \nablax )f^\e =\tfrac{1}{\e} (\Feq( \boldsymbol{U}^\e)-f^\e)$};

    \node (solution_approx) at (\colfourth,\rowfirst-1) {$f^\e(t+\h, \bx, \velocity)$};

    \node (discretization1) at (\colfirst-1.1,\rowsecond+1.1) {\textcolor{pnasred}{Discretization}};
    \node (discretization12) at (\colfirst-1.1,\rowsecond+0.8) {$\{\latticevelocity_1,\ldots, \latticevelocity_Q\}$};
    \node (discretization2) at (\colfourth-1.,\rowsecond+1.2) {\textcolor{pnasred}{Discretization}};

    \node (fdiscr) at (\colfirst,\rowsecond) {$\F(t, \bx) = (\F_1, \dots, \F_Q)^\top$};

    \node (SplittingMethod_discre) at (\colsec-1,\rowsecond+1.2) {\textbf{Splitting Method}};

    \node (collision_discr) at (\colsec,\rowsecond+0.8) {Collision};
    \node[draw=pnasblue, rectangle, fill=pnasblue!10, minimum width=4.5cm, minimum height=1.cm, text width=4.cm, text centered, rounded corners=2pt, line width=0.9pt] (streaming_Eq_discre) at (\colsec,\rowsecond) {$\F_i^{\mathrm{coll}} = \F_i + \frac{1}{\tau}\bigl[{\phiNN}_i(\mathcal{D}(\F)) - \F_i\bigr]$};

    \node (streaming_discr) at (\colthird,\rowsecond+0.8) {Streaming};
    \node[draw=pnasblue, rectangle, fill=pnasblue!10, minimum width=4.cm, minimum height=1.cm, text width=4.cm, text centered, rounded corners=2pt, line width=0.9pt] (collision_Eq_discr) at (\colthird,\rowsecond) {$\F_i(t+\h,\bx) = \F_i^{\mathrm{coll}}(t,\bx-\latticevelocity_i)$};

    \node (solution_approx_discr) at (\colfourth,\rowsecond) {$\F(t+\h, \bx)$};

    \draw[line width=0.9pt, dashed, dash pattern=on 4pt off 2pt, draw=pnasgray, rounded corners=2pt]
        (\colsec-2.5,\rowsecond+1.0)--(\colthird+2.5,\rowsecond+1.0)--(\colthird+2.5,\rowsecond-.8)--(\colsec-2.5,\rowsecond-.8)--(\colsec-2.5,\rowsecond+1.0);

    \node (Lifting) at (\colfirst-1.1,\rowthird+1.2) {\textcolor{pnasred}{Lifting ($\mathcal{E}$)}};
    \node (Observables) at (\colfirst-1.1,\rowthird+.5) {(Macroscopic)};
    \node (Projection) at (\colfourth-.8,\rowthird+1.) {\shortstack{\textcolor{pnasred}{Projection}\\ \textcolor{pnasred}{($\mathcal{D}$)}}};

    \node (U_obs) at (\colfirst,\rowthird) {$\boldsymbol{U}(t, \bx)$};
    \node (collision_discr_macro) at (\colsec-1,\rowthird+0.2) {\textbf{Conservation law}};

    \node[text=pnasgreen, font=\scriptsize\itshape] at (\colsec-1,\rowthird-0.2) {(Lax entropy pair)};

    \node[draw=pnasgreen, rectangle, fill=pnasgreen!10, minimum width=3.cm, minimum height=1.cm, text width=3.cm, text centered, rounded corners=2pt, line width=0.9pt] (macro_solution_Eq) at (0,\rowthird) {$\partial_t \boldsymbol{U} + \nablax \cdot \boldsymbol{F} ( \boldsymbol{U})=0$};
    \node (solution_macros) at (\colfourth,\rowthird) {$\boldsymbol{U}(t+\h, \bx)$};

    \draw[->, line width=1pt] (f) -- (BGK_Eq);
    \draw[->, line width=1pt] (BGK_Eq) -- (solution_approx);

    \draw[->, line width=1pt] (fdiscr) -- (streaming_Eq_discre);
    \draw[->, line width=1pt] (streaming_Eq_discre) -- (collision_Eq_discr);
    \draw[->, line width=1pt] (collision_Eq_discr) -- (solution_approx_discr);

    \draw[->, line width=1pt] (U_obs) -- (macro_solution_Eq);
    \draw[->, line width=1pt] (macro_solution_Eq) -- (solution_macros);

    \draw[->, line width=1pt, dashed, color=pnasgray] (domain) -- (fdiscr);
    \draw[->, line width=1pt] (U_obs) -- (fdiscr);
    \draw[->, line width=1pt, dashed, color=pnasgray] (solution_approx) -- (solution_approx_discr);
    \draw[->, line width=1pt] (solution_approx_discr) -- (solution_macros);

\end{tikzpicture}}
\vspace{-0.6em}
\caption{\textbf{LB+NeurDE solver.} A convex-entropy hyperbolic system is lifted to a Boltzmann--BGK kinetic representation, discretized in velocity, and advanced by a collision/streaming split. Transport remains a
fixed numerical operation; conservation is imposed by moment projection.}
\label{fig:architecture}
\vspace{-0.5em}
\end{figure}

\subsection{Hybrid compressible LB realization and training}

For compressible flow we use the two-population thermal LB scaffold of
\cite{karlin2013consistent,saadat2019lattice}, which decouples viscosity
and thermal conductivity by transporting mass/momentum in \(\F\) and
energy in a second population \(\G\):
\begin{subequations}
\begin{align}
\F_i(t{+}1,\bx{+}\latticevelocity_i)-\F_i(t,\bx) &= \tau_1^{-1}\bigl(\Feqlattice_i-\F_i\bigr),\\
\G_i(t{+}1,\bx{+}\latticevelocity_i)-\G_i(t,\bx) &= \tau_2^{-1}\bigl(\Geqlattice_i-\G_i\bigr) + \bigl(\tau_2^{-1}-\tau_1^{-1}\bigr)\bigl(\G^\ast_i-\G_i\bigr).
\end{align}
\end{subequations}
Macroscopic fields are recovered as
\begin{equation}
\rho=\textstyle\sum_i W_i \F_i,\quad
\rho\Velocity=\textstyle\sum_i W_i \latticevelocity_i \F_i,\quad
2\rho E=\textstyle\sum_i W_i \G_i,\quad
\temperature=\Cv^{-1}\bigl(E-\tfrac12\Velocity\!\cdot\!\Velocity\bigr).
\end{equation}
We retain the analytic extended equilibrium for \(\Feqlattice\) and the
quasi-equilibrium \(\G^\ast\), and let \NN{} learn only the energy-channel
closure \(\Geqlattice\). This is a compressible-flow design choice rather
than a restriction of the framework: the scalar conservation-law variants in
the appendix use a single population and learn \(\Feqlattice\) directly.
For the two-population compressible solver, learning \(\Geqlattice\) isolates
the high-Mach energy-closure bottleneck without asking the network to replace
mass and momentum transport, and is the setting for which
\Cref{cor:hybrid_thermal_specialization} gives the exact energy constraint
\(\sum_i W_i \G_i=2\rho E\).

Training proceeds in two stages. We first fit \NN{} as a local equilibrium
map on reference pairs \((\boldsymbol{U},\Geqlattice)\), then fine-tune it
inside the LB solver on short autoregressive rollouts so the closure sees the
states it induces. The experiment sections report the number of trajectory
samples used in each case; optimizer settings, rollout lengths, TVD
regularization, and per-case architectures are in
\Cref{sec:training,sec:experimental}.

\FloatBarrier
\section{Experiments}
\label{sec:experiments}
\label{sec:numerical}
\label{section:specific_datasets}

We evaluate NeurDE on six conserved systems with LB-host rollouts and a finite-volume DUGKS interface probe: LWR traffic flow, Buckley--Leverett flow, Burgers shock formation, subsonic Sod, transonic Sod, and supersonic flow past a cylinder.
The main baseline is the corresponding
polynomial-equilibrium LB host (LB+Poly), a standard numerical baseline.

The organizing principle is placement: the network learns
the local BE closure, while the host kinetic solver supplies transport and
conservation. LB and finite-volume (FV) DUGKS are two ways to query the same
interface. The experiments test five roles for this placement: stabilization,
acceleration, correction, transfer, and ablation against broader learned
updates. Because shock phase errors
can dominate pointwise norms, we report stability, positivity, energy-moment
residuals, wave-location errors, and runtime. Standard \(L^1\) and \(L^2\)
errors fail to properly quantify errors when shocks are present.

Unless noted otherwise, the main compressible experiments train compact
\(\mathcal{O}(10^3)\)-parameter MLP closures on the first \(500\) states of a
single trajectory and test autoregressively beyond that window. Exact
architectures, data splits, and solver settings are in \Cref{sec:experimental}.
\Cref{tab:variant_short} summarizes where learning enters and what each
experiment tests about that placement.

\begin{table}[t]
\centering
\scriptsize
\setlength{\tabcolsep}{3pt}
\renewcommand{\arraystretch}{0.88}
\begin{tabularx}{\linewidth}{@{}>{\raggedright\arraybackslash}p{0.15\linewidth}>{\raggedright\arraybackslash}p{0.38\linewidth}>{\raggedright\arraybackslash}X@{}}
\toprule
\textbf{Role} & \textbf{Where learning enters} & \textbf{Main signal} \\
\midrule
Stabilize & Local BE equilibrium in the host rollout: \(\Geqlattice\) for compressible tests, \(\Feqlattice\) for scalar laws & Sod/cylinder/late-start/OOD rollouts remain stable where LB+Poly fails or degrades \\
Accelerate & Same equilibrium map replaces cellwise Newton entropy solves & Sod cases 1/2 solver-only runtime improves by \(27.4\times/37.1\times\) \\
Correct & Learned Shakhov-type face closure inside FV/DUGKS transport & Contact error \(6.22\to3.22\) cells with \(929/929\) stable horizon \\
Ablate & Full post-collision map, deliberately learning beyond equilibrium & Full-collision surrogate develops unphysical temperature modes \\
Transfer & Conservative moment correction at the closure interface & LWR/Buckley--Leverett errors \(1.95{\times}10^{-2}/3.61{\times}10^{-2}\); Burgers shock improvement \\
\bottomrule
\end{tabularx}
\caption{Evidence organized by learned object. The common target is the local
BE closure/equilibrium; LB and finite-volume (FV) DUGKS are host transport
realizations. Detailed setups and metrics are in
\Cref{sec:experimental,appx:sod_metrics,appx:cylinder_metrics}; FV/DUGKS, FNO,
scalar-law variants, and fully neural ablations are in
\Cref{sec:fvfd_main,appx:FNO,appx:additional_conservation_laws,sec:ablation}.}
\label{tab:variant_short}
\end{table}

\subsection{Scalar conservation-law probes}
\label{sec:burgers_main}
Before the compressible benchmarks, three scalar laws probe whether the same
single-population closure \(\Feqlattice\) and moment-matching interface
transfer beyond Euler. LWR tracks the reference through \(t=0.4975\) with
final relative error \(1.95\times10^{-2}\); Buckley--Leverett stays close to
the smooth branch and sharp front with errors below \(3.61\times10^{-2}\);
Burgers isolates shock formation and benefits from exact moment enforcement
without TVD regularization. Full setups and plots are in
\Cref{appx:lwr_setup_conservative,appx:buckley_setup_conservative,appx:burgers_setup_conservative}.

\subsection{Sod shock tubes}
\label{section:Sod_cases}\label{subsection:Sod_case_1}\label{subsection:Sod_case_2}
\begin{figure}[htp!]
    \centering
    \includegraphics[width=\linewidth]{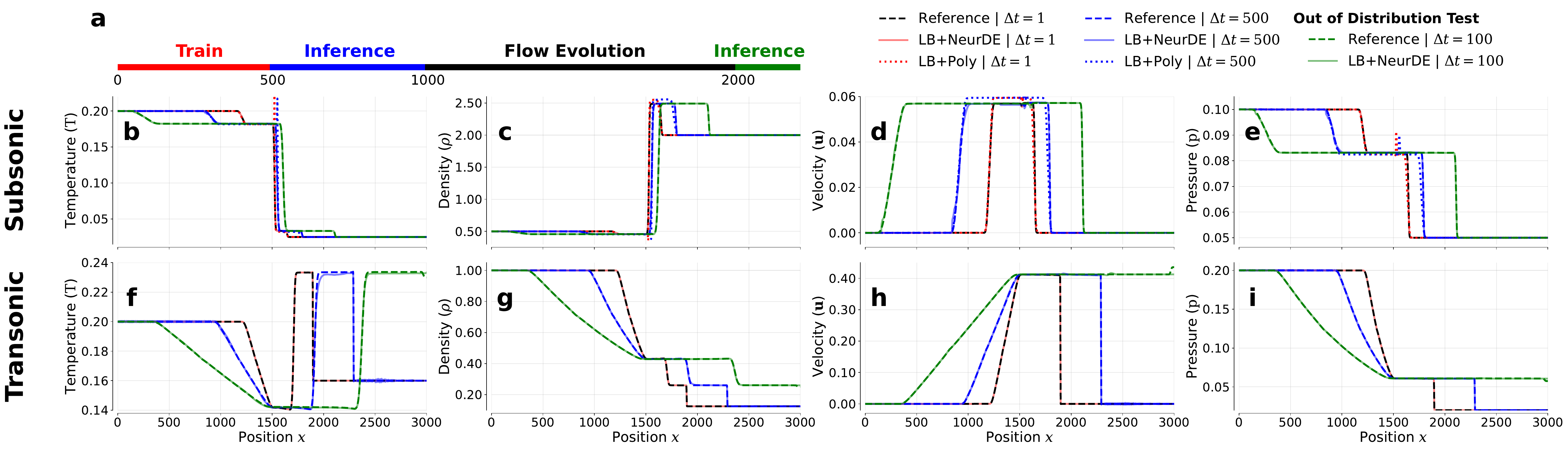}
    \vspace{-0.5em}
    \caption{Subsonic and transonic Sod rollouts after training on
    \(t<500\). The subsonic and transonic cases are shown on the left and
    right, respectively. LB+\NN{} tracks the reference through \(t=999\),
    while the polynomial LB closure degrades in the subsonic case and
    diverges in the transonic case. Thick solid, dotted, and thin solid curves
    denote \LBNN{}, reference, and polynomial LB, respectively; colors mark the
    reported rollout times in each panel.
    }
    \label{fig:sod_all}
    \vspace{-0.5em}
\end{figure}

\begin{figure}[h]
    \begin{minipage}{0.49\textwidth}
        \centering
        \includegraphics[width=\linewidth]{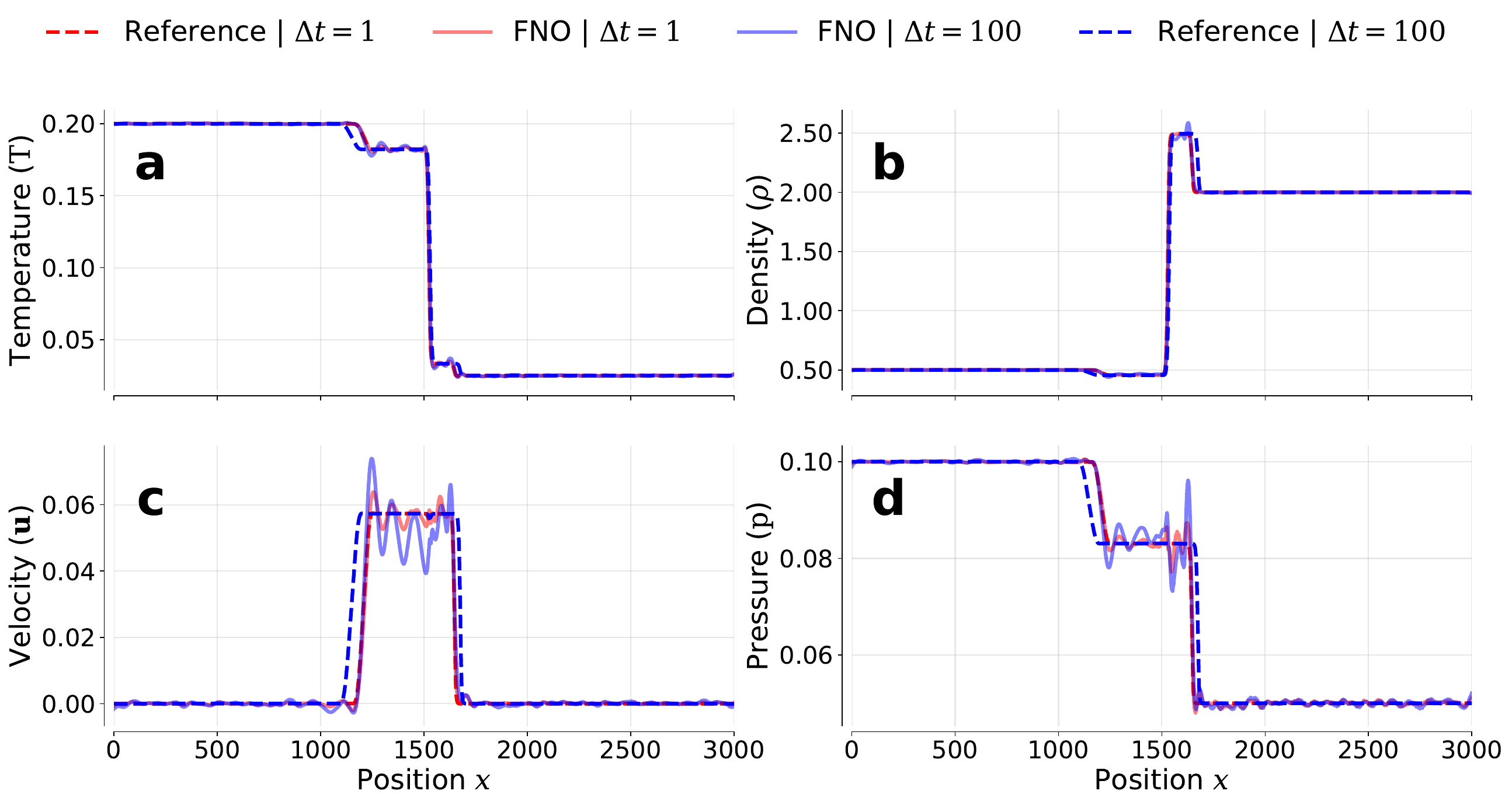}
        \caption{FNO benchmark results on subsonic sod}
        \label{fig:fno_sod}
    \end{minipage}
    \hfill
    \begin{minipage}{0.49\textwidth}
        \centering
        \includegraphics[width=\linewidth]{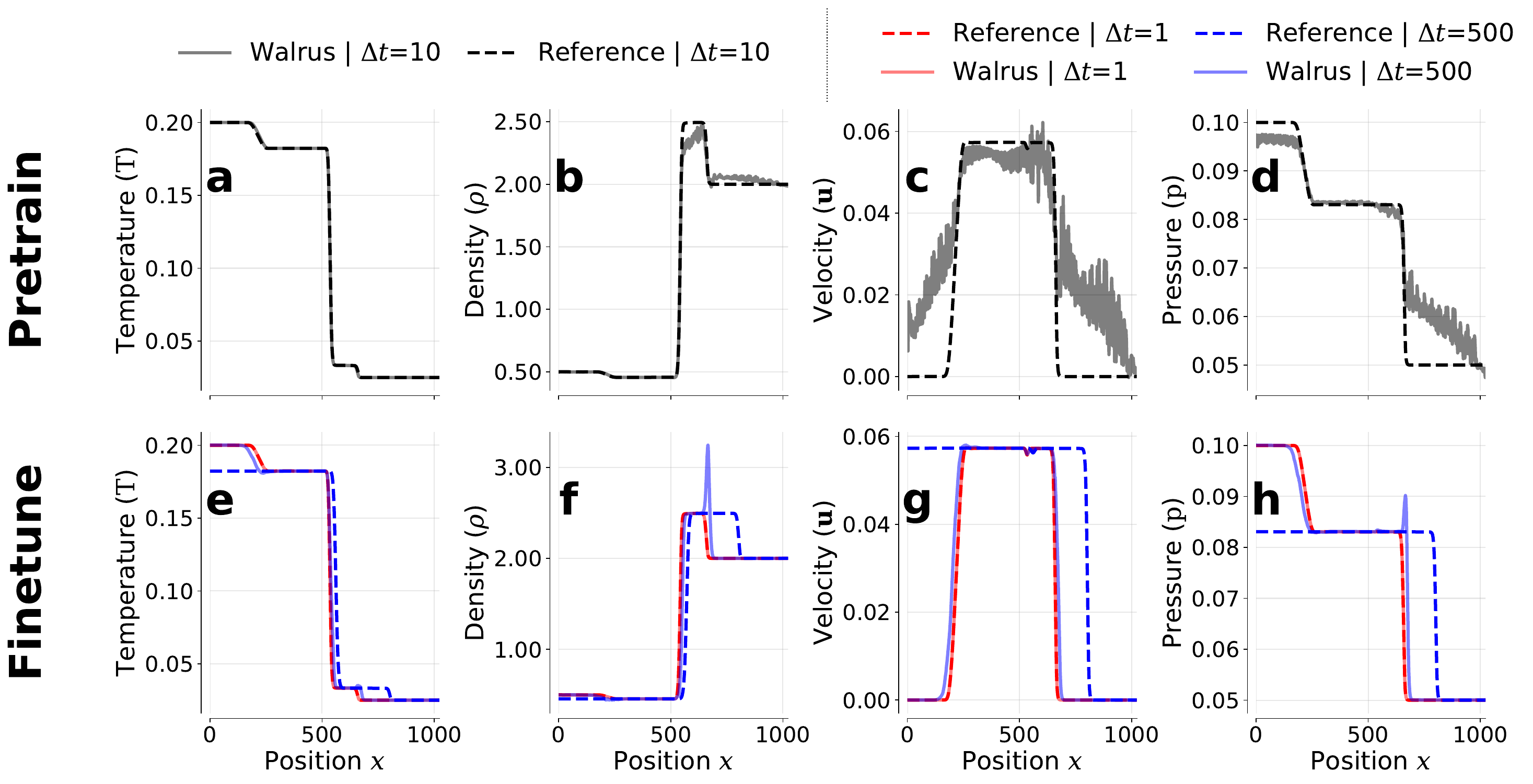}
        \caption{Walrus benchmark results on subsonic sod}
        \label{fig:walrus_sod}
    \end{minipage}
\end{figure}

The Sod tests isolate the placement question: can a
learned local equilibrium retain entropic wave structure, avoid
polynomial-closure failure, and remove the cellwise Newton solve? The benchmark is a Riemann problem with rarefaction, contact, and shocks. We compare against a standard
two-population polynomial LB closure
\cite{karlin2013consistent,saadat2019lattice}.
Both Sod models are trained on the first \(500\) time steps of one trajectory
and evaluated over \(499\)-step rollouts initialized at \(t=500\). The
subsonic case uses a width-\(32\) MLP closure, while the transonic case uses a
width-\(64\) closure with TVD regularization. The Riemann initial conditions
\eqref{eq:Sod_case_1}--\eqref{eq:Sod_case_2}, lattice parameters, and reference
solvers are listed in \Cref{sec:experimental}.
We evaluate \LBNN{} on both problems (Fig.~\ref{fig:sod_all}), and we evaluate FNO~\cite{li2021fourier} and Walrus~\cite{walrus} on the Sod subsonic problem (Figs.~\ref{fig:fno_sod} and \ref{fig:walrus_sod}, respectively).

The easier Sod subsonic problem clearly delineates stable and fragile models.
In Fig.~\ref{fig:sod_all} we see that \LBNN{} and the Newton reference are stable.
However, FNO becomes unstable by introducing large oscillations after a single step.
In the longer \(100\)-step rollout, the predicted discontinuities do not move within, indicating that FNO struggles to learn the dynamics.
While the pretrained Walrus model struggles with shocks, the fine-tuned model is much more stable and shows good agreement after a single step.
During longer rollouts, we again see the discontinuities hardly move, further demonstrating the challenges of learning shock dynamics.
Both the Walrus and the numerical LB+Poly methods demonstrate overshoots at the shock front, which are often caused by low-order approximations of the underlying system.
Thus the ML baselines are useful as stress tests rather
than matched long-rollout solvers here: FNO develops large oscillations almost
immediately and is stable for fewer than \(10\) autoregressive steps, while
fine-tuned Walrus remains finite in the plotted \(100\)-step rollout but still
struggles to advect the discontinuities over longer horizons.
\NN, however, stably captures the entire Sod subsonic dynamic, outperforming the much larger FNO (\(\mathcal{O}(10^4)\times\)) and the Walrus foundation model (\(\mathcal{O}(10^6)\times\)) pretrained on terabytes of data.
Due to FNO and Walrus's visibly poor performance on the simplest of our compressible datasets, we no longer include them in our further and more challenging comparisons.


Given NeurDE's improvement over the ML baselines, we next compare the learned closure against matched numerical closures.
The low-order polynomial LB baseline, LB+Poly, is stable during the Sod subsonic problem, although it overshoots in every variable (Fig.~\ref{fig:sod_all}(b-e)).
However, LB+Poly fails at harder problems, blowing up after \(10\) steps in the transonic Sod case.
The root-finding entropic reference is stable and accurate
in both Sod experiments (Fig.~\ref{fig:sod_all}). Table~\ref{tab:sod_main}
therefore reports Newton front metrics and solver-only runtime for both Sod
cases: \LBNN{} replaces the expensive local Newton--Raphson closure solve with
forward passes and obtains \(27.4\times\) and \(37.1\times\) solver-only
speedups in cases~1 and~2, respectively.

\begin{table}[t]
\centering
\scriptsize
\renewcommand{\arraystretch}{0.96}
\setlength{\tabcolsep}{3pt}
\resizebox{\linewidth}{!}{%
\begin{tabular}{@{}lcccc@{}}
\toprule
\textbf{Metric} & \shortstack{\textbf{Case 1}\\\textbf{\LBNN{}}} & \shortstack{\textbf{Case 1}\\\textbf{Newton}} & \shortstack{\textbf{Case 2}\\\textbf{\LBNN{}}} & \shortstack{\textbf{Case 2}\\\textbf{Newton}} \\
\midrule
Wave S/C/R error [cells] & \(17.61/7.53/29.12\) & \(17.61/1.47/23.12\) & \(2.20/1.35/30.10\) & \(3.20/9.65/29.10\) \\
Profile P/S/C error & \(0.0333/0.0243/0.0215\) & \(0.0330/{-}/{-}\) & \(0.0120/0.0092/0.0091\) & \(0.0139/{-}/{-}\) \\
Max energy-moment residual & \(1.90{\times}10^{-7}\) & --- & \(2.53{\times}10^{-7}\) & --- \\
Positivity violations & 0 & --- & 0 & --- \\
\midrule
Solver-only time [s] & \textbf{0.74} & 20.35 & \textbf{1.55} & 57.50 \\
Throughput [steps/s] & 672.16 & 24.52 & 321.89 & 8.68 \\
Runtime speedup & \(27.4\times\) & --- & \(37.1\times\) & --- \\
\bottomrule
\end{tabular}%
}
	\caption{Sod diagnostics from rollouts initialized at
	\(t=500\). S/C/R denote shock, contact-edge, and rarefaction-tail location
	errors; P/S/C denote post-shock plateau, shock-aligned, and contact-aligned
	profile errors. Wave, profile, positivity, and energy-moment rows summarize
	the learned rollouts at \(t=999\). Newton front metrics are from the
	case-specific NN/Newton/exact comparisons, and the runtime rows are repeated
	case-specific solver-only throughput benchmarks over the \(499\)-step
	rollout. Metric definitions are in \Cref{appx:sod_metrics}.}
\label{tab:sod_main}
\end{table}

\begin{figure}[htp!]
    \centering
    \begin{minipage}{0.30\linewidth}
    \centering
    \includegraphics[width=\linewidth,trim=2550 463 10 1140,clip]{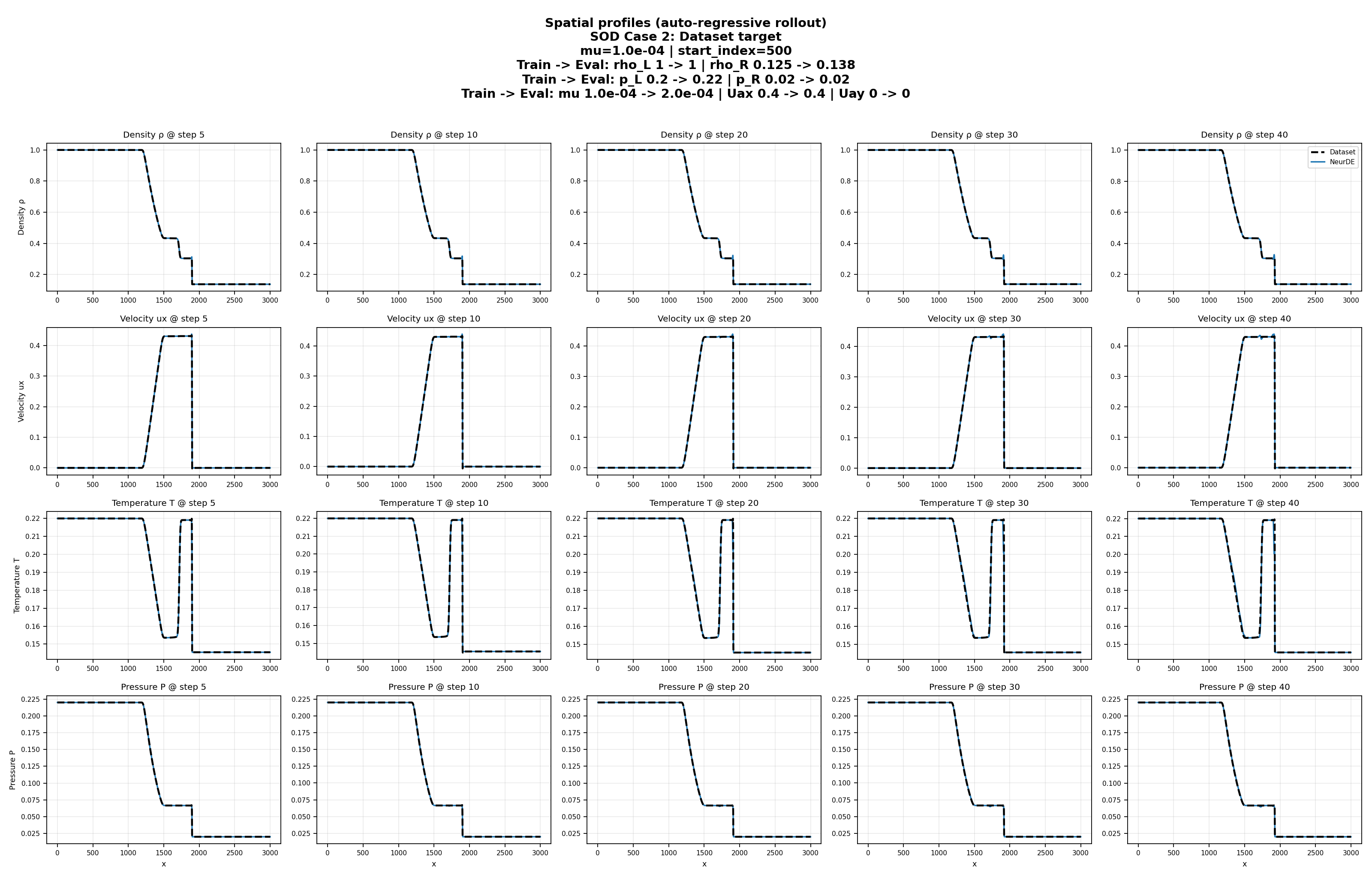}
    \end{minipage}\hspace{0.08\linewidth}
    \begin{minipage}{0.30\linewidth}
    \centering
    \includegraphics[width=\linewidth,trim=2550 463 10 1140,clip]{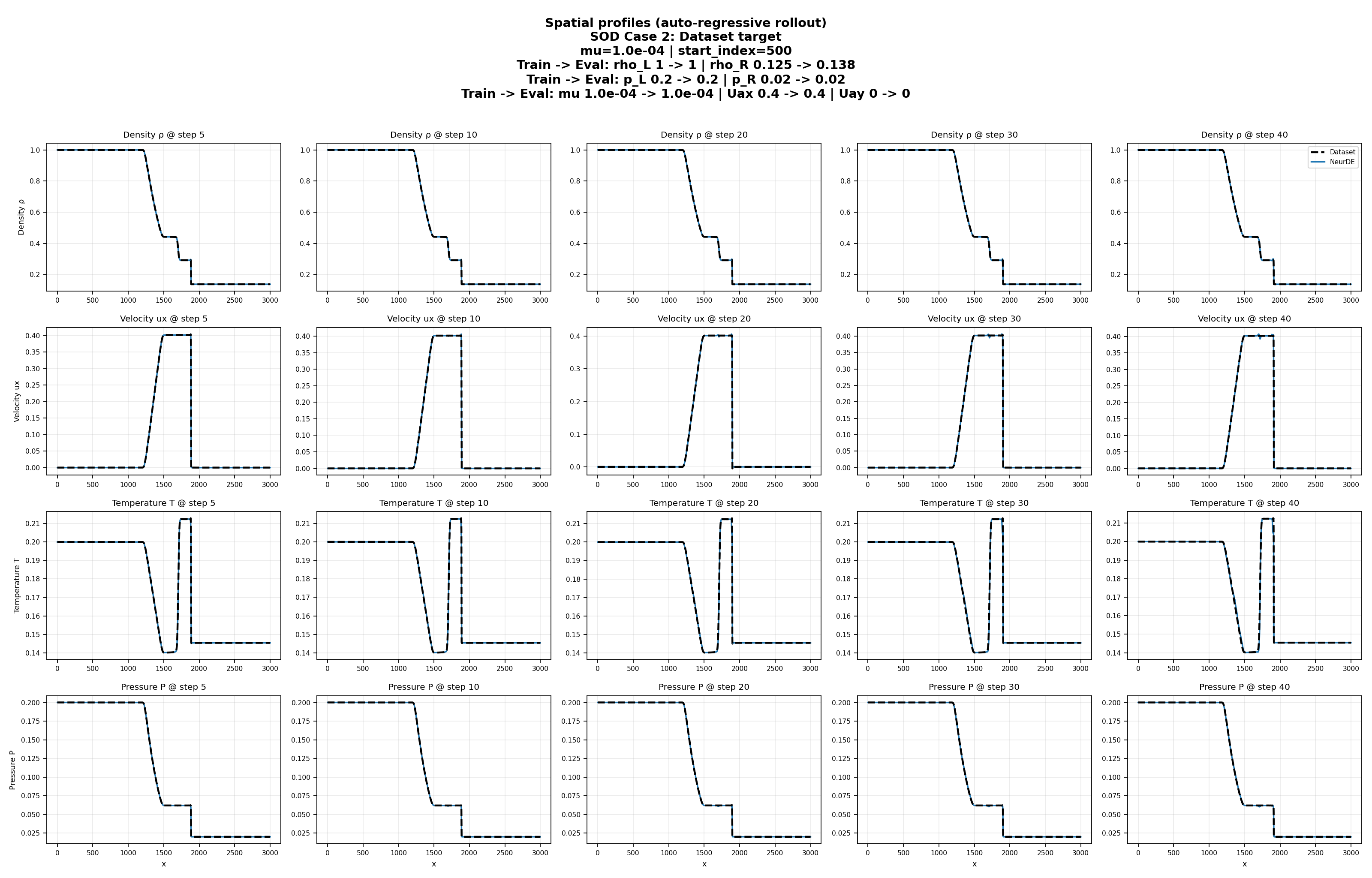}
    \end{minipage}
    \vspace{-0.5em}
    \caption{Final reported temperature samples from transonic Sod OOD
    probes without retraining. \emph{Left:} the moderate coupled shift
    \((\rho_R{+}10.4\%,p_L{+}10\%,\mu{+}100\%)\), cropped from
    \Cref{fig:ood_transonic_right_rho_left_pressure_viscosity_changed}.
    \emph{Right:} the single-parameter right-density shift
    \(\rho_R{+}10.4\%\), cropped from
    \Cref{fig:ood_transonic_right_density_changed}. The full multi-time
    density, velocity, temperature, and pressure panels are in the appendix.}
    \label{fig:ood_main}
    \vspace{-0.5em}
\end{figure}

\paragraph{Out of Distribution Tests.}
\label{subsub:shock_long}\label{sec:ood}
To test closure transfer, we reinitialize the fixed Sod
models at \(t=2000\) for \(100\) steps and perturb the transonic initial
conditions without retraining. Both late-start models remain stable, and
\Cref{fig:ood_main} shows stable tracking for harder and easier transonic
shifts, supporting the view that the BE closure, not a fitted trajectory map,
carries the transferable information.

\FloatBarrier
\subsection{Supersonic cylinder.}
\label{section:supersonic_flow}
The cylinder experiment asks whether the same learned
energy closure can recover a genuinely two-dimensional shock structure. Unlike
Sod, the dominant feature is a detached bow shock around a curved wall, so the
test separates what the closure should learn from what remains the host
solver's responsibility: transport and boundary treatment. At
\(\mathrm{Ma}_\infty=1.8\) and \(\mathrm{Re}=300\), LB+\NN{} is trained on the
first \(500\) steps and rolled forward to \(t=999\). The detected standoff
matches the reference grid location, pressure-jump errors remain below
\(2\%\), and density and temperature stay positive. The largest visible
deviations occur near boundary-influenced regions, which is consistent with
the fact that NeurDE does not learn a boundary closure.

We also use a harder late-start probe: a separate model is
trained on only the first \(150\) steps, before the developed bow-shock
structure is present, and then restarted from \(t_0=900\). Its \(t=999\)
visualization shows that the learned closure can reconstruct the developed
shock topology from states far outside the trajectory-training window. Full
setup and metric definitions are in \Cref{sec:experimental,appx:cylinder_metrics}.

\begin{figure}[htp!]
\centering
\begin{minipage}{0.54\linewidth}
    \includegraphics[width=\linewidth]{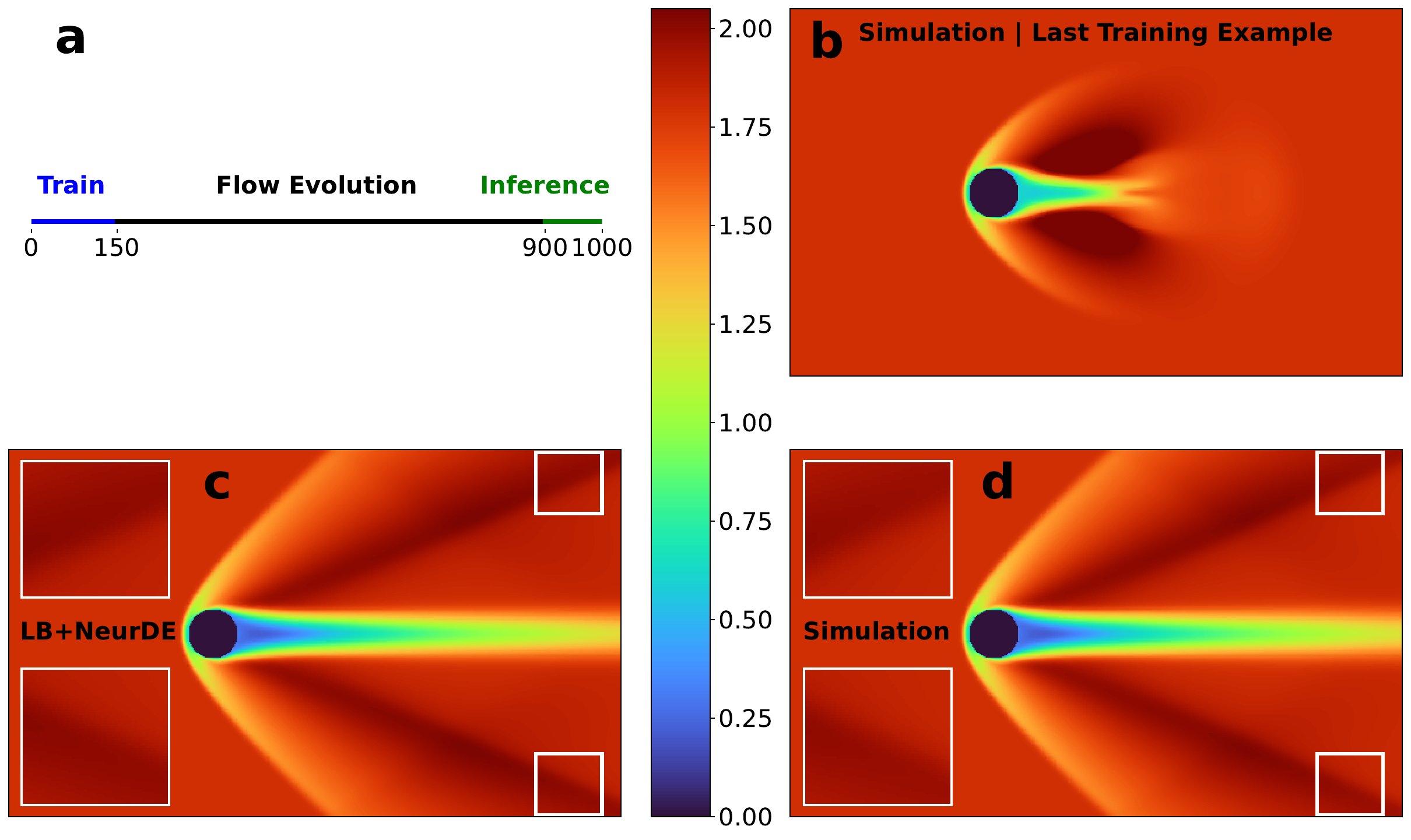}
\end{minipage}\hfill
\begin{minipage}{0.42\linewidth}
\centering
\small
\renewcommand{\arraystretch}{0.92}
\begin{tabular}{@{}lc@{}}
\toprule
Cylinder diagnostic & Value \\
\midrule
Stable horizon & 500/500 \\
Positivity violations & 0 \\
Min.\ \(\rho\) / \(T\) & 0.163 / 0.092 \\
Standoff error [cells] & 0.0 \\
Pressure-jump rel.\ error & \(1.79\%\) \\
Pressure-ratio rel.\ error & \(0.621\%\) \\
Centerline Mach error & \(2.65{\times}10^{-3}\) \\
\bottomrule
\end{tabular}
\end{minipage}
\caption{Supersonic flow past a cylinder at \(\mathrm{Ma}_\infty=1.8\),
\(\mathrm{Re}=300\). \emph{Left:} late-start Mach-number field at
\(t=999\), after training only on the first \(150\) steps and restarting
from \(t_0=900\). \emph{Right:} diagnostics for the matched \(500\)-step
rollout to \(t=999\). The zero standoff error means the discrete
centerline shock detector agrees with the reference grid detector.}
\label{fig:cylinder_main}
\vspace{-0.5em}
\end{figure}

\begin{figure}[htp!]
    \centering
    \includegraphics[width=0.68\linewidth]{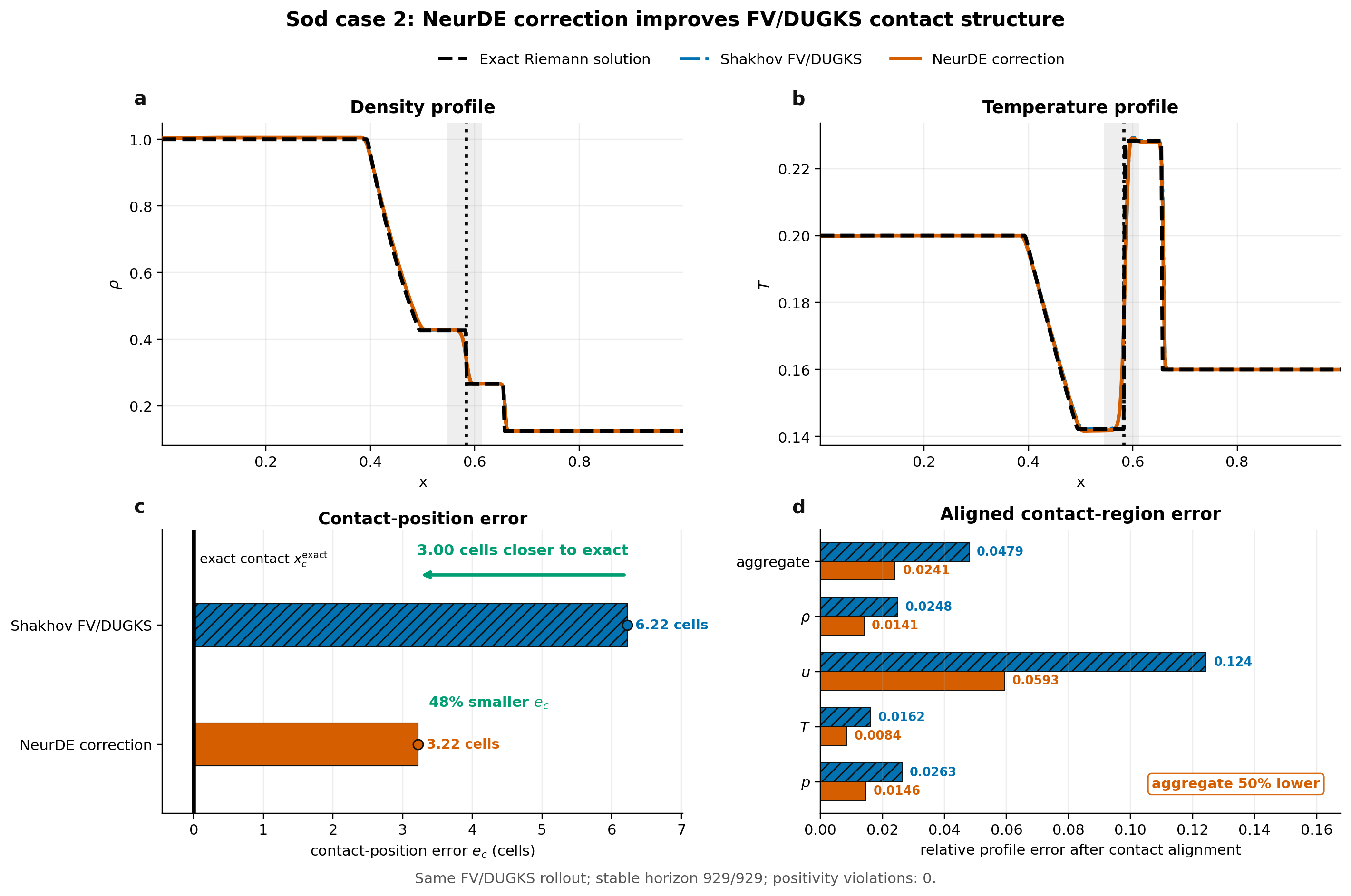}
    \vspace{-0.5em}
    \caption{FV/DUGKS contact correction on transonic Sod case 2. The
    exact-supervised NeurDE face-source correction moves the contact marker
    \(3.00\) cells closer to the exact Riemann contact than analytic Shakhov
    \cite{shakhov1968generalization}
    and halves the contact-aligned profile error, while preserving a
    \(929/929\) stable horizon with zero positivity violations. Estimator
    details are given in \Cref{appx:fvfd_dugks}.}
    \label{fig:fvfd_contact_main}
    \vspace{-0.5em}
\end{figure}

\FloatBarrier
\paragraph{Closure interface with fixed transport.}
\label{sec:fvfd_main}
The FV/DUGKS test is a portability and correction probe,
not another dataset row. DUGKS is a more general off-lattice finite-volume
kinetic update than collide--stream LB, so improving its closure without
replacing its transport step is a stronger test of the BE interface. On
transonic Sod case 2, FV+NeurDE uses the same local closure idea inside DUGKS
(face-level FV details are left to \Cref{appx:fvfd_dugks}) and improves the
analytic Shakhov closure~\cite{shakhov1968generalization}: the contact-location
error drops from \(6.22\) to \(3.22\) cells, the contact-aligned profile error
drops from \(4.79{\times}10^{-2}\) to \(2.41{\times}10^{-2}\), and the
\(929/929\)-step stable horizon with zero positivity violations is preserved
(\Cref{fig:fvfd_contact_main,tab:fvfd_dugks_metrics}). This matters because the
learned BE closure improves an advanced FV kinetic solver while leaving its
transport machinery intact.

\FloatBarrier
\paragraph{Additional probes.}
\label{section:evaluating_distant_time}\label{subsub:cylinder_long}
	The appendix gives later Sod failures, full OOD
	profiles, late-start cylinder probes, scalar-law details, and ablations
	against polynomial LB and broader neural replacements of the kinetic update.

\begin{ack}
J.A.L.B. thanks Jesse Chan and Khemraj Shukla for helpful discussions. J.A.L.B.
and K.H. acknowledge computational support from Lawrence Berkeley National
Laboratory. J.A.L.B. and K.H. thank Yotam Yaniv for assistance with the FNO
experiments. M.V.d.H. acknowledges support from the DOE under grant
DE-SC0020345, the NSF under grant DMS-2108175, the Simons Foundation under the
MATH+ X program (Award 271853, Project ``Simons Chair in Computational and
Applied Mathematics and Earth Science at Rice University''), and Occidental
Petroleum. M.W.M. acknowledges partial support from the DOE, NSF, and ONR. I.D.
acknowledges support from the European Research Council Starting Grant 852821
(SWING). The authors declare no competing interests.
\end{ack}

\label{sec:main_end}
\bibliographystyle{plainnat}
\bibliography{ref}

\appendix

\section{Supplementary Notes}
\label{app:details}
\label{app:theory}
\label{app:entropy_assumptions}

This appendix is organized in the order in which the corresponding material is
used in the main paper. We first give the kinetic notation and closure theory
supporting the Method section, then the implementation and training details,
and finally the empirical appendices in the same sequence as the Results
section: Sod and its baseline/component ablations, cylinder flow, FV/DUGKS, and
scalar conservation-law probes.

\begin{table}[h]
\centering
\small
\renewcommand{\arraystretch}{0.96}
\begin{tabularx}{\linewidth}{@{}p{0.23\linewidth}p{0.39\linewidth}p{0.11\linewidth}X@{}}
\toprule
Main-text item & What is documented here & Audience & Location \\
\midrule
Kinetic framework and closure theorem & Notation, BGK/discrete-velocity
background, entropy structure, maximum-entropy projection, entropy-defect
balance, conservative projection, and hybrid energy-channel specialization &
\(\text{theorem}\) &
\Cref{appendix:preliminaries,appx:conservative_neurde,thm:hybrid_two_population_neurde,cor:hybrid_thermal_specialization} \\
NeurDE architecture and training & Learned moment space, exponential-family
parameterization, compressible LB realization, two-stage training, and
experimental configurations &
\(\text{both}\) &
\Cref{appendix:methods_full,appendix:NeurDE,sec:experimental} \\
Sod shock tubes and ablations & Metrics, FNO baseline, polynomial-error
diagnostics, late-start and OOD probes, full-collision ablation, and learned
streaming ablation &
\(\text{experiment}\) &
\Cref{appx:sod_shock,appx:FNO,sec:ablation} \\
Cylinder flow & Boundary conditions, metrics, first-\(500\)-step rollout,
and late-start cylinder probe &
\(\text{experiment}\) &
\Cref{appendix:cylinder} \\
FV/DUGKS interface & Face-closure setup, analytic Shakhov comparison,
exact-supervised correction, metrics, and contact estimator &
\(\text{both}\) &
\Cref{appx:fvfd_dugks} \\
Scalar conservation laws & Burgers raw-vs-conservative comparison, LWR, and
Buckley--Leverett conservative rollouts &
\(\text{experiment}\) &
\Cref{appx:additional_conservation_laws} \\
\bottomrule
\end{tabularx}
\caption{Appendix roadmap. The table records where each main-text claim is
made precise, either as a formal closure statement or as an empirical
diagnostic.}
\label{tab:appendix_roadmap}
\end{table}

\Cref{appendix:notation,appendix:preliminaries} fix notation and recall the
kinetic background. The formal counterparts of
\Cref{thm:main_neurde_structure} are stated and proved in
\Cref{ass:discrete_admissible,thm:discrete_maxent,thm:discrete_defect,cor:discrete_approx_H,prop:two_state_weighted_projection,cor:two_state_projection_positivity,thm:two_state_conservative_neurde_entropy,thm:hybrid_two_population_neurde,cor:hybrid_thermal_specialization}.
\Cref{appendix:methods_full,appendix:NeurDE} then give the full implementation
and training record. The experiment appendices follow the Results section:
\Cref{appx:sod_shock,sec:ablation,appendix:cylinder,appx:fvfd_dugks,appx:additional_conservation_laws}.

\section{Notation Glossary}
\label{appendix:notation}
A summary of the notation used in this paper is presented in \cref{notation:glosary}.
\begin{table}[H]    
    \small
    \renewcommand{\arraystretch}{1.2} 
    
    \begin{tabularx}{\textwidth}{@{} l L l @{}}
        \toprule
        \textbf{Notation} & \textbf{Description} & \textbf{Reference} \\
        \midrule
        
        $\Omega, d$ & Spatial domain and number of components of spatial domain $\Omega \subset \real^d$ & \\
        $f^{\mathrm{MB}}$ & Maxwellian distribution & \Cref{eq:Maxwellian} \\
        $f^{\mathrm{eq}}$ or $\Feqlattice_i$ & Equilibrium distribution of $f$ or $\F_i$ & \\
        $\collision{\cdot}$ & Boltzmann collision operator & \Cref{eq:Boltzmann_Transport_eq} \\
        $\otimes, \, \osym$ & Tensor product and symmetric tensor product & \\
        
        \addlinespace[1ex]
        \rowcolor{headergray} 
        \multicolumn{3}{@{}l}{\textbf{Discrete velocity model and LB}} \\
        \addlinespace[0.5ex]
        
        $\Omega_{\{\F_i, \G_i\}}$ & BGK-type collision operator & \Cref{eq:LBM} \\
        $\boldsymbol{U}$ & Conserved macroscopic observables recovered by moment projection; in the compressible setting, $\boldsymbol{U}=(\rho,\rho \Velocity,\rho E)^\top$, with $E$ the specific total energy. & \Cref{eq:projection,eq:micro_into_macro} \\
        $\heatflux^{\mathrm{MB}}, \pressuretensor^{\mathrm{MB}}, \boldsymbol{R}^{\mathrm{MB}}$ & Maxwellian higher-order moments & \Cref{eq:higher-order} \\
        $\heatflux^{\mathrm{eq}}, \pressuretensor^{\mathrm{eq}}, \boldsymbol{R}^{\mathrm{eq}}$ & Equilibrium higher-order moments & \Cref{appendix:two_populations} \\
        $\{\velocity\}_{i=1}^Q$ & Discrete velocities & \Cref{eq:Boltzmann_BGK_discrete} \\
        $\{\latticevelocity_i\}_{i=1}^Q$ & Lattice velocities & \Cref{eq:LBM_algorithm_dimensionless} \\
        $\{\F_i, \G_i\}$ & Discrete velocity populations; $(t, \bx, \latticevelocity_i)\in \real^d_{\ge 0}\times \Omega \times \{\velocity_i\}_{i=1}^Q$ & \Cref{eq:Boltzmann_BGK_discrete} \\
        $\{W_i\}_{i=1}^Q$ & Temperature related weights & \Cref{appendix:two_populations} \\
        $\G_i^\ast$ & Quasi-equilibrium & \Cref{eq:quasiequilibrium} \\
        $\tau_1, \tau_2$ & Relaxation related viscosity and thermal conductivity & \Cref{eq:LBM} \\
        
        \addlinespace[1ex]
        \rowcolor{headergray}
        \multicolumn{3}{@{}l}{\textbf{Splitting algorithm}} \\
        \addlinespace[0.5ex]
        
        $\phiC$ & Solution operator of the collision problem & \Cref{eq:LBM_collision} \\
        $\phiS$ & Solution operator of the streaming (free flow) & \Cref{eq:LBM_streaming} \\
        $\phiS \phiC$ & LB algorithm & \Cref{eq:LBM_algorithm_dimensionless} \\

        \addlinespace[1ex]
        \rowcolor{headergray}
        \multicolumn{3}{@{}l}{\textbf{Surrogate model for the equilibrium}} \\
        \addlinespace[0.5ex]
        
        $\underbalpha, \underbvarphi$ & Neural networks & \Cref{eq:levermore_closure_NN} \\
        $\phiNN_i(\cdot)$ & Surrogate model for the equilibrium \NN{} \newline $\phiNN_i(\cdot)= \exp \left(\underbalpha \cdot \underbvarphi\right)(\latticevelocity_i)$ & \Cref{eq:levermore_closure_NN} \\
        $\phiCNN$ & Hybrid solution of the BGK-type collision with ML surrogate & \Cref{eq:collision_as_operator_LB} \\
        $\mathbb{M}$ & Moment space (span of $\underbvarphi$) & \Cref{eq:M_space} \\
        $\phiS \phiCNN$ & Hybrid model \LBNN{} & \Cref{alg:LBM_NN_algorithm} \\
        $\mathcal{D}$ & Operator mapping distributions to observables & \Cref{eq:operator_distr_to_eq} \\
        $(\rho,\Velocity,\temperature)^\top$ & Primitive variables used as neural-network inputs in the compressible-flow experiments. & \Cref{eq:macroscopical_LBM_2_pop,tab:ArchitectureSummary} \\

        \addlinespace[1ex]
        \rowcolor{headergray}
        \multicolumn{3}{@{}l}{\textbf{Experiments}} \\
        \addlinespace[0.5ex]

        $\pressure = R \rho \temperature$ & Pressure calculated by the ideal gas law & \\
        $\mathrm{Ma}$ & Local Mach number; $\mathrm{Ma} = (\Velocity \cdot \Velocity)^{1/2}(\gamma R \temperature )^{-1/2}$ & \Cref{section:supersonic_flow} \\
        $\gamma$ & Specific heat ratio & \Cref{sec:specific_arch} \\
        $\TV(\cdot)$ & Total variational principle & \Cref{eq:TVD_condition} \\
        $\mathrm{Re}$ & Reynolds number & \\
        $\mathrm{Ma}_\infty$ & Far-field Mach number & \Cref{section:supersonic_flow} \\
        $\Velocity_\infty$ & Far-field (flow) velocity & \Cref{section:supersonic_flow} \\
        
        \bottomrule
    \end{tabularx}
        \caption{\textbf{Glossary of mathematical notation.} Symbols are organized by topic: kinetic theory and lattice Boltzmann fundamentals, numerical algorithms and splitting schemes, neural network surrogate models, and experimental parameters.}
    \label{notation:glosary}
\end{table}

\newpage
\section{Preliminaries}\label{appendix:preliminaries}
This appendix provides supplementary details on the kinetic theory background introduced in the main text (\cref{sec:kinetic_background}) and develops the entropy-theoretic foundations of the NeurDE equilibrium ansatz.

The material is organized as follows. 
\Cref{subsection:Boltzmann_eqn} summarizes the structure of the general Boltzmann collision operator. 
\Cref{appendix:maxwellian} defines the Maxwellian equilibrium distribution (introduced in \cref{sec:kinetic_background}) in terms of its variational characterization as the entropy maximizer subject to conservation constraints.
Once this abstract equilibrium is established, we introduce in \Cref{appx:BGKmodel} the BGK relaxation model as a practical surrogate for the general collision operator. \Cref{appendix:discrete_kinetic} then discusses the discretization of the velocity space, leading to the so-called discrete velocity models for the Boltzmann equation. 
\Cref{appendix:fluid_dynamics_description} explains how the Boltzmann equation yields local conservation laws and, through constitutive relations, leads to macroscopic fluid dynamics. 
\Cref{appendix:M_space} reviews classical moment closure hierarchies following Grad~\cite{grad1949kinetic} and their systematic extension by Levermore~\cite{levermore1996moment}. 
\Cref{appendix:higher-order_moments} provides explicit expressions for higher-order moments of the Maxwell--Boltzmann distribution. 
\Cref{appendix:splitting_model} outlines the operator splitting technique used in the LBM; and \cref{appendix:LBM} summarizes the fundamental principles of LBM. 
\Cref{remark:lattices_closure} discusses closure (aliasing) errors introduced by discrete velocity sets, a key challenge in LB formulations also noted in the main text (see below \cref{eq:M_space}).
Finally, \cref{sec:entropy_structure_neurde} develops the entropy structure of the NeurDE equilibrium: we show that the exponential ansatz \cref{eq:levermore_closure_NN} arises from constrained entropy minimization on the discrete velocity set, prove that it defines a learned maximum-entropy closure with a strictly convex dual entropy and symmetrizable hyperbolicity, establish a formal $H$-theorem for the \LBNN{} collision, and quantify the entropy defect introduced by imperfect moment matching.

\subsection{The Boltzmann Transport Equation}
\label{subsection:Boltzmann_eqn}

Here, we provide an overview of classical kinetic theory. We consider the Boltzmann transport equation, see \cite{levermore1996moment, caflisch1982shock, ohwada1998higher} for references. Its dimensionless form is given by:
\begin{equation}\label{eq:Boltzmann_Transport_eq}
\left( \partial_t + \velocity \cdot \nablax \right) f(t, \bx, \velocity) = \dfrac{1}{\e}\collision{f}, \qquad t>0,\, (\bx, \velocity)\in \Omega \times \real^d   ,
\end{equation}
where $f(t, \bx, \velocity)$ represents a probability density distribution function, modeling the probability of finding a (gas) particle at time $t$, with position $\bx \in \Omega$, and velocity $\velocity\in \real^d$.
The parameter $\e$ is the Knudsen number, defined as the ratio of the mean free path over the length scale, which characterizes the degree of rarefaction of the gas. The collision operator is a quadratic integral operator over $\velocity$, whose domain $\mathscr{D}(\bC)$ is contained in the cone of nonnegative functions $f$.\ 

\paragraph{Characterization of the Collision Operator.}
The collision operator $\bC$ satisfies three properties, that relate with conservation laws, local dissipation, and symmetries. We recall from \cref{sec:kinetic_background} that we use the symbol $\bra \cdot \ket$ to represent integration over the velocity space; namely, $\bra  \psi(f) \ket = \int_{\real^d} \psi(f) d\velocity$. Let us briefly review these properties.  \
\begin{enumerate}
    \item In the collision process, \emph{mass, momentum, and energy are conserved}, i.e., for any distribution $f$, we~have:
\begin{equation}\label{eq:local_conservation}
    \left \langle \begin{pmatrix} 1,
      \velocity ,
     \dfrac{1}{2}\velocity \cdot \velocity 
     \end{pmatrix}^\top \collision{f}
     \right \rangle =  \begin{pmatrix}  0 ,
      0,
     0 \end{pmatrix}^\top, \qquad \text{ for every} f\in \mathscr{D}(\bC).
\end{equation}

\noindent Here, $\rho = \bra f \ket$ is the density, $\rho \Velocity =  \bra \velocity f \ket$ is the momentum, and $ \rho E = \bra \velocity\cdot \velocity f \ket/2$ is the energy density, with $E$ as the specific total energy.

Consequently, as a result of  \cref{eq:local_conservation}, solutions $f$ to the Boltzmann transport equation \cref{eq:Boltzmann_Transport_eq} satisfy (local) conservation laws. 
See \cref{appendix:fluid_dynamics_description} for the local conservation laws related to the solution of \cref{eq:Boltzmann_Transport_eq}. \ 
 
 Furthermore, $\langle \varphi(\velocity) \collision{f} \rangle =0,$
 for all $f \in \mathscr{D}(\bC)$ \emph{if, and only if,} $\varphi(\velocity)$ belongs to $\mathrm{span}\{1, \velocity ,\velocity \cdot \velocity \}$. This implies that there are no additional conservation laws beyond those given in \cref{eq:local_conservation}. \
 
 \item  The collision, $\bC$, satisfies the \emph{local dissipation} 
\begin{equation}\label{eq:local_dissipation}
\bra \log f \collision{f} \ket \le 0, \qquad f\in \mathscr{D}(\bC),     
\end{equation}

which implies Boltzmann's H-theorem, 
\begin{equation}\label{eq:H-theorem}
\partial_t \bra f(\log f -1) \ket + \nablax \cdot \bra \velocity f (\log f -1) \ket = \bra \log f \collision{f} \ket \le 0 ,  
\end{equation}
where $\bra f(\log f -1) \ket$ is the local entropy function, and  $ \bra \velocity f (\log f -1) \ket $ the local entropy flux. \ 

The total entropy is defined as 
$$\mathsf{s}=\int_\Omega \bra f(\log f -1) \ket d\bx,$$
and we then obtain the entropy inequality~\cite{le1997numerical},
$$\partial_t \mathsf{s}+ \int_{\partial \Omega} \bra \velocity f (\log f -1) \ket \cdot \nu d\sigma(\bx)\le 0.$$

\Cref{eq:local_dissipation} vanishes only at the local equilibrium (see \cref{appendix:maxwellian}).

    \item  The collision operator commutes with \emph{velocity translations and orthogonal transformations}. Let $\velocity' \in \real^d$, and $\boldsymbol{Q} \in \real^{D\times D}$ be an orthogonal matrix. Define $L_{\velocity'} f(\velocity) = f(\velocity -\velocity')$ and $L_{\boldsymbol{Q}}f(\velocity)= f(\boldsymbol{Q}^\top \velocity)$. 
    Then, 
\begin{equation}\label{eq:symmetries_collision}
    L_{\velocity'}\collision{f}= \collision{L_{\velocity'} f}, \qquad L_{\boldsymbol{Q}}\collision{f}= \collision{L_{\boldsymbol{Q}} f}.
\end{equation}

If $\Omega = \real^d$, then together with the transport operator, \cref{eq:symmetries_collision} implies Galilean and orthogonal invariance of the Boltzmann equation. \ 

This means that solutions $f$ of \cref{eq:Boltzmann_Transport_eq} transform covariantly under Galilean shifts and orthogonal changes of frame. Precisely, for any $f$ satisfying \cref{eq:Boltzmann_Transport_eq}, $\velocity' \in \real^d$ and $\boldsymbol{Q} \in \real^d \times \real^d$, we have 
\begin{align}
    \mathscr{A}_{\velocity'} f &= f(t, \bx-\velocity' t, \velocity-\velocity') \\
    \mathscr{A}_{{Q}} f &= f(t, {Q}^\top \bx, {Q}^\top\velocity).
\end{align}

\end{enumerate}

\subsection{Maxwellian Distribution---Equilibrium}\label{appendix:maxwellian}

The equilibrium distribution $\Feq$ (also known as Maxwellian) is defined as the kernel (null space) of the collision operator $\mathcal{C}$, i.e., the state for which collisions are in detailed balance,
\[
\mathcal{C}(\Feq)=0 .
\]
Boltzmann's H-theorem links this kinetic notion of equilibrium to thermodynamics.  Let the kinetic entropy (H-function) be
\begin{equation}\label{eq:kinetic_entropy}
    H(f)\;=\;\big\langle f\log f\big\rangle,
\end{equation}
where $\langle\cdot\rangle$ denotes integration over velocity space.  The H-theorem states that $H(f)$ is a Lyapunov functional for the collisional dynamics:
\[
\frac{d}{dt}H(f) \;=\; \big\langle \log f \,\mathcal{C}(f)\big\rangle \le 0,
\]
with equality if and only if $f$ is an equilibrium distribution (under the usual regularity and decay hypotheses) \cite{cercignani1988boltzmann,villani2002review}.  Hence the collision operator's null space coincides with the set of entropy-critical states.

Equivalently, the local Maxwellian $\Feq$ may be characterized \emph{variationally} as the unique minimizer of $H(f)$ subject to conservation of the collision invariants (mass, momentum, energy):
\begin{equation}\label{eq:entropy_constraint_app}
    \big\langle 
    \begin{pmatrix}
        1, \;    \velocity, \;
        \tfrac12|\velocity|^2
    \end{pmatrix} f \big\rangle ^\top
    =
    \begin{pmatrix}
        \rho,  \; \rho\Velocity,     \;   \rho E
    \end{pmatrix}^\top.
\end{equation}
Thus
\begin{equation}\label{eq:Maxwellian_variational_app}
    \Feq \;=\; \arg\min_{f\ge 0}\; \big\{ H(f)\;:\; \text{\cref{eq:entropy_constraint_app} holds} \big\},
\end{equation}
and solving this constrained minimization yields the classical Maxwell--Boltzmann form under standard hypotheses \cite{perthame1989global,levermore1996moment}:
\begin{equation}\label{eq:Maxwellian}
    f^{\mathrm{MB}}(t, \bx, \velocity) 
    = \frac{\rho(t, \bx)}{\bigl( 2 \pi R \temperature(t, \bx) \bigr)^{d/2}}
    \exp \left( -\frac{\bigl(\velocity - \Velocity(t, \bx)\bigr) \cdot \bigl(\velocity - \Velocity(t, \bx)\bigr)}{2 R \temperature(t, \bx)} \right),
\end{equation}
where $R$ is the gas constant and $d$ is the spatial dimension \cite{perthame1989global,levermore1996moment}.

The variational and dynamical characterizations are therefore consistent: the entropy minimizer subject to the collision-invariant constraints is exactly the null state of the collision operator (see \cite{cercignani1988boltzmann, villani2002review, levermore1996moment} for rigorous statements and hypotheses).

\paragraph{Discrete Velocities and Discrete Equilibrium.}
In lattice Boltzmann and other discrete-velocity methods (see \cref{appendix:discrete_kinetic}), the velocity space is replaced by a finite set $\{\velocity_i\}_{i=1}^Q$ with an associated discrete measure. Sampling the continuous Maxwellian at these nodes, $f^{\mathrm{MB}}(\velocity_i)$, does \emph{not} generally produce the true minimizer of the discrete entropy under the discrete moment constraints.  
The discrete equilibrium must instead be obtained by solving the corresponding constrained minimization problem on the discrete velocity set, or by solving a nonlinear system enforcing moment consistency \cite{perthame1989global,mieussens2000discrete}.  
This procedure is accurate but computationally expensive when performed at every space--time point.  
In this work, we replaced this inversion with a learned surrogate $\phiNN:\boldsymbol{U}\mapsto\Feqlattice$, which approximates the discrete entropy minimizer with high fidelity and negligible runtime cost.

\subsection{BGK Collision}\label{appx:BGKmodel}
A widely used surrogate for the full Boltzmann collision operator is the Bhatnagar--Gross--Krook (BGK) relaxation model \cite{bhatnagar1954model}. It replaces the nonlinear collision integral with a simple relaxation toward the equilibrium distribution \(\Feq \) over a characteristic time scale \(\tau\):
\begin{equation}
    \mathcal{C}(f) = \frac{1}{\tau} \bigl(\Feq - f\bigr).
\end{equation}
The BGK operator preserves the fundamental conservation laws of mass, momentum, and energy, since the Maxwellian \cref{eq:Maxwellian} satisfies
\begin{equation}
\bra \Feq  \ket = \bra f \ket, \qquad 
\bra \velocity \Feq  \ket = \bra \velocity f \ket, \qquad
\bra  \tfrac{1}{2} \velocity \cdot \velocity \Feq  \ket = \bra \tfrac{1}{2} \velocity \cdot \velocity  f \ket,
\end{equation}
where \(\bra \cdot \ket\) denotes integration over velocity space.  

Moreover, the BGK model enforces local entropy dissipation. Using the entropy functional \(H[f] = \bra f \log f \ket\), one obtains
\begin{align*} \bra \log f \, \tau^{-1} (\Feq - f) \ket &= \tau^{-1} \bra \log \tfrac{f}{\Feq} \, (\Feq - f) \ket +\tau^{-1} \bra \log \Feq \, (\Feq - f) \ket \\ &= \tau^{-1} \bra \log \tfrac{f}{\Feq} \, (\Feq - f) \ket = \tau^{-1} \bra \log \tfrac{f}{\Feq} \, \bigl(1 - \tfrac{f}{\Feq}\bigr) \Feq \ket \le 0, \end{align*}
since \(\log(x)(1 - x) \le 0\) for \(x>0\). This guarantees compliance with the \(H\)-theorem, and it ensures that \(f\) relaxes monotonically toward equilibrium. 

The BGK model thus provides a minimal yet physically consistent closure that retains the essential conservation and dissipation structure of the full Boltzmann collision operator, while being far simpler to evaluate.

\subsection{Discrete Velocities and Discrete Kinetic Equations}\label{appendix:discrete_kinetic}

To obtain a tractable kinetic scheme, the continuous velocity space is replaced by a finite set of discrete velocities $\Vcal \subset \mathbb{R}^d$. This amounts to approximating integrals in the Boltzmann or Boltzmann--BGK equation by a finite quadrature rule. Concretely, we introduce a discrete measure
\begin{equation}\label{eq:discrete_velocity_and_measure}
    \Vcal = \{\velocity_i\}_{i=1}^Q \subset \mathbb{R}^d,
    \qquad
    d\mu(\velocity) = \sum_{i=1}^Q W_i \,\delta(\velocity-\velocity_i),
\end{equation}
so that velocity integrals are approximated by
\[
\int_{\mathbb{R}^d} \psi(\velocity)\,d\velocity
\;\approx\;
\sum_{i=1}^Q W_i\,\psi(\velocity_i),
\]
where $\{W_i\}_{i=1}^Q$ are quadrature weights.  

The choice of $\Vcal$ and $W_i$ determines the accuracy with which velocity moments are reproduced and is central to the stability of the discrete kinetic scheme. Classical constructions rely on Gauss--Hermite quadrature or related cubature rules, which ensure exactness up to a prescribed polynomial degree \cite{shan1998discretization}. This discretization converts the continuous kinetic equation into a system of $Q$ coupled transport--relaxation equation---as we have seen in \cref{eq:Boltzmann_BGK_discrete}---one for each discrete velocity.


\subsection{Moment System of the Boltzmann Equation}
\label{appendix:fluid_dynamics_description}

Starting from the Boltzmann transport equation \cref{eq:Boltzmann_Transport_eq}, we obtain macroscopic conservation laws by multiplying by collision invariants $\varphi(\velocity)\in\mathrm{span}\{1,\velocity,\velocity\cdot\velocity\}$ and integrating over velocity space. Using the local conservation property of the collision operator (\cref{eq:local_conservation}), we arrive at
\begin{equation}\label{eq:general_consr}
\partial_t \big\langle \varphi(\velocity) f \big\rangle 
+ \nabla_{\bx} \cdot \big\langle \velocity \osym \varphi(\velocity) f \big\rangle = 0,
\end{equation}
where $\osym$ denotes the symmetric tensor product.  

Choosing $\varphi(\velocity)=1$, $\velocity$, and $\tfrac{1}{2}\velocity\cdot\velocity$ yields the balance laws for mass, momentum, and energy:
\begin{subequations}
\begin{align}
    \partial_t \big\langle f \big\rangle + \nabla_{\bx}\cdot \big\langle \velocity f \big\rangle &= 0,\\
    \partial_t \big\langle \velocity f \big\rangle + \nabla_{\bx}\cdot \big\langle \velocity\osym\velocity f \big\rangle &= 0,\\
    \partial_t \big\langle \tfrac{1}{2}\velocity\cdot\velocity\, f \big\rangle + \nabla_{\bx}\cdot \big\langle \tfrac{1}{2}(\velocity\cdot\velocity)\velocity f \big\rangle &= 0.
\end{align}
\end{subequations}
Identifying the macroscopic observables
\[
\big\langle f \big\rangle = \rho,\qquad
\big\langle \velocity f \big\rangle = \rho \Velocity,\qquad
\big\langle \tfrac{1}{2}\velocity\cdot\velocity\, f \big\rangle 
= E = \tfrac{1}{2}\rho|\Velocity|^2+\tfrac{d}{2}\rho\temperature,
\]
and introducing the flux decompositions
\begin{subequations}
\begin{align}
    \big\langle \velocity\osym\velocity f \big\rangle &= \rho\,\Velocity\osym\Velocity + \pressuretensor,\\
    \big\langle \tfrac{1}{2}(\velocity\cdot\velocity)\velocity f \big\rangle &= E\,\Velocity + \pressuretensor\,\Velocity + \mathbf{Q},
\end{align}
\end{subequations}
the moment equations can be written compactly as
\begin{equation}\label{eq:hydrodynamic_limit}
\partial_t 
\begin{pmatrix}
    \rho\\
    \rho\Velocity\\
    E
\end{pmatrix}
+\nabla_{\bx}\cdot
\begin{pmatrix}
    \rho\Velocity\\
    \rho\,\Velocity\osym\Velocity+\pressuretensor\\
    E\Velocity+\pressuretensor\Velocity+\mathbf{Q}
\end{pmatrix}
=
\begin{pmatrix}
    0\\
    0\\
    0
\end{pmatrix},
\end{equation}
where
\(
\pressuretensor=\big\langle (\velocity-\Velocity)\osym(\velocity-\Velocity)f\big\rangle
\)
and
\(
\mathbf{Q}=\tfrac{1}{2}\big\langle (\velocity-\Velocity)|\velocity-\Velocity|^2 f\big\rangle
\)
are the stress tensor and heat flux, respectively.  
We may further decompose
\begin{subequations}
\begin{align}
    \pressuretensor &= \rho\temperature\,\bI+\boldsymbol{\Sigma},\\
    \mathbf{Q} &= \rho\temperature\Velocity+\boldsymbol{\Sigma}\Velocity+\heatflux,
\end{align}
\end{subequations}
where $\boldsymbol{\Sigma}$ is the deviatoric stress tensor and $\heatflux$ the heat flux vector.

\begin{remark}[Hydrodynamic Closures]\label{remark:equation_of_state}
To close \cref{eq:hydrodynamic_limit}, constitutive relations must be specified for $\boldsymbol{\Sigma}$ and~$\heatflux$.  
\begin{itemize}
    \item For $\boldsymbol{\Sigma}=0$ and $\heatflux=0$, with $p=\rho R\temperature$, the system reduces to the \emph{compressible Euler equations}.
    \item For Newtonian viscous stress 
    \(
    \pressuretensor = \bigl(p - \zeta\,\nabla\cdot\Velocity\bigr)\bI 
    - \mu\bigl(\nabla\Velocity+\nabla\Velocity^\top - \tfrac{2}{d}(\nabla\cdot\Velocity)\bI\bigr)
    \)
    and Fourier heat flux $\heatflux = -\kappa\,\nabla\temperature$, one recovers the \emph{Navier--Stokes--Fourier equations}, where $\mu$, $\zeta$, and $\kappa$ denote dynamic viscosity, bulk viscosity, and thermal conductivity, respectively.
\end{itemize}
\end{remark}

\begin{remark}[Chapman--Enskog consistency for the thermal LB scaffold]\label{rem:chapman_enskog_consistency}
For the two-population thermal LB scheme used in the compressible experiments,
the learned closure replaces only the local energy equilibrium supplied to the
standard relaxation update. Under the usual diffusive/hydrodynamic scaling and
assuming the learned equilibrium matches the required conserved moments and
second-/third-order thermal moments to the accuracy stated in the training
loss, the standard Chapman--Enskog calculation for this scaffold recovers the
compressible Navier--Stokes--Fourier system at leading order. The transport
coefficients are determined by the relaxation times in the usual way:
viscous stresses are controlled by the \(\F\)-population relaxation
\(\tau_1\), while thermal conductivity is controlled by the \(\G\)-population
relaxation \(\tau_2\) and the Prandtl correction through \(\G^\ast\). Thus
NeurDE changes the closure used to evaluate the equilibrium moments; it does
not introduce a learned transport coefficient outside the host LB scaling.
\end{remark}

\subsection{Structure of the Moment Space}\label{appendix:M_space}

Let $\mathbb{M}$ be a finite-dimensional linear space of functions of $\velocity$ (often polynomial). Taking the moments of \cref{eq:Boltzmann_Transport_eq} with respect to $\boldsymbol{m}(\velocity)\in\mathbb{M}$ yields
\begin{equation}\label{eq:general_system_M}
\partial_t \big\langle \boldsymbol{m}(\velocity) f \big\rangle
+ \nabla_{\bx}\cdot \big\langle \velocity\osym\boldsymbol{m}(\velocity) f \big\rangle = 0.
\end{equation}
\Cref{eq:general_consr} is recovered by taking 
\(
\mathbb{M}=\mathrm{span}\{1,\velocity,\velocity\cdot\velocity\}.
\)

The \emph{moment closure problem} consists of expressing the flux terms in \cref{eq:general_system_M} in terms of finitely many moments determined by $\mathbb{M}$.  
When $\velocity$ is discretized, $\{\velocity_i\}_{i=1}^Q$, common choices for $\mathbb{M}$ include:

\begin{enumerate}
    \item Degree $\le 2$:
    \emph{Eulerian basis} $\mathbb{M}=\mathrm{span}\{1,\velocity_i,\velocity_i\cdot\velocity_i\}$,
    or \emph{Gaussian basis} $\mathbb{M}=\mathrm{span}\{1,\velocity_i,\velocity_i\osym\velocity_i\}$.
    \item Degree $\le 4$:
    \emph{Grad basis} \cite{grad1949kinetic}: $\mathbb{M}=\mathrm{span}\{1,\velocity_i,\velocity_i\osym\velocity_i,(\velocity_i\cdot\velocity_i)\velocity_i\}$,
    and \emph{Levermore basis} \cite{levermore1996moment}: $\mathbb{M}=\mathrm{span}\{1,\velocity_i,\velocity_i\osym\velocity_i,(\velocity_i\cdot\velocity_i)\velocity_i,(\velocity_i\cdot\velocity_i)^2\}$.
    \item Higher-order closures:
    tensorial bases involving cubic and quartic combinations of $\velocity_i$, e.g.,
    $$
    \mathrm{span}\{1,\velocity_i,\velocity_i^{\otimes 2},\velocity_i^{\otimes 3},\velocity_i^{\otimes 4}\}.
    $$
\end{enumerate}

These moment spaces form the foundation for classical closure hierarchies such as Grad's and Levermore's, and they underpin modern discrete kinetic methods and entropic closures.

\subsection{Higher-order moments of the Maxwell-Boltzmann distribution}
\label{appendix:higher-order_moments}
Some of the most commonly used higher-order moments of the Maxwellian \cref{eq:Maxwellian} include 
the pressure tensor \cref{eq:pressure_tensor}, the heat flux vector \cref{eq:heat_flux_vector}, and the contracted fourth-order tensor \cref{eq:forth_order_tensor}. These moments are defined as follows:
\begin{subequations}\label{eq:higher-order}
    \begin{align}
        \pressuretensor_{\alpha,\beta}^{\mathrm{MB}} &= \rho \Velocity_\alpha \Velocity_\beta + \rho \temperature \delta_{\alpha, \beta},\label{eq:pressure_tensor} \\ 
        {\heatflux}_\alpha^{\mathrm{MB}} &= 2\rho \Velocity_{\alpha}(E+ \temperature), \label{eq:heat_flux_vector}\\
        \boldsymbol{R}_{\alpha,\beta}^{\mathrm{MB}} &= 2 \rho E (\temperature \delta_{\alpha, \beta}+ \Velocity_\alpha \Velocity_\beta) + 2 \rho \temperature(\temperature \delta_{\alpha, \beta}+ 2 \Velocity_\alpha \Velocity_\beta). \label{eq:forth_order_tensor}
    \end{align}
\end{subequations}

\subsection{A splitting Method for \cref{eq:Boltzmann_BGK_discrete}} \label{appendix:splitting_model}
Here, we review the splitting approach for approximately solving kinetic equations. 
For a comprehensive introduction to splitting methods, see \cite{holden2010splitting, strang1968construction}. Rewriting \cref{eq:Boltzmann_BGK_discrete} as $$ \partial_t \F_i(t, \bx) = \underbrace{-\velocity_i \cdot \nabla \F_i(t, \bx)}_{\bS} + \underbrace{\tfrac{1}{\tau} \left( \Feqlattice_i (\boldsymbol{U}(t, \bx)) - \F_i(t, \bx)   \right)}_{\bC},$$
exposes a nonlinear collision and a linear transport (free-flow) parts of evolution. The splitting scheme can be represented by
means of the following steps
\begin{subequations}
\label{eq:splitting_discrete}
    \begin{alignat}{2}
        \partial_t \Tilde{\F}_i(t, \bx) &= \tfrac{1}{\tau} ( \Feqlattice_i (\boldsymbol{U}(t, \bx))- \tilde{\F}_i(t, \bx)), &\qquad \tilde{\F}_i(0, \bx) &= \F_i(0, \bx) \label{eq:collision}\\
        \partial_t \bar{\F}_i(t, \bx) &= - \velocity_i \cdot \nabla \bar{\F}_i(t,\bx), &\qquad \bar{\F}_i(0, \bx) &= \tilde{\F}_i(\h, \bx) ,
        \label{eq:streaming}
    \end{alignat}
\end{subequations}
where $\F(0, \bx)$ is the initial condition. 
Denoting the solutions operators to the collision subproblem (\cref{eq:collision}) by $\phiC$ and the streaming subproblem (\cref{eq:streaming}) by $\phiS$, we approximate the solution of Boltzmann-BGK equation (\cref{eq:Boltzmann_BGK_discrete}) from time $t$ to $t+\h$ as: $\F_i(t+\h, \bx) \approx\phiS \phiC\F_i(t, \bx)$.\footnote{The first numerical application to kinetic equations is attributed to \cite{aristov1976splitting}, building on Grad's idea \cite{grad1958principles}(p. 246-247) (see~\cite{walus1995computational}).}

\subsection{Lattice Boltzmann Scheme}\label{appendix:LBM}

Lattice Boltzmann (LB) algorithms provide a practical and efficient framework for approximating the Boltzmann--BGK equation \cite{bellotti2023truncation, bellotti2022finite}, and consequently, the original conservation law system \cref{eq:conservation} (see also \cite{dellar2013interpretation, schiller2014unified, hajabdollahi2018symmetrized}).  
A key feature of LB methods is that the streaming step of the kinetic equation becomes an exact lattice shift, leading to a simple and highly efficient numerical scheme.

The spatial domain $\Omega$ is discretized into a uniform lattice with spacing $\deltax$:
\[
\latticex = \deltax \mathbb{Z}^d \cap \Omega = \{ n \deltax \in \Omega : n \in \mathbb{Z}^d \}.
\]
For $T>0$ the time interval $[0,T]$ is similarly discretized as $\latticet = \h \mathbb{N} \cap [0,T]$. For a fixed pair $(\deltax,\h)$, the lattice speed is defined as $\deltax / \h$.  
The velocity set $\{\velocity_i\}_{i=1}^Q = \Vcal \subset \mathbb{R}^d$ (cf. \cref{eq:discrete_velocity_and_measure}) is chosen as integer multiples of a reference speed $c_s$: 
\[
\velocity_i = c_s \latticevelocity_i, \qquad \{\latticevelocity_i\}_{i=1}^Q \subset \mathbb{Z}^d,
\]
where $c_s$ is such that $c_s \h/\deltax = n_{\mathrm{ref}} \in \mathbb{N}$. This choice guarantees that after one time step $\h$, particles land exactly on lattice nodes, making the streaming step a pure index shift.

\paragraph{Collision Step.}
Following \cite{dellar2013interpretation}, the collision operator is discretized using a second-order trapezoidal quadrature, giving
\begin{equation}\label{eq:collision_operator}
\F_i^{\mathrm{coll}}(t,\bx) 
= \phiC \F_i(t,\bx) 
= \left(1 - \frac{1}{\tau}\right) \F_i(t,\bx) 
+ \frac{1}{\tau}\,\Feqlattice_i\bigl(\boldsymbol{U}(t,\bx)\bigr),
\qquad (t,\bx)\in \latticet\times\latticex,
\end{equation}
where $\tau$ is the relaxation time, $\F_i$ the $i$-th population, and $\Feqlattice_i$ the corresponding discrete equilibrium.

\paragraph{Streaming Step.}
The discrete streaming operator reads \cite{he1998discrete, dellar2002lattice}
\begin{equation}\label{eq:streaming_LBM}
\F_i(t+\h,\bx) 
= \phiS \F_i(t,\bx) 
= \F_i^{\mathrm{coll}}(t,\bx - \velocity_i \h),
\qquad (t,\bx)\in \latticet\times\latticex.
\end{equation}
Because $\bx - \velocity_i \h = \bx - c_s \latticevelocity_i \h = \bx - n_{\mathrm{ref}}\latticevelocity_i \deltax$ lies on the lattice, this step reduces to a simple shift in memory, avoiding \emph{costly interpolation}.

\paragraph{Equilibrium Distribution.}
The equilibrium $\Feqlattice_i$ is commonly approximated by truncating the Hermite expansion of the Maxwell--Boltzmann distribution \cref{eq:Maxwellian} at order $N$ \cite{grad1949kinetic, shan1998discretization, kruger2017lattice}:
\begin{equation}\label{eq:poly_equilibrium_hermite}
f^{\mathrm{MB}}(t,\bx,\velocity_i)
\;\approx\;
f^{\mathrm{MB}}_N(t,\bx,\velocity_i)
=
\omega(\velocity_i)
\sum_{k=0}^N
\frac{1}{k!}\,
\boldsymbol{a}^{\mathrm{eq},k}(t,\bx) : \sobolev^k(\velocity_i),
\end{equation}
where $\omega(\velocity) = \exp(-|\velocity|^2/2)(2\pi)^{-d/2}$ is the Gaussian weight and $\boldsymbol{a}^{\mathrm{eq},k}$ are the $k$-th order Hermite moments of the Maxwellian. The symbol $:$ denotes full tensor contraction.

Although this polynomial equilibrium is simple and efficient, it is well known to be accurate only at low Mach numbers and prone to instabilities when the flow deviates significantly from the reference state $(\Velocity,\temperature) = (0,\temperature_0)$ (see \cite{tran2022lattice}(Fig. 1)).  
More robust alternatives, including exponential or entropic closures \cite{latt2020efficient, tran2022lattice}, have been proposed to overcome these limitations, especially in high-speed flow regimes.

\paragraph{LB Scheme.}
The combined LB update is then
\[
\F_i(t+\h,\cdot) = \phiS\phiC\,\F_i(t,\cdot).
\]
With $n_{\mathrm{ref}}=1$, unit lattice speed, and the dimensionless variables $\bx\gets\bx/\deltax$, $t\gets t/\h$, the scheme reduces to
\begin{equation}\label{eq:Appendix_LBM_algorithm_dimensionless}
\F_i(t+1,\bx+\latticevelocity_i) - \F_i(t,\bx) 
= \frac{1}{\tau}\Bigl[\Feqlattice_i(\boldsymbol{U}(t,\bx)) - \F_i(t,\bx)\Bigr],
\qquad (t,\bx)\in\mathbb{N}_{\ge 0}\times\mathbb{Z}^d.
\end{equation}
We adopted this dimensionless formulation throughout the main text, see \cref{eq:LBM_algorithm_dimensionless}.

\subsection{Closure Relations in Lattice Velocities}\label{remark:lattices_closure}

Lattice velocity sets $\{\latticevelocity_i\}$ always satisfy an algebraic \emph{closure relation}.  
For simplicity, consider a one-dimensional velocity set $\Vcal = \{\latticevelocity_i\}_{i=1}^Q$.  
Because only $Q$ monomials $\{1,\latticevelocity_i,\dots,\latticevelocity_i^{Q-1}\}$ are linearly independent, the $Q$-th power can be expressed as a linear combination of lower-order powers:
\begin{equation}
\latticevelocity_i^Q = \mathrm{poly}\bigl(1,\latticevelocity_i,\ldots,\latticevelocity_i^{Q-1}\bigr).
\end{equation}
For instance, in the standard D1Q3 lattice (velocities $\latticevelocity_i\in\{0,\pm 1\}$), the closure relation is
\[
\latticevelocity_i^3 = \latticevelocity_i.
\]
Similar relations hold in higher dimensions ($d>1$), where velocity sets are Cartesian products of one-dimensional sets, $\{0,\pm 1\}^{\otimes d}$ \cite{karlin2010factorization}. These closure properties play a crucial role in determining the order of exact moment recovery and, consequently, the accuracy of LB schemes.

\subsection{Entropy Structure of the NeurDE Equilibrium}
\label{sec:entropy_structure_neurde}

We now develop the mathematical theory underlying the NeurDE equilibrium ansatz \cref{eq:levermore_closure_NN}. The central message is that the exponential branch--trunk form is not an ad hoc architectural choice but a necessary consequence of entropy minimization on the discrete velocity set. As a result, the exact entropy projector associated with the learned basis inherits the variational structure of Levermore's maximum-entropy closure program \cite{levermore1996moment,hauck2008convex}: a unique entropy minimizer, a strictly convex macroscopic entropy, symmetrizable hyperbolicity, and a formal $H$-theorem. We then show how the raw learned NeurDE model departs from that ideal theory through an explicit entropy-defect term.

The material is organized as follows. In \cref{subsec:derivation}, we derive the exponential ansatz from constrained entropy minimization on the lattice. In \cref{subsec:assumptions}, we state the admissibility conditions on the learned basis. In \cref{subsec:structural}, we prove the exact fixed-basis structural theorems. In \cref{subsec:learning_aware}, we return to the practical learned branch output and quantify how imperfect moment matching modifies the ideal entropy law.

\subsubsection{From entropy minimization to the exponential ansatz}
\label{subsec:derivation}

Let $\Vcal = \{\latticevelocity_i\}_{i=1}^Q$ be the discrete velocity set with quadrature weights $\{W_i\}_{i=1}^Q$ as in \cref{eq:discrete_velocity_and_measure}. Let
\[
\bm{\varphi}(\latticevelocity_i)
=
\bigl(\bvarphi_1(\latticevelocity_i;\theta^{\bvarphi}),\dots,\bvarphi_p(\latticevelocity_i;\theta^{\bvarphi})\bigr)^\top
\in \real^p
\]
denote the learned trunk basis, evaluated at the $i$-th lattice velocity. We require $\bvarphi_1(\latticevelocity_i;\theta^{\bvarphi})\equiv 1$ for all $i$, so that the constant function belongs to the span of the basis; this ensures that the total mass is among the constrained moments.

Recall from \cref{appendix:maxwellian} that the classical Maxwellian is characterized variationally as the minimizer of the kinetic entropy $H(f) = \bra f\log f\ket$ subject to conservation of mass, momentum, and energy \cref{eq:kinetic_entropy}. In the discrete-velocity setting, velocity integrals are replaced by weighted sums over the lattice, and the collision invariants $\{1,\latticevelocity_i,\tfrac{1}{2}\latticevelocity_i\cdot\latticevelocity_i\}$ are the natural moment basis. The NeurDE construction generalizes this by replacing the fixed collision invariants with the learned basis $\bm{\varphi}$, while retaining the entropy-minimization principle.

Concretely, given macroscopic observables $\boldsymbol{U}\in\real^p$, consider the constrained minimization problem
\begin{equation}\label{eq:discrete_entropy_min}
\min_{\F \ge 0}
\left\{
H(\F)
\eqdef
\sum_{i=1}^Q W_i\bigl(\F_i \log \F_i - \F_i\bigr)
\;:\;
\sum_{i=1}^Q W_i\,\bm{\varphi}(\latticevelocity_i)\,\F_i = \boldsymbol{U}
\right\}.
\end{equation}
This is a strictly convex program in $\F = (\F_1,\dots,\F_Q)^\top$ over the positive orthant. To solve it, we form the Lagrangian with multiplier vector $\bm{\alpha}=(\alpha_1,\dots,\alpha_p)^\top \in\real^p$:
\begin{equation}\label{eq:lagrangian_discrete}
\mathscr{L}(\F,\bm{\alpha})
=
\sum_{i=1}^Q W_i\bigl(\F_i\log\F_i - \F_i\bigr)
-
\bm{\alpha}\cdot
\left(
\sum_{i=1}^Q W_i\,\bm{\varphi}(\latticevelocity_i)\,\F_i - \boldsymbol{U}
\right).
\end{equation}
Taking the first variation with respect to $\F_i$ and setting it to zero gives
\[
\frac{\partial \mathscr{L}}{\partial \F_i}
=
W_i\bigl(\log\F_i - \bm{\alpha}\cdot\bm{\varphi}(\latticevelocity_i)\bigr)
= 0,
\qquad i=1,\dots,Q,
\]
from which we conclude that every stationary point must satisfy
\begin{equation}\label{eq:stationary_exp}
\F_i
=
\exp\!\bigl(\bm{\alpha}\cdot\bm{\varphi}(\latticevelocity_i)\bigr)
=
\exp\!\left(\sum_{k=1}^p \alpha_k\,\bvarphi_k(\latticevelocity_i;\theta^{\bvarphi})\right).
\end{equation}

The multiplier $\bm{\alpha}$ must be chosen so that the moment constraint is satisfied:
\begin{equation}\label{eq:moment_constraint_alpha}
\boldsymbol{U}
=
\sum_{i=1}^Q W_i\,\bm{\varphi}(\latticevelocity_i)\,\exp\!\bigl(\bm{\alpha}\cdot\bm{\varphi}(\latticevelocity_i)\bigr)
=
\nabla_{\bm{\alpha}}\Psi(\bm{\alpha}),
\end{equation}
where we have introduced the \emph{partition function}
\begin{equation}\label{eq:partition_fn}
\Psi(\bm{\alpha})
\eqdef
\sum_{i=1}^Q W_i\,\exp\!\bigl(\bm{\alpha}\cdot\bm{\varphi}(\latticevelocity_i)\bigr).
\end{equation}
This is a finite sum of exponentials, hence real-analytic on all of $\real^p$.

\begin{remark}[Absorption of quadrature weights in LBM]
\label{rem:weight_absorption}
If one passes to the usual lattice-Boltzmann populations
$\widetilde{\F}_i \eqdef W_i \F_i$, then macroscopic moments are recovered by bare summation,
\[
\sum_{i=1}^Q \bm{\varphi}(\latticevelocity_i)\,\widetilde{\F}_i = \boldsymbol{U}.
\]
However, the entropy does \emph{not} become weight-free. In the absorbed variables,
\[
H(\widetilde{\F})
=
\sum_{i=1}^Q
\left(
\widetilde{\F}_i \log\frac{\widetilde{\F}_i}{W_i}
-
\widetilde{\F}_i
\right),
\]
and the associated entropy minimizer satisfies
\[
\widetilde{\F}_i
=
W_i\,\exp\!\bigl(\bm{\alpha}^\star(\boldsymbol{U})\cdot\bm{\varphi}(\latticevelocity_i)\bigr).
\]
Equivalently, the factor $W_i$ may be viewed as a known velocity-dependent offset
$\log W_i$ in the log-equilibrium. Thus the quadrature weights are absorbed into the
reference measure, not removed from the variational structure.
\end{remark}

At the level of the exact entropy projector, the branch network would realize the multiplier map $\bm{\alpha}^\star(\boldsymbol{U})$. In that ideal case, \cref{eq:stationary_exp} recovers, up to the fixed weight absorption described in \Cref{rem:weight_absorption}, the branch--trunk exponential structure of \cref{eq:levermore_closure_NN}. In practice, the trained branch network produces an approximation to $\bm{\alpha}^\star$; the resulting structural defect is quantified in \cref{subsec:learning_aware}.

\begin{remark}[The exponential as structural constraint]
The derivation makes clear that the exponential output map in \cref{eq:levermore_closure_NN} is dictated by the variational problem, not by a desire for positivity alone. Any strictly positive output nonlinearity (e.g., softplus, squared activations) would guarantee $\F_i > 0$, but only the exponential produces a stationary point of the entropy Lagrangian. Replacing $\exp$ with a different activation would break the connection to entropy minimization and forfeit the structural guarantees that follow.
\end{remark}

\begin{remark}[Relation to Levermore's exponential closure]
\label{rem:levermore_relation}
The classical Levermore closure \cite{levermore1996moment} prescribes a fixed analytic basis---typically collision invariants or higher-order polynomial moments---and defines the equilibrium as the entropy minimizer under those moment constraints. The NeurDE construction follows the same variational principle but replaces the prescribed basis with a learned trunk $\bm{\varphi}(\cdot\,;\theta^{\bvarphi})$. When $\bm{\varphi}(\latticevelocity_i)$ reduces to $(1,\latticevelocity_i,\tfrac{1}{2}\latticevelocity_i\cdot\latticevelocity_i)^\top$, the absorbed-population form identified in \Cref{rem:weight_absorption} recovers the standard discrete entropy-minimizing equilibrium used in entropic lattice Boltzmann methods \cite{latt2020efficient,tran2022lattice}. In this sense, the learned basis enlarges the class of admissible closures while the exponential form preserves the variational backbone.
\end{remark}

\subsubsection{Admissibility of the learned basis}
\label{subsec:assumptions}

The structural theory requires the learned basis to satisfy a nondegeneracy condition that ensures the moment map $\bm{\alpha}\mapsto\nabla_{\bm{\alpha}}\Psi(\bm{\alpha})$ is injective.

\begin{assumption}[Discrete admissible basis]\label{ass:discrete_admissible}
Fix the trained parameters $\theta = (\theta^{\lambda},\theta^{\bvarphi})$. Assume:
\begin{enumerate}
    \item \textbf{(Constant component.)} $\bvarphi_1(\latticevelocity_i;\theta^{\bvarphi})\equiv 1$ for all $i=1,\dots,Q$.
    \item \textbf{(Nondegeneracy.)} The weighted basis vectors $\{W_i^{1/2}\,\bm{\varphi}(\latticevelocity_i)\}_{i=1}^Q$ span $\real^p$.
\end{enumerate}
\end{assumption}

The constant-component condition ensures that mass conservation (or, more generally, the zeroth moment) is among the constrained quantities. The nondegeneracy condition is equivalent to requiring that the $Q\times p$ matrix
\[
\boldsymbol{\Phi}
\eqdef
\begin{pmatrix}
W_1^{1/2}\,\bm{\varphi}(\latticevelocity_1)^\top \\
\vdots \\
W_Q^{1/2}\,\bm{\varphi}(\latticevelocity_Q)^\top
\end{pmatrix}
\in \real^{Q\times p}
\]
has rank $p$.

\begin{remark}[Training and admissibility]\label{rem:basis_admissibility_training}
In the experiments, the constant component and the Eulerian moment functions
are included in the learned moment space by construction; see
\Cref{sec:moment_space}. The rank condition is a finite-dimensional condition
on the trained trunk values at the chosen velocity set. It is therefore checked
after training on the fixed lattice rather than proved for arbitrary network
weights. If this matrix loses rank, the maximum-entropy projector and defect
theorems below should be read as applying only after restricting to a
nondegenerate subspace or retraining the basis.
\end{remark}

\begin{lemma}[Properties of the partition function]\label{lem:partition}
Under \Cref{ass:discrete_admissible}, the partition function $\Psi:\real^p\to\real$ defined in \cref{eq:partition_fn} satisfies:
\begin{enumerate}
    \item $\Psi$ is real-analytic on $\real^p$;
    \item $\Psi(\bm{\alpha})>0$ for all $\bm{\alpha}\in\real^p$;
    \item the gradient is
    \[
    \nabla_{\bm{\alpha}}\Psi(\bm{\alpha})
    =
    \sum_{i=1}^Q W_i\,\bm{\varphi}(\latticevelocity_i)\,\exp\!\bigl(\bm{\alpha}\cdot\bm{\varphi}(\latticevelocity_i)\bigr);
    \]
    \item the Hessian is
    \[
    \nabla_{\bm{\alpha}}^2\Psi(\bm{\alpha})
    =
    \sum_{i=1}^Q W_i\,\bm{\varphi}(\latticevelocity_i)\,\bm{\varphi}(\latticevelocity_i)^\top\,\exp\!\bigl(\bm{\alpha}\cdot\bm{\varphi}(\latticevelocity_i)\bigr)
    \succ 0
    \qquad \forall\,\bm{\alpha}\in\real^p.
    \]
\end{enumerate}
\end{lemma}

\begin{proof}
Properties (i)--(iii) are immediate from the fact that $\Psi$ is a finite sum of exponentials. For (iv), let $\xi\in\real^p\setminus\{0\}$. Then
\[
\xi^\top \nabla_{\bm{\alpha}}^2\Psi(\bm{\alpha})\,\xi
=
\sum_{i=1}^Q W_i\,\bigl(\xi\cdot\bm{\varphi}(\latticevelocity_i)\bigr)^2\,\exp\!\bigl(\bm{\alpha}\cdot\bm{\varphi}(\latticevelocity_i)\bigr).
\]
Each term is nonneg\-ative with strictly positive exponential weight. If the sum were zero, then $\xi\cdot\bm{\varphi}(\latticevelocity_i) = 0$ for all $i$ with $W_i>0$, contradicting the spanning condition in \Cref{ass:discrete_admissible}(ii). Hence $\nabla_{\bm{\alpha}}^2\Psi\succ 0$.
\end{proof}

Strict positive definiteness of the Hessian implies that $\Psi$ is strictly convex and that the moment map $\bm{\alpha}\mapsto\nabla_{\bm{\alpha}}\Psi(\bm{\alpha})$ is a smooth local diffeomorphism everywhere.

\begin{definition}[Interior realizable set]\label{def:discrete_realizable}
The \emph{interior realizable set} is the image of the moment map:
\[
\mathcal{R}^\circ
\eqdef
\nabla_{\bm{\alpha}}\Psi(\real^p)
=
\left\{
\sum_{i=1}^Q W_i\,\bm{\varphi}(\latticevelocity_i)\,\exp\!\bigl(\bm{\alpha}\cdot\bm{\varphi}(\latticevelocity_i)\bigr)
\;:\;
\bm{\alpha}\in\real^p
\right\}
\subset\real^p.
\]
A moment vector $\boldsymbol{U}\in\mathcal{R}^\circ$ is called \emph{realizable} (with respect to the learned basis).
\end{definition}

The set $\mathcal{R}^\circ$ is open by the inverse function theorem. In the present finite discrete exponential-family setting with $\bvarphi_1\equiv 1$, $\mathcal{R}^\circ$ is also convex by the standard geometry of full exponential families; equivalently, after normalizing by the zeroth component, one obtains the interior of the convex hull of the sufficient statistics.

\begin{remark}[Learned branch and conservative correction]\label{rem:learned_realizability}
The trained branch network outputs a positive population by construction, but
finite approximation error means that its learned moments need not exactly
equal the target moments and the associated target may lie near the boundary of
the realizable set. The projection results in
\Cref{prop:two_state_weighted_projection,cor:two_state_projection_positivity}
address the practical physical constraints: they move the raw positive output
onto the affine moment set \(C\Feqlattice=\boldsymbol{U}\), with an explicit
smallness condition under which positivity is retained. The exact
maximum-entropy statements are therefore local statements on realizable compact
sets, while the conservative projection supplies the implementation-level
moment correction.
\end{remark}

\subsubsection{Structural theorems}
\label{subsec:structural}

We now state and prove the exact fixed-basis structural results. Together, they establish that the entropy projector associated with the learned basis defines a maximum-entropy closure with the same mathematical backbone as the classical Levermore theory.

\begin{theorem}[Maximum-entropy characterization of the exact projector]\label{thm:discrete_maxent}
Under \Cref{ass:discrete_admissible}, for every $\boldsymbol{U}\in\mathcal{R}^\circ$ there exists a unique multiplier $\bm{\alpha}^\star(\boldsymbol{U})\in\real^p$ such that
\[
\boldsymbol{U}
=
\sum_{i=1}^Q W_i\,\bm{\varphi}(\latticevelocity_i)\,
\exp\!\bigl(\bm{\alpha}^\star(\boldsymbol{U})\cdot\bm{\varphi}(\latticevelocity_i)\bigr),
\]
and the exact entropy projector
\[
\phiNN_i(\boldsymbol{U})
=
\exp\!\bigl(\bm{\alpha}^\star(\boldsymbol{U})\cdot\bm{\varphi}(\latticevelocity_i)\bigr)
\]
is the unique minimizer of the discrete $H$-functional
\[
H(\F)
=
\sum_{i=1}^Q W_i\bigl(\F_i\log\F_i - \F_i\bigr)
\]
over all $\F\ge 0$ satisfying the learned moment constraint
$\sum_{i=1}^Q W_i\,\bm{\varphi}(\latticevelocity_i)\,\F_i = \boldsymbol{U}$.
Since $\boldsymbol{U}\in\mathcal{R}^\circ$, the minimizing state is the exponential family element above and is therefore strictly positive componentwise. Hence the minimization may equivalently be restricted to strictly positive feasible states.
Equivalently,
\[
\phiNN(\boldsymbol{U})
=
\argmin_{\F\in\mathcal{F}(\boldsymbol{U})} H(\F),
\]
where
\[
\mathcal{F}(\boldsymbol{U})
=
\left\{
\F\in\real_{>0}^Q
\;:\;
\sum_{i=1}^Q W_i\,\bm{\varphi}(\latticevelocity_i)\,\F_i = \boldsymbol{U}
\right\}.
\]
\end{theorem}

\begin{proof}
\textit{Existence and uniqueness of the multiplier.}
By \Cref{lem:partition}, $\Psi$ is strictly convex and smooth on $\real^p$, so its gradient $\nabla_{\bm{\alpha}}\Psi$ is injective. Since $\boldsymbol{U}\in\mathcal{R}^\circ = \nabla_{\bm{\alpha}}\Psi(\real^p)$, there exists $\bm{\alpha}^\star\in\real^p$ satisfying
\[
\boldsymbol{U}
=
\nabla_{\bm{\alpha}}\Psi(\bm{\alpha}^\star)
=
\sum_{i=1}^Q W_i\,\bm{\varphi}(\latticevelocity_i)\,\exp\!\bigl(\bm{\alpha}^\star\cdot\bm{\varphi}(\latticevelocity_i)\bigr).
\]
Injectivity of the gradient map gives uniqueness.

\textit{Entropy minimality.}
Let $\F\in\mathcal{F}(\boldsymbol{U})$ be any feasible distribution, and write $\phiNN_i=\phiNN_i(\boldsymbol{U})$ for brevity. Then
\begin{align}
H(\F)-H(\phiNN)
&=
\sum_{i=1}^Q W_i
\Bigl[
\F_i\log\F_i-\F_i-\phiNN_i\log\phiNN_i+\phiNN_i
\Bigr]
\notag\\
&=
\sum_{i=1}^Q W_i
\left[
\F_i\log\frac{\F_i}{\phiNN_i}
-
(\F_i-\phiNN_i)
+
(\F_i-\phiNN_i)\log\phiNN_i
\right].
\label{eq:entropy_split_corrected}
\end{align}
Since $\log\phiNN_i=\bm{\alpha}^\star\cdot\bm{\varphi}(\latticevelocity_i)$, the last term is
\[
\sum_{i=1}^Q W_i(\F_i-\phiNN_i)\log\phiNN_i
=
\bm{\alpha}^\star\cdot
\sum_{i=1}^Q W_i\,\bm{\varphi}(\latticevelocity_i)(\F_i-\phiNN_i)
=
\bm{\alpha}^\star\cdot(\boldsymbol{U}-\boldsymbol{U})
=0.
\]
Therefore,
\begin{equation}\label{eq:bregman_identity}
H(\F)-H(\phiNN)
=
\sum_{i=1}^Q W_i
\left[
\F_i\log\frac{\F_i}{\phiNN_i}
- \F_i + \phiNN_i
\right]
=
D_{\mathrm{KL}}^W(\F\|\phiNN),
\end{equation}
where $D_{\mathrm{KL}}^W$ denotes the weighted Kullback--Leibler divergence. Since
$s\log s-s+1\ge 0$ for all $s>0$, with equality iff $s=1$, the right-hand side is nonnegative and vanishes iff $\F_i=\phiNN_i$ for all $i$.
\end{proof}

\begin{corollary}[Information-projection identity]\label{cor:info_projection}
Under the assumptions of \Cref{thm:discrete_maxent}, for every $\F\in\mathcal{F}(\boldsymbol{U})$,
\[
H(\F)-H(\phiNN(\boldsymbol{U}))
=
D_{\mathrm{KL}}^W\bigl(\F\,\|\,\phiNN(\boldsymbol{U})\bigr)
\ge 0.
\]
Thus $\phiNN(\boldsymbol{U})$ is the information projection of any feasible $\F$ onto the exponential family
\[
\mathcal{E}
=
\left\{
\bigl(\exp(\bm{\alpha}\cdot\bm{\varphi}(\latticevelocity_i))\bigr)_{i=1}^Q
\;:\;
\bm{\alpha}\in\real^p
\right\}.
\]
\end{corollary}

\begin{proposition}[Dual entropy and entropy variables]\label{prop:discrete_dual}
Define the Legendre--Fenchel dual of $\Psi$ by
\[
h(\boldsymbol{U})
\eqdef
\sup_{\bm{\alpha}\in\real^p}
\bigl\{
\bm{\alpha}\cdot\boldsymbol{U} - \Psi(\bm{\alpha})
\bigr\},
\qquad \boldsymbol{U}\in\mathcal{R}^\circ.
\]
Then:
\begin{enumerate}
    \item $h$ is real-analytic and strictly convex on $\mathcal{R}^\circ$;
    \item the supremum is attained uniquely at $\bm{\alpha}^\star(\boldsymbol{U})$, so that
    \[
    h(\boldsymbol{U})
    =
    \bm{\alpha}^\star(\boldsymbol{U})\cdot\boldsymbol{U}
    -
    \Psi\bigl(\bm{\alpha}^\star(\boldsymbol{U})\bigr);
    \]
    \item
    \[
    \nabla_{\boldsymbol{U}} h(\boldsymbol{U}) = \bm{\alpha}^\star(\boldsymbol{U}),
    \qquad
    \nabla_{\boldsymbol{U}}^2 h(\boldsymbol{U})
    =
    \bigl[\nabla_{\bm{\alpha}}^2\Psi(\bm{\alpha}^\star(\boldsymbol{U}))\bigr]^{-1}
    \succ 0;
    \]
    \item $h$ coincides with the minimum entropy value:
    \[
    h(\boldsymbol{U})
    =
    H\bigl(\phiNN(\boldsymbol{U})\bigr)
    =
    \sum_{i=1}^Q W_i\bigl(\phiNN_i\log\phiNN_i - \phiNN_i\bigr).
    \]
\end{enumerate}
In particular, $\bm{\alpha}^\star(\boldsymbol{U})=\nabla_{\boldsymbol{U}}h(\boldsymbol{U})$ are the entropy variables associated with the macroscopic entropy $h$.
\end{proposition}

\begin{proof}
Items (i)--(iii) follow from standard Legendre duality for strictly convex smooth potentials. For (iv),
\begin{align*}
H(\phiNN)
&=
\sum_{i=1}^Q W_i
\bigl(\phiNN_i\log\phiNN_i - \phiNN_i\bigr)
\\
&=
\sum_{i=1}^Q W_i
\bigl(\bm{\alpha}^\star\cdot\bm{\varphi}(\latticevelocity_i)\bigr)
\exp\!\bigl(\bm{\alpha}^\star\cdot\bm{\varphi}(\latticevelocity_i)\bigr)
-
\sum_{i=1}^Q W_i
\exp\!\bigl(\bm{\alpha}^\star\cdot\bm{\varphi}(\latticevelocity_i)\bigr)
\\
&=
\bm{\alpha}^\star\cdot\nabla_{\bm{\alpha}}\Psi(\bm{\alpha}^\star) - \Psi(\bm{\alpha}^\star)
=
\bm{\alpha}^\star\cdot\boldsymbol{U} - \Psi(\bm{\alpha}^\star)
=
h(\boldsymbol{U}).
\qedhere
\end{align*}
\end{proof}

\begin{proposition}[Symmetrizable hyperbolicity of the induced moment system]\label{prop:discrete_sym_hyp}
Consider the closed moment system
\begin{equation}\label{eq:closed_moment_system}
\partial_t \boldsymbol{U}
+
\sum_{j=1}^d \partial_{x_j}\boldsymbol{F}_j(\boldsymbol{U}) = 0,
\qquad
\boldsymbol{F}_j(\boldsymbol{U})
\eqdef
\sum_{i=1}^Q W_i\,(\latticevelocity_i)_j\,\bm{\varphi}(\latticevelocity_i)\,\phiNN_i(\boldsymbol{U}).
\end{equation}
Define the flux-generating potentials
\begin{equation}\label{eq:flux_generating}
G_j(\bm{\alpha})
\eqdef
\sum_{i=1}^Q W_i\,(\latticevelocity_i)_j\,\exp\!\bigl(\bm{\alpha}\cdot\bm{\varphi}(\latticevelocity_i)\bigr),
\qquad j=1,\dots,d.
\end{equation}
Then:
\begin{enumerate}
    \item $h(\boldsymbol{U})$ is a convex entropy for \cref{eq:closed_moment_system}, with associated entropy flux
    \[
    q_j(\boldsymbol{U})
    =
    \bm{\alpha}^\star(\boldsymbol{U})\cdot\boldsymbol{F}_j(\boldsymbol{U})
    -
    G_j\bigl(\bm{\alpha}^\star(\boldsymbol{U})\bigr);
    \]
    \item in entropy variables $\bm{\alpha}=\nabla_{\boldsymbol{U}}h(\boldsymbol{U})=\bm{\alpha}^\star(\boldsymbol{U})$, the system takes the symmetric form
    \begin{equation}\label{eq:symmetric_form}
    \nabla_{\bm{\alpha}}^2\Psi(\bm{\alpha})\,\partial_t\bm{\alpha}
    +
    \sum_{j=1}^d \nabla_{\bm{\alpha}}^2G_j(\bm{\alpha})\,\partial_{x_j}\bm{\alpha}
    =0;
    \end{equation}
    \item the leading matrix $\nabla_{\bm{\alpha}}^2\Psi$ is symmetric positive definite and each flux Jacobian in entropy variables is symmetric, so the system is symmetrizable hyperbolic on $\mathcal{R}^\circ$.
\end{enumerate}
\end{proposition}

\begin{proof}
Since $\boldsymbol{F}_j(\boldsymbol{U})=\nabla_{\bm{\alpha}}G_j(\bm{\alpha})$ with $\bm{\alpha}=\bm{\alpha}^\star(\boldsymbol{U})$, we compute at the level of differentials
\[
dq_j
=
d\bm{\alpha}^\star\cdot \boldsymbol{F}_j
+
\bm{\alpha}^\star\cdot d\boldsymbol{F}_j
-
\nabla_{\bm{\alpha}}G_j(\bm{\alpha}^\star)\cdot d\bm{\alpha}^\star
=
\bm{\alpha}^\star\cdot d\boldsymbol{F}_j.
\]
Hence
\[
\nabla_{\boldsymbol{U}} q_j
=
\bigl[D_{\boldsymbol{U}}\boldsymbol{F}_j(\boldsymbol{U})\bigr]^\top
\bm{\alpha}^\star(\boldsymbol{U})
=
\bigl[D_{\boldsymbol{U}}\boldsymbol{F}_j(\boldsymbol{U})\bigr]^\top
\nabla_{\boldsymbol{U}} h(\boldsymbol{U}),
\]
which is the entropy-flux compatibility condition.

Next, since $\boldsymbol{U}=\nabla_{\bm{\alpha}}\Psi(\bm{\alpha})$,
\[
\partial_t\boldsymbol{U}
=
\nabla_{\bm{\alpha}}^2\Psi(\bm{\alpha})\,\partial_t\bm{\alpha},
\qquad
\partial_{x_j}\boldsymbol{F}_j
=
\nabla_{\bm{\alpha}}^2G_j(\bm{\alpha})\,\partial_{x_j}\bm{\alpha}.
\]
Substituting these identities into \cref{eq:closed_moment_system} yields \cref{eq:symmetric_form}. By \Cref{lem:partition}, $\nabla_{\bm{\alpha}}^2\Psi(\bm{\alpha})\succ 0$, and each $\nabla_{\bm{\alpha}}^2G_j(\bm{\alpha})$ is symmetric because it is a Hessian.
\end{proof}

\begin{proposition}[Formal $H$-theorem for exact BGK relaxation]\label{prop:discrete_H}
Let $\F_i=\F_i(t,\bx)>0$ be a sufficiently regular solution of
\begin{equation}\label{eq:bgk_neurde}
\partial_t \F_i + \latticevelocity_i\cdot\nabla_{\bx}\F_i
=
\frac{1}{\tau}\bigl(\phiNN_i(\boldsymbol{U}_\F)-\F_i\bigr),
\qquad i=1,\dots,Q,
\end{equation}
where
\[
\boldsymbol{U}_\F(t,\bx)
=
\sum_{i=1}^Q W_i\,\bm{\varphi}(\latticevelocity_i)\,\F_i(t,\bx)
\]
and assume $\boldsymbol{U}_\F(t,\bx)\in\mathcal{R}^\circ$ for all $(t,\bx)$ under consideration. Assume periodic boundary conditions or boundary conditions under which spatial flux terms vanish after integration. Then the discrete kinetic entropy
\[
\mathscr{H}[\F](t)
\eqdef
\int_\Omega \sum_{i=1}^Q W_i\bigl(\F_i\log\F_i - \F_i\bigr)\,d\bx
\]
satisfies
\begin{equation}\label{eq:H_theorem_discrete}
\frac{d}{dt}\mathscr{H}[\F](t)
=
-\frac{1}{\tau}\int_\Omega
\sum_{i=1}^Q W_i\,
(\F_i - \phiNN_i)\log\frac{\F_i}{\phiNN_i}
\,d\bx
\le 0.
\end{equation}
The inequality is strict unless $\F_i = \phiNN_i(\boldsymbol{U}_\F)$ for all $i$, almost everywhere in $\Omega$.
\end{proposition}

\begin{proof}
Since
\[
\frac{d}{d\F_i}\bigl(\F_i\log\F_i-\F_i\bigr)=\log\F_i,
\]
we multiply \cref{eq:bgk_neurde} by $W_i\log\F_i$, sum over $i$, and integrate over $\Omega$.

\textit{Left-hand side.}
Using
\[
\log\F_i\,\partial_t\F_i
=
\partial_t(\F_i\log\F_i-\F_i),
\qquad
\log\F_i\,\latticevelocity_i\cdot\nabla_{\bx}\F_i
=
\nabla_{\bx}\cdot\Bigl(\latticevelocity_i(\F_i\log\F_i-\F_i)\Bigr),
\]
we obtain, after summing over $i$ and integrating over $\Omega$,
\[
\frac{d}{dt}\mathscr{H}[\F](t),
\]
because the divergence term vanishes under the assumed boundary conditions.

\textit{Right-hand side.}
The collision contribution is
\[
\frac{1}{\tau}\int_\Omega
\sum_{i=1}^Q W_i\,
(\phiNN_i-\F_i)\log\F_i
\,d\bx.
\]
Decompose
\[
\log\F_i
=
\log\frac{\F_i}{\phiNN_i}
+
\log\phiNN_i.
\]
Since
\[
\log\phiNN_i
=
\bm{\alpha}^\star(\boldsymbol{U}_\F)\cdot\bm{\varphi}(\latticevelocity_i),
\]
the second term becomes
\[
\sum_{i=1}^Q W_i(\phiNN_i-\F_i)\log\phiNN_i
=
\bm{\alpha}^\star(\boldsymbol{U}_\F)\cdot
\sum_{i=1}^Q W_i\,\bm{\varphi}(\latticevelocity_i)(\phiNN_i-\F_i)
=0,
\]
because $\phiNN(\boldsymbol{U}_\F)$ and $\F$ have the same learned moments. Therefore,
\[
\frac{d}{dt}\mathscr{H}[\F](t)
=
-\frac{1}{\tau}\int_\Omega
\sum_{i=1}^Q W_i\,
(\F_i - \phiNN_i)\log\frac{\F_i}{\phiNN_i}
\,d\bx.
\]
Since $(s-1)\log s\ge 0$ for all $s>0$, each summand is nonnegative.
\end{proof}

\subsubsection{Learning-aware entropy-defect analysis}
\label{subsec:learning_aware}

The results in \cref{subsec:structural} concern the \emph{exact entropy projector} associated with the learned basis. The practical NeurDE model, however, does not in general solve the nonlinear moment-inversion problem exactly. We therefore return to the raw learned branch output
\begin{equation}\label{eq:raw_multiplier}
\widehat{\bm{\alpha}}(\boldsymbol{U})
\eqdef
\bigl(
\balpha_1(\boldsymbol{U};\theta^\lambda),\dots,
\balpha_p(\boldsymbol{U};\theta^\lambda)
\bigr)^\top
\in\real^p
\end{equation}
and define the raw learned equilibrium
\begin{equation}\label{eq:raw_learned_equilibrium}
\widehat{\phiNN}_i(\boldsymbol{U})
\eqdef
\exp\!\bigl(
\widehat{\bm{\alpha}}(\boldsymbol{U})\cdot
\bm{\varphi}(\latticevelocity_i)
\bigr).
\end{equation}
Unlike the exact projector $\phiNN_i(\boldsymbol{U})$ of \Cref{thm:discrete_maxent}, the raw learned equilibrium $\widehat{\phiNN}_i(\boldsymbol{U})$ need not reproduce the target moments exactly.

\begin{definition}[Moment defect]\label{def:moment_defect}
The \emph{moment defect} of the raw learned equilibrium is
\begin{equation}\label{eq:moment_defect}
m(\boldsymbol{U})
\eqdef
\sum_{i=1}^Q W_i\,\bm{\varphi}(\latticevelocity_i)\,\widehat{\phiNN}_i(\boldsymbol{U})
-
\boldsymbol{U}
\;\in\real^p.
\end{equation}
\end{definition}

When $m\equiv 0$, the raw learned equilibrium coincides with the exact moment-matching projector on its domain of definition, and the ideal $H$-theorem is recovered. When $m\not\equiv 0$, the entropy law acquires an explicit defect term.

\begin{theorem}[Exact entropy balance for the raw learned equilibrium]\label{thm:discrete_defect}
Let $\F_i=\F_i(t,\bx)>0$ be a sufficiently regular solution of
\begin{equation}\label{eq:bgk_raw_neurde}
\partial_t \F_i + \latticevelocity_i\cdot\nabla_{\bx}\F_i
=
\frac{1}{\tau}\bigl(\widehat{\phiNN}_i(\boldsymbol{U}_\F)-\F_i\bigr),
\qquad i=1,\dots,Q,
\end{equation}
where
\[
\boldsymbol{U}_\F(t,\bx)
=
\sum_{i=1}^Q W_i\,\bm{\varphi}(\latticevelocity_i)\,\F_i(t,\bx).
\]
Assume periodic boundary conditions or boundary conditions under which spatial flux terms vanish after integration. Then the discrete kinetic entropy
\[
\mathscr{H}[\F](t)
=
\int_\Omega \sum_{i=1}^Q W_i\bigl(\F_i\log\F_i-\F_i\bigr)\,d\bx
\]
satisfies the exact identity
\begin{equation}\label{eq:defect_identity}
\frac{d}{dt}\mathscr{H}[\F](t)
=
\underbrace{
-\frac{1}{\tau}\int_\Omega
\sum_{i=1}^Q W_i\,
(\F_i-\widehat{\phiNN}_i)\log\frac{\F_i}{\widehat{\phiNN}_i}
\,d\bx
}_{\le 0\ \text{(dissipation)}}
\;+\;
\underbrace{
\frac{1}{\tau}\int_\Omega
\widehat{\bm{\alpha}}(\boldsymbol{U}_\F)\cdot m(\boldsymbol{U}_\F)\,d\bx
}_{\text{defect}}.
\end{equation}
\end{theorem}

\begin{proof}
As in \Cref{prop:discrete_H}, multiply \cref{eq:bgk_raw_neurde} by $W_i\log\F_i$, sum over $i$, and integrate over $\Omega$. The transport term gives $d\mathscr{H}[\F]/dt$. For the collision term,
\[
\frac{1}{\tau}\int_\Omega
\sum_{i=1}^Q W_i\,
(\widehat{\phiNN}_i-\F_i)\log\F_i
\,d\bx.
\]
Decompose
\[
\log\F_i
=
\log\frac{\F_i}{\widehat{\phiNN}_i}
+
\log\widehat{\phiNN}_i.
\]
The first piece yields the dissipative term in \cref{eq:defect_identity}. Since
\[
\log\widehat{\phiNN}_i(\boldsymbol{U}_\F)
=
\widehat{\bm{\alpha}}(\boldsymbol{U}_\F)\cdot\bm{\varphi}(\latticevelocity_i),
\]
the second piece becomes
\begin{align*}
\sum_{i=1}^Q W_i(\widehat{\phiNN}_i-\F_i)\log\widehat{\phiNN}_i
&=
\widehat{\bm{\alpha}}(\boldsymbol{U}_\F)\cdot
\sum_{i=1}^Q W_i\,\bm{\varphi}(\latticevelocity_i)(\widehat{\phiNN}_i-\F_i)
\\
&=
\widehat{\bm{\alpha}}(\boldsymbol{U}_\F)\cdot
\left(
\sum_{i=1}^Q W_i\,\bm{\varphi}(\latticevelocity_i)\,\widehat{\phiNN}_i
-
\boldsymbol{U}_\F
\right)
\\
&=
\widehat{\bm{\alpha}}(\boldsymbol{U}_\F)\cdot m(\boldsymbol{U}_\F).
\end{align*}
Integrating over $\Omega$ yields \cref{eq:defect_identity}.
\end{proof}

\begin{remark}[Boundary entropy flux]\label{rem:entropy_boundary_flux}
For non-periodic domains the same calculation retains the boundary entropy
flux. Defining
\[
\mathcal B_H(t)
=
\int_{\partial\Omega}
\nu(\bx)\cdot
\sum_{i=1}^Q W_i\,\latticevelocity_i
\bigl(\F_i\log\F_i-\F_i\bigr)
\,dS,
\]
\Cref{eq:defect_identity} becomes
\[
\frac{d}{dt}\mathscr H[\F](t)
=
-\mathcal B_H(t)
-\frac{1}{\tau}\int_\Omega
\sum_{i=1}^Q W_i(\F_i-\widehat{\phiNN}_i)
\log\frac{\F_i}{\widehat{\phiNN}_i}\,d\bx
+\frac{1}{\tau}\int_\Omega
\widehat{\bm{\alpha}}(\boldsymbol{U}_\F)\cdot m(\boldsymbol{U}_\F)\,d\bx .
\]
Thus inlet/outlet or wall treatments add the expected boundary flux, while the
bulk dissipation/defect decomposition is unchanged. The cylinder experiments
therefore use the theorem as a local bulk-closure diagnostic, with boundary
handling supplied by the host solver.
\end{remark}

\begin{corollary}[Approximate $H$-theorem]\label{cor:discrete_approx_H}
Let $K\subset\real^p$ be compact. If $\|m(\boldsymbol{U})\|\le \delta$ for all $\boldsymbol{U}\in K$ (in any norm $\|\cdot\|$ on $\real^p$, with dual norm $\|\cdot\|_\ast$), and if $\boldsymbol{U}_\F(t,\bx)\in K$ throughout the evolution, then
\begin{equation}\label{eq:approx_H}
\frac{d}{dt}\mathscr{H}[\F](t)
\le
-\frac{1}{\tau}\int_\Omega
\sum_{i=1}^Q W_i\,
(\F_i-\widehat{\phiNN}_i)\log\frac{\F_i}{\widehat{\phiNN}_i}
\,d\bx
+
\frac{|\Omega|}{\tau}
\sup_{\boldsymbol{U}\in K}\|\widehat{\bm{\alpha}}(\boldsymbol{U})\|_\ast\,\delta.
\end{equation}
If $\delta=0$, the defect term vanishes and one recovers the formal $H$-theorem.
\end{corollary}

\begin{proof}
Apply \Cref{thm:discrete_defect} and estimate pointwise
\[
\bigl|
\widehat{\bm{\alpha}}(\boldsymbol{U})\cdot m(\boldsymbol{U})
\bigr|
\le
\|\widehat{\bm{\alpha}}(\boldsymbol{U})\|_\ast\,\|m(\boldsymbol{U})\|
\le
\sup_{\boldsymbol{U}\in K}\|\widehat{\bm{\alpha}}(\boldsymbol{U})\|_\ast\,\delta.
\]
Then integrate over $\Omega$.
\end{proof}

\begin{remark}[Interpretation of the defect term]\label{rem:defect_bound}
The defect term in \cref{eq:defect_identity} is explicit and structurally meaningful:
\begin{enumerate}
    \item it vanishes exactly when the raw learned equilibrium matches the prescribed learned moments;
    \item it is linear in the moment residual $m(\boldsymbol{U})$;
    \item it is weighted by the learned multiplier $\widehat{\bm{\alpha}}(\boldsymbol{U})$, i.e., by the dual variables that parameterize the log-equilibrium.
\end{enumerate}
Thus the raw learned NeurDE model is not an exact maximum-entropy closure in general, but it is an \emph{approximate} entropy closure whose departure from the ideal law is explicit, measurable, and trainable.
\end{remark}

\begin{remark}[What the structural theory guarantees]
\label{rem:summary_guarantees}
Taken together, \Cref{thm:discrete_maxent,prop:discrete_dual,prop:discrete_sym_hyp,prop:discrete_H,thm:discrete_defect} show:
\begin{enumerate}
    \item the exponential branch--trunk form is the variational form singled out by entropy minimization;
    \item for a fixed admissible learned basis, the exact entropy projector is a genuine maximum-entropy closure with strictly convex dual entropy and symmetrizable hyperbolicity;
    \item the practical raw learned model satisfies an exact entropy identity with an explicit defect term;
    \item the ideal $H$-theorem is recovered precisely when the learned moment defect vanishes.
\end{enumerate}
This is the correct way to read the NeurDE structure: the fixed-basis exact projector inherits classical Levermore geometry, while the trained neural model inherits an approximate entropy law whose defect is explicit.
\end{remark}

\section{Methods (full version)}
\label{appendix:methods_full}

\subsection{Constructing the neural moment space}
\label{sec:moment_space}
The core of the \NN{} framework is the parameterization of the equilibrium manifold. We define the network's expressivity through a learned moment space
\begin{equation}\label{eq:M_space}
    \mathbb{M} \eqdef \mathrm{span}\left\{\bvarphi_k(\latticevelocity_i): i=1, \ldots, Q, k=1, \ldots, p\right\},
\end{equation}
where $\latticevelocity_i$ represents the discrete lattice velocities. The design of this space is governed by two competing requirements: conservation and expressivity.\ 

To ensure that the learned space can exactly recover the Eulerian moment basis, we explicitly enforce that the standard Eulerian moments, $\mathrm{span}\{1,\latticevelocity_i,\latticevelocity_i\!\cdot\!\latticevelocity_i\}$, are contained in $\mathbb{M}$ during training. Exact conservation of the network output is then imposed separately through the conservative constructions described in \cref{appx:conservative_neurde}. Simultaneously, to represent non-equilibrium features that are poorly captured by classical low-order closures, we select the dimension $p$ to be sufficiently large. This lets the network learn higher-order basis functions for flux and stress terms, trading closure expressivity against model complexity (see \cref{appendix:M_space}).

\subsection{Compressible lattice-Boltzmann realization}
\label{sec:specific_arch}
With the mathematical structure of the equilibrium defined, we now turn to its practical implementation within a numerical solver. To demonstrate the framework's capability in challenging regimes (\cref{section:Sod_cases,section:supersonic_flow}), we implement \NN{} within a compressible LB solver. We adopt a two-population thermal formulation to accommodate variable specific heat ratios ($\gamma$) and Prandtl numbers ($\mathrm{Pr}$) \cite{tran2022lattice, saadat2019lattice, frapolli2015entropic}, which are restricted to fixed values in standard kinetic schemes. \ 

The evolution of the system is split into two coupled relaxation processes, following the formulation of \cite{karlin2013consistent}(Eqs.74-75). The population $\F_i$ transports mass and momentum, while a separate population $\G_i$ transports energy. Their dynamics are governed by:
\begin{subequations}\label{eq:LBM}
   \begin{align}
        \F_i(t+1, \bx+ \latticevelocity_i ) - \F_i(t, \bx) &= \tfrac{1}{\tau_1}(\Feqlattice_i - \F_i), \label{eq:LBM-momentum} \\
        \begin{split}
        \G_i(t+ 1, \bx+ \latticevelocity_i )  - \G_i(t, \bx) &=   \tfrac{1}{\tau_2} (\Geqlattice_i - \G_i) \\
        & + (\tfrac{1}{\tau_2}- \tfrac{1}{\tau_1})(\G^\ast_i - \G_i). 
        \end{split} \label{eq:LBM-energy}
  \end{align}
\end{subequations}
This decoupling, mediated by the intermediate quasi-equilibrium $\G^\ast$ (see \cref{eq:quasiequilibrium})
allows us to independently set the viscosity ($\mu \propto \tau_1$) and the thermal conductivity ($\kappa \propto \tau_2$) \cite{saadat2019lattice}.The macroscopic quantities---density $\rho$, velocity $u$, and temperature $T$---are recovered by taking moments over these populations 
\begin{equation}
\label{eq:macroscopical_LBM_2_pop}
\begin{split}
\rho = \bra \F \ket, &\quad \rho\Velocity = \bra \latticevelocity \F \ket, \quad E = \tfrac{1}{2\rho}\bra \G \ket , \\
\temperature&= \tfrac{1}{\Cv}\left( E - \tfrac{1}{2}\Velocity\cdot \Velocity\right)
\end{split}
\end{equation}
with $\Cv$ and $\Cp$ the specific heats and $\gamma = \Cp/\Cv$, $\mathrm{Pr} = \Cp \mu / \kappa$. In the absorbed-population convention, the energy channel carried by $\G$ is
\[
\bra \G \ket = 2\rho E.
\]
This is the quantity enforced by the conservative hybrid energy correction in \Cref{thm:hybrid_two_population_neurde,cor:hybrid_thermal_specialization}. See \cref{appendix:two_populations}.

\subsection{Hybrid closure and training}
\label{sec:training}
Targeting the specific challenges of high-Mach flows, we implement a \textbf{hybrid closure strategy} within the double-population formulation (\cref{eq:LBM}) that separates the learned and analytic components. We use the analytic extended equilibrium (\cref{eq:Feq_prod}) for the momentum population $\Feqlattice_i$, while \NN{} learns the energy-channel equilibrium $\Geqlattice_i$. Traditional high-Mach LB solvers often add explicit correction terms and non-local operations to improve stability \cite{saadat2019lattice}; here the learned local energy closure replaces that part of the construction in the reported experiments.

To promote physical consistency across scales, the training procedure for \NN{} (\Cref{alg:LBM_NN_algorithm_full_training}) proceeds in two stages, aligning the optimization with both the mesoscopic statistical description and the macroscopic continuum dynamics (see panel (a) in \cref{alg:LBM_NN_algorithm}).

\begin{enumerate}
    \item \textbf{Static Pre-training (Mesoscopic Prior):} First, we train \NN{} to map known macroscopic states $\boldsymbol{U}$ to high-fidelity equilibrium targets. In the compressible experiments these targets are $\Geqlattice_i$; in the single-population scalar-law variants they are $\Feqlattice_i$. This gives the network a physically plausible local closure before temporal dynamics are introduced.
    \item \textbf{Dynamic Training (Macroscopic Consistency):} Second, we train the solver dynamics through temporal forecasting. We integrate the pre-trained \NN{} into the LB solver and perform multi-step rollouts, comparing the predicted macroscopic trajectories against high-fidelity simulations. Minimizing the cumulative rollout error encourages the locally accurate equilibria learned in stage one to compose into stable global dynamics, as in \cite{latt2020efficient}.
\end{enumerate}

\subsection{Experimental configurations}
\label{sec:experimental}
Having established the solver architecture and training protocol, we validate the framework against three distinct flow regimes, each designed to probe specific failure modes of classical schemes.

\subsection{Sod shock tube}\label{appendix:general_sod}

The Sod shock tube is a canonical one-dimensional Riemann problem for the
compressible Euler equations and remains one of the standard benchmarks for
shock-capturing methods \cite{sod1978survey}. The domain is initially divided by a
diaphragm at \(x=x_0\) into a high-pressure left state and a low-pressure right
state, both at rest,
\[
(\rho,u,p)(x,0)=
\begin{cases}
(\rho_L,0,p_L), & x < x_0,\\[2mm]
(\rho_R,0,p_R), & x > x_0.
\end{cases}
\]
After the diaphragm is removed, the discontinuous initial data evolves into a
self-similar solution depending on \(\xi=(x-x_0)/t\). For the classical Sod
configuration with \(p_L>p_R\) and \(\rho_L>\rho_R\), the solution consists of
three elementary wave families: a left-propagating rarefaction fan, a contact
discontinuity moving to the right, and a right-propagating shock
\cite{sod1978survey}.

Figure~\ref{fig:sod_illustration} summarizes the wave structure that emerges in
the Sod shock tube after the diaphragm is removed. In
Fig.~\ref{fig:sod_illustration}\textbf{a}, the initial condition consists of
two constant states separated at \(x_0\): a high-pressure region on the left
and a low-pressure region on the right. Once the diaphragm is removed, this
imbalance generates waves that propagate away from the initial discontinuity.

Fig.~\ref{fig:sod_illustration}\textbf{b} shows the corresponding
space--time diagram, with position \(x\) on the horizontal axis and time \(t\)
on the vertical axis. In this representation, each slanted line or curve traces
the location of a wave as time evolves. These wave paths divide the
\(x\)--\(t\) plane into five regions, each corresponding to a different local
fluid state: the undisturbed left state, the rarefaction fan, the uniform state
behind the rarefaction, the uniform state between the contact discontinuity and
the shock, and the undisturbed right state. The rarefaction occupies a finite
wedge-shaped region bounded by its head and tail characteristics, whereas the
contact discontinuity and shock appear as sharp trajectories moving to the
right.

Each of these waves reflects a different aspect of compressible flow. Across
the rarefaction fan, the solution changes smoothly and, for an ideal gas, the
expansion is isentropic. Across the contact discontinuity, pressure and
velocity remain continuous while density changes abruptly. Across the shock, the
state changes discontinuously according to the Rankine--Hugoniot jump
conditions. As a result, the Sod problem provides a compact but demanding test:
a numerical method must simultaneously resolve a smooth expansion, sharply
capture a contact discontinuity, and predict the correct shock speed
\cite{sod1978survey}.

This wave structure is seen directly in the density profile in
Fig.~\ref{fig:sod_illustration}\textbf{c}. Starting from the left, the first
plateau corresponds to the undisturbed driver gas. This is followed by the
smooth decrease in density across the rarefaction fan, then by a nearly
constant intermediate region. The contact discontinuity produces a density jump
between two intermediate plateaus, while the shock creates the final sharp jump
back to the undisturbed right state. Because all of these canonical features
appear in a single exact solution, the Sod shock tube remains a standard and
informative benchmark for compressible solvers \cite{sod1978survey}.

\begin{figure}[htp]
    \centering
\definecolor{natureBlue}{RGB}{31, 119, 180}
\definecolor{natureRed}{RGB}{214, 39, 40}
\definecolor{natureGreen}{RGB}{44, 160, 44}
\definecolor{natureOrange}{RGB}{255, 127, 14}
\definecolor{natureGray}{RGB}{120, 120, 120}
\definecolor{natureDarkGray}{RGB}{50, 50, 50} 

\begin{tikzpicture}[
    x=1.2cm, y=1.2cm, 
    every node/.style={text=natureDarkGray},
    line cap=round, line join=round
]

    \node[anchor=west, text=black] at (-1.5, 10.5) {{\Large\textbf{a}} \quad \textbf{Initial State} ($t = 0$)};
    
    \draw[thick, natureDarkGray] (0, 8.5) rectangle (12, 9.5);
    
    \fill[natureBlue, opacity=0.15] (0, 8.5) rectangle (6, 9.5);
    \node at (3, 9.2) {\textbf{Driver Section} ($L$)};
    \node at (3, 8.8) {\small $\rho_\text{L}, P_\text{L}, u_\text{L} = 0$};
    
    \fill[natureRed, opacity=0.1] (6, 8.5) rectangle (12, 9.5);
    \node at (9, 9.2) {\textbf{Driven Section} ($R$)};
    \node at (9, 8.8) {\small $\rho_\text{R}, P_\text{R}, u_\text{R} = 0$};

    \draw[ultra thick, natureDarkGray] (6, 8.3) -- (6, 9.7) node[above, text=black] {\small \textbf{Diaphragm} ($x_0$)};

    \node[anchor=west, text=black] at (-1.5, 7.5) {{\Large\textbf{b}} \quad \textbf{Space-Time Diagram} ($x$-$t$)};
    
    \draw[thick, natureDarkGray] (0, 3.5) -- (12.2, 3.5) node[right, text=black] {$x$};
    \draw[thick, natureDarkGray] (6, 3.5) -- (6, 7.2) node[above, text=black] {$t$};
    
    \draw[natureDarkGray] (6, 3.5) -- (6, 3.4) node[below] {\small $x_0$};
    \draw[natureDarkGray] (0, 3.5) -- (0, 3.4);
    \draw[natureDarkGray] (12, 3.5) -- (12, 3.4);

    \fill[natureBlue, opacity=0.15] (6, 3.5) -- (3, 6.5) -- (4.5, 6.5) -- cycle;
    \draw[natureBlue, thick] (6, 3.5) -- (3, 6.5) node[above left, text=natureBlue, font=\footnotesize] {Tail};
    \draw[natureBlue!50, thin] (6, 3.5) -- (3.5, 6.5); 
    \draw[natureBlue!50, thin] (6, 3.5) -- (4.0, 6.5); 
    \draw[natureBlue, thick] (6, 3.5) -- (4.5, 6.5) node[above, text=natureBlue, xshift=-0.2cm, font=\footnotesize] {Head};
    
    \draw[natureOrange, ultra thick] (6, 3.5) -- (8.5, 6.5) node[above, text=natureOrange, font=\footnotesize] {Contact};
    
    \draw[natureRed, ultra thick] (6, 3.5) -- (11, 6.5) node[above right, text=natureRed, font=\footnotesize] {Shock};
    
    \node[circle, draw=natureDarkGray, inner sep=1pt, minimum size=4mm] at (1.5, 5.5) {\footnotesize 1};
    \node[circle, draw=natureDarkGray, inner sep=1pt, minimum size=4mm] at (3.8, 6.1) {\footnotesize 2};
    \node[circle, draw=natureDarkGray, inner sep=1pt, minimum size=4mm] at (7.0, 5.5) {\footnotesize 3};
    \node[circle, draw=natureDarkGray, inner sep=1pt, minimum size=4mm] at (9.6, 5.5) {\footnotesize 4};
    \node[circle, draw=natureDarkGray, inner sep=1pt, minimum size=4mm] at (11.6, 5.5) {\footnotesize 5};

    \node[anchor=west, text=black] at (-1.5, 2.5) {{\Large\textbf{c}} \quad \textbf{Fluid Profile} ($t = t_1$)};
    
    \draw[thick, natureDarkGray] (0, -2.0) -- (12.2, -2.0) node[right, text=black] {$x$};
    \draw[thick, natureDarkGray] (0, -2.0) -- (0, 1.0) node[above, text=black, yshift=5.5pt] {$\rho(x)$};
    
    \draw[natureDarkGray, dashed, opacity=0.3] (3, 6.5) -- (3, -2.0);
    \draw[natureDarkGray, dashed, opacity=0.3] (4.5, 6.5) -- (4.5, -2.0);
    \draw[natureDarkGray, dashed, opacity=0.3] (8.5, 6.5) -- (8.5, -2.0);
    \draw[natureDarkGray, dashed, opacity=0.3] (11, 6.5) -- (11, -2.0);

    
    \draw[ultra thick, natureDarkGray] (0, 0.5) -- (3, 0.5);
    
    \draw[ultra thick, natureBlue] (3, 0.5) .. controls (3.8, 0.5) and (4.2, -0.3) .. (4.5, -0.3);
    
    \draw[ultra thick, natureDarkGray] (4.5, -0.3) -- (8.5, -0.3);
    
    \draw[ultra thick, natureOrange, dashed] (8.5, -0.3) -- (8.5, -0.9);
    \draw[ultra thick, natureDarkGray] (8.5, -0.9) -- (11, -0.9);
    
    \draw[ultra thick, natureRed, dashed] (11, -0.9) -- (11, -1.3);
    \draw[ultra thick, natureDarkGray] (11, -1.3) -- (12, -1.3);
    
    \node[text=natureBlue, below] at (3.75, -2.15) {\footnotesize \textbf{Expansion}};
    \node[text=natureOrange, below] at (8.5, -2.15) {\footnotesize \textbf{Contact}};
    \node[text=natureRed, below] at (11, -2.15) {\footnotesize \textbf{Shock}};
    
\end{tikzpicture}
      \caption[Conceptual diagram of the Sod shock tube problem.]{\textbf{Conceptual diagram of the Sod shock tube problem.} \textbf{a}, Initial high-pressure and low-pressure states separated by a diaphragm at \(x_0\). \textbf{b}, Space--time wave diagram showing the rarefaction fan, contact discontinuity, and primary shock. \textbf{c}, Density profile at \(t=t_1\), showing how these waves appear as spatial flow features.}
      \label{fig:sod_illustration}
\end{figure}

\subsubsection{Subsonic Baseline}
We first calibrate the method on a standard subsonic Sod shock tube to verify the accurate capture of rarefaction waves, contact discontinuities, and shocks. We employ a specific heat ratio $\gamma=2.0$ and Prandtl number $\mathrm{Pr}=0.71$. The flow is initialized with a density ratio of $5:1$ and a temperature ratio of $8:1$:
\begin{equation}\label{eq:Sod_case_1}
    \left( \rho, \Velocity_x, \Velocity_y, \temperature \right) = 
    \begin{cases}
        (0.5, 0, 0, 0.2),&  x/ L_x \le 1/2,\\ 
        (2.5, 0, 0, 0.025),&  x/ L_x > 1/2
    \end{cases}    
\end{equation}
with $L_x = 3001$ \cite{sod1978survey}. To mitigate Galilean invariance errors inherent to discrete velocity lattices, we apply a small reference frame shift $\Velocity_{\mathrm{shift}}=(\tfrac{3}{50},0)$ in \LBNN{}, following \cite{frapolli2016lattice}. 

\paragraph{Benchmarking against Operator Learning:} To isolate the benefits of the hybrid kinetic architecture, we establish a comparative baseline using a Fourier Neural Operator (FNO) trained on this same configuration. This comparison highlights the distinction between the physics-embedded \LBNN{} and purely data-driven operator learning, particularly regarding conservation and shock resolution. The FNO baseline employs a standard 2D architecture trained with the pointwise MSE objective detailed in \Cref{appx:FNO}. Capturing shock dynamics is challenging for Neural Operators~\cite{mcgreivy2024weak,FalsePromizeZeroShot_TR}, and is known to be challenging for spectral methods like FNO. In practice, the tested autoregressive FNO rollouts destabilize rapidly on this benchmark, so we use FNO only as a short-horizon auxiliary baseline. To make the comparison nontrivial, we tune the FNO over high spectral resolutions, embedding dimensions, and depths; \Cref{appx:FNO} reports the full grid, hardware-limited capacity study, and boundary treatment.

While FNO is a data-driven architecture, its reliance on spectral convolutions to achieve a global receptive field introduces a specific inductive bias towards periodic continuous functions. To address the non-periodic boundaries of the Sod shock tube and mitigate spectral leakage, we implement a \emph{Palindromic Edge-Padded Extension}  (\Cref{fig:boundaries}). This forces $C^1$ continuity at the periodic wrap of the fast Fourier Transform (FFT) inputs. However, despite this rigorous pre-processing, the global support of the underlying Fourier basis functions persists within the learned operator, resulting in characteristic Gibbs phenomenon oscillations near shock discontinuities (\Cref{fig:fno_1,fig:fno_2,fig:fno_n512}). This isolates the difficulty of capturing sharp hyperbolic features using global frequency-domain approximations.  Comprehensive details regarding the boundary handling are provided in \cref{appx:FNO}.

\subsubsection{Transonic Baseline} To probe stability limits, we simulate a transonic shock tube with low viscosity ($\mu = 10^{-4}$), approaching the theoretical stability bound of the collision operator ($\tau_1 \!\to\! 1/2$, \cref{eq:viscosity}). In this stiff regime, standard polynomial closures typically fail due to the under-resolution of high-order moments. We set $\gamma=1.4$ and initialize the flow with: 
\begin{equation}\label{eq:Sod_case_2}
    \left( \rho/\rho_0, \Velocity_x/\sqrt{\Velocity_0\cdot \Velocity_0} , \Velocity_y/\sqrt{\Velocity_0\cdot \Velocity_0}, \pressure/\pressure_0 \right) = 
    \begin{cases}
        (1.0, 0, 0, 1.0), & \quad x/L_x \le 1/2,\\[0.3em]
        (0.125, 0, 0, 0.1), & \quad x/L_x > 1/2.
    \end{cases}    
\end{equation}
and $\gamma = 1.4$, $\mathrm{Pr} = 0.71$, and $L_x = 3001$.
 To maintain stability in the presence of strong discontinuities, we apply a larger velocity shift of $(\tfrac{2}{5}, 0)$ and regularize the training with a Total Variation Diminishing (TVD) penalty to suppress spurious Gibbs oscillations (see \cref{appendix:TVD}) \cite{frapolli2016lattice}. Near Mach~1 (\cref{fig:Mach_case2_700_Mach}), the polynomial equilibria fail to capture high-order moments, leading to breakdowns unrelated to $\tau\!\to\!1/2$ but to closure errors at transonic speeds.



\subsection{2D supersonic flow past a cylinder}\label{appendix:supersonic_shock}

The two-dimensional supersonic flow past a cylinder provides a useful benchmark
for compressible solvers because it combines several challenging features in a
single configuration, including a detached bow shock, strong compression, and
rapid flow turning around a curved obstacle. In the present setup, a uniform
supersonic inflow enters from the left at Mach
\(\mathrm{Ma}_\infty = 1.8\) and Reynolds number \(\mathrm{Re}=300\), and
interacts with a circular cylinder embedded in a rectangular computational
domain.

Figure~\ref{fig:cylinder-bc-schematic} summarizes the geometry, boundary
conditions, and the main qualitative flow structures expected in this regime.
The left boundary prescribes the incoming freestream state through Dirichlet
data for \(\rho\), \(\mathbf{u}\), and \(T\), while the right boundary uses a
first-order Neumann outflow condition. The top and bottom boundaries are treated
as free-streaming boundaries so that outgoing disturbances can leave the domain
with minimal artificial reflection. At the cylinder surface, a bounce-back wall
condition enforces the solid no-penetration constraint.

As the supersonic inflow encounters the cylinder, it cannot adjust smoothly
upstream and instead forms a detached bow shock ahead of the body. This shock
compresses and decelerates the incoming gas before the flow moves around the
cylinder. Downstream of the shock, the fluid undergoes strong turning around the
curved surface, which generates substantial gradients in pressure, density, and
velocity. Farther downstream, the interaction of these compressed flow regions
can produce additional weak compressive structures, often described as
recompression waves or shocks.

This configuration is more demanding than a one-dimensional Riemann problem
because the dominant structures are genuinely two-dimensional and curved.
Accurate simulation requires the method to predict the stand-off distance and
shape of the bow shock, maintain a stable post-shock solution around the
cylinder, and resolve the downstream compressive features without introducing
spurious oscillations. For this reason, supersonic flow past a cylinder provides
a stringent test of the robustness and resolution properties of compressible
flow solvers.

\begin{figure}[htp]
    \centering
    \definecolor{dombg}{HTML}{FFFFFF}    
\definecolor{cylfill}{HTML}{222222}  
\definecolor{bowcol}{HTML}{D55E00}   
\definecolor{shearcol}{HTML}{0072B2} 
\definecolor{reccol}{HTML}{E69F00}   

\begin{tikzpicture}[
    >=Stealth, 
    font=\sffamily\small,
    boundary/.style={thick, black!85},
    freestream/.style={thick, black!60, dashed},
    inletarrow/.style={->, thick, black!75},
    featurelabel/.style={fill=white, inner sep=3pt, align=center, draw=black!20, rounded corners=2pt},
    walllabel/.style={fill=white, inner sep=2pt, align=center, font=\footnotesize, draw=black!20, rounded corners=2pt}
]
    
    \def\L{11.5} 
    \def\H{7.0}  
    \def\Cx{3.0} 
    \def\Cy{3.5} 
    \def\R{0.6}  
    
    \draw[thick, fill=dombg, draw=white] (0,0) rectangle (\L,\H);
    
    \draw[->, thick, black!70] (0.4, 0.4) -- (1.2, 0.4) node[right, font=\itshape] {x};
    \draw[->, thick, black!70] (0.4, 0.4) -- (0.4, 1.2) node[above, font=\itshape] {y};

    \draw[boundary] (0,0) -- (0,\H);
    \foreach \y in {0.7, 1.82, 2.94, 4.06, 5.18, 6.3} {
        \draw[inletarrow] (-0.8, \y) -- (0, \y);
    }
    \node[anchor=east, align=right] at (-1.2, \Cy) {
        \textbf{Inlet}\\[4pt]
        Dirichlet:\\[2pt]
        $\rho, \mathbf{u}, T$\\[4pt]
        $\mathrm{Ma}_\infty = 1.8$
    };
    
    \draw[boundary] (\L,0) -- (\L,\H);
    \node[anchor=west, align=left] at (\L+0.3, \Cy) {
        \textbf{Outlet}\\[4pt]
        $1^{\text{st}}$-order\\[2pt]
        Neumann\\[4pt]
        $\frac{\partial \phi}{\partial x} = 0$
    };
    
    \draw[freestream] (0,\H) -- (\L,\H);
    \draw[freestream] (0,0) -- (\L,0);
    \node[anchor=south, text=black!80] at (\L/2, \H+0.1) {\textbf{Top boundary:} free-streaming};
    \node[anchor=north, text=black!80] at (\L/2, -0.1) {\textbf{Bottom boundary:} free-streaming};

    
    \draw[thick, bowcol, dashed] (\Cx-1.1, \Cy) .. controls (\Cx-1.1, \Cy+2.5) and (\Cx+2.0, \H-0.5) .. (10.5, \H);
    \draw[thick, bowcol, dashed] (\Cx-1.1, \Cy) .. controls (\Cx-1.1, \Cy-2.5) and (\Cx+2.0, 0.5) .. (10.5, 0);
    
    \node[featurelabel, draw=bowcol!50] (bowtext) at (\Cx-1.5, \H-1.2) {\textbf{Bow shock}};
    \draw[->, bowcol, semithick] (bowtext.south) -- (\Cx-0.75, \Cy+1.2);
    
    \draw[thick, shearcol, dash pattern=on 6pt off 3pt] (\Cx+0.2, \Cy+\R-0.1) 
        .. controls (\Cx+1.5, \Cy+0.3) .. (\L, \Cy+0.3);
    \draw[thick, shearcol, dash pattern=on 6pt off 3pt] (\Cx+0.2, \Cy-\R+0.1) 
        .. controls (\Cx+1.5, \Cy-0.3) .. (\L, \Cy-0.3);
    
    \node[featurelabel, draw=shearcol!50] (shear) at (\Cx+2.1, \Cy+1.7) {Shear-layer separation};
    \draw[->, shearcol, semithick] (shear.south) -- (\Cx+0.8, \Cy+0.45);

    \draw[->, semithick, shearcol] (\Cx+0.85, \Cy+0.22) arc (180:-90:0.35 and 0.15);
    \draw[->, semithick, shearcol] (\Cx+0.85, \Cy-0.22) arc (180:450:0.35 and 0.15);
    
    \node[featurelabel, draw=shearcol!50] (recirc) at (\Cx+3.6, \Cy) {Recirculation\\bubble};
    \draw[->, shearcol, semithick] (recirc.west) -- (\Cx+1.5, \Cy);
    
    \draw[semithick, reccol, loosely dashed] (\Cx+1.5, \Cy+0.5) -- (\L, \Cy+2.5);
    \draw[semithick, reccol, loosely dashed] (\Cx+1.5, \Cy-0.5) -- (\L, \Cy-2.5);
    
    \draw[<->, reccol, semithick] (\Cx+6.8, \Cy-2.0) -- (\Cx+6.8, \Cy+2.0) node[midway, featurelabel, draw=reccol!50] {\textbf{Recompression}\\ \textbf{shocks}};

    \filldraw[thick, fill=cylfill, draw=black] (\Cx,\Cy) circle (\R);
    \node[font=\small, text=white] at (\Cx,\Cy) {\textbf{Wall}};
    
    \node[walllabel] (wallnode) at (\Cx+0.4, \Cy-\R-1.0) {bounce-back};
    \draw[->, thick, black!70] (wallnode.north) -- (\Cx+0.15, \Cy-\R+0.02);

\end{tikzpicture}
    \caption[Schematic of the 2D supersonic flow past a cylinder]{\textbf{Schematic
    of the computational domain, boundary conditions, and dominant flow features}
    for the 2D supersonic flow past a cylinder at \(\mathrm{Re}=300\) and
    \(\mathrm{Ma}_\infty=1.8\). The illustration highlights the detached bow shock,
    separated shear layers, recirculation region, and downstream recompression
    shocks observed in the reference and LB+NeurDE solutions (see
    Fig.~\ref{fig:cylinder_main}).}
    \label{fig:cylinder-bc-schematic}
\end{figure}

\subsubsection{2D Supersonic Interaction}
Finally, we test geometric and supersonic flow using flow past a cylinder. The Reynolds number is $\mathrm{Re}=300$ (based on cylinder diameter), the far-field Mach number $\mathrm{Ma}_\infty=1.8$, temperature $\temperature_\infty=0.2$, and $\gamma=1.4$. The local Mach number is $\mathrm{Ma}=(\Velocity \cdot \Velocity)^{1/2}(\gamma R \temperature)^{-1/2}$, assuming an ideal gas with $R=1$. We apply a streamwise lattice velocity shift $\Velocity_{\textrm{shift}} = \tfrac{3}{5}\mathrm{Ma}_\infty\sqrt{\gamma R\temperature_\infty}\,(1,0)$, with the $3/5$ coefficient chosen empirically to reduce spurious bow-shock oscillations. 

Dirichlet conditions are imposed at the inlet for $\rho$, $\Velocity$, and $\temperature$, first-order Neumann conditions at the outlet, free-streaming conditions along the top and bottom boundaries, and no-slip/no-penetration on the cylinder wall via bounce-back (\cref{fig:cylinder_main}). The reference solution employs an exponential moment closure for $\Geqlattice_i$ obtained via root finding~\cite{latt2020efficient}, combined with the extended equilibrium $\Feqlattice_i$ (\cref{eq:Feq_prod}).



\section{\NN{} General Setting and Training Algorithm} \label{appendix:NeurDE}
This section records the operator-level formulation behind the \LBNN{} implementation used in the main text.
We distinguish three objects: the fixed moment projection that maps kinetic populations to macroscopic variables, the fixed streaming operator supplied by the kinetic scheme, and the learned local equilibrium map used inside collision.
This separation is the organizing principle for the ablations in \Cref{sec:ablation}: changing the equilibrium closure is the main method, while learning collision or streaming wholesale is treated as a stress test.

The material is organized as follows.
\Cref{subxn:structure_} introduces the general formulation of the operators, and \cref{appx:general_setting} interprets the model as a coupled macroscopic--mesoscopic system.
\Cref{appendix:streaming_cnn} shows that the lattice Boltzmann streaming operator can be represented exactly as a non-learnable depthwise convolution, providing the basis for viewing the entire lattice Boltzmann scheme as a neural network architecture in \cref{appx:LNN_architecture}.
\Cref{sec:moment_space} describes the construction of the learned moment space $\mathbb{M}$ that determines the expressivity of the \NN{} equilibrium.
\Cref{appx:conservative_neurde} addresses the gap between approximate and exact conservation: it presents three constructive routes to machine-precision conservation of mass, momentum, and energy, including a differentiable projection, a null-space parameterization, and a constrained exponential-family construction.
\Cref{appx:training} outlines the two-stage training strategy---pretraining followed by joint trajectory optimization---with the pretraining procedure detailed in \cref{appx:pretraining} and the full training algorithm in \cref{appx:training_alg}.
Finally, \cref{sec:training} describes the hybrid closure strategy used in the compressible-flow experiments, in which the analytic extended equilibrium is retained for the momentum population while the energy closure is delegated entirely to \NN{}.

\paragraph{Scope.} The main experiments keep streaming fixed and learn only the equilibrium mapping within collision.
\Cref{appendix:streaming} explores a learnable streaming surrogate as an ablation; its poorer behavior supports the design choice but is not part of the main method.

\subsection{Operator Structure}
\label{subxn:structure_}
Let \(\F=(f_{1},\dots,f_{Q})^{\top}\) denote the lattice populations and \(\boldsymbol{U}(t,\bx)=\M[\F]\) the conserved macroscopic observables (density, momentum, and energy density). We represent the equilibrium via the composite map
\begin{equation}\label{eq:operator_distr_to_eq}
\F \xlongrightarrow{\ \M\ }\boldsymbol{U}(t,\bx)
\xlongrightarrow{\ \bm{\phi}\ }\Feqlattice(t,\bx),    
\end{equation}

\noindent where \(\bm{\phi}:\boldsymbol{U}\mapsto\Feqlattice\) assigns lattice populations consistent with \(\boldsymbol{U}\).  In classical LB, \(\bm{\phi}\) is a fixed, low-order polynomial Maxwellian; here, we replace \(\bm{\phi}\) by a neural network \(\phiNN\) and express the collision (\cref{eq:LBM_collision}) as
\begin{equation}\label{eq:collision_as_operator_LB}
\phiCNN
= \Bigl(1-\tfrac{1}{\tau}\Bigr)\bI 
+ \tfrac{1}{\tau}\,\bigl(\phiNN\circ\M\bigr),
\end{equation}
an affine relaxation map combining the identity \(\bI\) and the learned equilibrium. The streaming operator \(\phiS\) (\cref{eq:LBM_streaming}) is an exact lattice shift, introducing no trainable parameters or numerical dissipation. It remains fixed, and only \(\phiCNN\) is learned. The complete update is given by the composition \(\phiS\,\phiCNN\) (\cref{eq:LBM_algorithm_dimensionless}).

\subsection{General Formulation and Interpretation}\label{appx:general_setting}
The \LBNN{} framework couples a macroscopic conservation law with its mesoscopic (kinetic) representation.  
At the macroscopic level, the system evolves in terms of conserved observables $\boldsymbol{U}(t,\bx)$ that satisfy a nonlinear conservation law (cf. the green region in \cref{fig:architecture}(a)).  
At the kinetic level, the dynamics are described by the discrete particle populations
\[
\F(t,\bx) = [f_i(t,\bx)]_{i=1}^Q,
\]
which evolve according to a discrete Boltzmann--BGK process (cf. the blue region in \cref{fig:architecture}(a)).  

The bridge between these two levels is established by two fixed operators: a \emph{lifting operator}, which maps macroscopic observables $\boldsymbol{U}$ to a consistent kinetic state $\F$; and a \emph{projection operator}, which recovers $\boldsymbol{U}$ as velocity moments of $\F$:
\begin{equation}\label{eq:moment_projection_appendix}
\boldsymbol{U}(t,\bx) = \mathcal{D}[\F](t,\bx),
\end{equation}
where $\mathcal{D}$ is the moment map. For instance, 

\begin{equation}\label{eq:micro_into_macro}
\mathcal{D}[f](t, \bx)
=
\left\langle
\begin{pmatrix}
1, \velocity, \tfrac{1}{2}\velocity \cdot \velocity
\end{pmatrix}^{\top}
f(t, \bx, \velocity)
\right\rangle
=
\begin{pmatrix}
\rho, \rho \Velocity, \rho E
\end{pmatrix}^{\top}.
\end{equation}
Here \(E\) denotes the specific total energy.

During training and inference, \LBNN{} evolves the kinetic populations forward in time using the learned equilibrium closure (\NN{}), embedded within the standard lattice Boltzmann collision--streaming step.  
The updated state is then projected back to obtain $\boldsymbol{U}$, completing one lifting--evolution--projection cycle.  
In ML terms, this structure is analogous to an \emph{encoder--decoder model}, with analytically known and fixed encoding (lifting) and decoding (projection) operators.  

This coupling between the kinetic and macroscopic levels is illustrated schematically in \cref{fig:training_micro_macro}, where training occurs in the kinetic space while observables are recovered by projection at each step.
\begin{figure}[htb!]
    \centering
        \sffamily 
    
    \begin{tikzpicture}[>=Stealth, scale=0.95, transform shape]
    
    \pgfdeclarelayer{background}
    \pgfsetlayers{background,main}

    \pgfmathsetmacro{\colfirst}{0}
    \pgfmathsetmacro{\colsec}{5}     
    \pgfmathsetmacro{\colthird}{10}    
    \pgfmathsetmacro{\rowfirst}{0}
    \pgfmathsetmacro{\rowsecond}{2.2}
     
    \tikzset{
        timeline/.style={line width=2.5pt, line cap=round},
        arrow/.style={->, line width=1pt, pnasgray, dashed, -{Stealth[length=2.5mm, width=1.8mm]}},
        label/.style={font=\small\sffamily, inner sep=2pt},
        box/.style={
            rectangle, 
            rounded corners=1pt, 
            draw=black!40, 
            thick, 
            fill=white, 
            minimum width=1.6cm, 
            minimum height=0.8cm, 
            font=\small\sffamily, 
            align=center
        }
    }
     
    \begin{pgfonlayer}{background}
        \shade[top color=pnasblue!5, bottom color=pnasgreen!5, rounded corners=5pt] 
            (\colfirst-0.5, \rowsecond+0.5) rectangle (\colthird+0.5, \rowfirst-0.5);
        
        \draw[dashed, line width=0.8pt, gray!40] (\colsec, \rowsecond+1.5) -- (\colsec, \rowfirst-1.5);
    \end{pgfonlayer}
     
    \node[font=\bfseries\sffamily, anchor=east, text=pnasblue] at (-0.8, \rowsecond+0.15) {Discrete kinetic level};
    \node[font=\large, anchor=east] at (-0.8, \rowsecond-0.35) 
        {$\mathbf{f} = [f(t,\mathbf{x}, \mathbf{c}_j)]_{j=1}^Q$};
     
    \node[font=\bfseries\sffamily, anchor=east, text=pnasgreen] at (-0.8, \rowfirst+0.15) {Macroscopic};
    \node[font=\large, anchor=east] at (-0.8, \rowfirst-0.35) 
        {$\boldsymbol{U}(t,\bx) = (\rho, \rho\Velocity, E)$};
     
    \draw[timeline, pnasblue!60, dashed] (\colfirst, \rowsecond) -- (\colsec, \rowsecond);
    \draw[timeline, pnasblue!60] (\colsec, \rowsecond) -- (\colthird, \rowsecond);
     
    \draw[timeline, pnasgreen] (\colfirst, \rowfirst) -- (\colsec, \rowfirst);
    \draw[timeline, pnasgreen, dashed] (\colsec, \rowfirst) -- (\colthird, \rowfirst);
    
    \node[font=\small\bfseries\sffamily, text=black!60] at (2.5, \rowsecond+0.6) {Training Phase};
    \node[font=\small\bfseries\sffamily, text=black!60] at (7.5, \rowsecond+0.6) {Testing Phase};

    \node[box, fill=pnasred!10, draw=pnasred!50] at (1.5, \rowsecond+1.4) {Equilibrium\\$\Feqlattice$};
    \node[box, fill=pnasblue!10, draw=pnasblue!50] at (3.5, \rowsecond+1.4) {Collision\\+Stream};
     
    \node[box, fill=pnasred!10, draw=pnasred!50] at (6.5, \rowsecond+1.4) {NN\\Predict};
    \node[box, fill=pnasblue!10, draw=pnasblue!50] at (8.5, \rowsecond+1.4) {Collision\\+Stream};
     
    \foreach \x in {0.5, 1.8, 3.2, 4.7} {
        \draw[arrow, ->] (\x, \rowfirst+0.15) -- (\x, \rowsecond-0.15);
    }
    \node[box, fill=pnasorange!20, minimum width=1.0cm, minimum height=0.6cm] at (2.5, 1.1) {$\mathcal{E}$};
    \node[font=\scriptsize\sffamily\bfseries, text=black!60] at (2.5, 0.65) {Lifting};
     
    \foreach \x in {5.5, 6.8, 8.2, 9.7} {
        \draw[arrow, <-] (\x, \rowfirst+0.15) -- (\x, \rowsecond-0.15);
    }
    \node[box, fill=pnasblue!10, minimum width=1.0cm, minimum height=0.6cm] at (7.5, 1.1) {$\mathcal{D}$};
    \node[font=\scriptsize\sffamily\bfseries, text=black!60] at (7.5, 0.65) {Projection};
     
    \foreach \x/\label in {\colfirst/0, \colsec/500, \colthird/1000} {
        \draw[gray!50] (\x, \rowfirst-0.1) -- (\x, \rowfirst-0.3);
        \node[font=\footnotesize\sffamily, text=black!60] at (\x, \rowfirst-0.65) {$t=\label$};
    }
    \node[font=\footnotesize\sffamily, text=black!60] at (2.5, \rowfirst-0.65) {$t=250$};
    \node[font=\footnotesize\sffamily, text=black!60] at (6.25, \rowfirst-0.65) {$t=750$};
     
    \draw[->, line width=1.2pt, pnasred!60, -{Stealth}] (0, \rowfirst+0.48) -- (0, \rowfirst-0.05);
    \node[font=\scriptsize\sffamily\bfseries, text=pnasred!60] at (-0.1, \rowfirst+0.58) {Initialize};
     
    \draw[->, line width=1.2pt, pnasred!60, bend left=20, -{Stealth}] (10.1, \rowsecond) to (10.6, \rowsecond-0.5);
    \node[font=\scriptsize\sffamily\bfseries, text=pnasred!60, anchor=west] at (10.65, \rowsecond-0.6) {\shortstack[l]{Final\\Prediction}};
     
    \node[box, fill=yellow!10, draw=yellow!40!orange, minimum width=3.2cm, font=\scriptsize, align=center] 
        at (2.5, \rowfirst-1.3) 
        {$\boldsymbol{U} = \sum_j \mathbf{p}(\latticevelocity_j) \F_j$\\ \textbf{Moment map}};
     
    \node[box, fill=yellow!10, draw=yellow!40!orange, minimum width=3.2cm, font=\scriptsize, align=center] 
        at (7.5, \rowfirst-1.3) 
        {$\Feqlattice = \phiNN(\boldsymbol{U};\theta)$\\ \textbf{\NN{} prediction}};
     
    \begin{scope}[shift={(10.8, \rowsecond+0.8)}]
        \draw[rounded corners=2pt, fill=white, draw=black!20, thick] 
            (-0.1, -0.7) rectangle (2.2, 0.6);
        \node[font=\footnotesize\bfseries\sffamily, anchor=west] at (0.0, 0.4) {Legend};
        
        \draw[timeline, pnasgreen] (0.1, 0.1) -- (0.5, 0.1);
        \node[font=\scriptsize\sffamily, anchor=west] at (0.6, 0.1) {Observed};
        
        \draw[timeline, pnasblue!60] (0.1, -0.2) -- (0.5, -0.2);
        \node[font=\scriptsize\sffamily, anchor=west] at (0.6, -0.2) {Predicted};
        
        \draw[timeline, gray!50, dashed] (0.1, -0.5) -- (0.5, -0.5);
        \node[font=\scriptsize\sffamily, anchor=west] at (0.6, -0.5) {Derived};
    \end{scope}
     
    \end{tikzpicture}
\caption{\textbf{Detailed interplay between macroscopic and kinetic representations.}
The diagram illustrates the complete training and testing workflow (extension of panel~a in \cref{alg:LBM_NN_algorithm}). The bottom timeline shows macroscopic observables $\boldsymbol{U}$ (density, momentum, energy density), while the top timeline represents the kinetic distribution $\mathbf{f}=[f(t,\mathbf{x},\mathbf{c}_j)]_{j=1}^Q$.
\textbf{Training phase (left):} Macroscopic data (solid green) is lifted to the kinetic level via operator $\mathcal{E}$, where equilibrium distributions and collision-streaming operators evolve the system. 
\textbf{Testing phase (right):} The trained neural network predicts equilibrium states (solid blue), which are then projected back via $\mathcal{D}$ to obtain macroscopic predictions. 
Dashed lines indicate derived quantities. Yellow boxes show key mathematical operations: moment projection maps kinetic to macroscopic variables, while the neural network learns the equilibrium mapping. The vertical dashed line separates observed training data from predictions.}
\label{fig:training_micro_macro}
\end{figure}

If the governing macroscopic equation \cref{eq:conservation} is known exactly, the lifting step requires only the initial and boundary conditions.  
In more general cases---when the governing dynamics are partially known or approximate---multiple lifting procedures may be employed, and their adequacy must be checked against the observed system behavior.

\subsection{Streaming as a Convolutional Neural Network}
\label{appendix:streaming_cnn}

The lattice Boltzmann streaming step shifts each population $\F_i$ along its discrete velocity $\latticevelocity_i$. 
For a general discrete velocity set $\Vcal = \{\latticevelocity_i\}_{i=1}^{Q}$, the streaming operator---for a dimensionless LB scheme---acts as
\begin{equation}\label{eq:streaming_general}
    \F_i(t+1,\, \bx)
    = \F_i^{\mathrm{coll}}(t,\, \bx - \latticevelocity_i),
    \qquad i = 1,\ldots,Q,
\end{equation}
where $\F_i^{\mathrm{coll}}$ denotes the post-collision distribution (see \cref{eq:LBM_collision}). 
This operator is linear and translational, meaning it can be represented exactly as a \emph{depthwise convolution} in a convolutional neural network (CNN).

\paragraph{General Convolutional Form.}
Let $\F^{\mathrm{coll}}(t,\bx) = (\F_1^{\mathrm{coll}}, \ldots, \F_Q^{\mathrm{coll}})^\top$ be the $Q$-channel tensor of post-collision distributions. 
For each discrete velocity direction $i$, we define a fixed convolution kernel $K^{(i)}$ such that
\begin{equation}\label{eq:kernel_general}
K^{(i)}(\ell) =
\begin{cases}
1, & \text{if } \ell = -\latticevelocity_i,\\[4pt]
0, & \text{otherwise},
\end{cases}
\qquad \ell \in \{-L_{\max}, \ldots, L_{\max}\}^d,
\end{equation}
where $L_{\max}$ is the largest lattice velocity magnitude in any spatial direction (for D2Q9, $L_{\max}=1$; for higher-order or thermal lattices, $L_{\max}=2$ or $3$). 
The streaming operator \cref{eq:streaming_general} is then equivalent to a discrete convolution:
\begin{equation}\label{eq:streaming_convolution}
\F_i(t+1,\, \bx)
= (K^{(i)} \ast \F_i^{\mathrm{coll}})(t,\, \bx)
= \sum_{\ell \in \{-L_{\max}, \ldots, L_{\max}\}^{\otimes^d}} 
K^{(i)}(\ell)\,
\F_i^{\mathrm{coll}}(t,\, \bx - \ell),
\end{equation}
which corresponds to a \emph{depthwise convolution} with $Q$ input and $Q$ output channels, where each kernel acts independently on one population channel. 
All kernel coefficients are fixed (\emph{non-trainable}) and contain a single nonzero entry located according to the discrete velocity $\latticevelocity_i$.

\paragraph{Example: D2Q9 Lattice.}
For the D2Q9 model ($Q=9$), the discrete velocities are
\begin{equation*}
\Vcal = \bigl\{ 
(0,0),\, (1,0),\, (-1,0),\, (0,1),\, (0,-1),\, 
(1,1),\, (-1,1),\, (-1,-1),\, (1,-1)
\bigr\}.
\end{equation*}
Here, $L_{\max}=1$, so the convolution kernels $K^{(i)}$ are defined on a $3\times3$ stencil. 
Each kernel has a single one-hot entry at the location $(-\latticevelocity_i)$, and zero elsewhere. 
Explicitly, the nine kernels correspond to center, right, left, up, down, and the four diagonal shifts. 

For example, for the population moving rightward ($\latticevelocity_2 = (1,0)$), the kernel is
\begin{equation*}
K^{(2)} = 
\begin{bmatrix}
0 & 0 & 0\\
1 & 0 & 0\\
0 & 0 & 0
\end{bmatrix},
\end{equation*}
while for the stationary population ($\latticevelocity_1 = (0,0)$), the kernel is simply the identity:
\begin{equation*}
K^{(1)} = 
\begin{bmatrix}
0 & 0 & 0\\
0 & 1 & 0\\
0 & 0 & 0
\end{bmatrix}.
\end{equation*}
All other kernels follow analogously by shifting the single nonzero entry according to $-\latticevelocity_i$.

The streaming operation for all populations can then be expressed compactly as
\begin{equation}\label{eq:streaming_cnn_compact}
\F(t+1,\, \cdot)
=
\mathrm{DWConv}\!\left(
\F^{\mathrm{coll}}(t,\, \cdot);\,
\{K^{(i)}\}_{i=1}^9
\right),
\end{equation}
where $\mathrm{DWConv}$ denotes a \emph{depthwise convolution} layer with $9$ input and output channels, kernel size $3\times3$, stride $1$, and $\texttt{groups}=9$. 
Boundary conditions (e.g., periodic, bounce-back, or open) are applied through the convolution padding rule.

\paragraph{Generalization to Higher-Order Lattices.}
For higher-dimensional or multi-speed lattices, the same structure applies with a larger kernel size.  
If the discrete velocity components satisfy $\latticevelocity_{i,\alpha} \in \{-L_{\max}, \ldots, L_{\max}\}$ for $\alpha=1,\ldots,d$, then the corresponding kernel $K^{(i)}$ occupies a $(2L_{\max}+1)^d$ stencil.  
For instance:
\begin{itemize}
    \item \textbf{D3Q27:} $L_{\max}=1$, kernel size $3\times3\times3$;
    \item \textbf{D2Q37 or thermal models:} $L_{\max}=3$, kernel size $7\times7$;
    \item \textbf{Generalized high-speed lattices:} kernel size $(2L_{\max}+1)^d$.
\end{itemize}
Each kernel remains one-hot, representing a deterministic lattice shift. 

Thus, for any discrete velocity model, the LBM streaming step can be written as a parameter-free, depthwise convolutional layer:
\begin{equation}
\F(t+1) = \mathrm{DWConv}\!\left(\F^{\mathrm{coll}}(t); \{K^{(i)}\}_{i=1}^{Q}\right),
\end{equation}
where the kernels $\{K^{(i)}\}$ encode the discrete velocity set $\Vcal$. 
This establishes a direct correspondence between the streaming operation in kinetic theory and a fixed, sparse convolutional mapping, while the collision operator provides the nonlinear local transformation that may be learned or modeled.

\subsection{\LBNN{} as an Architecture}\label{appx:LNN_architecture}
\begin{figure}[ht!]
\centering
\input{figs/appendix_D_neural_lattice}
\caption{\textbf{Neural lattice Boltzmann architecture.} 
\textbf{a.} The lattice Boltzmann update as an \emph{autoregressive neural network}. Collision and streaming act as sequential layers. The output $\F(t{+}1)$ recursively serves as the next input. 
\textbf{b.} The learnable equilibrium $\Feqlattice$. Small neural networks (gray/green/red icons) map macroscopic observables $\boldsymbol{U}$ and lattice velocities onto the moment space $\mathbb{M}$. The inner product of these features followed by an exponential renormalization ensures thermodynamic consistency.
\textbf{c.} The streaming operator as fixed, sparse depthwise convolution kernels $\{K_\ell\}$.}
\label{fig:lbm_architecture}
\end{figure}

The decomposition in panel~a of \cref{fig:lbm_architecture} 
highlights the equivalence between the lattice Boltzmann update and an autoregressive neural architecture: the streaming operator performs a linear shift (non-learnable convolution), and the collision operator performs a nonlinear local transformation that can be replaced or augmented by a neural network. 
In our formulation, the equilibrium mapping within the collision operator (\cref{fig:lbm_architecture}b) is parameterized by a neural network, while the streaming operation remains fixed. 
This modular view implies that, in principle, both operators can be learned end-to-end: replacing the streaming step by learnable convolutional kernels \(K_\ell\) (as explored in \cref{appendix:streaming}) yields a fully neural LBM. 
However, our results indicate that learning the collision operator alone suffices to capture non-equilibrium dynamics, while maintaining the physical stability ensured by the fixed streaming stencil.
Algorithm~\ref{alg:LBM_NN_algorithm} provides a detailed illustration of the collision and stream operators.

\vspace{0.5em}
\begin{figure}[H]
\centering
\sffamily 

\textbf{{\Large a} \hspace{ 0.5em }  \large Hybrid Temporal Propagation Scheme} \\[0.5em]
\begin{tikzpicture}[>=Latex, scale=0.95, transform shape]
    \tikzset{
        timeline/.style={line width=2pt, line cap=round},
        arrow/.style={->, line width=1.0pt, pnasgray!60, dashed},
        marker/.style={circle, fill, inner sep=2pt},
        labeltext/.style={font=\sffamily\bfseries, anchor=east},
        smalltext/.style={font=\sffamily\scriptsize, text=pnasgray}
    }
    
    \def\yKin{1.5} \def\yMac{0} \def\xStart{0} \def\xTrain{5} \def\xEnd{10}

    \node[labeltext, text=pnasblue] at (-0.5, \yKin) {Kinetic ($\F$)};
    \node[labeltext, text=pnasgreen!80!black] at (-0.5, \yMac) {Macroscopic ($\boldsymbol{U}$)};
    
    \draw[timeline, pnasblue!50, dashed] (\xStart, \yKin) -- (\xTrain, \yKin);
    \draw[timeline, pnasblue] (\xTrain, \yKin) -- (\xEnd, \yKin);
    \draw[timeline, pnasgreen] (\xStart, \yMac) -- (\xTrain, \yMac);
    \draw[timeline, pnasgreen, dashed] (\xTrain, \yMac) -- (\xEnd, \yMac);
    
    \foreach \x in {\xStart, \xTrain} \node[marker, pnasgreen] at (\x, \yMac) {};
    \foreach \x in {\xTrain, \xEnd} \node[marker, pnasblue] at (\x, \yKin) {};
    
    \draw[decorate, decoration={brace, amplitude=5pt, mirror}, line width=0.8pt, pnasgray] 
        (\xStart, \yMac-0.5) -- (\xTrain, \yMac-0.5);
    \node[font=\sffamily\small, text=pnasgray] at (2.5, \yMac-0.9) {Training Phase};
    
    \foreach \x/\step in {1/n, 2.5/n+1, 4/n+2} {
        \draw[arrow, ->] (\x, \yMac+0.15) -- (\x, \yKin-0.15);
        \node[smalltext] at (\x, \yMac-0.25) {$\step$};
    }
    \node[font=\large, fill=white, inner sep=1pt, text=pnasgray] at (2.5, 0.75) {$\mathcal{E}$ (Lifting)};

    \foreach \x/\step in {6/n+4, 7.5/n+5, 9/n+6} {
        \draw[arrow, <-] (\x, \yMac+0.15) -- (\x, \yKin-0.15);
        \node[smalltext] at (\x, \yMac-0.25) {$\step$};
    }
    \node[font=\large, fill=white, inner sep=1pt, text=pnasgray] at (7.5, 0.75) {$\mathcal{D}$ (Project)};

    \draw[->, line width=0.5pt, pnasgray] (\xStart, -1.3) -- (\xEnd+0.5, -1.3) node[right, font=\small] {Time $t$};
    \foreach \x/\label in {\xStart/t_0, \xTrain/t_{\mathrm{train}}, \xEnd/t_{\mathrm{end}}} {
        \draw[pnasgray] (\x, -1.3) -- (\x, -1.4);
        \node[font=\sffamily\small, text=pnasgray, anchor=north] at (\x, -1.4) {$\label$};
    }

    \node[anchor=north west, draw=gray!20, rounded corners, fill=white, inner sep=5pt] at (\xEnd+0.5, \yKin) {
        \begin{tikzpicture}[scale=0.8, baseline]
            \node[anchor=west, font=\sffamily\footnotesize] at (0,0.8) {\textbf{Legend}};
            \draw[line width=2pt, pnasgreen] (0,0.4) -- (0.5,0.4) node[right, black, font=\sffamily\footnotesize] {Observed};
            \draw[line width=2pt, pnasblue] (0,0) -- (0.5,0) node[right, black, font=\sffamily\footnotesize] {Predicted};
            \draw[line width=2pt, pnasgray!50, dashed] (0,-0.4) -- (0.5,-0.4) node[right, black, font=\sffamily\footnotesize] {Derived};
        \end{tikzpicture}
    };
\end{tikzpicture}

\vspace{1em}

\RestyleAlgo{plain} 
\SetAlgoNlRelativeSize{0}
\SetNlSty{textbf}{}{.}
\SetAlFnt{\sffamily\small} 
\SetKwSty{textbf}

\begin{minipage}[t]{0.98\textwidth}
    \textbf{{ \Large b} \hspace{ .5em } \large  Main Routine} \hfill \textbf{ {\Large c} \hspace{ .5em }  \large Subroutines} \\[2pt]
    \hrule height 1pt \vspace{2pt} 
    
    \begin{minipage}[t]{0.48\textwidth}
        \vspace{0pt} 
        \begin{algorithm}[H]
            \SetAlgoLined
            \DontPrintSemicolon
            \SetKwFunction{FLBNDEQ}{\textit{LB\_NeurDE}}
            \SetKwProg{Fn}{Function}{:}{}
            
            \Fn{\FLBNDEQ{$t, t_\mathrm{end}, \boldsymbol{U}(t, \bx)$}}{
                Initialize $\mathbf{F}^\mathrm{pred}_\mathrm{hist}, \mathbf{F}^\mathrm{eq}_\mathrm{hist}$\;
                \label{line:init_lists}
                $\F \leftarrow \texttt{init\_pop}(\boldsymbol{U}(t_0))$\tcp*{Lifting $\mathcal{E}$}
                \label{line:init_pop}
                \While{$t < t_{\mathrm{end}}$}{
                    $\Feqlattice \leftarrow \texttt{get\_NeurDE}(\F)$ \tcp*{See c.1}
                    \label{line:call_neurde}
                    $\F \leftarrow \texttt{splitting}(\F, \Feqlattice)$ \tcp*{See c.2}
                    \label{line:call_splitting}
                    Store $\F, \Feqlattice$ in history\;
                    $t\leftarrow t + 1$\;
                }
                \KwRet $\mathbf{F}^\mathrm{pred}_\mathrm{hist}, \mathbf{F}^\mathrm{eq}_\mathrm{hist}$
            }
        \end{algorithm}
    \end{minipage}%
    \hfill\vrule width 0.5pt\hfill 
    \begin{minipage}[t]{0.48\textwidth}
        \vspace{0pt}
        \textbf{{\Large{c}}.1 \NN{}} (\texttt{get\_NeurDE})\\
        \begin{algorithm}[H]
            \SetAlgoLined
            \DontPrintSemicolon
            \KwRet $\Feqlattice \leftarrow \phiNN(\mathcal{D}[\F]; \theta)$
            \label{line:neurde}
        \end{algorithm}
        \vspace{0.5em}
        \textbf{{\Large{c}}.2 Operator Splitting} (\texttt{splitting})\\
        \begin{algorithm}[H]
            \SetAlgoLined
            \DontPrintSemicolon
            $\F^{\mathrm{coll}} \leftarrow \F + \tau^{-1}[\Feqlattice-\F]$ \tcp*{Collision}
            \label{alg:collision}
            \KwRet $\F^{\mathrm{coll}}_i[\boldsymbol{x} - \latticevelocity_i]$ \tcp*{Stream}
            \label{line:streaming}
        \end{algorithm}
    \end{minipage}
    
    \vspace{2pt} \hrule height 1pt 
\end{minipage}

\caption{\textbf{Hybrid kinetic-macroscopic propagation.} 
\textbf{(a)} Kinetic distributions $\F$ (blue) and macroscopic observables $\boldsymbol{U}$ (green) evolve via lifting operator $\mathcal{E}$ and projection $\mathcal{D}$. The model trains on observed data ($t_0 \to t_{\mathrm{train}}$) before predicting future states ($t > t_{\mathrm{train}}$).
\textbf{(b)} The main routine alternates between Neural Network equilibrium prediction (Line \ref{line:call_neurde}) and operator splitting (Line \ref{line:call_splitting}).
\textbf{(c)} Subroutines detailing the projection-learning step (\texttt{get\_NeurDE}) and the collision-streaming step (\texttt{splitting}).}
\label{alg:LBM_NN_algorithm}
\end{figure}


\subsection{Conservative \NN{} and Machine-Precision Conservation}
\label{appx:conservative_neurde}

\NN{} in \cref{eq:levermore_closure_NN} is designed to approximate the equilibrium map while preserving the kinetic structure of the model.  
However, conservation is not automatic unless it is imposed explicitly.  
In practice, a standard training loss may drive the conservation residual to a small value, but not to zero, and such residuals may accumulate over long autoregressive rollouts.  
For this reason, it is useful to distinguish between an approximately conservative equilibrium predictor and a \emph{conservative \NN{}}, by which we mean a construction for which the conserved moments are satisfied up to machine precision at every evaluation point.

To state this precisely, let
\[
\Feqlattice=({\Feqlattice}_1,\dots,{\Feqlattice}_Q)^\top\in\mathbb{R}^Q
\]
denote an equilibrium population associated with the lattice velocities
\[
\{\latticevelocity_i\}_{i=1}^Q.
\]
Let
\[
\bm{U}=(U_1,\dots,U_{n_c})^\top
\]
be the vector of conserved macroscopic variables, where \(n_c\) is the number of conserved moments.  
For each conserved quantity, let \(m_\alpha(\latticevelocity_i)\) denote the corresponding collision invariant evaluated at \(\latticevelocity_i\), and define the constraint matrix
\[
\bm{C}\in\mathbb{R}^{n_c\times Q},
\qquad
C_{\alpha i}=m_\alpha(\latticevelocity_i).
\]
Then the exact conservation requirement is
\[
\bm{C}\Feqlattice=\bm{U}.
\]
A \emph{conservative \NN{}} is therefore any neural equilibrium construction for which this identity holds up to floating-point roundoff.

There are at least three constructive ways to enforce this property.

\paragraph{(a) Differentiable projection onto the conservation affine space.}
Let \(\tilde{\Feqlattice}(\bm{U})\) denote the raw output of \NN{}, with components
\[
{\tilde{\Feqlattice}}_i
=
\exp\!\bigl(\underbalpha(\bm{U})\cdot \underbvarphi(\latticevelocity_i)\bigr),
\qquad i=1,\dots,Q.
\]
Because this output is only approximate, it generally satisfies the
conservation constraints only up to a residual,
\[
\bm{C}\tilde{\Feqlattice}(\bm{U}) \approx \bm{U},
\]
rather than exactly. A simple way to restore conservation is to project
\(\tilde{\Feqlattice}\) onto the affine set
\[
\{\Feqlattice\in\mathbb{R}^Q:\bm{C}\Feqlattice=\bm{U}\}.
\]
Given a symmetric positive definite weight matrix
\(\bm{W}\in\mathbb{R}^{Q\times Q}\), define the corrected equilibrium by
\[
\Feqlattice
=
\argmin_{\bm{C}\Feqlattice=\bm{U}}
\frac12(\Feqlattice-\tilde{\Feqlattice})^\top
\bm{W}
(\Feqlattice-\tilde{\Feqlattice}).
\]
A standard Lagrange-multiplier calculation gives the explicit solution
\[
\Feqlattice
=
\tilde{\Feqlattice}
+
\bm{W}^{-1}\bm{C}^\top
\bigl(\bm{C}\bm{W}^{-1}\bm{C}^\top\bigr)^{-1}
\bigl(\bm{U}-\bm{C}\tilde{\Feqlattice}\bigr).
\]
By construction,
\[
\bm{C}\Feqlattice=\bm{U}.
\]
This correction is attractive because it is explicit, inexpensive, and
fully differentiable. Indeed, the matrix
\(\bm{C}\bm{W}^{-1}\bm{C}^\top\) has size only \(n_c\times n_c\), so the
additional cost is negligible compared with that of the full kinetic
rollout.

Moreover, when
\[
\bm{W}=\mathrm{diag}\!\bigl(1/{\tilde{\Feqlattice}}_i\bigr),
\]
the correction admits a natural \emph{local entropic interpretation}.
Since \(\tilde{\Feqlattice}\) is strictly positive and is generated by an
exponential ansatz, a natural notion of proximity is the relative
entropy, or Kullback--Leibler divergence, with respect to the reference
state \(\tilde{\Feqlattice}\)
\cite{banerjee2005clustering,levermore1996moment,hauck2008convex}. In
particular, one may consider the nonlinear constrained minimization
problem
\[
\min_{\bm{C}\Feqlattice=\bm{U}}
\sum_{i=1}^Q
\left(
{\Feqlattice}_i
\log\frac{{\Feqlattice}_i}{{\tilde{\Feqlattice}}_i}
-
{\Feqlattice}_i
+
{\tilde{\Feqlattice}}_i
\right),
\]
whose minimizer is the exact KL projection of
\(\tilde{\Feqlattice}\) onto the affine constraint set
\[
\{\Feqlattice\in\mathbb{R}^Q:\bm{C}\Feqlattice=\bm{U}\}.
\]
The weighted least-squares correction above is the corresponding
quadratic approximation. Indeed, expanding the KL functional around
\(\tilde{\Feqlattice}\) yields
\[
\sum_{i=1}^Q
\left(
{\Feqlattice}_i
\log\frac{{\Feqlattice}_i}{{\tilde{\Feqlattice}}_i}
-
{\Feqlattice}_i
+
{\tilde{\Feqlattice}}_i
\right)
=
\frac12
\sum_{i=1}^Q
\frac{\bigl({\Feqlattice}_i-{\tilde{\Feqlattice}}_i\bigr)^2}
{{\tilde{\Feqlattice}}_i}
+
\mathcal{O}\!\left(\|\Feqlattice-\tilde{\Feqlattice}\|^3\right),
\]
so choosing
\[
\bm{W}=\mathrm{diag}\!\bigl(1/{\tilde{\Feqlattice}}_i\bigr)
\]
recovers exactly the Hessian of the relative entropy at the reference
state $\tilde{\Feqlattice}$ \cite{dabak2002relations}. Hence this projection is not an
arbitrary additive adjustment in population space: it is the smallest
correction in the local information geometry induced by the exponential
model. Equivalently, perturbations of populations with small
\({\tilde{\Feqlattice}}_i\) are penalized more strongly, whereas larger
populations can absorb more of the conservation correction. Therefore,
when the residual \(\bm{U}-\bm{C}\tilde{\Feqlattice}\) is small, the
corrected equilibrium remains close to the raw exponential ansatz in the
entropic sense most natural for kinetic equilibria.

The main limitation is that the correction remains linear in population
space. Although \({\tilde{\Feqlattice}}_i>0\) for all \(i\) by
construction, the additive term
\[
\bm{W}^{-1}\bm{C}^\top
\bigl(\bm{C}\bm{W}^{-1}\bm{C}^\top\bigr)^{-1}
\bigl(\bm{U}-\bm{C}\tilde{\Feqlattice}\bigr)
\]
has no sign constraint, so some entries of \(\Feqlattice\) may become
negative after projection. When the conservation residual is already
small, such violations are often small in magnitude, but positivity is
not guaranteed structurally. Thus this construction enforces the
conserved moments exactly, but it does not preserve the positive cone
\[
\{\Feqlattice(t,\bx)\in\mathbb{R}^Q : {\Feqlattice}_i(t,\bx)>0
\text{ for all } i\}.
\]
If strict positivity is essential, one must instead use a genuinely
nonlinear entropic projection or a constrained exponential-family
construction, as in part~(c).

\paragraph{(b) Exact conservative parameterization in the null space of the conserved moments.}
A stronger approach is to impose conservation directly at the architectural level, so that the network never predicts in the conserved directions.  
Let
\[
r=\operatorname{rank}(\bm{C}),
\]
and let
\[
\bm{N}\in\mathbb{R}^{Q\times(Q-r)}
\]
be a matrix whose columns form a basis of \(\ker(\bm{C})\).  
Also choose any particular map
\[
{\Feqlattice}^{\mathrm{p}}(\bm{U})\in\mathbb{R}^Q
\qquad\text{such that}\qquad
\bm{C}{\Feqlattice}^{\mathrm{p}}(\bm{U})=\bm{U}.
\]
Then every conservative equilibrium can be written as
\[
\Feqlattice
=
{\Feqlattice}^{\mathrm{p}}(\bm{U})
+
\bm{N}\bm{z}(\bm{U}),
\]
where the neural network predicts only the reduced coordinates
\[
\bm{z}(\bm{U})\in\mathbb{R}^{Q-r}.
\]
Since \(\bm{C}\bm{N}=0\), one has
\[
\bm{C}\Feqlattice
=
\bm{C}{\Feqlattice}^{\mathrm{p}}(\bm{U})
+
\bm{C}\bm{N}\bm{z}(\bm{U})
=
\bm{U}.
\]
Thus conservation holds identically, independent of the network output.

This construction has two conceptual advantages.  
First, the network is not asked to learn the conserved moments from data; they are built in exactly.  
Second, the effective output dimension is reduced from \(Q\) to \(Q-r\), so the network learns only the higher-order, non-conserved degrees of freedom that determine the detailed shape of the equilibrium.  
In this sense, one obtains a genuinely \emph{conservative \NN{}} by design.  
Its limitation is similar to that of the projection method: exact conservation alone does not imply positivity unless the basis and parameterization are chosen with additional care.

\paragraph{(c) Constrained exponential-family construction.}
The most natural route, when positivity is also essential, is to enforce conservation directly inside the exponential ansatz.  
Let \(\{s_\beta(\latticevelocity_i)\}_{\beta=1}^{n_s}\) be additional basis functions describing the non-conserved part of the equilibrium, and let the neural network predict coefficients
\[
\bm{\eta}(\bm{U})=(\eta_1(\bm{U}),\dots,\eta_{n_s}(\bm{U})).
\]
For each \(i=1,\dots,Q\), define
\[
{\Feqlattice}_i(\bm{U};\bm{\lambda})
=
\exp\!\Bigl(
\sum_{\alpha=1}^{n_c}\lambda_\alpha\, m_\alpha(\latticevelocity_i)
+
\sum_{\beta=1}^{n_s}\eta_\beta(\bm{U})\, s_\beta(\latticevelocity_i)
\Bigr),
\]
and set
\[
\Feqlattice(\bm{U};\bm{\lambda})
=
\bigl(
{\Feqlattice}_1(\bm{U};\bm{\lambda}),
\dots,
{\Feqlattice}_Q(\bm{U};\bm{\lambda})
\bigr)^\top .
\]
The conserved multipliers \(\bm{\lambda}=(\lambda_1,\dots,\lambda_{n_c})\) are then chosen so that
\[
\bm{C}\Feqlattice(\bm{U};\bm{\lambda})=\bm{U},
\]
or, equivalently,
\[
\sum_{i=1}^Q m_\alpha(\latticevelocity_i)\,
{\Feqlattice}_i(\bm{U};\bm{\lambda})
=
U_\alpha,
\qquad \alpha=1,\dots,n_c.
\]
For fixed \(\bm{\eta}(\bm{U})\), this is a nonlinear system of dimension \(n_c\) for the conserved multipliers \(\bm{\lambda}\).  
Provided \(\bm{U}\) lies in the interior of the realizable moment set, this system can be solved locally by Newton iteration or a similar method.  
Because the equilibrium remains exponential, positivity is automatic, and because \(\bm{\lambda}\) is chosen so that the moment equations are satisfied exactly, conservation is enforced up to solver tolerance, which can be driven to machine precision.

This construction is therefore the closest to an entropic or maximum-entropy viewpoint: the network learns only the unconstrained part of the exponential family, while the conserved dual variables are fixed implicitly by the exact moment constraints.  
Its only additional cost is the local nonlinear solve, but this solve is very small and is often negligible compared with the total cost of the kinetic update.

In summary, there are three practical ways to obtain a conservative \NN{}.  
The projection construction in part~(a) is the simplest retrofit for an existing model; the null-space parameterization in part~(b) builds conservation directly into the architecture and reduces the number of learned degrees of freedom; and the constrained exponential construction in part~(c) is the most natural choice when one wants exact conservation together with strict positivity.  
All three routes produce machine-precision conservation, but they differ in how strongly they preserve positivity, thermodynamic structure, and architectural simplicity. The scalar Burgers, LWR, and Buckley--Leverett examples in \Cref{appx:additional_conservation_laws} illustrate how these conservative constructions transfer beyond the compressible-flow setting of the main text.

\paragraph{A two-state bridge to the entropy-defect theory.}
The conservative constructions above are expressed in terms of the physical
conserved state \(\bm{U}\in\mathbb{R}^{n_c}\), whereas the entropy-defect
analysis of \Cref{subsec:learning_aware} is written for the learned-moment
state \(\boldsymbol{\Xi}\in\mathbb{R}^p\). To connect these viewpoints without
identifying the two state spaces, we introduce
\[
\bm{U}_\F
:=
\bm{C}\F \in \mathbb{R}^{n_c},
\qquad
\boldsymbol{\Xi}_\F
:=
\sum_{i=1}^Q W_i\,\bm{\varphi}(\latticevelocity_i)\,\F_i \in \mathbb{R}^p,
\]
and interpret the raw reference equilibrium in this subsection as
\[
\tilde{\Feqlattice}(\boldsymbol{\Xi})
:=
\bigl(
\widehat{\phiNN}_1(\boldsymbol{\Xi}),\dots,\widehat{\phiNN}_Q(\boldsymbol{\Xi})
\bigr)^\top,
\]
where \(\widehat{\phiNN}_i\), \(\widehat{\bm{\alpha}}\), and the moment defect
\(m(\boldsymbol{\Xi})\) are exactly those of
\Cref{eq:raw_multiplier,eq:raw_learned_equilibrium,def:moment_defect}.

\begin{proposition}[Two-state conservative projection]
\label{prop:two_state_weighted_projection}
Let \(K_{\Xi}\subset\mathbb{R}^p\) be compact, and let
\(\mathcal A\subset\mathbb{R}^{n_c}\times K_{\Xi}\) be compact. Let
\[
\tilde{\Feqlattice}:K_{\Xi}\to\mathbb{R}_{>0}^Q
\]
be the raw reference equilibrium, and let
\[
\bm{W}:K_{\Xi}\to\mathbb{R}^{Q\times Q}
\]
be such that \(\bm{W}(\boldsymbol{\Xi})\) is symmetric positive definite for
every \(\boldsymbol{\Xi}\in K_{\Xi}\). Define
\[
\bm{A}(\boldsymbol{\Xi})
:=
\bm{C}\bm{W}(\boldsymbol{\Xi})^{-1}\bm{C}^\top,
\qquad
r_c(\bm{U},\boldsymbol{\Xi})
:=
\bm{U}-\bm{C}\tilde{\Feqlattice}(\boldsymbol{\Xi}),
\]
and assume
\[
\inf_{\boldsymbol{\Xi}\in K_{\Xi}}
\lambda_{\min}\!\bigl(\bm{A}(\boldsymbol{\Xi})\bigr)\ge \mu_0>0.
\]
For \((\bm{U},\boldsymbol{\Xi})\in\mathcal A\), define
\begin{equation}
\label{eq:two_state_weighted_projection}
\Feqlattice_{\mathrm c}(\bm{U},\boldsymbol{\Xi})
=
\tilde{\Feqlattice}(\boldsymbol{\Xi})
+
\bm{W}(\boldsymbol{\Xi})^{-1}\bm{C}^\top
\bm{A}(\boldsymbol{\Xi})^{-1}
r_c(\bm{U},\boldsymbol{\Xi}).
\end{equation}
Then, for every \((\bm{U},\boldsymbol{\Xi})\in\mathcal A\):
\begin{enumerate}
\item \textbf{Exact conservation:}
\[
\bm{C}\Feqlattice_{\mathrm c}(\bm{U},\boldsymbol{\Xi})=\bm{U}.
\]

\item \textbf{Optimality of the correction:}
\(\Feqlattice_{\mathrm c}(\bm{U},\boldsymbol{\Xi})\) is the unique minimizer of
\[
\min_{\bm{C}\Feqlattice=\bm{U}}
\frac12\,
\bigl(\Feqlattice-\tilde{\Feqlattice}(\boldsymbol{\Xi})\bigr)^\top
\bm{W}(\boldsymbol{\Xi})\,
\bigl(\Feqlattice-\tilde{\Feqlattice}(\boldsymbol{\Xi})\bigr).
\]

\item \textbf{Projection identity:}
if
\[
\Delta(\bm{U},\boldsymbol{\Xi})
:=
\Feqlattice_{\mathrm c}(\bm{U},\boldsymbol{\Xi})
-
\tilde{\Feqlattice}(\boldsymbol{\Xi}),
\qquad
|z|_{\bm{W}(\boldsymbol{\Xi})}^2
:=
z^\top \bm{W}(\boldsymbol{\Xi}) z,
\]
then
\begin{equation}
\label{eq:two_state_projection_identity}
|\Delta(\bm{U},\boldsymbol{\Xi})|_{\bm{W}(\boldsymbol{\Xi})}^2
=
r_c(\bm{U},\boldsymbol{\Xi})^\top
\bm{A}(\boldsymbol{\Xi})^{-1}
r_c(\bm{U},\boldsymbol{\Xi}).
\end{equation}
Consequently, if
\[
\sup_{(\bm{U},\boldsymbol{\Xi})\in\mathcal A}|r_c(\bm{U},\boldsymbol{\Xi})|
\le \delta_c,
\]
then
\begin{equation}
\label{eq:two_state_projection_bound}
|\Delta(\bm{U},\boldsymbol{\Xi})|_{\bm{W}(\boldsymbol{\Xi})}^2
\le
\mu_0^{-1}\delta_c^2
\qquad \forall\,(\bm{U},\boldsymbol{\Xi})\in\mathcal A.
\end{equation}
\end{enumerate}
\end{proposition}

\begin{proof}
Fix \((\bm{U},\boldsymbol{\Xi})\in\mathcal A\) and abbreviate
\[
\tilde{\Feqlattice}:=\tilde{\Feqlattice}(\boldsymbol{\Xi}),
\qquad
\bm{W}:=\bm{W}(\boldsymbol{\Xi}),
\qquad
\bm{A}:=\bm{A}(\boldsymbol{\Xi}),
\qquad
r:=r_c(\bm{U},\boldsymbol{\Xi}).
\]
Write the constrained minimization problem in the increment variable
\[
\Delta=\Feqlattice-\tilde{\Feqlattice},
\]
so that it becomes
\[
\min_{\bm{C}\Delta=r}\frac12\,\Delta^\top\bm{W}\Delta.
\]
Its Lagrangian is
\[
\mathcal L(\Delta,\lambda)
=
\frac12\,\Delta^\top\bm{W}\Delta
-
\lambda^\top(\bm{C}\Delta-r).
\]
Stationarity with respect to \(\Delta\) gives
\[
\bm{W}\Delta-\bm{C}^\top\lambda=0
\quad\Longrightarrow\quad
\Delta=\bm{W}^{-1}\bm{C}^\top\lambda.
\]
Imposing the constraint \(\bm{C}\Delta=r\) yields
\[
\bm{C}\bm{W}^{-1}\bm{C}^\top\lambda
=
\bm{A}\lambda
=
r.
\]
Since \(\bm{A}\) is symmetric positive definite by assumption, it is invertible,
and
\[
\lambda=\bm{A}^{-1}r,
\qquad
\Delta=\bm{W}^{-1}\bm{C}^\top\bm{A}^{-1}r.
\]
Therefore
\[
\Feqlattice
=
\tilde{\Feqlattice}+\Delta
=
\tilde{\Feqlattice}+\bm{W}^{-1}\bm{C}^\top\bm{A}^{-1}r,
\]
which is exactly \eqref{eq:two_state_weighted_projection}. Because the
objective is strictly convex, this minimizer is unique. This proves item~2.

For item~1,
\[
\bm{C}\Feqlattice_{\mathrm c}(\bm{U},\boldsymbol{\Xi})
=
\bm{C}\tilde{\Feqlattice}
+
\bm{C}\bm{W}^{-1}\bm{C}^\top\bm{A}^{-1}r
=
\bm{C}\tilde{\Feqlattice}
+
\bm{A}\bm{A}^{-1}r
=
\bm{C}\tilde{\Feqlattice}+r
=
\bm{U}.
\]

For item~3,
\[
\Delta=\bm{W}^{-1}\bm{C}^\top\bm{A}^{-1}r,
\]
hence
\[
|\Delta|_{\bm{W}}^2
=
\Delta^\top\bm{W}\Delta
=
r^\top\bm{A}^{-1}\bm{C}\bm{W}^{-1}\bm{W}\bm{W}^{-1}\bm{C}^\top\bm{A}^{-1}r
=
r^\top\bm{A}^{-1}\bm{C}\bm{W}^{-1}\bm{C}^\top\bm{A}^{-1}r
=
r^\top\bm{A}^{-1}r,
\]
which is \eqref{eq:two_state_projection_identity}. Since
\[
\bm{A}(\boldsymbol{\Xi})\succeq \mu_0 I
\quad\Longrightarrow\quad
\bm{A}(\boldsymbol{\Xi})^{-1}\preceq \mu_0^{-1}I,
\]
we obtain
\[
|\Delta(\bm{U},\boldsymbol{\Xi})|_{\bm{W}(\boldsymbol{\Xi})}^2
=
r_c(\bm{U},\boldsymbol{\Xi})^\top
\bm{A}(\boldsymbol{\Xi})^{-1}
r_c(\bm{U},\boldsymbol{\Xi})
\le
\mu_0^{-1}|r_c(\bm{U},\boldsymbol{\Xi})|^2
\le
\mu_0^{-1}\delta_c^2.
\qedhere
\]
\end{proof}

\begin{corollary}[Smallness criterion for positivity]
\label{cor:two_state_projection_positivity}
Under the assumptions of \Cref{prop:two_state_weighted_projection}, assume in
addition that \(\bm{W}\) is continuous on \(K_{\Xi}\), and define
\[
\underline w
:=
\inf_{\boldsymbol{\Xi}\in K_{\Xi}}
\lambda_{\min}\!\bigl(\bm{W}(\boldsymbol{\Xi})\bigr)>0.
\]
Then
\[
\|\Delta(\bm{U},\boldsymbol{\Xi})\|_\infty
\le
\underline w^{-1/2}\,|\Delta(\bm{U},\boldsymbol{\Xi})|_{\bm{W}(\boldsymbol{\Xi})}
\le
\underline w^{-1/2}\mu_0^{-1/2}\delta_c
\qquad
\forall\,(\bm{U},\boldsymbol{\Xi})\in\mathcal A.
\]
Hence if
\[
\inf_{\boldsymbol{\Xi}\in K_{\Xi}}
\min_{1\le i\le Q}\tilde{\Feqlattice}_i(\boldsymbol{\Xi})
>
\underline w^{-1/2}\mu_0^{-1/2}\delta_c,
\]
then
\[
\Feqlattice_{\mathrm c}(\bm{U},\boldsymbol{\Xi})\in\mathbb{R}_{>0}^Q
\qquad \forall\,(\bm{U},\boldsymbol{\Xi})\in\mathcal A.
\]
\end{corollary}

\begin{proof}
Since \(\bm{W}(\boldsymbol{\Xi})\succeq \underline w I\) on \(K_{\Xi}\),
\[
|z|_{\bm{W}(\boldsymbol{\Xi})}^2
\ge
\underline w\,\|z\|_2^2
\ge
\underline w\,\|z\|_\infty^2,
\]
hence
\[
\|z\|_\infty \le \underline w^{-1/2}|z|_{\bm{W}(\boldsymbol{\Xi})}.
\]
Apply this with \(z=\Delta(\bm{U},\boldsymbol{\Xi})\) and use
\eqref{eq:two_state_projection_bound}. The positivity conclusion follows from
\[
\bigl(\Feqlattice_{\mathrm c}(\bm{U},\boldsymbol{\Xi})\bigr)_i
=
\tilde{\Feqlattice}_i(\boldsymbol{\Xi})+\Delta_i(\bm{U},\boldsymbol{\Xi})
\ge
\tilde{\Feqlattice}_i(\boldsymbol{\Xi})-\|\Delta(\bm{U},\boldsymbol{\Xi})\|_\infty.
\qedhere
\]
\end{proof}

\begin{remark}[Connection with the KL-quadratic correction]
\label{rem:two_state_KL_metric}
The proposition above recovers the local KL-quadratic correction discussed in
part~(a) when
\[
\bm{W}(\boldsymbol{\Xi})
=
\operatorname{diag}\!\bigl(1/\tilde{\Feqlattice}_i(\boldsymbol{\Xi})\bigr).
\]
\end{remark}

The next result combines this two-state conservative correction with the raw
entropy-defect identity of \Cref{thm:discrete_defect,cor:discrete_approx_H}. It
certifies a corrected single-population BGK-type NeurDE closure, not the full
two-population hybrid compressible implementation used in Sec.~4. The hybrid
two-population extension for that compressible scaffold is stated later in
\Cref{thm:hybrid_two_population_neurde,cor:hybrid_thermal_specialization}.

\begin{theorem}[Two-state conservative NeurDE with controlled entropy defect]
\label{thm:two_state_conservative_neurde_entropy}
Assume the hypotheses of \Cref{prop:two_state_weighted_projection} with
\[
\tilde{\Feqlattice}(\boldsymbol{\Xi})
=
\bigl(
\widehat{\phiNN}_1(\boldsymbol{\Xi}),\dots,\widehat{\phiNN}_Q(\boldsymbol{\Xi})
\bigr)^\top.
\]
Assume in addition that \(\bm{W}\) is continuous on \(K_{\Xi}\), and define
\[
A_\alpha
:=
\sup_{\boldsymbol{\Xi}\in K_{\Xi}}
\|\widehat{\bm{\alpha}}(\boldsymbol{\Xi})\|_\ast
<\infty,
\qquad
\delta_m
:=
\sup_{\boldsymbol{\Xi}\in K_{\Xi}}
\|m(\boldsymbol{\Xi})\|
<\infty,
\]
\[
\delta_c
:=
\sup_{(\bm{U},\boldsymbol{\Xi})\in\mathcal A}
|r_c(\bm{U},\boldsymbol{\Xi})|
<\infty,
\qquad
\underline w
:=
\inf_{\boldsymbol{\Xi}\in K_{\Xi}}
\lambda_{\min}\!\bigl(\bm{W}(\boldsymbol{\Xi})\bigr)
>0,
\]
\[
W_{(1)} := \sum_{i=1}^Q W_i,
\qquad
\delta := \max\{\delta_c,\delta_m\}.
\]

Let \(\F_i=\F_i(t,\bx)>0\) be a sufficiently regular solution of the corrected
BGK system
\begin{equation}
\label{eq:two_state_corrected_bgk}
\partial_t \F_i + \latticevelocity_i\cdot\nabla_{\bx}\F_i
=
\frac{1}{\tau}\Bigl(
\bigl(\Feqlattice_{\mathrm c}(\bm{U}_\F,\boldsymbol{\Xi}_\F)\bigr)_i-\F_i
\Bigr),
\qquad i=1,\dots,Q,
\end{equation}
where
\[
\bm{U}_\F(t,\bx)
:=
\bm{C}\F(t,\bx)\in\mathbb{R}^{n_c},
\qquad
\boldsymbol{\Xi}_\F(t,\bx)
:=
\sum_{i=1}^Q W_i\,\bm{\varphi}(\latticevelocity_i)\,\F_i(t,\bx)\in\mathbb{R}^p.
\]
Assume periodic boundary conditions, or boundary conditions under which the
transport entropy flux vanishes after integration, and assume that
\[
\bigl(\bm{U}_\F(t,\bx),\boldsymbol{\Xi}_\F(t,\bx)\bigr)\in\mathcal A
\qquad \text{for all }(t,\bx)
\]
under consideration. Assume moreover that there exist constants
\(0<\eta_-<\eta_+\) such that
\[
\eta_-
\le
\F_i(t,\bx),\,
\tilde{\Feqlattice}_i(\boldsymbol{\Xi}_\F(t,\bx)),\,
\bigl(\Feqlattice_{\mathrm c}(\bm{U}_\F(t,\bx),\boldsymbol{\Xi}_\F(t,\bx))\bigr)_i
\le
\eta_+
\]
for all \(i,t,\bx\).

Then:
\begin{enumerate}
\item \textbf{Exact conservation and optimality.}
For every \((\bm{U},\boldsymbol{\Xi})\in\mathcal A\),
\[
\bm{C}\Feqlattice_{\mathrm c}(\bm{U},\boldsymbol{\Xi})=\bm{U},
\]
and \(\Feqlattice_{\mathrm c}(\bm{U},\boldsymbol{\Xi})\) is the unique
minimizer of
\[
\min_{\bm{C}\Feqlattice=\bm{U}}
\frac12\,
\bigl(\Feqlattice-\tilde{\Feqlattice}(\boldsymbol{\Xi})\bigr)^\top
\bm{W}(\boldsymbol{\Xi})\,
\bigl(\Feqlattice-\tilde{\Feqlattice}(\boldsymbol{\Xi})\bigr).
\]

\item \textbf{Size of the correction.}
If
\[
\Delta(\bm{U},\boldsymbol{\Xi})
:=
\Feqlattice_{\mathrm c}(\bm{U},\boldsymbol{\Xi})
-
\tilde{\Feqlattice}(\boldsymbol{\Xi}),
\qquad
|z|_{\bm{W}(\boldsymbol{\Xi})}^2
:=
z^\top \bm{W}(\boldsymbol{\Xi})z,
\]
then
\begin{equation}
\label{eq:two_state_theorem_projection_bound}
|\Delta(\bm{U},\boldsymbol{\Xi})|_{\bm{W}(\boldsymbol{\Xi})}^2
=
r_c(\bm{U},\boldsymbol{\Xi})^\top
\bm{A}(\boldsymbol{\Xi})^{-1}
r_c(\bm{U},\boldsymbol{\Xi})
\le
\mu_0^{-1}\delta_c^2
\qquad
\forall\,(\bm{U},\boldsymbol{\Xi})\in\mathcal A.
\end{equation}

\item \textbf{Near-entropy law.}
Define the discrete kinetic entropy and weighted dissipation functional by
\[
\mathscr{H}[\F](t)
:=
\int_\Omega
\sum_{i=1}^Q
W_i\bigl(\F_i\log\F_i-\F_i\bigr)\,d\bx,
\]
\[
\mathscr D(\F\|\mathbf H)
:=
\int_\Omega
\sum_{i=1}^Q
W_i\,(\F_i-\mathbf H_i)\log\frac{\F_i}{\mathbf H_i}
\,d\bx.
\]
Then
\begin{equation}
\label{eq:two_state_entropy_ineq}
\frac{d}{dt}\mathscr{H}[\F](t)
\le
-\frac{1}{\tau}\,
\mathscr D\!\bigl(
\F\,\|\,\Feqlattice_{\mathrm c}(\bm{U}_\F,\boldsymbol{\Xi}_\F)
\bigr)
+
\frac{C_K}{\tau}\,\delta,
\end{equation}
where
\begin{equation}
\label{eq:two_state_LK}
L_K
:=
\sup_{a,z\in[\eta_-,\eta_+]}
\left|
\log z + 1 - \frac{a}{z}
\right|
\le
\max\!\bigl\{|\log\eta_-|,|\log\eta_+|\bigr\}
+
1
+
\frac{\eta_+}{\eta_-},
\end{equation}
and one may take
\begin{equation}
\label{eq:two_state_CK}
C_K
=
|\Omega|
\left(
A_\alpha
+
L_K\,W_{(1)}\,(\underline w\,\mu_0)^{-1/2}
\right).
\end{equation}
In particular, if \(\delta_c=\delta_m=0\), then
\begin{equation}
\label{eq:two_state_exact_H}
\frac{d}{dt}\mathscr{H}[\F](t)
=
-\frac{1}{\tau}\,
\mathscr D\!\bigl(
\F\,\|\,\Feqlattice_{\mathrm c}(\bm{U}_\F,\boldsymbol{\Xi}_\F)
\bigr)
\le 0.
\end{equation}
\end{enumerate}
\end{theorem}

\begin{proof}
Items~1 and~2 are exactly the conclusions of
\Cref{prop:two_state_weighted_projection}.

For item~3, write
\[
h(t,\bx)
:=
\Feqlattice_{\mathrm c}(\bm{U}_\F(t,\bx),\boldsymbol{\Xi}_\F(t,\bx)),
\qquad
\tilde h(t,\bx)
:=
\tilde{\Feqlattice}(\boldsymbol{\Xi}_\F(t,\bx)),
\]
and
\[
\Delta(t,\bx):=h(t,\bx)-\tilde h(t,\bx).
\]

Multiply \eqref{eq:two_state_corrected_bgk} by \(W_i\log\F_i\), sum over \(i\),
and integrate over \(\Omega\). The transport term gives only a spatial
divergence,
\[
W_i\log\F_i\,\latticevelocity_i\cdot\nabla_{\bx}\F_i
=
W_i\nabla_{\bx}\cdot\Bigl(
\latticevelocity_i(\F_i\log\F_i-\F_i)
\Bigr),
\]
which vanishes after integration under the assumed boundary conditions.
Therefore
\[
\frac{d}{dt}\mathscr{H}[\F](t)
=
\frac{1}{\tau}
\int_\Omega
\sum_{i=1}^Q
W_i\,(h_i-\F_i)\log\F_i\,d\bx.
\]

Add and subtract \(\log h_i\):
\[
\frac{d}{dt}\mathscr{H}[\F](t)
=
-\frac{1}{\tau}
\int_\Omega
\sum_{i=1}^Q
W_i\,(\F_i-h_i)\log\frac{\F_i}{h_i}\,d\bx
+
\frac{1}{\tau}
\int_\Omega
\sum_{i=1}^Q
W_i\,(h_i-\F_i)\log h_i\,d\bx.
\]
Hence
\begin{equation}
\label{eq:two_state_entropy_split}
\frac{d}{dt}\mathscr{H}[\F](t)
=
-\frac{1}{\tau}\,\mathscr D(\F\|h)
+
\frac{1}{\tau}\int_\Omega E_h(\bx,t)\,d\bx,
\end{equation}
where
\[
E_h(\bx,t)
:=
\sum_{i=1}^Q W_i\,(h_i-\F_i)\log h_i.
\]

Define the corresponding raw term
\[
E_{\tilde h}(\bx,t)
:=
\sum_{i=1}^Q W_i\,(\tilde h_i-\F_i)\log \tilde h_i.
\]
Since
\[
\log \tilde h_i
=
\widehat{\bm{\alpha}}(\boldsymbol{\Xi}_\F)\cdot
\bm{\varphi}(\latticevelocity_i),
\]
we obtain
\begin{align*}
E_{\tilde h}
&=
\sum_{i=1}^Q
W_i\,(\tilde h_i-\F_i)\,
\widehat{\bm{\alpha}}(\boldsymbol{\Xi}_\F)\cdot
\bm{\varphi}(\latticevelocity_i)
\\
&=
\widehat{\bm{\alpha}}(\boldsymbol{\Xi}_\F)\cdot
\left(
\sum_{i=1}^Q
W_i\,\bm{\varphi}(\latticevelocity_i)\,\tilde h_i
-
\sum_{i=1}^Q
W_i\,\bm{\varphi}(\latticevelocity_i)\,\F_i
\right)
\\
&=
\widehat{\bm{\alpha}}(\boldsymbol{\Xi}_\F)\cdot
\left(
\sum_{i=1}^Q
W_i\,\bm{\varphi}(\latticevelocity_i)\,
\tilde{\Feqlattice}_i(\boldsymbol{\Xi}_\F)
-
\boldsymbol{\Xi}_\F
\right)
\\
&=
\widehat{\bm{\alpha}}(\boldsymbol{\Xi}_\F)\cdot m(\boldsymbol{\Xi}_\F).
\end{align*}
Therefore, pointwise,
\begin{equation}
\label{eq:two_state_raw_defect_bound}
|E_{\tilde h}(\bx,t)|
\le
\|\widehat{\bm{\alpha}}(\boldsymbol{\Xi}_\F(\bx,t))\|_\ast\,
\|m(\boldsymbol{\Xi}_\F(\bx,t))\|
\le
A_\alpha\,\delta_m
\le
A_\alpha\,\delta.
\end{equation}

It remains to estimate \(E_h-E_{\tilde h}\). Define
\[
F(a,z):=(z-a)\log z,
\qquad a,z>0.
\]
Then
\[
E_h-E_{\tilde h}
=
\sum_{i=1}^Q
W_i\Bigl(F(\F_i,h_i)-F(\F_i,\tilde h_i)\Bigr).
\]
A direct computation gives
\[
\partial_z F(a,z)=\log z + 1 - \frac{a}{z}.
\]
By the definition of \(L_K\) and the mean value theorem,
\[
|F(a,z_1)-F(a,z_2)|
\le
L_K\,|z_1-z_2|
\qquad
\forall\,a,z_1,z_2\in[\eta_-,\eta_+].
\]
Hence
\[
|E_h-E_{\tilde h}|
\le
L_K\sum_{i=1}^Q W_i\,|\Delta_i|
\le
L_K\,W_{(1)}\,\|\Delta\|_\infty.
\]
Since \(\bm{W}(\boldsymbol{\Xi}_\F)\succeq \underline w I\),
\[
\|\Delta\|_\infty
\le
\underline w^{-1/2}|\Delta|_{\bm{W}(\boldsymbol{\Xi}_\F)}.
\]
By item~2, applied to the admissible pair
\((\bm{U}_\F(\bx,t),\boldsymbol{\Xi}_\F(\bx,t))\in\mathcal A\),
\[
|\Delta|_{\bm{W}(\boldsymbol{\Xi}_\F)}
\le
\mu_0^{-1/2}\delta_c
\le
\mu_0^{-1/2}\delta.
\]
Therefore
\begin{equation}
\label{eq:two_state_correction_defect_bound}
|E_h-E_{\tilde h}|
\le
L_K\,W_{(1)}\,(\underline w\,\mu_0)^{-1/2}\,\delta.
\end{equation}

Combining \eqref{eq:two_state_raw_defect_bound} and
\eqref{eq:two_state_correction_defect_bound}, we obtain
\[
E_h(\bx,t)
\le
\left(
A_\alpha
+
L_K\,W_{(1)}\,(\underline w\,\mu_0)^{-1/2}
\right)\delta.
\]
Insert this bound into \eqref{eq:two_state_entropy_split} and integrate over
\(\Omega\):
\[
\frac{d}{dt}\mathscr{H}[\F](t)
\le
-\frac{1}{\tau}\,\mathscr D(\F\|h)
+
\frac{|\Omega|}{\tau}
\left(
A_\alpha
+
L_K\,W_{(1)}\,(\underline w\,\mu_0)^{-1/2}
\right)\delta.
\]
Since \(h=\Feqlattice_{\mathrm c}(\bm{U}_\F,\boldsymbol{\Xi}_\F)\), this is
exactly \eqref{eq:two_state_entropy_ineq} with \eqref{eq:two_state_CK}.

Finally, if \(\delta_c=\delta_m=0\), then
\[
r_c(\bm{U},\boldsymbol{\Xi})=0
\qquad\text{for all }(\bm{U},\boldsymbol{\Xi})\in\mathcal A,
\]
and
\[
m(\boldsymbol{\Xi})=0
\qquad\text{for all }\boldsymbol{\Xi}\in K_{\Xi}.
\]
The first identity implies \(\Delta=0\) on the admissible pairs, hence
\(h=\tilde h\). The second implies \(E_{\tilde h}=0\). Therefore \(E_h=0\),
and \eqref{eq:two_state_exact_H} follows.
\end{proof}

\begin{remark}[Why this theorem is stated in two-state form]
\label{rem:two_state_why}
The theorem keeps the physical conserved block
\(\bm{U}_\F=\bm{C}\F\) separate from the learned-moment state
\(\boldsymbol{\Xi}_\F=\sum_{i=1}^Q W_i\,\bm{\varphi}(\latticevelocity_i)\F_i\).
A one-state theorem would require an additional anchoring hypothesis relating
the learned latent moment sector to the physical conserved sector.
\end{remark}

\subsection{Training Scheme}\label{appx:training}
As we described in the main text, \cref{sec:NN_LBNN}, the training of \LBNN{} consists of two consecutive stages:  
(1) a \emph{pretraining stage}, where \NN{} is learned independently to predict equilibria from macroscopic observables, and  
(2) a \emph{joint training stage}, where \NN{} is integrated into the kinetic solver $(\phiS\phiC)$---forming what we called \LBNN{}---and optimized through multi-step trajectory prediction.  
This hierarchical procedure is intended to give a locally plausible closure before it is exposed to autoregressive rollout errors during joint training.

\subsubsection{Pretraining Procedure}\label{appx:pretraining}

Pretraining provides a useful initialization for \NN{} by fitting the learned equilibrium distributions before they are embedded into the kinetic dynamics.  
A randomly initialized model often leads to unstable behavior during joint training; pretraining alleviates this issue.

Given a dataset of macroscopic--equilibrium pairs
\(\{(\boldsymbol{U}_n, \mathcal{E}_n)\}_{n=1}^{N}\), where
\(\mathcal{E}_n\) denotes the equilibrium target used by the chosen
realization (\(\Geqlattice_n\) for the compressible energy-channel
experiments and \(\Feqlattice_n\) for the single-population scalar-law
experiments), we train \NN{} using supervised regression:
\begin{equation}
    \min_{\theta} 
    \sum_{n=1}^{N} 
    \ell\!\left(\phiNN(\boldsymbol{U}_n;\theta),\,\mathcal{E}_n\right),
\end{equation}
where $\ell$ denotes an appropriate loss function (e.g., an $\Lp$ norm).  
The targets may be obtained either from high-resolution kinetic simulations or by solving Levermore's exponential closure problem \cite{levermore1996moment, latt2020efficient}.  

Alternatively, synthetic training data can be generated from the exponential ansatz
\[
\F_n = \left[\exp\!\left(\sum_{k=1}^p \balpha_{n,k}\,\bvarphi_k(\latticevelocity_i)\right)\right]_{i=1}^Q,
\]
where $\bvarphi_k$ are chosen moment basis functions (e.g., Eulerian or Levermore bases; cf.~\cref{appendix:M_space}), $\alpha_{n,k}\sim\mathbb{P}$ are sampled coefficients, and the associated observables are computed as $\boldsymbol{U}_n = \mathcal{D}[\F_n]$.

\subsubsection{Training Algorithm}\label{appx:training_alg}
In the second stage, the pretrained \NN{} is embedded into the kinetic update operator $\phiS\phiC$---yielding the full \LBNN{} scheme---and optimized end-to-end by comparing predicted trajectories against high-fidelity reference simulations.  
The training objective minimizes discrepancies in the transported populations and in the corresponding equilibrium targets over prediction horizons of length \(N_r\). In the single-population notation this reads
\[
L = 
\sum_{r=t}^{t+N_r}
\alpha\,\ell\!\left(\F(r), \F^{\mathrm{pred}}(r)\right)
+ (1-\alpha)\,\ell\!\left(\Feqlattice(r), {\Feqlattice}^{\mathrm{pred}}(r)\right),
\]
where $\alpha\in[0,1]$ controls the trade-off between mesoscopic and equilibrium-level losses. For the hybrid compressible experiments, the same objective is applied to the learned energy-channel equilibrium \(\Geqlattice\) and its transported population \(\G\).

The complete algorithm is outlined below (see \cref{alg:LBM_NN_algorithm_full_training}).
\begin{algorithm}[ht!]
    \SetAlgoLined
    \SetNlSty{textbf}{\{}{\}}
    \KwData{$\tau,\, \{\latticevelocity_i\}_{i=1}^Q, \,\{W_i\}_{i=1}^Q, \alpha\in [0,1],\, \eta,\, N_r,\, \alpha' = 1- \alpha$ \tcp*{Set parameters, boundary conditions} 
    $\theta \leftarrow \texttt{\textit{pretraining}} (\theta\sim \mathrm{random})$\tcp*{Perform pre-training as previously described}
    $\big \{\{\F(0, \bx), \Feqlattice(0, \bx)\}, \ldots,  \{\F(t_{N_{\mathrm{train}}}, \bx), \Feqlattice(t_{N_{\mathrm{train}}}, \bx)\}\big\}$ \tcp*{Load training data}} \label{alg:training_load_equilibrium}
    \For{ $0\le$ epoch $\le N$}{
        \For{$0 \le t \le  t_{N_{\mathrm{train}}} $}{ 
            $t_\mathrm{end} = \min(t_{N_\text{train}}, t + N_r)$ \;
            $\mathbf{F}^\mathrm{pred}_\mathrm{hist},\mathbf{F}^\mathrm{eq}_\mathrm{hist}  = \texttt{\textit{LB\_\NN}}(t, t_{\text{end}}, \mathcal{D}[\F](t, \bx))$\tcp*{Make temporal prediction}
            $L \leftarrow \sum_{r=t}^{t_\text{end}} \alpha \ell\left (\F(r, \bx), \mathbf{F}^\mathrm{pred}_\mathrm{hist}[r,\bx] \right) + \alpha' \ell \left(\Feqlattice(r, \bx), \mathbf{F}^\mathrm{eq}_\mathrm{hist} [r,\bx] \right)$\label{alg:loss_accumulate}\tcp*{Accumulate loss}
            $\theta \leftarrow (\theta - \eta \partial_{\theta} L)$\tcp*{Update the parameters}
                        $ t \leftarrow t+1$\;
        }
        $epoch\leftarrow epoch+1$;
    }
    \KwOut{ $\phiNN(\cdot ; \theta)$}
    \caption{Training \LBNN{} $(\phiS \phiCNN)$}
    \label{alg:LBM_NN_algorithm_full_training}
\end{algorithm}

\vspace{0.5 em}

\paragraph{Training and optimization.}
The gradient-based optimization of \LBNN{} proceeds through \emph{backpropagation through time} (BPTT) applied to the unrolled kinetic solver.  
Each training trajectory of length $N_r$ corresponds to an unrolled sequence of discrete-time operators $(\phiS \phiC)$, with the neural closure $\phiNN$ (\NN{}) embedded within the collision step.  
The loss in line~\ref{alg:loss_accumulate} of \cref{alg:LBM_NN_algorithm_full_training} is differentiated through this sequence using automatic differentiation, propagating gradients across both streaming and collision operations.  
This structure exposes the learned closure to rollout error while keeping the physical symmetries and lattice topology in the fixed transport operator.

Similar unrolled training strategies have been employed in differentiable flow simulations \cite{kochkov2021machine, fan2024differentiable} and more general differentiable PDE solvers~\cite{brandstetter2022message, lippe2023pde}.  
The hyperparameter $N_r$ determines the prediction horizon (i.e., the number of unrolled steps in BPTT).  
Its choice depends on both the dataset characteristics and the initialization of the network parameters $\theta$ in \NN{}.  
In our empirical evaluation (\cref{sec:numerical}), $N_r$ ranged from $10$ to $25$, beyond which improvements became marginal while computational cost increased substantially.

The loss weighting parameter $\alpha\in[0,1]$ controls the balance between kinetic and equilibrium-level objectives.  
For the tests in \cref{sec:numerical}, setting $\alpha=0$ already yielded stable and accurate training.  
Regularization can further improve rollout behavior: for instance, by imposing soft constraints on the conservation of moments in the predicted equilibrium.  
When higher-order moment closures are introduced~\cite{latt2020efficient}, relaxing them via inequality-based penalties (e.g.,
\(\relu(\bra\bvarphi(\latticevelocity_i)\phiNN\ket_\mu - \bra\bvarphi(\velocity)f^{\mathrm{MB}}\ket)\))
can alleviate non-existence issues for feasible equilibrium solutions~\cite{pavan2011general}.  
Further implementation and optimization details are provided in \cref{sec:numerical,appendix:optimization}.


\newpage

\section{Empirical Evaluation}
\label{appendix:experiments_details}
This appendix provides supplementary details for the empirical evaluation presented in the main text (\cref{sec:numerical}). It focuses on methods and parameters applicable to all test cases; experiment-specific configurations for the Sod shock tube and the flow over a cylinder are given in \cref{appx:sod_shock} and \cref{appendix:cylinder}, respectively.

We first elaborate on the underlying physics model: the two-population thermal lattice Boltzmann formulation (\cref{appendix:two_pop}) and the polynomial expansion of the $\F$ and $\G$ population equilibria (\cref{appendix:poly_F_pop,appendix:poly_g_pop}). The $\G$ equilibrium is used in the subsonic Sod (\cref{section:Sod_cases}) as a benchmark following \cite{karlin2013consistent,saadat2019lattice}. As discussed in the main text, the polynomial approximation deteriorates at high Mach numbers, becoming numerically unstable in transonic and supersonic regimes. We then describe the neural network implementation and training details, including the architecture (\cref{sec:specific_arch}), optimization parameters (\cref{appendix:arch_params,appendix:optimization}), and dataset composition and computational setup (\cref{appendix:train_time_dataset}).

\subsection{Two-Population Thermal LB Scheme}\label{appendix:two_pop}
\label{appendix:two_populations}

This subsection expands on the consistent two-population model introduced in \cref{sec:specific_arch} (see \cite{karlin2013consistent, saadat2019lattice}). In particular, we specify the temperature-dependent weights $W_i$, the quasi-equilibrium state defined in \cref{eq:LBM-energy}, and in \cref{appendix:poly_F_pop,appendix:poly_g_pop} we describe the polynomial expansion of the $\F$ and $\G$ populations.\ 

We recall from the main text that the relaxation parameters $\tau_1$ and $\tau_2$ in \cref{eq:LBM} correspond to the dynamic viscosity $\mu$ and thermal conductivity $\kappa$, which are defined as,
\begin{equation}\label{eq:viscosity}
    \mu= (\tau_1-\tfrac{1}{2}) \rho \temperature,
\end{equation}
and $$\kappa= \Cp\,(\tau_2-\tfrac{1}{2}) \rho \temperature,$$
where $\Cv$ and $\Cp$ are the specific heats at constant volume and pressure (\cref{sec:specific_arch}), respectively, and all quantities are expressed in lattice units. The Prandtl number is then $\mathrm{Pr} = \Cp\mu/\kappa$.

As a consequence of the variable Prandtl number, the intermediate quasi-equilibrium state $\G_i^\ast$ is introduced in \cref{eq:LBM-energy} (cf.~\cite{karlin2013consistent}) and is defined as
\begin{align}\label{eq:quasiequilibrium}
    \G_i^\ast \eqdef \Geqlattice_i +  \tfrac{2}{\temperature}  W_i\,{ \Velocity_\beta \left( \pressuretensor_{\alpha, \beta} - \pressuretensor_{\alpha, \beta}^{\eq} \right) \latticevelocity_{i, \alpha} },
\end{align}
where the $W_i$ are temperature-dependent weights \cite{saadat2019lattice}, given by: 
\begin{equation}\label{eq:temperature_weights} 
    W_i = \prod_{\alpha} W_{i,\alpha}, \qquad W_{\pm 1} = \tfrac{1}{2}\temperature, \quad W_0 = 1 - \temperature \qquad \qquad \alpha\in \{x,y\} \text{ or } \alpha\in\{x,y,z\}.
\end{equation}
The pressure tensors $\pressuretensor$ and $\pressuretensor^{\eq}$ in \cref{eq:quasiequilibrium} are higher-order moments, defined as $\pressuretensor = \bra \latticevelocity \osym \latticevelocity \F\ket$ and $\pressuretensor^{\eq} = \bra \latticevelocity \osym \latticevelocity \Feqlattice \ket$.\footnote{Following previous work~\cite{chen1998lattice}, Greek indices are used for vector components, while Roman indices are used to label discrete lattices vector. The Einstein's convention summation is applied to Greek indices (e.g., \cref{eq:quasiequilibrium}). } \ 

Other relevant higher-order moments include: the heat flux vector, defined as $\heatflux = \bra \latticevelocity  \G \ket $, and  its equilibrium counterpart $ \heatflux^{\eq} = \bra \latticevelocity \Geqlattice \ket$; and the contracted fourth-order moment tensor $\boldsymbol{R} = \bra \latticevelocity \osym \latticevelocity \G\ket$, with its equilibrium form being $\boldsymbol{R}^{\mathrm{eq}} = \bra \latticevelocity \osym \latticevelocity \Geqlattice\ket$. See, e.g., \cref{appendix:higher-order_moments}.

\subsubsection{Hybrid Conservative Entropy Structure}
\label{appendix:hybrid_two_pop_theory}

\begin{remark}[Absorbed-population convention for the hybrid model]
\label{rem:hybrid_two_pop_convention}
In the compressible solver we work with the weight-absorbed populations,
consistent with \Cref{eq:macroscopical_LBM_2_pop,rem:weight_absorption}. The
conserved physical state is written in the manuscript convention as
\[
\bm{U}=(\rho,\rho\Velocity,\rho E)\in\mathbb{R}^{d+2},
\]
while the absorbed energy population carries the channel
\[
\rho = \sum_{i=1}^Q \F_i,
\qquad
\rho\Velocity = \sum_{i=1}^Q \latticevelocity_i\,\F_i,
\qquad
\sum_{i=1}^Q \G_i = 2\rho E.
\]
On any admissible set with \(\rho>0\), this convention is equivalent to the
solver variables \((\rho,\Velocity,\temperature)\) in
\Cref{eq:macroscopical_LBM_2_pop}.
Bare sums below are understood in this absorbed convention; the explicit
temperature-dependent weights \(W_i\) remain only in the thermal
quasi-equilibrium correction \Cref{eq:quasiequilibrium}.
\end{remark}

The next proposition gives the exact conservative retrofit for the learned
energy population alone, while leaving the analytic \(\F\)-population
unchanged.

\begin{proposition}[Energy-channel conservative correction for the hybrid model]
\label{prop:hybrid_energy_projection}
Let \(K_{\mathrm{phys}}\subset\mathbb{R}^{d+2}\) be compact, and let
\[
\bm{U}=(\rho,\rho\Velocity,\rho E)\in K_{\mathrm{phys}}.
\]
Let
\[
\widetilde{\Geqlattice}_\theta:K_{\mathrm{phys}}\to\mathbb{R}_{>0}^Q
\]
be a raw learned energy equilibrium. Define
\[
C_g \eqdef (1,\dots,1)\in\mathbb{R}^{1\times Q},
\qquad
C_g\G = 2\rho E.
\]
Let
\[
\bm{W}_g:K_{\mathrm{phys}}\to\mathbb{R}^{Q\times Q}
\]
be continuous and such that \(\bm{W}_g(\bm{U})\) is symmetric positive
definite for every \(\bm{U}\in K_{\mathrm{phys}}\). Define
\[
A_g(\bm{U})
\eqdef
C_g\,\bm{W}_g(\bm{U})^{-1}C_g^\top \in \mathbb{R},
\qquad
r_g^{\mathrm{phys}}(\bm{U})
\eqdef
2\rho E-C_g\widetilde{\Geqlattice}_\theta(\bm{U}),
\]
and assume
\[
\inf_{\bm{U}\in K_{\mathrm{phys}}}A_g(\bm{U})\ge \mu_0>0.
\]
Define the corrected learned energy equilibrium by
\begin{equation}
\label{eq:hybrid_energy_projection}
\Geqlattice_{\mathrm c}(\bm{U})
=
\widetilde{\Geqlattice}_\theta(\bm{U})
+
\bm{W}_g(\bm{U})^{-1}C_g^\top A_g(\bm{U})^{-1}
r_g^{\mathrm{phys}}(\bm{U}).
\end{equation}
Then, for every \(\bm{U}\in K_{\mathrm{phys}}\),
\begin{enumerate}
\item \textbf{Exact energy consistency:}
\[
C_g\,\Geqlattice_{\mathrm c}(\bm{U})=2\rho E.
\]

\item \textbf{Optimality of the correction:}
\(\Geqlattice_{\mathrm c}(\bm{U})\) is the unique minimizer of
\[
\min_{C_g\G=2\rho E}
\frac12\,
\bigl(\G-\widetilde{\Geqlattice}_\theta(\bm{U})\bigr)^\top
\bm{W}_g(\bm{U})\,
\bigl(\G-\widetilde{\Geqlattice}_\theta(\bm{U})\bigr).
\]

\item \textbf{Projection identity:}
if
\[
\Delta_g(\bm{U})
\eqdef
\Geqlattice_{\mathrm c}(\bm{U})-\widetilde{\Geqlattice}_\theta(\bm{U}),
\qquad
|z|_{\bm{W}_g(\bm{U})}^2 \eqdef z^\top \bm{W}_g(\bm{U})z,
\]
then
\begin{equation}
\label{eq:hybrid_energy_projection_identity}
|\Delta_g(\bm{U})|_{\bm{W}_g(\bm{U})}^2
=
r_g^{\mathrm{phys}}(\bm{U})\,A_g(\bm{U})^{-1}\,r_g^{\mathrm{phys}}(\bm{U}).
\end{equation}
Consequently, if
\[
\sup_{\bm{U}\in K_{\mathrm{phys}}}|r_g^{\mathrm{phys}}(\bm{U})|
\le
\delta_E,
\]
then
\begin{equation}
\label{eq:hybrid_energy_projection_bound}
|\Delta_g(\bm{U})|_{\bm{W}_g(\bm{U})}^2
\le
\mu_0^{-1}\delta_E^2
\qquad \forall\,\bm{U}\in K_{\mathrm{phys}}.
\end{equation}
\end{enumerate}
\end{proposition}

\begin{proof}
Fix \(\bm{U}\in K_{\mathrm{phys}}\) and abbreviate
\[
\widetilde{\Geqlattice}\eqdef \widetilde{\Geqlattice}_\theta(\bm{U}),
\qquad
\bm{W}\eqdef \bm{W}_g(\bm{U}),
\qquad
A\eqdef A_g(\bm{U}),
\qquad
r\eqdef r_g^{\mathrm{phys}}(\bm{U}).
\]
Write the constrained minimization problem in the increment variable
\[
\Delta=\G-\widetilde{\Geqlattice},
\]
so that it becomes
\[
\min_{C_g\Delta=r}\frac12\,\Delta^\top\bm{W}\Delta.
\]
Its Lagrangian is
\[
\mathcal L(\Delta,\lambda)
=
\frac12\,\Delta^\top\bm{W}\Delta
-
\lambda(C_g\Delta-r).
\]
Stationarity with respect to \(\Delta\) gives
\[
\bm{W}\Delta-C_g^\top\lambda=0
\quad\Longrightarrow\quad
\Delta=\bm{W}^{-1}C_g^\top\lambda.
\]
Imposing the constraint \(C_g\Delta=r\) yields
\[
C_g\bm{W}^{-1}C_g^\top\lambda
=
A\lambda
=
r.
\]
Since \(A>0\), it is invertible and
\[
\lambda=A^{-1}r,
\qquad
\Delta=\bm{W}^{-1}C_g^\top A^{-1}r.
\]
Therefore
\[
\G
=
\widetilde{\Geqlattice}+\bm{W}^{-1}C_g^\top A^{-1}r,
\]
which is exactly \eqref{eq:hybrid_energy_projection}. Since the objective is
strictly convex, the minimizer is unique. This proves item~2.

For item~1,
\[
C_g\Geqlattice_{\mathrm c}(\bm{U})
=
C_g\widetilde{\Geqlattice}
+
C_g\bm{W}^{-1}C_g^\top A^{-1}r
=
C_g\widetilde{\Geqlattice}+AA^{-1}r
=
C_g\widetilde{\Geqlattice}+r
=
2\rho E.
\]

For item~3,
\[
\Delta=\bm{W}^{-1}C_g^\top A^{-1}r,
\]
hence
\[
|\Delta|_{\bm{W}}^2
=
\Delta^\top\bm{W}\Delta
=
r\,A^{-1}C_g\bm{W}^{-1}\bm{W}\bm{W}^{-1}C_g^\top A^{-1}r
=
r\,A^{-1}C_g\bm{W}^{-1}C_g^\top A^{-1}r
=
r\,A^{-1}r,
\]
which is \eqref{eq:hybrid_energy_projection_identity}. Since
\[
A_g(\bm{U})\ge \mu_0
\quad\Longrightarrow\quad
A_g(\bm{U})^{-1}\le \mu_0^{-1},
\]
we obtain
\[
|\Delta_g(\bm{U})|_{\bm{W}_g(\bm{U})}^2
=
r_g^{\mathrm{phys}}(\bm{U})\,A_g(\bm{U})^{-1}\,r_g^{\mathrm{phys}}(\bm{U})
\le
\mu_0^{-1}|r_g^{\mathrm{phys}}(\bm{U})|^2
\le
\mu_0^{-1}\delta_E^2.
\qedhere
\]
\end{proof}

The next theorem states the corresponding formal continuous-time entropy
estimate for the hybrid two-population scaffold underlying \Cref{eq:LBM}.

\begin{theorem}[Hybrid two-population conservative NeurDE with controlled energy-channel entropy defect]
\label{thm:hybrid_two_population_neurde}
Let \(K_{\mathrm{phys}}\subset\mathbb{R}^{d+2}\) be compact, and let
\[
\bm{U}=(\rho,\rho\Velocity,\rho E)\in K_{\mathrm{phys}}.
\]
Assume:
\begin{enumerate}
\item \textbf{Analytic momentum equilibrium.}
There is an analytic equilibrium
\[
\mathbf{f}^{\mathrm{an}}:K_{\mathrm{phys}}\to\mathbb{R}_{>0}^Q
\]
such that, with
\[
C_f\in\mathbb{R}^{(d+1)\times Q},
\qquad
C_f\F=(\rho,\rho\Velocity),
\]
one has
\[
C_f\mathbf{f}^{\mathrm{an}}(\bm{U})=(\rho,\rho\Velocity)
\qquad \forall\,\bm{U}\in K_{\mathrm{phys}}.
\]

\item \textbf{Anchored learned energy basis.}
There is a learned energy-channel basis
\[
\bm{\psi}_g(\latticevelocity_i)
=
\bigl(1,\bm{\chi}_g(\latticevelocity_i)\bigr)\in\mathbb{R}^{p_g},
\qquad i=1,\dots,Q,
\]
and the associated moment map
\[
M_g[\mathbf h]
\eqdef
\sum_{i=1}^Q \bm{\psi}_g(\latticevelocity_i)\,\mathbf h_i
\in \mathbb{R}^{p_g}
\]
has first component equal to the absorbed energy-channel moment,
\[
(M_g[\mathbf h])_1 = C_g\mathbf h = \sum_{i=1}^Q \mathbf h_i.
\]

\item \textbf{Raw learned energy equilibrium.}
The raw learned energy equilibrium has exponential form
\[
\widetilde{\Geqlattice}_{\theta,i}(\bm{U})
=
\exp\!\bigl(
\widehat{\bm{\alpha}}_g(\bm{U})\cdot \bm{\psi}_g(\latticevelocity_i)
\bigr),
\qquad
\bm{U}\in K_{\mathrm{phys}},
\quad i=1,\dots,Q,
\]
with continuous multiplier map
\[
\widehat{\bm{\alpha}}_g:K_{\mathrm{phys}}\to\mathbb{R}^{p_g}.
\]

\item \textbf{Completed target in the learned energy-moment space.}
There is a prescribed completion map
\[
\widetilde{\boldsymbol{\Xi}}_g^\star:K_{\mathrm{phys}}\to\mathbb{R}^{p_g},
\qquad
\widetilde{\boldsymbol{\Xi}}_g^\star(\bm{U})
=
\bigl(2\rho E,\bm{\Lambda}_g^\star(\bm{U})\bigr),
\]
whose first component is the absorbed energy-channel constraint. A natural choice is
\[
\widetilde{\boldsymbol{\Xi}}_g^\star(\bm{U})
=
M_g[\mathbf{g}^\star(\bm{U})],
\]
with \(\mathbf{g}^\star\) the reference energy equilibrium from
\Cref{eq:Geq_Levermore}.

\item \textbf{Completed residual bound.}
The completed residual
\[
r_g^{\mathrm{comp}}(\bm{U})
\eqdef
M_g[\widetilde{\Geqlattice}_\theta(\bm{U})]
-
\widetilde{\boldsymbol{\Xi}}_g^\star(\bm{U})
\in \mathbb{R}^{p_g}
\]
satisfies
\[
\sup_{\bm{U}\in K_{\mathrm{phys}}}\|r_g^{\mathrm{comp}}(\bm{U})\|
\le
\delta_{\mathrm{comp}}.
\]

\item \textbf{Energy correction.}
The corrected energy equilibrium \(\Geqlattice_{\mathrm c}\) is given by
\eqref{eq:hybrid_energy_projection}, and the hypotheses of
\Cref{prop:hybrid_energy_projection} hold. In particular,
\[
\sup_{\bm{U}\in K_{\mathrm{phys}}}|r_g^{\mathrm{phys}}(\bm{U})|
\le
\delta_E,
\qquad
r_g^{\mathrm{phys}}(\bm{U})
=
2\rho E-C_g\widetilde{\Geqlattice}_\theta(\bm{U}).
\]

\item \textbf{Anchoring mismatch along the rollout.}
Along the evolution under consideration, the mismatch
\[
\ell_g(\F,\G)
\eqdef
\widetilde{\boldsymbol{\Xi}}_g^\star(\bm{U}_{\F,\G})-M_g[\G]
\]
satisfies
\[
\|\ell_g(\F(t,\bx),\G(t,\bx))\|
\le
\delta_{\mathrm{anc}}
\qquad \text{for all }(t,\bx),
\]
where
\[
\bm{U}_{\F,\G}(t,\bx)
\eqdef
\left(
\sum_{i=1}^Q \F_i(t,\bx),
\sum_{i=1}^Q \latticevelocity_i\,\F_i(t,\bx),
\frac12\sum_{i=1}^Q \G_i(t,\bx)
\right).
\]

\item \textbf{Uniform bounds.}
There exist constants \(0<\eta_-<\eta_+\) such that
\[
\eta_-
\le
\G_i(t,\bx),\,
\widetilde{\Geqlattice}_{\theta,i}(\bm{U}_{\F,\G}(t,\bx)),\,
\bigl(\Geqlattice_{\mathrm c}(\bm{U}_{\F,\G}(t,\bx))\bigr)_i
\le
\eta_+
\]
for all \(i,t,\bx\). Also define
\[
A_{\alpha,g}
\eqdef
\sup_{\bm{U}\in K_{\mathrm{phys}}}\|\widehat{\bm{\alpha}}_g(\bm{U})\|_\ast
<\infty,
\qquad
\underline\lambda_g
\eqdef
\inf_{\bm{U}\in K_{\mathrm{phys}}}
\lambda_{\min}\!\bigl(\bm{W}_g(\bm{U})\bigr)
>0.
\]

\item \textbf{Energy-preserving analytic coupling.}
Let
\[
S_\ast(\F,\G;\bm{U})\in\mathbb{R}^Q
\]
be an analytic coupling term satisfying
\[
C_g\,S_\ast(\F,\G;\bm{U})=0
\]
for every admissible \((\F,\G,\bm{U})\).
\end{enumerate}

Consider the formal continuous-time analogue of \Cref{eq:LBM},
\begin{align}
\partial_t \F_i + \latticevelocity_i\cdot\nabla_{\bx}\F_i
&=
\frac{1}{\tau_1}\Bigl(
\mathbf{f}_i^{\mathrm{an}}(\bm{U}_{\F,\G})-\F_i
\Bigr),
\label{eq:hybrid_f_eqn}
\\
\partial_t \G_i + \latticevelocity_i\cdot\nabla_{\bx}\G_i
&=
\frac{1}{\tau_2}\Bigl(
\bigl(\Geqlattice_{\mathrm c}(\bm{U}_{\F,\G})\bigr)_i-\G_i
\Bigr)
+
S_{\ast,i}(\F,\G;\bm{U}_{\F,\G}),
\label{eq:hybrid_g_eqn}
\end{align}
with periodic boundary conditions, or boundary conditions under which the
transport entropy flux vanishes after integration. Assume also that
\[
\bm{U}_{\F,\G}(t,\bx)\in K_{\mathrm{phys}}
\qquad \text{for all }(t,\bx)
\]
under consideration.

Then:
\begin{enumerate}
\item \textbf{Exact conservation of the hybrid collision operator.}
The collision terms in \eqref{eq:hybrid_f_eqn}--\eqref{eq:hybrid_g_eqn}
preserve mass, momentum, and total energy exactly:
\[
C_f\Bigl(\mathbf{f}^{\mathrm{an}}(\bm{U}_{\F,\G})-\F\Bigr)=0,
\qquad
C_g\Bigl(\Geqlattice_{\mathrm c}(\bm{U}_{\F,\G})-\G\Bigr)=0,
\qquad
C_gS_\ast(\F,\G;\bm{U}_{\F,\G})=0.
\]
Hence, after spatial integration under the assumed boundary conditions, the
total invariants
\(\int_\Omega \rho\,d\bx\),
\(\int_\Omega \rho\Velocity\,d\bx\), and
\(\int_\Omega \rho E\,d\bx\) are conserved.

\item \textbf{Optimality and size of the energy correction.}
For every \(\bm{U}\in K_{\mathrm{phys}}\),
\(\Geqlattice_{\mathrm c}(\bm{U})\) is the unique minimizer of
\[
\min_{C_g\G=2\rho E}
\frac12\,
\bigl(\G-\widetilde{\Geqlattice}_\theta(\bm{U})\bigr)^\top
\bm{W}_g(\bm{U})\,
\bigl(\G-\widetilde{\Geqlattice}_\theta(\bm{U})\bigr),
\]
and its correction
\[
\Delta_g(\bm{U})
\eqdef
\Geqlattice_{\mathrm c}(\bm{U})-\widetilde{\Geqlattice}_\theta(\bm{U})
\]
satisfies
\[
|\Delta_g(\bm{U})|_{\bm{W}_g(\bm{U})}^2
=
r_g^{\mathrm{phys}}(\bm{U})\,A_g(\bm{U})^{-1}\,r_g^{\mathrm{phys}}(\bm{U})
\le
\mu_0^{-1}\delta_E^2.
\]

\item \textbf{Energy-channel near-entropy law.}
Define
\[
\mathscr{H}_g[\G](t)
\eqdef
\int_\Omega
\sum_{i=1}^Q \bigl(\G_i\log\G_i-\G_i\bigr)\,d\bx,
\]
and
\[
\mathscr D_g(\G\|\mathbf H)
\eqdef
\int_\Omega
\sum_{i=1}^Q
(\G_i-\mathbf H_i)\log\frac{\G_i}{\mathbf H_i}\,d\bx.
\]
Then
\begin{equation}
\label{eq:hybrid_energy_entropy_ineq}
\frac{d}{dt}\mathscr{H}_g[\G](t)
\le
-\frac{1}{\tau_2}\,
\mathscr D_g\!\bigl(\G\,\|\,\Geqlattice_{\mathrm c}(\bm{U}_{\F,\G})\bigr)
+
\frac{|\Omega|}{\tau_2}
\left(
A_{\alpha,g}\bigl(\delta_{\mathrm{comp}}+\delta_{\mathrm{anc}}\bigr)
+
L_g\sqrt{\frac{Q}{\underline\lambda_g\,\mu_0}}\,\delta_E
\right)
+
\bigl|\mathscr R_\ast(t)\bigr|,
\end{equation}
where
\[
\mathscr R_\ast(t)
\eqdef
\int_\Omega
\sum_{i=1}^Q
S_{\ast,i}(\F,\G;\bm{U}_{\F,\G})\,\log\G_i\,d\bx,
\]
and
\begin{equation}
\label{eq:hybrid_Lg}
L_g
\eqdef
\sup_{a,z\in[\eta_-,\eta_+]}
\left|\log z + 1 - \frac{a}{z}\right|
\le
\max\!\bigl\{|\log\eta_-|,|\log\eta_+|\bigr\}
+
1
+
\frac{\eta_+}{\eta_-}.
\end{equation}
\end{enumerate}
\end{theorem}

\begin{proof}
Item~1 follows directly from the moment constraints. For the
\(\F\)-equation,
\[
C_f\Bigl(\mathbf{f}^{\mathrm{an}}(\bm{U}_{\F,\G})-\F\Bigr)
=
C_f\mathbf{f}^{\mathrm{an}}(\bm{U}_{\F,\G})-C_f\F
=
(\rho,\rho\Velocity)-(\rho,\rho\Velocity)
=
0.
\]
For the \(\G\)-equation,
\[
C_g\Bigl(\Geqlattice_{\mathrm c}(\bm{U}_{\F,\G})-\G\Bigr)
=
2\rho E-C_g\G
=
2\rho E-2\rho E
=
0
\]
by \Cref{prop:hybrid_energy_projection}, and
\(C_gS_\ast(\F,\G;\bm{U}_{\F,\G})=0\) by assumption. Integrating the transport
equations against the corresponding conserved rows gives only spatial
divergence terms, which vanish under the stated boundary conditions.

Item~2 is exactly \Cref{prop:hybrid_energy_projection}.

It remains to prove item~3. Write, for brevity,
\[
h(t,\bx)\eqdef \Geqlattice_{\mathrm c}(\bm{U}_{\F,\G}(t,\bx)),
\qquad
\widetilde h(t,\bx)\eqdef \widetilde{\Geqlattice}_\theta(\bm{U}_{\F,\G}(t,\bx)).
\]
Since
\[
\frac{d}{dz}(z\log z-z)=\log z,
\]
multiply \eqref{eq:hybrid_g_eqn} by \(\log \G_i\), sum over \(i\), and
integrate over \(\Omega\). The transport term contributes only a divergence,
which vanishes after integration. Hence
\[
\frac{d}{dt}\mathscr{H}_g[\G](t)
=
\frac{1}{\tau_2}
\int_\Omega
\sum_{i=1}^Q
(h_i-\G_i)\log\G_i\,d\bx
+
\mathscr R_\ast(t).
\]
Add and subtract \(\log h_i\):
\[
\frac{d}{dt}\mathscr{H}_g[\G](t)
=
-\frac{1}{\tau_2}
\int_\Omega
\sum_{i=1}^Q
(\G_i-h_i)\log\frac{\G_i}{h_i}\,d\bx
+
\frac{1}{\tau_2}
\int_\Omega E_c(\bx,t)\,d\bx
+
\mathscr R_\ast(t),
\]
where
\[
E_c(\bx,t)\eqdef \sum_{i=1}^Q (h_i-\G_i)\log h_i.
\]
Thus
\begin{equation}
\label{eq:hybrid_entropy_split}
\frac{d}{dt}\mathscr{H}_g[\G](t)
=
-\frac{1}{\tau_2}\mathscr D_g(\G\|h)
+
\frac{1}{\tau_2}\int_\Omega E_c(\bx,t)\,d\bx
+
\mathscr R_\ast(t).
\end{equation}

We now compare \(E_c\) to the corresponding raw term
\[
E_{\widetilde h}(\bx,t)\eqdef \sum_{i=1}^Q (\widetilde h_i-\G_i)\log \widetilde h_i.
\]
Since
\[
\log \widetilde{\Geqlattice}_{\theta,i}(\bm{U}_{\F,\G})
=
\widehat{\bm{\alpha}}_g(\bm{U}_{\F,\G})\cdot\bm{\psi}_g(\latticevelocity_i),
\]
we obtain
\begin{align*}
E_{\widetilde h}
&=
\sum_{i=1}^Q
(\widetilde h_i-\G_i)\,
\widehat{\bm{\alpha}}_g(\bm{U}_{\F,\G})\cdot\bm{\psi}_g(\latticevelocity_i)
\\
&=
\widehat{\bm{\alpha}}_g(\bm{U}_{\F,\G})\cdot
\left(
\sum_{i=1}^Q \bm{\psi}_g(\latticevelocity_i)\,\widetilde h_i
-
\sum_{i=1}^Q \bm{\psi}_g(\latticevelocity_i)\,\G_i
\right)
\\
&=
\widehat{\bm{\alpha}}_g(\bm{U}_{\F,\G})\cdot
\left(
M_g[\widetilde{\Geqlattice}_\theta(\bm{U}_{\F,\G})]
-
\widetilde{\boldsymbol{\Xi}}_g^\star(\bm{U}_{\F,\G})
\right)
\\
&\quad
+
\widehat{\bm{\alpha}}_g(\bm{U}_{\F,\G})\cdot
\left(
\widetilde{\boldsymbol{\Xi}}_g^\star(\bm{U}_{\F,\G})-M_g[\G]
\right)
\\
&=
\widehat{\bm{\alpha}}_g(\bm{U}_{\F,\G})\cdot
r_g^{\mathrm{comp}}(\bm{U}_{\F,\G})
+
\widehat{\bm{\alpha}}_g(\bm{U}_{\F,\G})\cdot
\ell_g(\F,\G).
\end{align*}
Therefore, pointwise,
\begin{equation}
\label{eq:hybrid_raw_term_bound}
|E_{\widetilde h}(\bx,t)|
\le
A_{\alpha,g}\bigl(\delta_{\mathrm{comp}}+\delta_{\mathrm{anc}}\bigr).
\end{equation}

It remains to estimate \(E_c-E_{\widetilde h}\). Define
\[
F(a,z)\eqdef (z-a)\log z,
\qquad a,z>0.
\]
Then
\[
E_c-E_{\widetilde h}
=
\sum_{i=1}^Q
\Bigl(F(\G_i,h_i)-F(\G_i,\widetilde h_i)\Bigr).
\]
Since
\[
\partial_zF(a,z)=\log z + 1 - \frac{a}{z},
\]
the definition \eqref{eq:hybrid_Lg} and the mean-value theorem yield
\[
|F(a,z_1)-F(a,z_2)|
\le
L_g\,|z_1-z_2|
\qquad
\forall\,a,z_1,z_2\in[\eta_-,\eta_+].
\]
Hence
\[
|E_c-E_{\widetilde h}|
\le
L_g\,\|h-\widetilde h\|_1
=
L_g\,\|\Delta_g(\bm{U}_{\F,\G})\|_1.
\]
Now
\[
\|\Delta_g(\bm{U}_{\F,\G})\|_1
\le
\sqrt{Q}\,\|\Delta_g(\bm{U}_{\F,\G})\|_2
\le
\sqrt{\frac{Q}{\underline\lambda_g}}\,
|\Delta_g(\bm{U}_{\F,\G})|_{\bm{W}_g(\bm{U}_{\F,\G})}.
\]
By item~2,
\[
|\Delta_g(\bm{U}_{\F,\G})|_{\bm{W}_g(\bm{U}_{\F,\G})}
\le
\mu_0^{-1/2}\delta_E.
\]
Therefore
\begin{equation}
\label{eq:hybrid_projection_difference_bound}
|E_c-E_{\widetilde h}|
\le
L_g\sqrt{\frac{Q}{\underline\lambda_g\,\mu_0}}\,\delta_E.
\end{equation}

Combining \eqref{eq:hybrid_raw_term_bound} and
\eqref{eq:hybrid_projection_difference_bound}, we obtain
\[
E_c(\bx,t)
\le
A_{\alpha,g}\bigl(\delta_{\mathrm{comp}}+\delta_{\mathrm{anc}}\bigr)
+
L_g\sqrt{\frac{Q}{\underline\lambda_g\,\mu_0}}\,\delta_E.
\]
Insert this estimate into \eqref{eq:hybrid_entropy_split} and integrate over
\(\Omega\):
\[
\frac{d}{dt}\mathscr{H}_g[\G](t)
\le
-\frac{1}{\tau_2}\mathscr D_g(\G\|h)
+
\frac{|\Omega|}{\tau_2}
\left(
A_{\alpha,g}\bigl(\delta_{\mathrm{comp}}+\delta_{\mathrm{anc}}\bigr)
+
L_g\sqrt{\frac{Q}{\underline\lambda_g\,\mu_0}}\,\delta_E
\right)
+
\mathscr R_\ast(t).
\]
Replacing \(\mathscr R_\ast(t)\) by \(|\mathscr R_\ast(t)|\) gives
\eqref{eq:hybrid_energy_entropy_ineq}.
\end{proof}

\begin{corollary}[Specialization to the thermal two-population scaffold of \Cref{eq:LBM,eq:quasiequilibrium}]
\label{cor:hybrid_thermal_specialization}
In addition to the assumptions of \Cref{thm:hybrid_two_population_neurde},
suppose the analytic coupling term is
\[
S_{\ast,i}(\F,\G;\bm{U}_{\F,\G})
=
\Bigl(\frac{1}{\tau_2}-\frac{1}{\tau_1}\Bigr)
\bigl(\G_i^\ast-\G_i\bigr),
\]
with
\begin{equation}
\label{eq:thermal_gstar_corrected}
\G_i^\ast
=
\bigl(\Geqlattice_{\mathrm c}(\bm{U}_{\F,\G})\bigr)_i
+
\frac{2}{\temperature}\,
W_i\,\Velocity_\beta\,
\bigl(\pressuretensor_{\alpha,\beta}-\pressuretensor_{\alpha,\beta}^{\eq}\bigr)\,
\latticevelocity_{i,\alpha},
\end{equation}
where \(\pressuretensor\) and \(\pressuretensor^{\eq}\) are the pressure
tensors built from the \(\F\)-population and its analytic equilibrium, and
assume the lattice symmetry
\[
\sum_{i=1}^Q W_i\,\latticevelocity_{i,\alpha}=0
\qquad
\text{for each }\alpha.
\]
Then:
\begin{enumerate}
\item \textbf{The corrected quasi-equilibrium preserves energy:}
\[
C_g\G^\ast
=
C_g\Geqlattice_{\mathrm c}(\bm{U}_{\F,\G})
=
2\rho E,
\]
hence
\[
C_gS_\ast(\F,\G;\bm{U}_{\F,\G})=0.
\]

\item \textbf{If}
\[
\|\G^\ast(t,\bx)-\G(t,\bx)\|_1 \le \delta_\ast
\qquad \text{for all }(t,\bx),
\]
\textbf{then}
\begin{equation}
\label{eq:hybrid_Rstar_bound}
\bigl|\mathscr R_\ast(t)\bigr|
\le
\Bigl|\frac{1}{\tau_2}-\frac{1}{\tau_1}\Bigr|\,
|\Omega|\,L_\ast\,\delta_\ast,
\qquad
L_\ast \eqdef \max\!\bigl\{|\log\eta_-|,|\log\eta_+|\bigr\}.
\end{equation}
Consequently,
\begin{align}
\frac{d}{dt}\mathscr{H}_g[\G](t)
\le\;&
-\frac{1}{\tau_2}\,
\mathscr D_g\!\bigl(\G\,\|\,\Geqlattice_{\mathrm c}(\bm{U}_{\F,\G})\bigr)
\notag\\
&+
\frac{|\Omega|}{\tau_2}
\left(
A_{\alpha,g}\bigl(\delta_{\mathrm{comp}}+\delta_{\mathrm{anc}}\bigr)
+
L_g\sqrt{\frac{Q}{\underline\lambda_g\,\mu_0}}\,\delta_E
\right)
\notag\\
&+
\Bigl|\frac{1}{\tau_2}-\frac{1}{\tau_1}\Bigr|\,
|\Omega|\,L_\ast\,\delta_\ast.
\label{eq:hybrid_final_thermal_entropy_ineq}
\end{align}
\end{enumerate}
\end{corollary}

\begin{proof}
Summing \eqref{eq:thermal_gstar_corrected} over \(i\) gives
\[
C_g\G^\ast
=
C_g\Geqlattice_{\mathrm c}(\bm{U}_{\F,\G})
+
\frac{2}{\temperature}\,\Velocity_\beta\,
\bigl(\pressuretensor_{\alpha,\beta}-\pressuretensor_{\alpha,\beta}^{\eq}\bigr)
\sum_{i=1}^Q W_i\,\latticevelocity_{i,\alpha}.
\]
By the symmetry assumption, the second term vanishes, so
\[
C_g\G^\ast
=
C_g\Geqlattice_{\mathrm c}(\bm{U}_{\F,\G})
=
2\rho E.
\]
Hence
\[
C_gS_\ast
=
\Bigl(\frac{1}{\tau_2}-\frac{1}{\tau_1}\Bigr)
\bigl(C_g\G^\ast-C_g\G\bigr)
=
\Bigl(\frac{1}{\tau_2}-\frac{1}{\tau_1}\Bigr)
(2\rho E-2\rho E)
=
0.
\]

Next,
\[
\mathscr R_\ast(t)
=
\Bigl(\frac{1}{\tau_2}-\frac{1}{\tau_1}\Bigr)
\int_\Omega
\sum_{i=1}^Q
(\G_i^\ast-\G_i)\log \G_i\,d\bx.
\]
Since \(\G_i\in[\eta_-,\eta_+]\),
\[
|\log \G_i|
\le
L_\ast
\qquad \forall\,i,
\]
and therefore
\[
\bigl|\mathscr R_\ast(t)\bigr|
\le
\Bigl|\frac{1}{\tau_2}-\frac{1}{\tau_1}\Bigr|
\int_\Omega
L_\ast\,\|\G^\ast-\G\|_1\,d\bx
\le
\Bigl|\frac{1}{\tau_2}-\frac{1}{\tau_1}\Bigr|\,
|\Omega|\,L_\ast\,\delta_\ast.
\]
Substituting this into \eqref{eq:hybrid_energy_entropy_ineq} yields
\eqref{eq:hybrid_final_thermal_entropy_ineq}.
\end{proof}

\begin{remark}[Interpretation]
\label{rem:hybrid_two_pop_interpretation}
The theorem certifies the actual hybrid learned object at the level of the
energy channel: the \(\F\)-population remains analytic and exactly
conservative, the learned \(\G\)-population is corrected to exact energy
consistency, and the energy-channel entropy law degrades only through three
explicit, monitorable quantities,
\[
\delta_{\mathrm{comp}}
\quad\text{(completed learned residual),}\qquad
\delta_{\mathrm{anc}}
\quad\text{(anchoring mismatch),}\qquad
\delta_E
\quad\text{(energy projection residual),}
\]
plus the analytic coupling remainder \(\mathscr R_\ast\) inherited from the
two-population thermal scaffold.
\end{remark}

\begin{remark}[When this becomes a full hybrid entropy theorem]
\label{rem:full_hybrid_entropy_upgrade}
If, in addition, the analytic \(\F\)-population closure satisfies a
dissipative entropy identity
\[
\frac{d}{dt}\mathscr{H}_f[\F](t)
\le
-\frac{1}{\tau_1}\mathscr D_f\!\bigl(\F\,\|\,\mathbf{f}^{\mathrm{an}}(\bm{U}_{\F,\G})\bigr),
\]
and the chosen thermal coupling satisfies \(\mathscr R_\ast(t)\le 0\), then
adding this to \Cref{thm:hybrid_two_population_neurde} yields a full hybrid
entropy inequality for \(\mathscr{H}_f[\F]+\mathscr{H}_g[\G]\). Thus
\Cref{thm:hybrid_two_population_neurde} isolates exactly what remains to be
shown to upgrade the hybrid compressible realization to a full two-population
\(H\)-theorem.
\end{remark}

\subsubsection{Polynomial Equilibrium for the $\F$ Population}
\label{appendix:poly_F_pop}
For the equilibrium of the first population---in the experiments of the main text---we employ the extended equilibrium distribution \cite{karlin2010factorization},
\begin{equation}
\label{eq:Feq_prod}
\Feqlattice = \rho \Psi \otimes \Psi,
\end{equation}

where $\Psi= (\Psi_0, \Psi_1, \Psi_{-1})^\top$,\footnote{The order may vary based on the chosen enumeration of the lattice velocities.} 
with
\[
\Psi_{\pm 1} = \tfrac{1}{2}  \!\left[ \pm (\latticevelocity_{\alpha} - \Velocity_{\alpha}) 
+ (\latticevelocity_{\alpha} - \Velocity_{\alpha})^2 
+ \temperature \right],
\qquad 
\Psi_0 = 1 - \left[ (\latticevelocity_{\alpha} - \Velocity_{\alpha})^2 + \temperature \right],
\]
for $\alpha = x, y$. This formulation is derived and discussed in \cite{prasianakis2007lattice, karlin2010factorization}.

\subsubsection{Polynomial Equilibrium for the $\G$ Population}
\label{appendix:poly_g_pop}

In \cite{karlin2013consistent}, the equilibrium distribution for the $\G$ population is formulated as a polynomial expansion of the energy. The expansion originates from the Maxwellian form
\[
\Geqlattice \sim (\velocity \cdot \velocity)\,
\exp\!\left[-\tfrac{(\velocity - \Velocity)\cdot(\velocity - \Velocity)}{2\temperature}\right]
= (\velocity \cdot \velocity)\, \G^{\mathrm{MB}},
\]
where $\G^{\mathrm{MB}}$ denotes the Maxwellian distribution given in \cref{eq:Maxwellian}. This can be expressed as

\begin{equation}\label{eq:Geq_poly}
    \Geqlattice_{i,\mathrm{poly}}
    = W_i 
    \left( 
        2 \rho E
        + \dfrac{\heatflux^{\mathrm{MB}}_\alpha \latticevelocity_{i, \alpha}}{\temperature}
        + \dfrac{
            (\boldsymbol{R}_{\alpha, \beta}^{\mathrm{MB}}-2 \rho E \temperature \delta_{\alpha,\beta})
            (\latticevelocity_{i, \alpha} \latticevelocity_{i, \beta}- \temperature \delta_{\alpha, \beta})
        }{2 \temperature^2} 
    \right),
\end{equation}
where $W_i$ is defined in \cref{eq:temperature_weights}, $\heatflux^{\mathrm{MB}}$ is the Maxwellian heat flux, and $\boldsymbol{R}^{\mathrm{MB}}$ the contracted fourth-order moment of the Maxwellian distribution (see \cref{appendix:higher-order_moments}).

\subsection{Datasets}
\label{appendix:datasets}

All datasets considered in \cref{sec:numerical} are generated using the kinetic scheme in \cref{eq:LBM}. 
For $\Geqlattice_i$, we employ Levermore's model:
\begin{equation}\label{eq:Geq_Levermore}
\Geqlattice_i = \rho W_i \exp\!\left(\alpha_1 + \alpha_{\latticevelocity_i} \cdot \latticevelocity_i\right),    
\end{equation}
where $W_i$ is a temperature-related weight (see \cref{eq:temperature_weights} in \Cref{appendix:two_pop}) and $\rho$ is the local density. 
The parameters $\underbalpha = (\alpha_1, \alpha_{\latticevelocity_{i,x}}, \alpha_{\latticevelocity_{i,y}})$ are determined using the Newton--Raphson method at each spatio-temporal point, as detailed in \cite{latt2020efficient}. 
The iteration terminates when either the maximum absolute difference between consecutive $\underbalpha$ values is below $10^{-6}$, or after $20$ iterations if convergence is not achieved.%
\footnote{All experiments are performed in single precision for computational efficiency.}

For the equilibrium of the first population, we employ the extended equilibrium distribution defined in \cref{appendix:poly_F_pop}. 
While computing the numerical equilibrium for both populations could yield slightly more accurate results, the computational cost of applying Newton's method twice per time step renders this approach impractical. 
The extended equilibrium (\cref{eq:Feq_prod}) is already fourth-order accurate with respect to the Mach number and does not significantly affect overall accuracy.

This configuration offers two main advantages:  
(1) it simplifies dataset generation by requiring Newton's method for only one population, and  
(2) it mitigates errors associated with the equilibrium of the second population, as analyzed in \cite{saadat2019lattice}(Eq.~26) through a correction term.  
Instead of introducing such corrections explicitly, we rely on the learned surrogate \NN{} to approximate them implicitly.

\subsection{Architecture Parameters}
\label{appendix:arch_params}

\Cref{tab:ArchitectureSummary} summarizes the neural network architectures used in the experiments of \cref{sec:numerical}.
For the compressible-flow experiments, the branch network takes the primitive variables $(\rho,\Velocity,\temperature)^\top$ as input; this should be distinguished from the conserved state vector $\boldsymbol{U}$ used elsewhere in the paper.

\begin{table}[!h]
    \centering
    \renewcommand{\arraystretch}{0.9} 
    \begin{tabular}{cccccc}
        \toprule
        \small Experiment & \small Network & \small Activation & \small Layer size & \small Input & \small Renormalization \\ 
        \small  & \small  & \small  & \small  & \small  & \small map $\beta(\cdot)$ \\ 
        \midrule
        \small SOD case 1 & \small $\underbalpha$ & \multirow{2}{*}{\small GELU} & \small 4x32, 32x32, 32x32, 32x32 & \small $(\rho, \Velocity, \temperature)^\top$ & \multirow{2}{*}{\small $\exp(\cdot)$} \\
        \small (\cref{eq:Sod_case_1}) & \small $\underbvarphi$ &  & \small 9x32, 32x32, 32x32, 32x32 & \small $\{\latticevelocity_i: i=1, \ldots, 9\}$ & \\
        \midrule
        \small SOD case 2 & \small $\underbalpha$ & \multirow{2}{*}{\small GELU} & \small 4x64, 64x64, 64x64, 64x64 & \small $(\rho, \Velocity, \temperature)^\top$ & \multirow{2}{*}{\small $\exp(\cdot)$} \\
        \small (\cref{eq:Sod_case_2}) & \small $\underbvarphi$ &  & \small 9x64, 64x64, 64x64, 64x64 & \small $\{\latticevelocity_i: i=1, \ldots, 9\}$ & \\
        \midrule
        \small Cylinder & \small $\underbalpha$ & \multirow{2}{*}{\small GELU} & \small 4x32, 32x32, 32x32, 32x32 & \small $(\rho, \Velocity, \temperature)^\top$ & \multirow{2}{*}{\small $\exp(\cdot)$} \\
        \small (\cref{section:supersonic_flow}) & \small $\underbvarphi$ &  & \small 9x32, 32x32, 32x32, 32x32 & \small $\{\latticevelocity_i: i=1, \ldots, 9\}$ & \\
        \bottomrule
    \end{tabular}
    \caption{Neural network architectures used in \cref{section:Sod_cases,section:supersonic_flow}.}
    \label{tab:ArchitectureSummary}
\end{table}

\subsection{Optimization Algorithm}
\label{appendix:optimization}

All experiments employ the AdamW optimizer. 
During the first (pre-training) stage, the learning rate is initialized at $10^{-3}$ and reduced by a factor of $\tfrac{1}{2}$ every $100$ epochs over a total of $500$ epochs. 
In the second training stage, where \LBNN{} learns from problem-specific trajectories (\cref{alg:LBM_NN_algorithm_full_training}), the learning rate is fixed at $10^{-4}$ with the same linear scheduler and $N_r = 25$. 
Empirically, the results show minimal sensitivity to the parameter $\alpha \in [0, 1]$ in \cref{alg:LBM_NN_algorithm_full_training}; hence, we set $\alpha = 0$ for simplicity. 
The $\Lp$-norm is used as the loss $\ell$ in all experiments.

For both the Sod shock tube (\cref{section:Sod_cases}) and the 2D supersonic flow (\cref{section:supersonic_flow}), the trajectory dataset is defined as
\begin{equation}\label{eq:dataset}
    \mathsf{D} \eqdef 
    \Bigl\{ 
        \{\F_i(t, \bx), \G_i(t, \bx), \Geqlattice_i(t, \bx)\}_{i=1}^9 : 
        t = 0, \ldots, t_N,\ N > 500,\ \bx \text{ in the computational domain} 
    \Bigr\}.
\end{equation}
We use the first $500$ time steps from $\mathsf{D}$ for training.
In \cref{subsub:cylinder_long}, training uses the first $150$ temporal points. 

\subsection{Training and Computational Setup}
\label{appendix:train_time_dataset}

All LB experiments (assuming a known equilibrium) were implemented as standalone Python codes. 
The pre-training and main training phases both utilize the dataset defined in \cref{eq:dataset}. 
During pre-training, samples are used independently of their temporal order. 
In contrast, the second stage explicitly incorporates temporal evolution, as described in \cref{alg:LBM_NN_algorithm_full_training}. 

The randomly generated parameters $\lambda$ introduced in \cref{sec:NN_LBNN} were not required for the main experiments (\cref{section:Sod_cases,section:supersonic_flow}), as the existing trajectory data were sufficient. 
Each experiment uses a single trajectory representing the evolution of \cref{eq:conservation} under given initial and boundary conditions. 
Typically, the first $500$ time steps are used for training, except in the case of \cref{subsub:cylinder_long}, where only $150$ samples are available. 
To augment this smaller dataset during pre-training, an additional $350$ randomly generated $\alpha_n$ values are used to construct the pairs
\[
\{\boldsymbol{U}_n,\ [W_i \exp(\alpha_{n,1} + \alpha_{n,\latticevelocity_i}\cdot \latticevelocity_i)]_{i=1}^9\},
\]
while reserving the original $150$ samples of \cref{eq:dataset} for the second training phase.

Training was performed on a Tesla V100-DGXS-32GB GPU. 
Typical total training times were under one day per model, with most of the time spent in the second-stage trajectory-based training (\cref{alg:LBM_NN_algorithm_full_training}). 
Pre-training typically completed within a few hours.

\section{Shock Tube}
\label{appx:sod_shock}
This appendix details the numerical setup for the Sod shock tube experiments (\cref{appendix:Sod_case_1,appendix:Sod_case_2}). 
For both the subsonic and transonic configurations, the \emph{computational domain} consists of a grid of $3001 \times 5$. 
The model is trained over the first $500$ time steps and subsequently evaluated in an autoregressive manner for the next $500$ steps.

We present analyses common to both flow regimes. 
We first examine the local Mach number distributions (\cref{appendix:local_mach_number,subsection:local_mach_number_case2}) to confirm the subsonic and transonic character of the respective compressible flows discussed in the main text (\cref{section:Sod_cases}). 
We then extend the long-horizon evaluation of \cref{section:evaluating_distant_time}, specifically \cref{subsub:shock_long}, by initializing the simulation at progressively later time points---without retraining \NN{}---to assess its extrapolation capabilities beyond the training window (\cref{appendix:module_failure,appendix:module_failure_2}). 
In both cases, the model eventually fails when initialized more than $2500$ time steps away from the training data ($3900$ for the subsonic and $2500$ for the transonic case). 
Notably, this degradation is gradual over the reported horizon, indicating delayed loss of accuracy rather than immediate numerical divergence. 
For completeness, and to complement our comparison with FNO, additional details for the subsonic case are provided in \cref{appx:FNO}.

For the transonic case, we further detail the modified training protocol, which differs from \cref{alg:LBM_NN_algorithm_full_training} through the inclusion of a total variation diminishing (TVD) regularizer. 
In \cref{subsection:training_w_TVD}, we describe this TVD regularization procedure and demonstrate its effectiveness in suppressing oscillations and improving the smoothness of the predicted solutions. 
\Cref{alg:LBM_NN_algorithm_full_training_w_reg} shows the extension of the baseline training algorithm to incorporate the TVD regularization term. 
\Cref{appendix:poly_error_case2} reports the equilibrium errors observed in the transonic regime when using the $\G$-population polynomial equilibrium described in \cref{eq:Geq_poly}.
Finally, \cref{appendix:ood_transonic} presents a systematic suite of out-of-distribution experiments for the transonic configuration, in which the Riemann data and viscosity are perturbed at inference time---without retraining---to probe the observed stability limits of the learned closure under parameter shifts ranging from mild ($1\%$ pressure change) to severe (compound perturbations of density, pressure, and viscosity exceeding $100\%$).

\subsection{Sod Evaluation Protocol and Metrics}
\label{appx:sod_metrics}
For all Sod tables in the paper, the rollout is initialized from the dataset state at $t=500$, compared against the exact Euler Riemann solution for $499$ autoregressive steps, and reported at final time $t=999$. Stability is probed separately over a $1000$-step rollout.

We use the following metric definitions.
\begin{itemize}[leftmargin=*]
    \item \textbf{Shock, contact-edge, and rarefaction-tail location errors.} These are absolute location errors in grid cells. The shock position is detected as the midpoint where the smoothed pressure profile has maximal absolute gradient inside a shock window. The contact-edge position is detected as the midpoint maximizing
    \[
        \frac{|\partial_x \rho|}{|\partial_x p|/\max |\partial_x p| + 10^{-6}}
    \]
    inside the star-region search window, favoring a density jump with little pressure jump. The rarefaction-tail position is the first cell where the density drops below \(\rho_L^\ast + 0.05(\rho_L-\rho_L^\ast)\) inside the rarefaction search window.
    \item \textbf{Post-shock plateau error.} This measures profile fidelity in the constant star region between the contact and the shock. The interval is the interior of \([x^\ast_{\mathrm{contact}},x^\ast_{\mathrm{shock}}]\), trimmed by \(\max(10,\mathrm{width}/8)\) cells on both sides, and the reported value is the mean relative error over \(q\in\{\rho,u_x,T\}\) on that interval.
    \item \textbf{Shock-aligned and contact-aligned profile errors.} These shift the predicted final profile by the detected shock offset or detected contact-edge offset, respectively, and then average the relative profile errors over \(q\in\{\rho,u_x,T,P\}\). They therefore measure shape error after removing the corresponding phase drift.
    \item \textbf{Stable rollout horizon.} This is the number of valid rollout steps completed before the first step with any non-finite value, \(\rho\le 0\), or \(T\le 0\). If a rollout remains valid throughout the probe, the stable horizon is \(1000\).
    \item \textbf{Conservation drift.} We track \(M(t)=\sum_x \rho\), \(P_x(t)=\sum_x \rho u_x\), and \(E_{\mathrm{tot}}(t)=\sum_x \rho(c_vT+\tfrac{1}{2}(u_x^2+u_y^2))\). The reported drifts are \(\max_t |M(t)-M(0)|/|M(0)|\), \(\max_t |E_{\mathrm{tot}}(t)-E_{\mathrm{tot}}(0)|/|E_{\mathrm{tot}}(0)|\), and \(\max_t |P_x(t)-P_x(0)|/|M(0)|\).
    \item \textbf{Positivity violations and extrema.} We record the minimum density and minimum temperature encountered anywhere in the rollout, together with the total count of rollout steps having at least one non-finite value, one cell with \(\rho\le 0\), or one cell with \(T\le 0\).
    \item \textbf{Energy-moment residual.} At each rollout step we evaluate
    \[
        r(t)=\max_x \frac{\left|\sum_q G_q(x,t)-2\rho(x,t)E(x,t)\right|}{|2\rho(x,t)E(x,t)|+\varepsilon},
        \qquad
        E=c_vT+\tfrac{1}{2}(u_x^2+u_y^2).
    \]
    The table reports \(\max_t r(t)\), and the corresponding diagnostic curve plots \(r(t)\) versus rollout step.
    \item \textbf{Runtime.} This is the wall-clock time measured around the rollout loop only; we report both total seconds and successful steps per second.
\end{itemize}

\begin{table}[H]
    \centering
    \small
    \renewcommand{\arraystretch}{0.95}
    \begin{tabular}{@{}llcc@{}}
        \toprule
        Probe & Metric & \LBNN{} & Newton \\
        \midrule
        Case 1 & Solver-only rollout time [s] & 0.74 & 20.35 \\
        Case 1 & Solver-only throughput [steps/s] & 672.16 & 24.52 \\
        Case 2 & Solver-only rollout time [s] & 1.55 & 57.50 \\
        Case 2 & Solver-only throughput [steps/s] & 321.89 & 8.68 \\
        Case 2 & Recorded rollout time [s] & 3.03 & 76.36 \\
        Case 2 & Recorded throughput [steps/s] & 164.43 & 6.53 \\
        \bottomrule
    \end{tabular}
    \caption{Runtime modes for Sod Newton comparisons over
    \(499\) steps from \(t=500\). Solver-only rows measure the rollout loop used
    in the main text; recorded rows include trajectory recording overhead and
    are available for case~2.}
    \label{tab:sod_runtime_modes}\label{tab:sod_case2_runtime_modes}
\end{table}

\subsection{Subsonic Shock Tube Case (from \cref{section:Sod_cases})}
\label{appendix:Sod_case_1}
This subsection provides supplementary information regarding subsonic Sod shock tube case (as introduced in \cref{section:Sod_cases}). \Cref{appendix:local_mach_number} illustrates the local Mach number distribution within the shock tube at time $t=700$; and \cref{appendix:module_failure} presents the model failure of this case at different initial conditions, expanding on the analysis presented in \cref{subsub:shock_long} for the subsonic case.

\subsubsection{Subsonic Nature of the Experiment}
\label{appendix:local_mach_number}

\begin{figure}[H]
    \centering
    \includegraphics[width=0.5\linewidth]{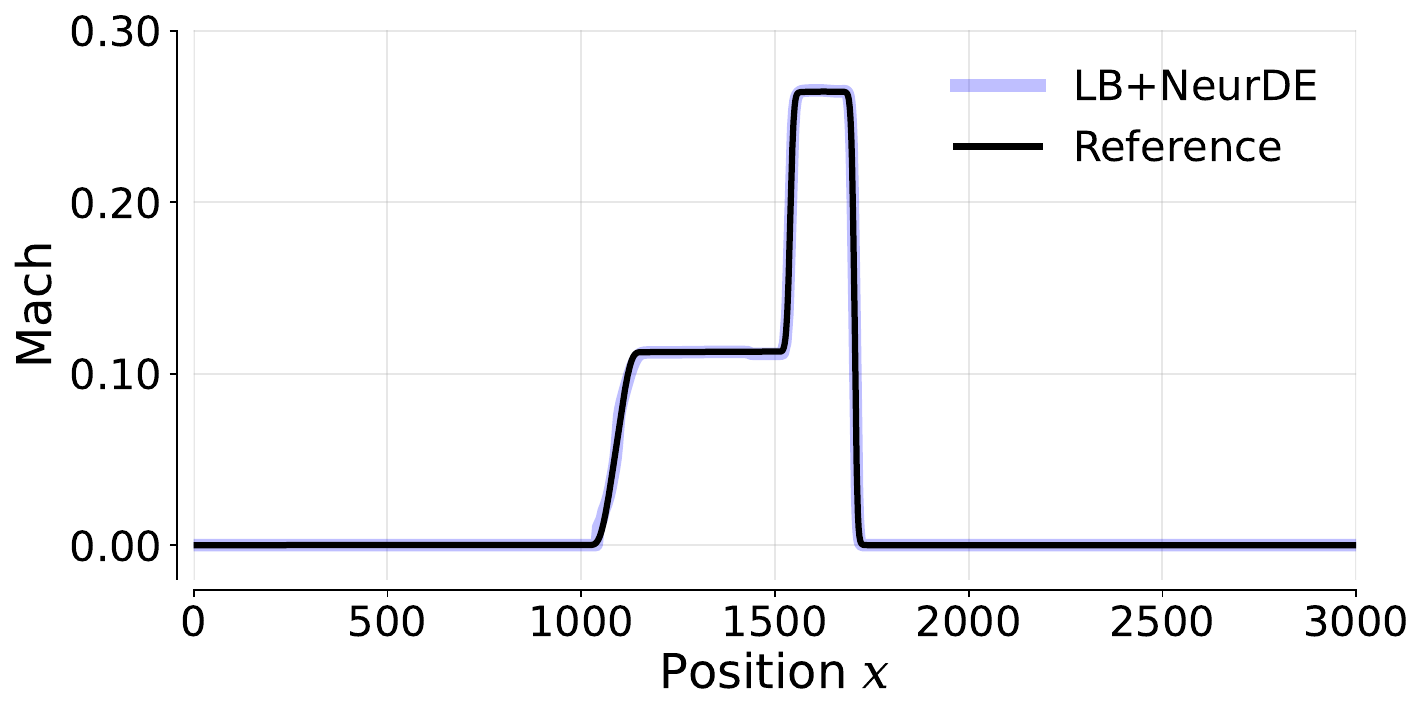}
    \caption{Comparison of the local \emph{Mach} number for the subsonic Sod shock tube (case 1 \cref{eq:Sod_case_1}) between \LBNN{} and simulation results. The black line represents the numerical reference, while the blue line depicts the flow predicted by \LBNN. This snapshot is taken at time-step $700$.}
    \label{fig:Mach_case1_700_Mach}
\end{figure}

\Cref{fig:Mach_case1_700_Mach} visually confirms the subsonic nature of the case, a characteristic we have emphasized throughout the primary portion of this work. For the local Mach number we use 
\[\mathrm{Ma} = (\Velocity \cdot \Velocity)^{1/2}(\gamma R \temperature )^{-1/2}.\]

\subsubsection{Model Failure (from \cref{subsub:shock_long})}
\label{appendix:module_failure}
\begin{figure}[htb!]
    \centering
    \includegraphics[width=\linewidth]{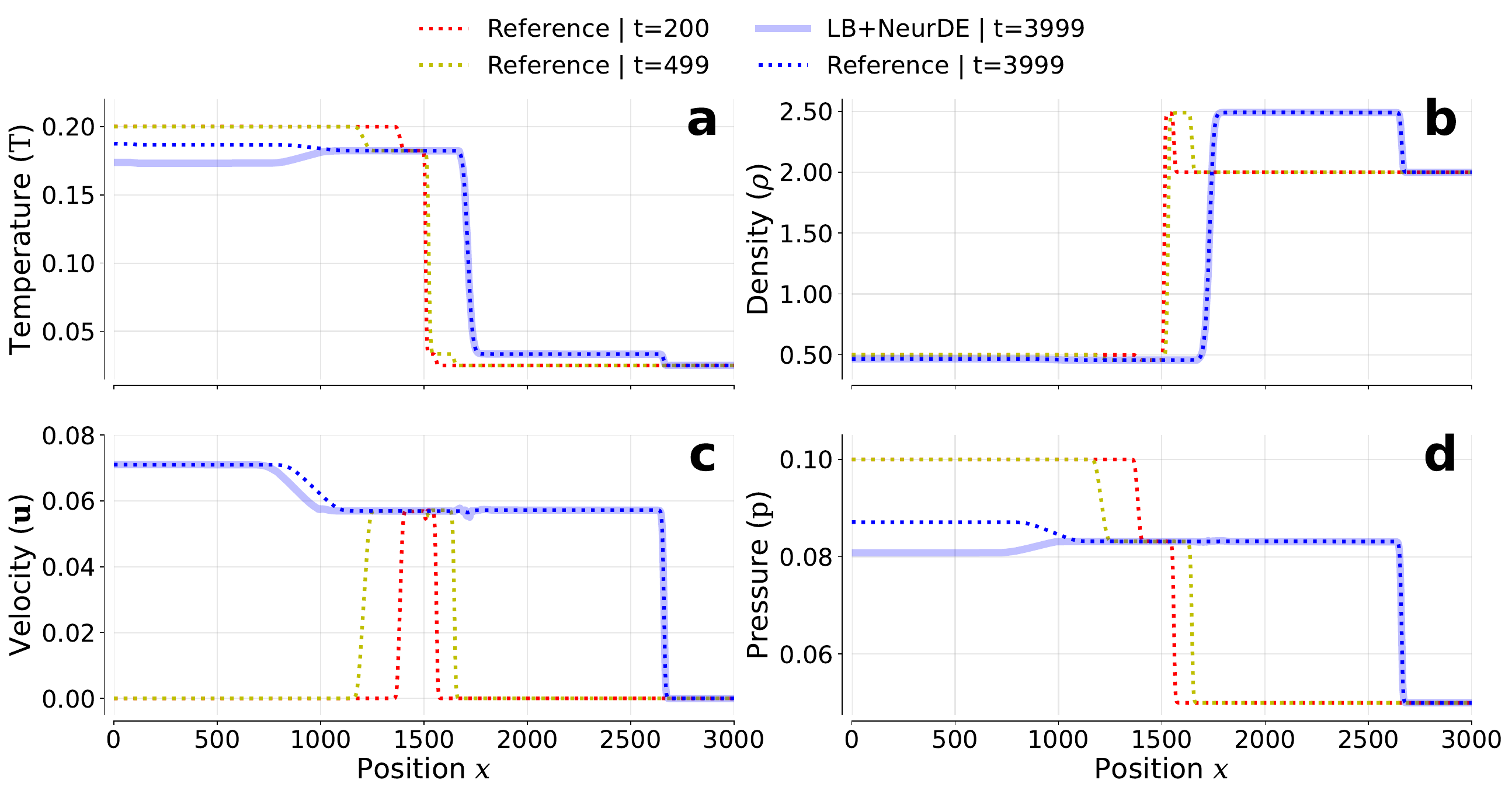}
    \caption{\LBNN's failure on the subsonic Sod shock tube when predicting $t=3999$ after being initialized at $t_0=3900$. The solid blue line represents the \LBNN{} prediction and the dotted lines represent simulated results at different times. Panels a, b, c, and d report \emph{temperature}, \emph{density}, \emph{velocity}, and \emph{pressure}, respectively.}
    \label{fig:model_failure_SOD_Case_1_long_times}
\end{figure}

Here, we explore the limits of the model when initialized with very long time-steps, a strategy presented in \cref{subsub:shock_long}. We recall that the architecture was previously trained in \cref{section:Sod_cases} using only the first 500 time-steps of its dataset. In \cref{subsub:shock_long}, we demonstrated that the model could predict the next 100 time-steps of the flow evolution starting at time $t=2000$. Here, we consider a more extreme case by using time $t=3900$ as an initial condition and utilizing our \LBNN{} to predict the subsequent 100 time-steps. In \cref{fig:model_failure_SOD_Case_1_long_times}, we observe that the model \emph{gracefully deviates} from the reference solution.

\subsubsection{FNO Benchmark}\label{appx:FNO}

\begin{figure}[htp]
    \centering
    \input{figs/appendix_F_FNO_BC}
\caption{\textbf{Boundary extensions for the FNO baseline.} \textbf{a}, A periodic wrap of non-periodic data creates a jump and slow \(O(k^{-1})\) spectral decay. \textbf{b}, Symmetric reflection gives \(C^0\) continuity but leaves derivative kinks, limiting decay to \(O(k^{-2})\). \textbf{c}, The palindromic edge-padded extension duplicates edge values to impose an approximately zero-gradient periodic interface, giving effective \(C^1\) continuity and reducing spectral leakage.}
    \label{fig:boundaries}
\end{figure}

To establish a rigorous baseline, we trained and evaluated the Fourier Neural Operator (FNO)~\cite{li2021fourier} on the subsonic Sod shock tube problem.
We note that resolving sharp features (particularly shock discontinuities) is very challenging for any spectral method, including FNO~\cite{mcgreivy2024weak,FalsePromizeZeroShot_TR}.
To address this, we target large numbers of Fourier modes to the point where the model's size makes its utility unfeasible.
The hyperparameter space was designed to maximize spectral resolution while respecting the memory bandwidth limits of high-performance accelerators.
Training was conducted on compute nodes equipped with $4\times$ NVIDIA A100 GPUs with 40--80~GB HBM2e per device.

We explored the following configuration grid:
\begin{itemize}
    \item \textbf{Lifting Dimension ($d_v$):} $\{128, 256\}$
    \item \textbf{Fourier Modes ($\mathbf{k}_{\text{max}}$):} $\{(128, 3), (256, 3), (512, 3)\}$ 
    \item \textbf{Network Depth ($L$):} $\{4, 8\}$ layers
\end{itemize}

The computational scaling of the FNO is dominated by the spectral layers, where the memory overhead of the fast Fourier transform (FFT) operations limits the achievable modal size. The number of FFT modes also limits FNO's ability to resolve sharp features like shocks. To strictly maximize the baseline's representational capacity, we prioritized spectral bandwidth over training throughput. Specifically, targeting $\mathbf{k}_{\text{max}}=512$ Fourier modes with a lifting dimension of $d_v=256$ required saturating the combined $320$\,GB memory envelope of the 4 A100 ($320$\,GB) accelerators. Consequently, we reduced the effective batch size (from $32$ to $4$ per GPU) to accommodate the dense optimizer states and intermediate activations. This configuration represents the maximal parameter density feasible on this hardware, ensuring the baseline is limited by its inductive bias rather than insufficient capacity.

\paragraph{Boundary Conditions.}
FNOs assume periodicity due to their use of spectral convolutions, which conflicts with the non-periodic Neumann boundaries of the Sod shock tube. A naive periodic wrap introduces a jump discontinuity and associated $O(k^{-1})$ spectral decay. To mitigate this artifact, we apply a \emph{Palindromic Edge-Padded Extension}, which mirrors the domain and duplicates the boundary values. This enforces an approximately zero-gradient condition at the periodic interface, yielding an effectively $C^{1}$ periodic extension. The resulting spectral decay improves to $O(k^{-3})$, substantially reducing leakage and aliasing (\Cref{fig:boundaries}).

\begin{table}[ht]
\centering
\begin{tabular}{@{}llcrc lcS[table-format=1.2e-1]S[table-format=1.2e-1]@{}}
\toprule
\multicolumn{3}{c}{Configuration} & \multicolumn{2}{c}{Model Stats} & \multicolumn{4}{c}{Error Metrics (MSE)} \\
\cmidrule(r){1-3} \cmidrule(lr){4-5} \cmidrule(l){6-9}
$d_v$ & $\mathbf{k}_{\max}$ $(x, y)$ & $L$ & \multicolumn{1}{c}{Params} & Epoch & \multicolumn{1}{c}{Val Loss} & \multicolumn{1}{c}{$\rho$} & \multicolumn{1}{c}{$\Velocity$} & \multicolumn{1}{c}{$\temperature$} \\
\midrule
128 & (128, 3) & 4 & 33.7M & 499 & \bfseries 3.20e-4 & \bfseries 2.69e-4 & 3.75e-5 & 1.35e-5 \\
128 & (128, 3) & 8 & 67.3M & 131 & 1.42e-3 & 1.30e-3 & 8.78e-5 & 3.96e-5 \\
256 & (128, 3) & 4 & 134.6M & 499 & 4.23e-4 & 3.83e-4 & \bfseries 3.08e-5 & \bfseries 9.00e-6 \\
256 & (128, 3) & 8 & 269.1M & 155 & 4.48e-1 & 4.37e-1 & 4.77e-3 & 6.61e-3 \\
\addlinespace
128 & (256, 3) & 4 & 67.2M & 182 & 4.52e-4 & 3.52e-4 & 7.48e-5 & 2.50e-5 \\
128 & (256, 3) & 8 & 134.4M & 147 & 1.11e-3 & 9.87e-4 & 8.28e-5 & 3.86e-5 \\
256 & (256, 3) & 4 & 268.8M & 168 & 5.15e-4 & 4.29e-4 & 6.38e-5 & 2.29e-5 \\
256 & (256, 3) & 8 & 537.5M & 141 & 1.12e-3 & 1.02e-3 & 7.75e-5 & 2.28e-5 \\
\addlinespace
128 & (512, 3) & 4 & 134.4M & 147 & 6.41e-4 & 6.00e-4 & 3.28e-5 & 8.05e-6 \\
128 & (512, 3) & 8 & 268.6M & 147 & 8.16e-4 & 7.46e-4 & 5.77e-5 & 1.16e-5 \\
256 & (512, 3) & 4 & 537.3M & 139 & 1.16e-3 & 1.10e-3 & 3.98e-5 & 1.44e-5 \\
256 & (512, 3) & 8 & 1.07B & 131 & 1.05e-3 & 9.63e-4 & 6.61e-5 & 2.02e-5 \\
\bottomrule
\end{tabular}
\caption{\textbf{Performance of Fourier Neural Operator variants.} We compare models with varying hidden channel width ($h$), Fourier modes ($\mathbf{k}_{\max}$), and depth ($L$). We also report the number of parameters and the epoch at which the best validation loss was achieved. Training utilized early stopping with a patience of 100 epochs, causing larger models to terminate earlier than the maximum 500 epochs once validation loss plateaued. The best result for each metric is highlighted in bold.}
\label{tab:fno_results}
\end{table}

\paragraph{FNO Architecture and Training Protocol.}
To evaluate the limits of the spectral approach, we deployed a high-capacity FNO architecture consisting of $8$ Fourier layers with a hidden channel width of $256$. We utilized a 2D spectral convolution retaining the top $512$ modes in the spatial dimension to maximize the capture of high-frequency shock content. As detailed in the previous section, we applied reflective (palindromic) boundary padding to mitigate spectral leakage at the non-periodic boundaries.

The model was trained to map the primitive variables $(\rho, u, T)$ at step $t$ to step $t+1$ using a dataset derived from the kinetic simulation (Compressible regime, $\text{Pr}=0.71$, $\nu=0.025$ as specified in \cref{eq:Sod_case_1}, see \cref{sec:experimental}). The data was split $50/20/30$ (train/validation/test) with a batch size of $4$. We minimized the Mean Squared Error (MSE) using the AdamW optimizer ($\beta_1=0.9, \beta_2=0.999$, weight decay $1\times 10^{-4}$) over $500$ epochs. We emphasize, however, that for shock-dominated rollouts this pointwise loss is mainly an optimization surrogate rather than the most physically meaningful evaluation metric, since small phase errors in the discontinuities can dominate it. In the tested autoregressive setting, these FNO rollouts destabilize within a small number of steps, which is why we report only the 1- and 10-step forecasts below. To ensure stable convergence, we employed a OneCycle learning rate scheduler, linearly warming up to a maximum learning rate of $1\times 10^{-3}$ for the first $30\%$ of training before annealing down to $1\times 10^{-6}$. Gradients were clipped at a norm of $1.0$ to prevent divergence during the initial training phase.

\begin{figure}
    \centering
    \includegraphics[width=\linewidth]{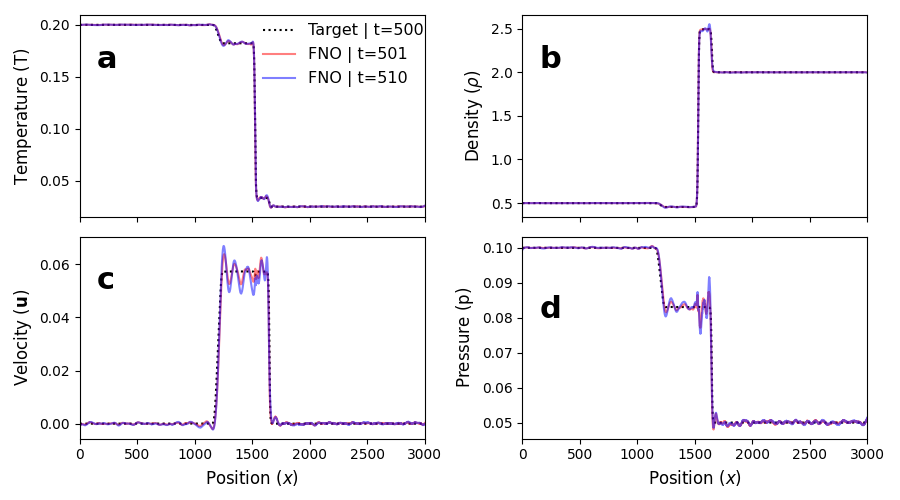}
    \caption{We train and evaluate the FNO model on the subsonic Sod shock tube with 128 Fourier modes, 128 embedding dimensions, and 4 layers. We evaluate (a) Temperature, (b) Density, (c) Velocity, and (d) Pressure for only 1 and 10 time steps.}
    \label{fig:fno_1}
\end{figure}

\begin{figure}
    \centering
    \includegraphics[width=\linewidth]{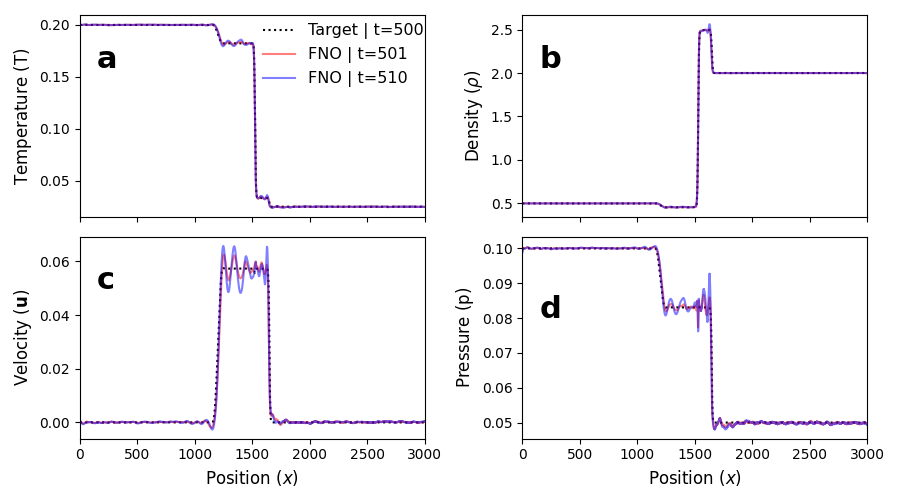}
    \caption{We train and evaluate the FNO model on the subsonic Sod shock tube with 128 Fourier modes, 256 embedding dimensions, and 4 layers. We evaluate (a) Temperature, (b) Density, (c) Velocity, and (d) Pressure for only 1 and 10 time steps.}
    \label{fig:fno_2}
\end{figure}

\begin{figure}
    \centering
    \includegraphics[width=\linewidth]{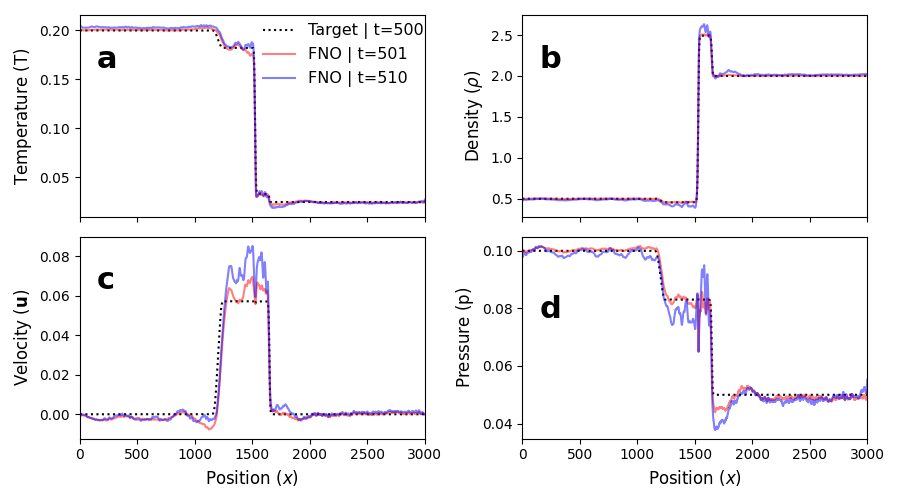}
    \caption{We train and evaluate the FNO model on the subsonic Sod shock tube with 512 Fourier modes, 256 embedding dimensions, and 8 layers. We evaluate (a) Temperature, (b) Density, (c) Velocity, and (d) Pressure for only 1 and 10 time steps.}
    \label{fig:fno_n512}
\end{figure}

\paragraph{FNO Experimental Results.}
Our empirical results show that the tested autoregressive FNO rollouts destabilize rapidly on this benchmark and already struggle to accurately predict the next time step ($\Delta t=1$).
Table~\ref{tab:fno_results} shows the validation results for our hyperparameter tuning.
Under the fixed optimization budget and early-stopping protocol used here, the 128-mode configurations generally achieve the best validation results.
Figures~\ref{fig:fno_1} and \ref{fig:fno_2} show the best two FNO temporal forecasting models with 128 Fourier modes, and 128 and 256 embedding dimensions, respectively, and 4 layers.
Figure~\ref{fig:fno_n512} shows the results from our largest model with 512 Fourier modes, 256 embedding dimensions, and 8 layers.
This comparison should be read as an auxiliary short-horizon baseline rather than a matched long-rollout comparison to the shock-aware Sod metrics used for \LBNN{}. Even before the rollout destabilizes, FNO shows significant deviations after a single time step.
This is expected as shocks are very challenging for any model, often requiring special methods to address them.

We provide the MSE on the test dataset for all of our trained models in \Cref{tab:fno_results}.
The larger-mode models are substantially more expensive and, within this training budget, do not yield better validation performance.
Unfortunately, the model's memory scaling as a function of the number of Fourier modes inhibits us from exploring further.
Regardless, the model's size with 512 Fourier modes limits its practicality given this level of accuracy.

\subsection{Transonic Shock Tube Case (from \cref{section:Sod_cases})}
\label{appendix:Sod_case_2}
\begin{figure}[bh!]
    \centering
    \includegraphics[width=\linewidth]{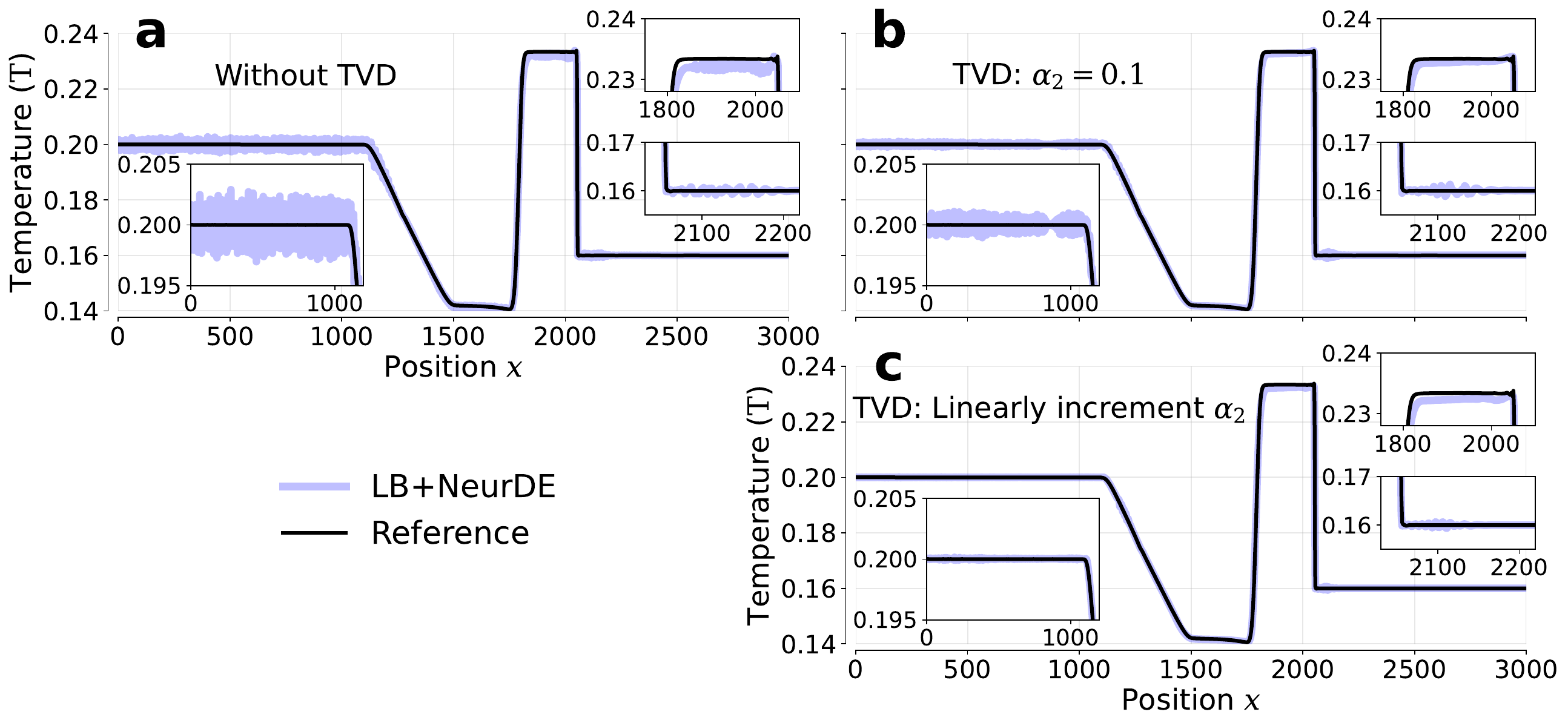}
    \caption{Improvements in \LBNN{} \emph{temperature} profile predictions through the inclusion of TVD regularization during training. Panel a shows the \LBNN{} performance without TVD regularization. Panel b shows the \LBNN{} performance when TVD regularization is included and weighted by a constant $\alpha_2$, see \cref{alg:LBM_NN_algorithm_full_training_w_reg}. Panel c shows the \LBNN{} performance when TVD regularization is included with a linearly incremented weight for $\alpha_2$. The black line represents the numerical simulation, and the blue line represents the \LBNN{} prediction. These snapshots are taken at $t=700$.}
    \label{fig:TVD_regularizer}
\end{figure}

This subsection provides additional details for the transonic Sod shock tube case (\cref{eq:Sod_case_2}). 
Unlike the subsonic configuration, the transonic Sod shock tube exhibits pronounced oscillatory behavior, highlighting the increased numerical difficulty of this regime. 
These oscillations are shown in \cref{fig:TVD_regularizer}a. 
As discussed in the main text, this case poses a greater challenge than \cref{eq:Sod_case_1} and therefore requires additional regularization to suppress the oscillations. 

To address the artifacts observed in \cref{fig:TVD_regularizer}a, we incorporate a total variation diminishing (TVD) regularization term, detailed in \cref{subsection:training_w_TVD}, with the corresponding modified algorithm presented in \cref{alg:LBM_NN_algorithm_full_training_w_reg}, see \cref{algorithm:TVD}. 
To confirm the near-sonic nature of this setup, \cref{subsection:local_mach_number_case2} visualizes the local Mach number at lattice time $t=700$. 
\Cref{appendix:poly_error_case2} provides an error analysis of the polynomial baseline in this flow regime, motivating its exclusion from \cref{fig:sod_all}(e -- h). 
Finally, \cref{appendix:module_failure_2} illustrates model degradation under different initialization times, complementing the long-horizon stability analysis discussed in \cref{subsub:shock_long}.

\subsubsection{Training with Regularization by the Total Variation Diminishing Principle} \label{subsection:training_w_TVD}

To mitigate numerical oscillations, particularly in regions with steep gradients, we add a soft total-variation regularizer during training.
For one-dimensional scalar conservation laws, TVD schemes are designed to respect the total-variation bounds used in classical shock-capturing analysis \cite{harten1997high}.
In the compressible two-population experiments this term is used as a training heuristic rather than as a proof that the full learned solver is TVD.
For an observable $\boldsymbol{U}$ of interest, we penalize increases in total variation with the $\relu(\cdot)$ function:

\begin{equation}\label{eq:relu_regularization}\relu\left(\TV(\boldsymbol{U}(t+1, \cdot)) - \TV(\boldsymbol{U}(t, \cdot))\right).
\end{equation}

In \cref{appendix:TVD}, we briefly review key concepts of the TVD principle, following Harten \cite{harten1997high}. The training algorithm is modified to incorporate this regularizer, as detailed in \cref{algorithm:TVD}. Finally, in \cref{fig:TVD_regularizer}, see \cref{algorithm:TVD}, we demonstrate the impact of the TVD regularization on the temperature profile at time $t=700$.\

\subsubsection{Total Variation Diminishing Principle} \label{appendix:TVD}

Consider a function $w(t, x)$ and an operator $L$ tied to a numerical scheme such that $w(t_{n+1}, x) =L w(t_n, x)$ (e.g., $L$ could be a point finite-difference scheme). We state only the discrete TVD condition needed to define the regularizer used in training.

A scheme is considered TVD, if for any function $w(t, x)$ of bounded total variation, the following inequality holds:
   \begin{equation}\label{eq:TVD_condition}
    \TV(L w) \le \TV(w),
    \end{equation}
where, 
$$
\TV(w(t_n, \cdot)) = \sum_{j=-\infty}^\infty \vert w(t_n, x_{j+1})- w(t_n, x_{j})\vert.
$$
For monotone schemes applied to one-dimensional scalar conservation laws, such bounds are a standard route to nonlinear stability and convergence; see \cite{harten1997high}(Theorem 2.1).

\subsubsection{Adding the TVD in the Training Algorithm}
\label{algorithm:TVD}

Here, for the sake of completeness, we present the modification of \cref{alg:LBM_NN_algorithm_full_training} to incorporate TV regularization. For simplicity, we demonstrate the algorithm for a single population $\Hlattice_i \in \{\F_i, \G_i\}$. 
\begin{algorithm}[ht!]
    \SetAlgoLined
    \SetNlSty{textbf}{\{}{\}}
    \KwData{$\tau,\, \{\latticevelocity_i\}_{i=1}^Q, \,\{W_i\}_{i=1}^Q, \alpha\in [0,1], \alpha_2,\, \eta,\, N_r$} 
    $\theta \leftarrow \texttt{\textit{pretraining}} (\theta\sim \mathrm{random})$\tcp*{Perform pre-training}
    $\big \{\{\Hlattice(0, \bx), \Heqlattice(0, \bx)\}, \ldots,  \{\Hlattice(t_{N_{\mathrm{train}}}, \bx), \Heqlattice(t_{N_{\mathrm{train}}}, \bx)\}\big\}$ \tcp*{Load trajectories} 
    \For{$0 \le \text{epoch} \le N$}{
        \For{$0 \le t \le t_{N_{\mathrm{train}}}$}{ 
            $t_\mathrm{end} = \min(t_{N_\text{train}}, t + N_r)$ \;
            $\mathbf{H}^\mathrm{pred}_\mathrm{hist},\mathbf{H}^\mathrm{eq}_\mathrm{hist}  = \texttt{\textit{LB\_NDEQ}}(t, t_{\text{end}}, \boldsymbol{M}[\Hlattice](t, \bx))$\tcp*{Make temporal prediction}
            $L \leftarrow \sum_{r=t}^{t_\text{end}} \alpha \ell\left (\Hlattice(r, \bx), \mathbf{H}^\mathrm{pred}_\mathrm{hist}[r,\bx] \right) + \alpha' \ell \left(\Heqlattice(r, \bx), \mathbf{H}^\mathrm{eq}_\mathrm{hist} [r,\bx] \right)$\tcp*{Accumulate loss}
            $L \leftarrow \sum_{r=t+1}^{t_\text{end}} \alpha_2\, \relu\bigg( \TV\left(\boldsymbol{M}[\mathbf{H}^\mathrm{eq}_\mathrm{hist}[r,\bx]]\right)- \TV\left(\boldsymbol{M}[\mathbf{H}^\mathrm{eq}_\mathrm{hist}[r-1,\bx]]  \right) \bigg)$\label{alg:reg_accumulate}\tcp*{regularizer} \label{alg:TVD1}
 $\theta \leftarrow (\theta - \eta \partial_{\theta} L)$\tcp*{Update the parameters}
                        $ t \leftarrow t+1$\;
        }
        $epoch\leftarrow epoch+1$;
    }
    \KwOut{ $\phiNN(\cdot ; \theta)$}
    \caption{Second stage of training $\phiS \phiCNN$ with $\TV$ regularizer.}
    \label{alg:LBM_NN_algorithm_full_training_w_reg}
\end{algorithm}

We see that \cref{alg:LBM_NN_algorithm_full_training_w_reg} is similar to \cref{alg:LBM_NN_algorithm_full_training}, with the main differences in Line~\ref{alg:TVD1} of \cref{alg:LBM_NN_algorithm_full_training_w_reg}, where the TVD condition is added. Specifically, for our applications, the condition in \cref{eq:TVD_condition} becomes:
    $$\TV(\boldsymbol{U}(t,\bx)) \le \TV (\boldsymbol{U}(t-1,\bx)).$$
    
The inequality is soft-enforced by using the $\relu$ function, as shown in \cref{eq:relu_regularization}. We note that while \cite{patel2022thermodynamically}(Eq. 11) suggests using $\left(\relu\left(\TV(Lw) - \TV(w)\right)\right)^2$ as the regularizer, we found that this did not significantly improve the results in our numerical experiments. Indeed, the results obtained with this alternative regularization were found to be indistinguishable from those obtained with \cref{eq:relu_regularization} in terms of solution accuracy and model performance. More generally, any regularizer of the form:
$$ \big\Vert \relu\left(\TV(Lw) - \TV(w) \right)\big\Vert,$$
could potentially be explored, although our preliminary investigations did not reveal significant gains with this more general class of regularizers.\ 

\paragraph{Effects of TVD on shock tube case 2.}

After incorporating TVD regularization, we observe in \cref{fig:TVD_regularizer}b and \cref{fig:TVD_regularizer}c that the oscillations present in \cref{fig:TVD_regularizer}a are dampened or eliminated. 
Our results indicate that the application of TVD is beneficial, but the scaling of the regularization parameter is crucial to fully mitigate the oscillations that are observed in the absence of regularization. 
Specifically, we found that a linear TVD schedule successfully suppresses these oscillations. An adaptive approach allows the model to explore a broader solution space initially, before progressively imposing stricter constraints, thus achieving a balance between flexibility and stability. 

\subsubsection{Local Mach Number Transonic Case}
\label{subsection:local_mach_number_case2}

\begin{figure}[ht!]
    \centering
    \includegraphics[width=0.5\linewidth]{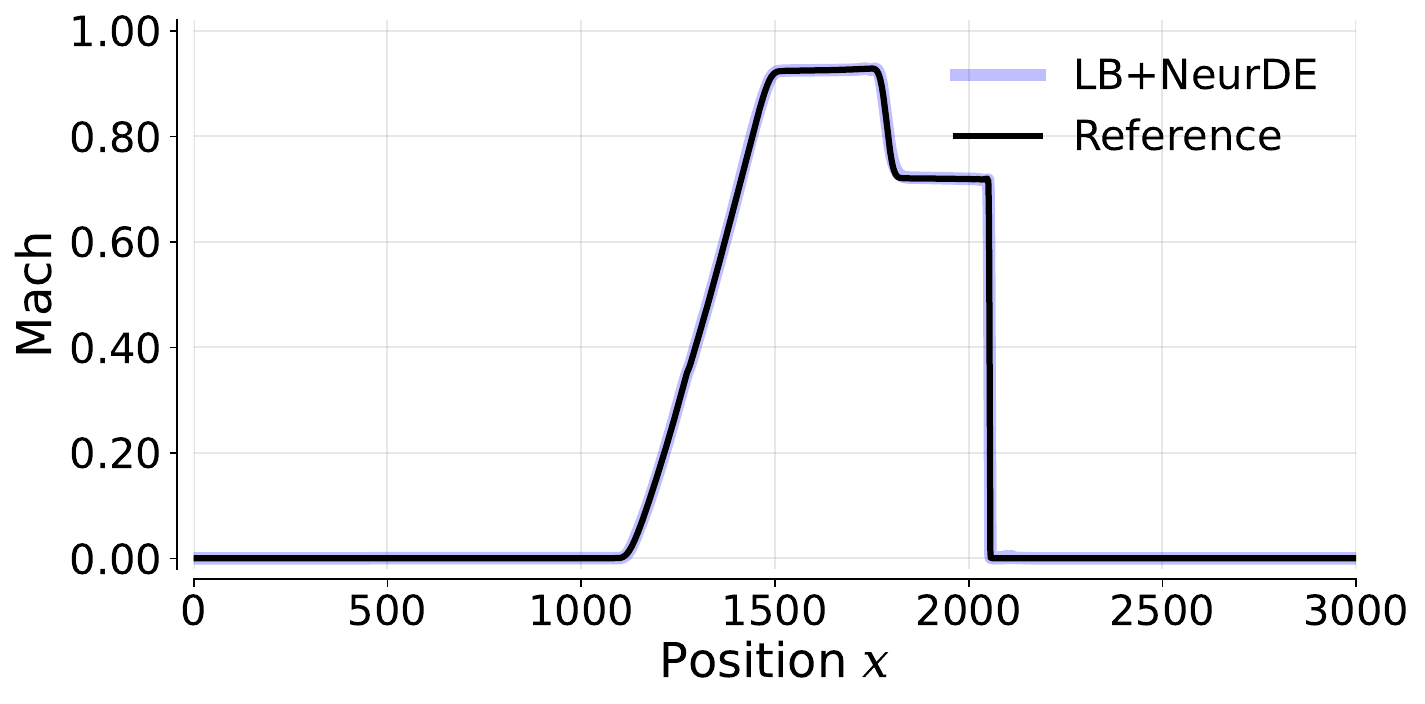}
    \caption{Comparison of the local \emph{Mach} number for the near-sonic Sod shock tube (case 2 \cref{eq:Sod_case_2}) between \LBNN{} and simulation results. The black line represents the numerical reference, while the blue line depicts the flow predicted by \LBNN{} trained with TVD. This snapshot is taken at time-step $700$. We observe that the local Mach number of the tube is close to $\mathrm{Ma}=1$ at around $\bx=1500$.}
    \label{fig:Mach_case2_700_Mach}
\end{figure}

In \cref{fig:Mach_case2_700_Mach}, we present a comparison between the predicted local Mach number, $\mathrm{Ma}$ (blue), and the reference value (in black). 
This analysis is based on \LBNN{} trained with TVD.
Assuming an ideal gas, the local Mach number is calculated using  $\mathrm{Ma} = (\Velocity \cdot \Velocity)^{1/2}(\gamma R \temperature )^{-1/2}$ (see, \cref{section:supersonic_flow}), consistent with the method used in the previous case.
The flow in the tube approaches $\mathrm{Ma}=1$. This near-sonic condition is particularly challenging for D2Q9 lattices and motivates the separate stability diagnostics reported for this case. Compare this with the simpler subsonic case presented in \cref{appendix:local_mach_number}.

\subsubsection{Errors of the Polynomial for the Transonic Case}\label{appendix:poly_error_case2}

Quantitatively, by the seventh time step, we already observe:
$\Vert \mathbf{R}_{\alpha,\alpha}^{\mathrm{eq, poly}} - \mathbf{R}_{\alpha,\alpha}^{\mathrm{MB}}\Vert_{\Lp} \ge \bigoh (10^2)$ for $\alpha \in\{x, y\}$; see \cref{tab:error_in_R_moment}. 
These rapidly growing higher-order moment errors make the polynomial comparison unusable in this regime and explain why the transonic rollout becomes unstable within roughly the first ten steps. This highlights the difficulty of maintaining stability for high-speed flows with the polynomial equilibrium and shows that the operator network in \cref{eq:levermore_closure_NN} plays a crucial role in the simulation.

\begin{table}[ht!]
    \centering
    \resizebox{\textwidth}{!}{%
    \begin{tabular}{@{}llllllll@{}}
        \toprule
        Iteration:  & $1$ & $2$  & $3$ & $4$ & $5$ & $6$  & $7$ \\
        \midrule
        $\Vert \mathbf{R}_{x,x}^{\mathrm{eq, poly}} - \mathbf{R}_{x,x}^{\mathrm{MB}}\Vert_{\Lp}$ & $0.09163$ & $0.1284$ & $0.1508$  & $0.1667$ &$0.1790$ &$6.554$& $24354.799$ \\
        $\Vert \mathbf{R}_{y,y}^{\mathrm{eq, poly}} - \mathbf{R}_{y,y}^{\mathrm{MB}}\Vert_{\Lp}$ & $0.0063$ & $0.0048$ & $0.0025$  & $0.0005$ &$0.0007$ &$0.095$& $473.0$\\
        \midrule
        $\Vert \mathbf{R}_{x,x}^{\mathrm{eq, NN}} - \mathbf{R}_{x,x}^{\mathrm{MB}}\Vert_{\Lp}$ & $(1.458) 10^{-4}$ & $(1.454)10^{-4}$ & $(1.447) 10^{-4}$  & $(1.454)10^{-4}$ &$(1.450)10^{-4}$ &$(1.453)10^{-4}$& $(1.452) 10^{-4}$ \\
        $\Vert \mathbf{R}_{y,y}^{\mathrm{eq, NN}} - \mathbf{R}_{x,x}^{\mathrm{MB}}\Vert_{\Lp}$ & $(2.823) 10^{-3}$ & $(2.8277)10^{-3}$ & $(2.822) 10^{-3}$  & $(2.8249)10^{-3}$ &$(2.8219)10^{-3}$ &$(8.228)10^{-3}$& $(2.821) 10^{-3}$\\
        \bottomrule
    \end{tabular}%
    }
     \caption{The $\Lp$-norm errors between the polynomial $\G$-population equilibrium (\cref{eq:Geq_poly}) and the Maxwellian higher-order moments (\cref{eq:higher-order}) for Sod case 2 (\cref{eq:Sod_case_2}). These errors already become large by step seven and precede the transonic rollout instability discussed in the main text. Compare with Levermore's moment system and the \LBNN{} hybrid model in \cref{fig:sod_all}(e -- h).}
     \label{tab:error_in_R_moment}
 \end{table}

The comparison in \cref{tab:error_in_R_moment} is based on the initial snapshot of the simulation, indicating the model's ability to represent the correct distribution rather than its generalization capabilities. Unfortunately, an assessment at time-step 700 (in lattice units) is not possible due to the instability of the polynomial LB scheme.   
\subsubsection{Model Failure (cf. \cref{subsub:shock_long})}\label{appendix:module_failure_2}

\begin{figure}[H]
    \centering
    \includegraphics[width=\linewidth]{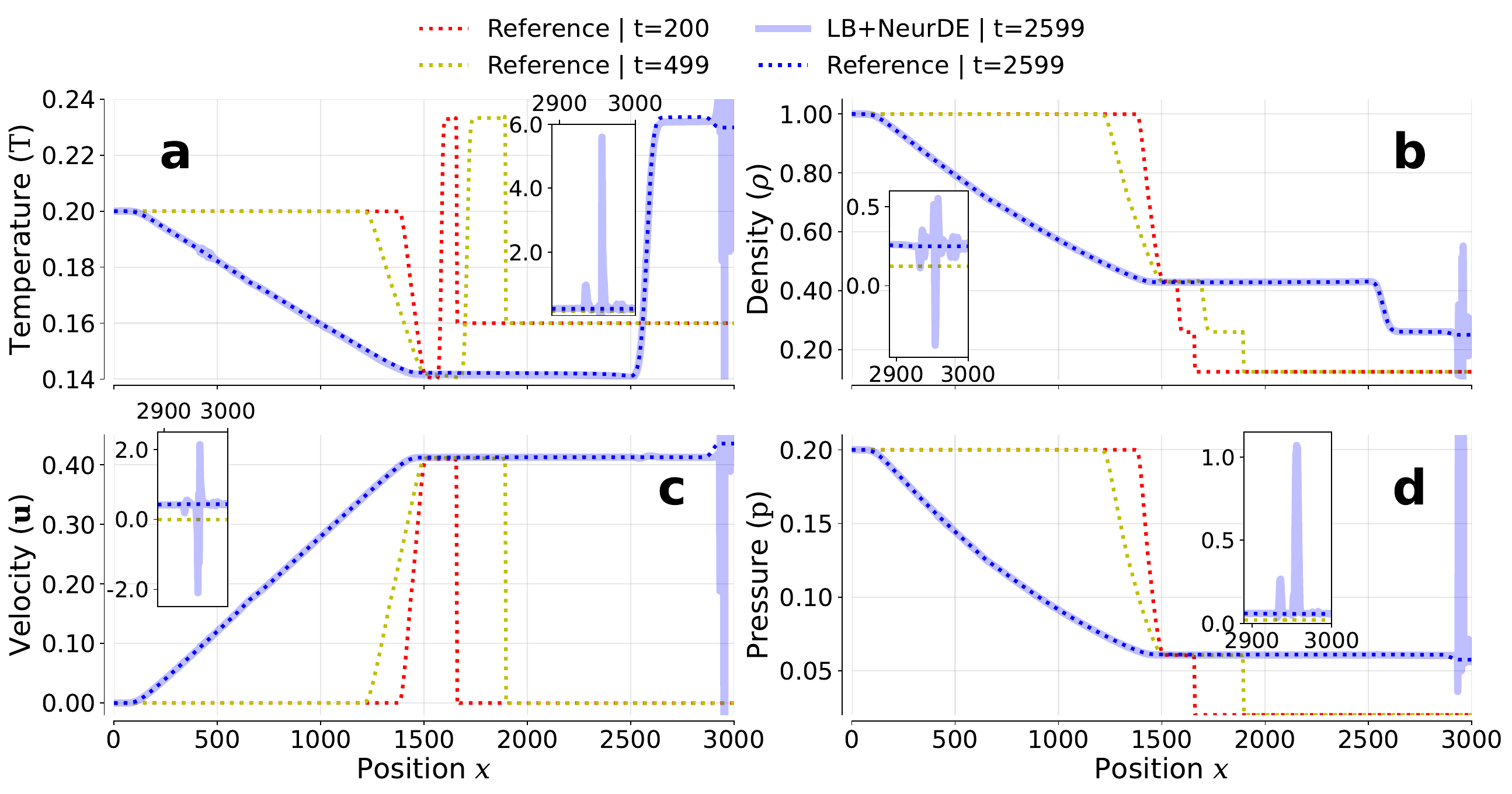}
    \caption{\LBNN's failure on the near-sonic Sod shock tube when predicting the $t=2599$ time-step after being initialized at $t_0=2500$. The solid blue line represents the \LBNN{} prediction and the dotted lines represent simulated results at different times. Panels a, b, c, and d report \emph{temperature}, \emph{density}, \emph{velocity}, and \emph{pressure}, respectively.}
    \label{fig:sod_Case_2_fail_5299}
\end{figure}

\begin{figure}[H]
    \centering
    \includegraphics[width=\linewidth]{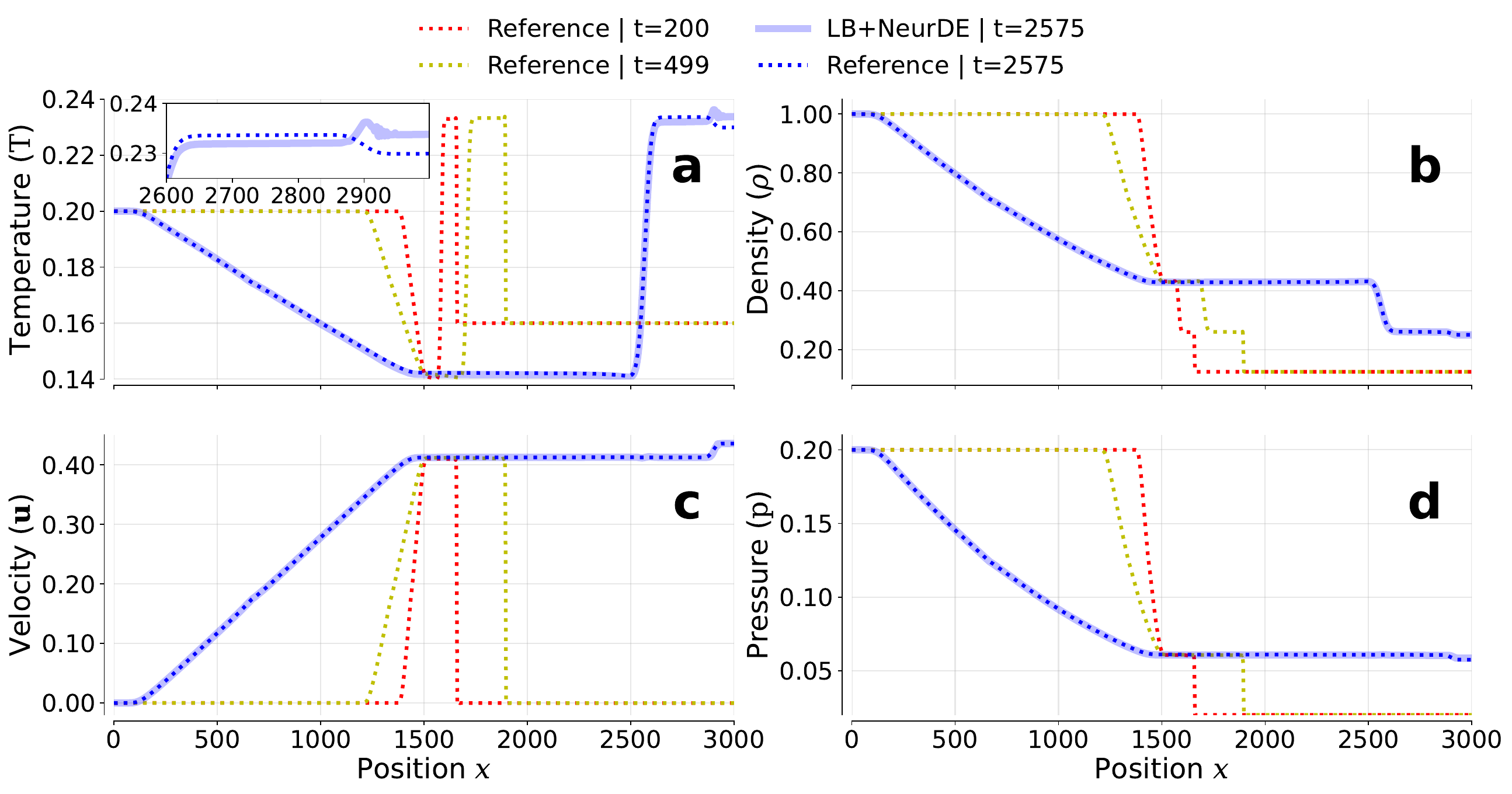}
    \caption{The onset of \LBNN's failure on the near-sonic Sod shock tube when predicting the $t=2575$ time-step after being initialized at $t_0=2500$. The solid blue line represents the \LBNN{} prediction and the dotted lines represent simulated results at different times. Panels a, b, c, and d report \emph{temperature}, \emph{density}, \emph{velocity}, and \emph{pressure}, respectively.}
    \label{fig:sod_Case_2_fail_2575}
\end{figure}

In the transonic shock tube case (\cref{eq:Sod_case_2}), which is more challenging than case 1, \LBNN{} fails when initialized at $t_0=2500$ and propagated for 100 time-steps. The failure manifests as large oscillations in all macroscopic variables, particularly at the right-hand side of the domain.
This is shown in \cref{fig:sod_Case_2_fail_5299}. These oscillations begin to appear around 75 time-steps after initialization ($t_0=2500$), as shown in \cref{fig:sod_Case_2_fail_2575}, and increase abruptly by $t=2599$, as indicated in \cref{fig:sod_Case_2_fail_5299}.

\subsubsection{Out-of-Distribution Perturbations}
\label{appendix:ood_transonic}

We next assess the behavior of \LBNN{} under out-of-distribution (OOD) perturbations of the transonic Sod shock tube. 
Throughout this experiment, the network is kept fixed (see \cref{eq:Sod_case_2} and  \cref{appendix:Sod_case_2}) and evaluated in the same autoregressive rollout setting used elsewhere in this appendix; only the inference-time Riemann data and/or viscosity are changed. 
The purpose is to determine how far the learned closure can be pushed beyond the training configuration before visible degradation appears.

\begin{figure}[ht!]
    \centering
    \includegraphics[width=\linewidth]{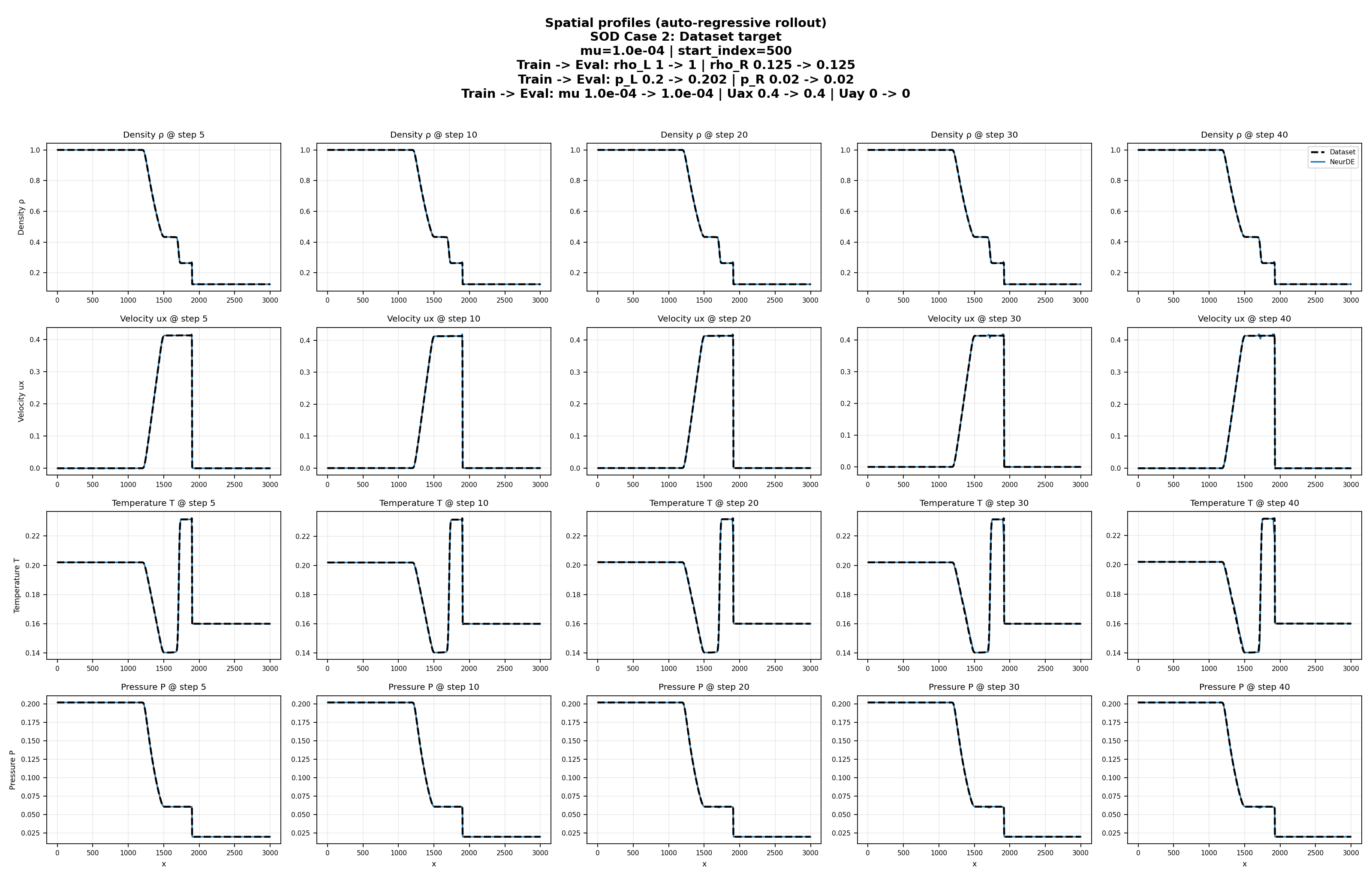}
    \caption{OOD evaluation for the transonic Sod shock tube with a $1\%$ increase in the left pressure $p_L$. This is the mildest single-parameter perturbation considered, and the rollout shows no visible discrepancy from the reference at the plotted resolution.}
    \label{fig:ood_transonic_pl_changed}
\end{figure}

We begin with isolated perturbations of individual parameters. 
In \cref{fig:ood_transonic_pl_changed}, increasing the left pressure $p_L$ by $1\%$ produces profiles with no visible discrepancy from the reference solution at the plotted resolution over the reported rollout. 

\begin{figure}[ht!]
    \centering
    \includegraphics[width=\linewidth]{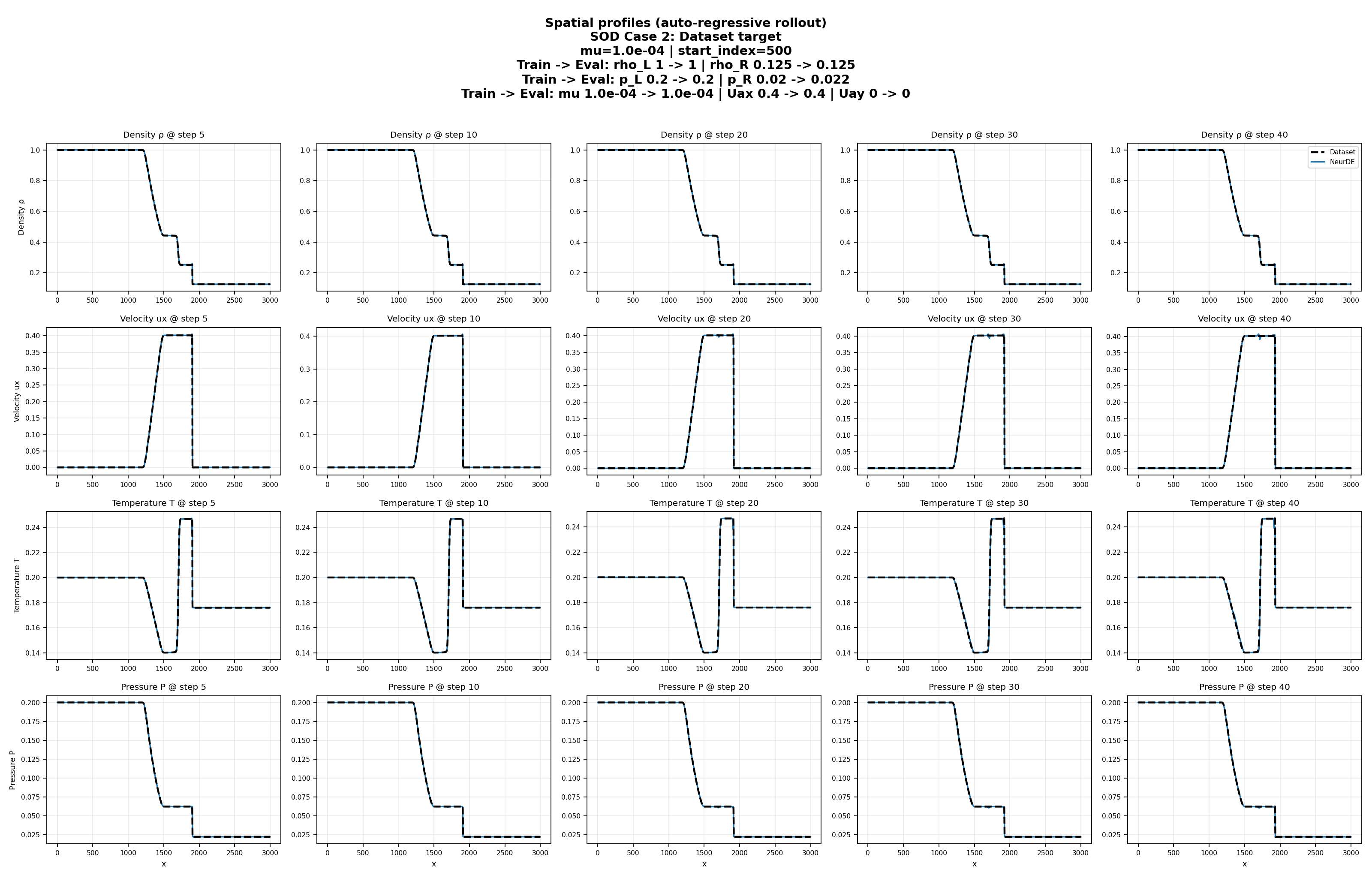}
    \caption{OOD evaluation for the transonic Sod shock tube with a $10\%$ increase in the right pressure $p_R$. The predicted macroscopic profiles remain stable and in close agreement with the target solution.}
    \label{fig:ood_transonic_pr_changed}
\end{figure}

\begin{figure}[ht!]
    \centering
    \includegraphics[width=\linewidth]{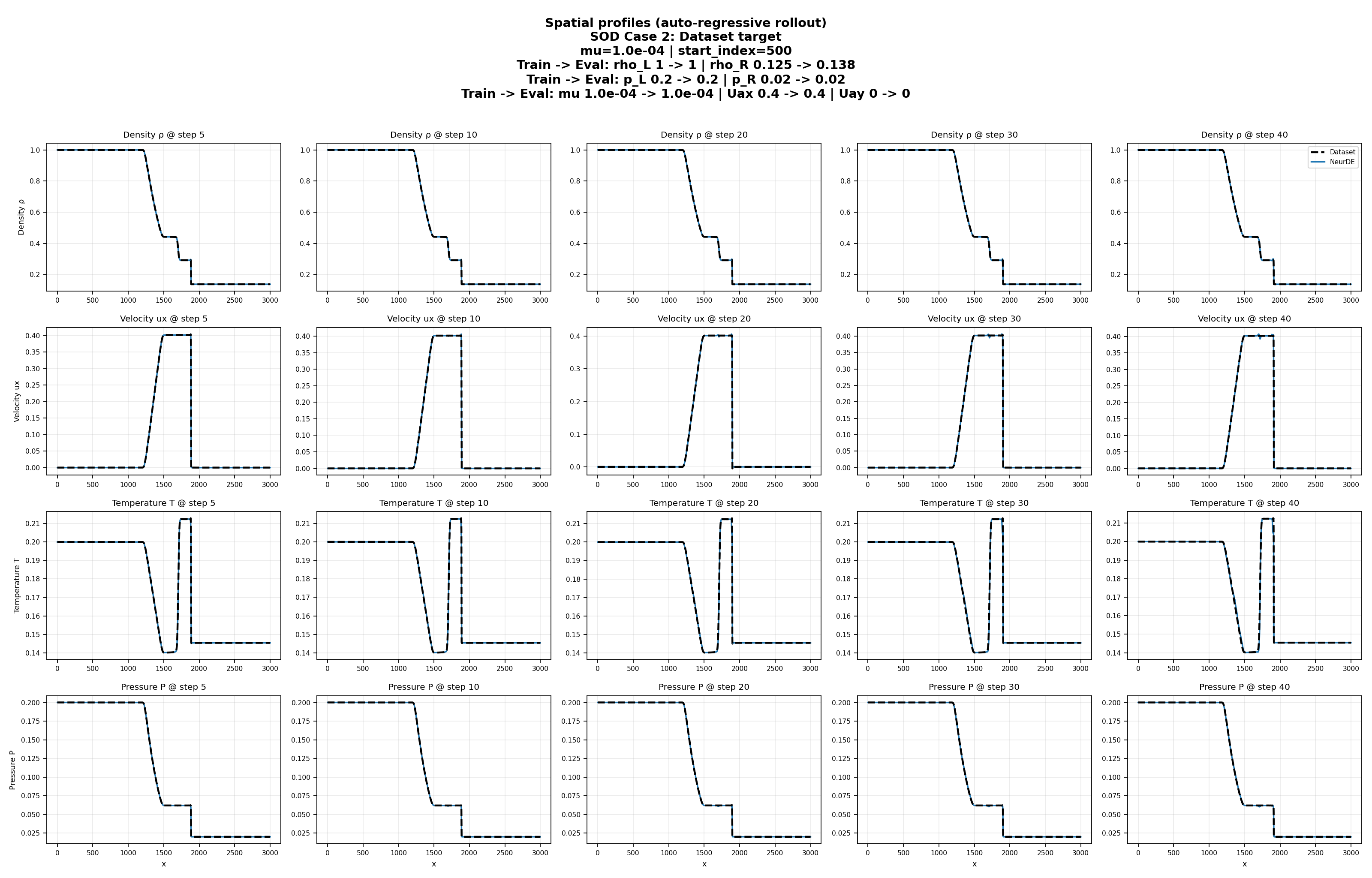}
    \caption{OOD evaluation for the transonic Sod shock tube with a $10.4\%$ increase in the right density $\rho_R$. The rollout continues to capture the wave structure at the plotted resolution.}
    \label{fig:ood_transonic_right_density_changed}
\end{figure}

\begin{figure}[ht!]
    \centering
    \includegraphics[width=\linewidth]{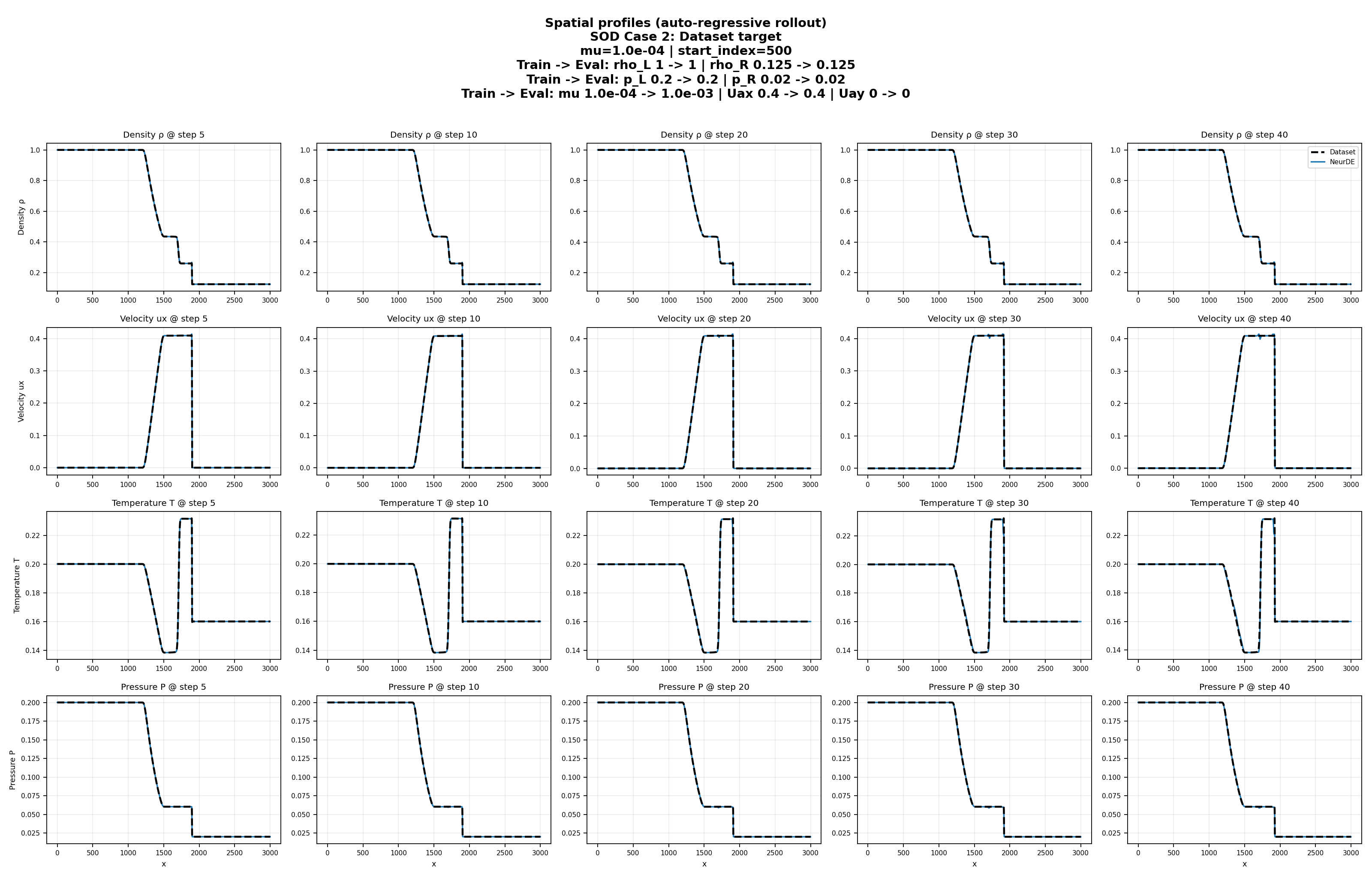}
    \caption{OOD evaluation for the transonic Sod shock tube with a $900\%$ increase in viscosity $\mu$. Despite this large transport perturbation, the rollout remains stable and closely follows the reference profiles.}
    \label{fig:ood_transonic_viscosity_changed}
\end{figure}

A similarly stable behavior is observed when increasing the right pressure $p_R$ by $10\%$ in \cref{fig:ood_transonic_pr_changed}, and when increasing the right density $\rho_R$ by $10.4\%$ in \cref{fig:ood_transonic_right_density_changed}. 
The same conclusion holds even for a much larger change in the transport coefficient: in \cref{fig:ood_transonic_viscosity_changed}, the viscosity $\mu$ is increased by $900\%$, yet the rollout remains stable and the density, velocity, temperature, and pressure profiles continue to closely track the target solution. 
Taken together, these single-parameter experiments show that the model is not restricted to an infinitesimal neighborhood of the training state.

\begin{figure}[ht!]
    \centering
    \includegraphics[width=\linewidth]{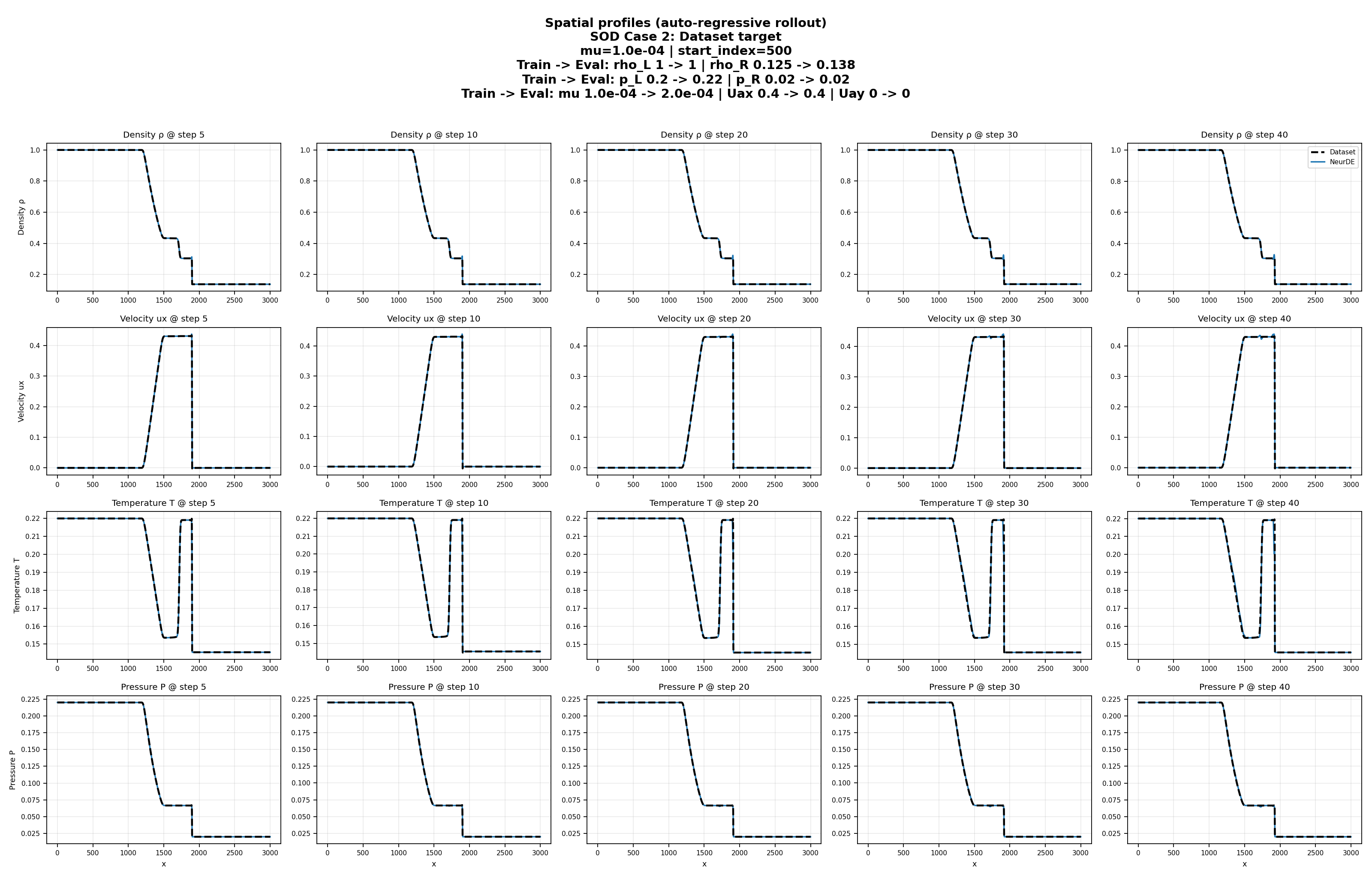}
    \caption{OOD evaluation for the transonic Sod shock tube under a moderate combined perturbation: $\rho_R$ increased by $10.4\%$, $p_L$ increased by $10\%$, and $\mu$ increased by $100\%$. The rollout remains stable and close to the reference in the plotted profiles.}
    \label{fig:ood_transonic_right_rho_left_pressure_viscosity_changed}
\end{figure}

We then consider a moderate coupled shift in several parameters simultaneously. 
In \cref{fig:ood_transonic_right_rho_left_pressure_viscosity_changed}, we increase $\rho_R$ by $10.4\%$, $p_L$ by $10\%$, and $\mu$ by $100\%$. 
Despite this combined perturbation, the predicted macroscopic fields remain in close agreement with the reference profiles throughout the rollout. 
This indicates that the learned closure retains nontrivial tolerance to simultaneous perturbations of multiple physically relevant quantities in this benchmark.

\begin{figure}[ht!]
    \centering
    \includegraphics[width=\linewidth]{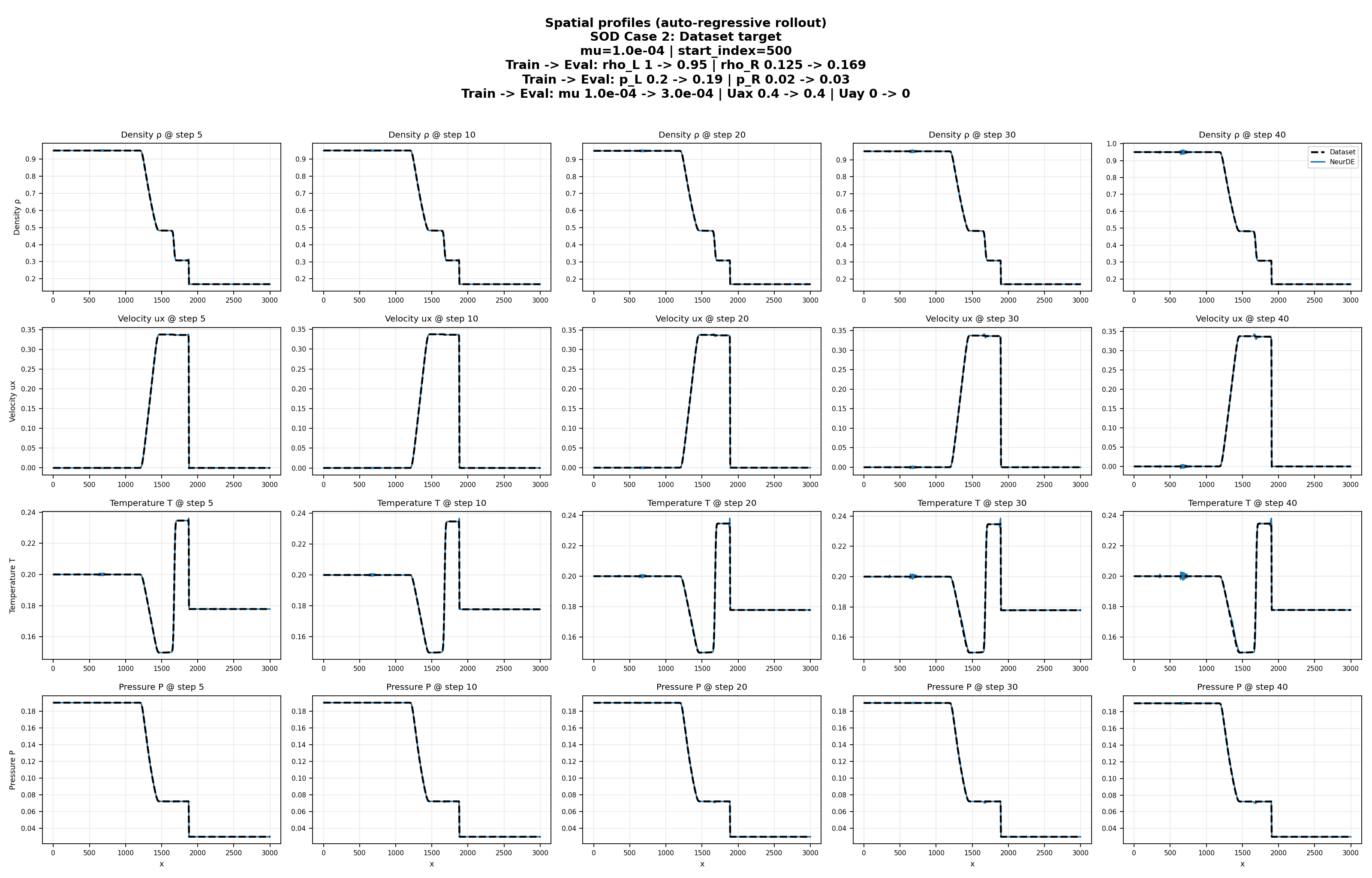}
    \caption{OOD evaluation for the transonic Sod shock tube under a stronger compound perturbation: $\rho_L$ decreased by $5\%$, $\rho_R$ increased by $35.2\%$, $p_L$ decreased by $5\%$, $p_R$ increased by $50\%$, and $\mu$ increased by $200\%$. The rollout remains stable, with localized deviations, indicating delayed degradation rather than immediate blow-up.}
    \label{fig:ood_module_failure_1}
\end{figure}

\begin{figure}[ht!]
    \centering
    \includegraphics[width=\linewidth]{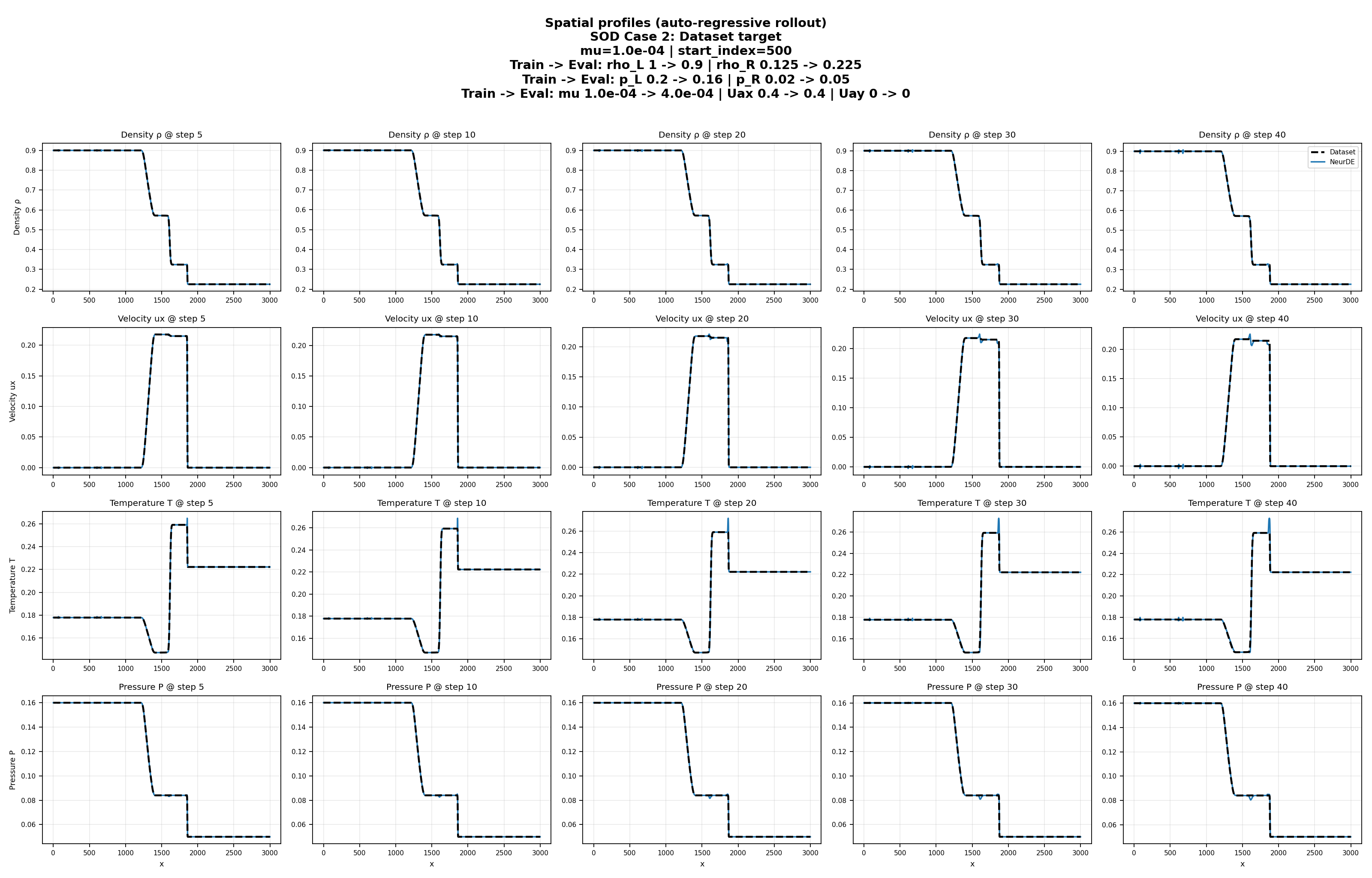}
    \caption{OOD evaluation for the transonic Sod shock tube under the most severe perturbation considered: $\rho_L$ decreased by $10\%$, $\rho_R$ increased by $80\%$, $p_L$ decreased by $20\%$, $p_R$ increased by $150\%$, and $\mu$ increased by $300\%$. Visible oscillations appear at later rollout times, especially in temperature and pressure, marking the onset of OOD degradation.}
    \label{fig:ood_module_failure_2}
\end{figure}

The limits of extrapolation begin to appear only under substantially stronger compound perturbations. 
In \cref{fig:ood_module_failure_1}, we decrease $\rho_L$ by $5\%$, increase $\rho_R$ by $35.2\%$, decrease $p_L$ by $5\%$, increase $p_R$ by $50\%$, and increase $\mu$ by $200\%$. 
Even under this shift, the rollout remains stable and the discrepancies remain localized near the wave structures. 
A clearer loss of accuracy is seen only in the most severe case, shown in \cref{fig:ood_module_failure_2}, where $\rho_L$ is decreased by $10\%$, $\rho_R$ is increased by $80\%$, $p_L$ is decreased by $20\%$, $p_R$ is increased by $150\%$, and $\mu$ is increased by $300\%$. 
Here, visible oscillations emerge at later rollout times, most notably in the temperature and pressure profiles.

Overall, these OOD experiments exhibit the same qualitative behavior observed in the distant-time initialization tests: the model does not immediately diverge when pushed beyond the training distribution, but instead shows delayed degradation as the perturbation size increases. 
The rollout first develops localized errors and mild oscillations under strong shifts, while the weaker and moderate perturbations remain stable and close to the reference in the reported profiles.

\clearpage
\FloatBarrier

\section{Ablation Study: Surrogate Models for Collision and Streaming}
\label{sec:ablation}

In this appendix, we present \emph{ablation studies} examining the role of ML surrogates within the \emph{splitting operator formulation} described in \cref{eq:collision,eq:streaming}. 
Specifically, we assess the effect of replacing either the collision operator $\phiC$ or the streaming operator $\phiS$---or both---with learned models, as motivated by the decomposition of the lattice Boltzmann (LB) scheme in \cref{eq:LBM_algorithm_dimensionless} (cf. \cref{eq:LBM_collision,eq:LBM_streaming}). 
Each experiment isolates a component of the LB update to quantify the impact of learned surrogates relative to the baseline physics-based formulation.

In the main study, \LBNN{} replaces only the equilibrium mapping within the collision operator. 
By contrast, several recent works have explored broader uses of ML in kinetic theory by substituting the entire collision operator with neural approximations \cite{miller2022neural, xiao2021using, xiao2023relaxnet, corbetta2023toward}. 
Here, we systematically extend this idea by testing both operators. 
Our goal is to elucidate the individual contributions of each operator to overall stability, accuracy, and physical consistency.

We focus on the compressible, subsonic Sod shock tube configuration (see \cref{eq:Sod_case_1}), extending the parameter range beyond the optimal LB conditions explored in prior studies \cite{corbetta2023toward}. 
Section~\ref{subsection:replacing collision} implements the surrogate-collision framework of \cite{corbetta2023toward}, both with and without explicit enforcement of conservation laws and symmetry constraints. 
Section~\ref{appendix:streaming} then investigates a neural surrogate for the streaming operator $\phiS$, following the approach of \cite{krishnapriyan2023learning}. 
Together, these experiments provide a comparative perspective on how learned operators can augment---or destabilize---classical kinetic solvers for compressible flows.

\subsection{Surrogate Model for Collision ($\phiC$)} 
\label{subsection:replacing collision}

Among the approaches that fully replace the collision operator with ML models, the work of Corbetta et al.~\cite{corbetta2023toward} is the closest comparison for our ablation. 
They substituted the BGK collision operator with a surrogate neural architecture for weakly compressible isothermal flows---regimes well aligned with standard LB formulations. 
By explicitly incorporating physical constraints such as conservation laws and lattice symmetries into the neural architecture, they improved accuracy in weakly compressible isothermal tests. 
However, their results were primarily obtained under conditions optimized for the LB method, potentially limiting their applicability to broader or higher-speed regimes.

Here, we compare our formulation of the equilibrium state (\cref{eq:levermore_closure_NN}) with the surrogate collision approach of \cite{corbetta2023toward}. 
In \cref{appendix:symCons}, we summarize their symmetry and conservation strategy. 
In \cref{subsection:ablation_details} and \cref{subsection:results_symm}, we provide the corresponding training details and simulation results for the subsonic shock tube case. 
Finally, \cref{appendix:without_algebraic} presents the results obtained after removing the symmetry and conservation constraints imposed in \cite{corbetta2023toward}.

\subsubsection{Enforcing Symmetry and Conservation as in Corbetta et al. \cite{corbetta2023toward}}\label{appendix:symCons}
Corbetta et al.~\cite{corbetta2023toward} parameterized the entire collision operator using a symmetry- and conservation-preserving multi-layer perceptron (MLP), hereafter denoted as $\MLPSymCons(\cdot;\theta)$. 
As discussed in \cref{subsection:Boltzmann_eqn} (cf.~\cite{levermore1996moment}), the collision operator $\collision{\cdot}$ satisfies three key physical principles:  
(i) local conservation of mass, momentum, and energy (\cref{eq:local_conservation});  
(ii) the local dissipation law implied by Boltzmann's H-theorem (\cref{eq:H-theorem}); and  
(iii) symmetry, including rotational and translational equivariance (\cref{eq:symmetries_collision}). \

To ensure that $\MLPSymCons(\cdot;\theta)$ inherits the correct symmetries, Corbetta et al.~\cite{corbetta2023toward} employed group averaging over the dihedral group $\mathrm{D}_8$, which captures rotations and reflections of the D2Q9 lattice.  
This yields a modified collision surrogate $\overline{\phiC}$ that satisfies the symmetry properties by construction. That is (cf., \cite{corbetta2023toward}(Eq. 21)), 
$$
\overline{\phiCNN}(\F_i)=\dfrac{1}{|\mathrm{D}_8|}\sum_{\sigma \in \mathrm{D}_8} \sigma^{-1} \phiCNN(\sigma\, \F_i).
$$
They further introduced linear transformations $\boldsymbol{A}$ and $\boldsymbol{B}$ to ensure mass and momentum conservation (energy was not explicitly enforced). The resulting post-collision distribution is then given by 
\begin{equation}\label{eq:appendix_postcollision}
    \Hlattice_i^{\mathrm{coll}} = \phiCNN(\Hlattice_i)=\boldsymbol{A} \Hlattice_i + \boldsymbol{B} \phiCNN(\Hlattice_i) ,
\end{equation}
where $\Hlattice_i \in \{ \F_i, \G_i \}$, and 
\begin{equation*}
\boldsymbol{A} =
\begin{pmatrix}
0 & 0 & 0 & 0 & 0 & 0 & 0 & 0 & 0 \\
0 & 0 & 0 & 0 & 0 & 0 & 0 & 0 & 0 \\
1 & 0 & 1 & 2 & 1 & 0 & 2 & 2 & 0 \\
0 & 0 & 0 & 0 & 0 & 0 & 0 & 0 & 0 \\
0 & 0 & 0 & 0 & 0 & 0 & 0 & 0 & 0 \\
-\frac{1}{2} & \frac{1}{2} & 0 & -\frac{3}{2} & -1 & 1 & -1 & -2 & 0 \\
0 & 0 & 0 & 0 & 0 & 0 & 0 & 0 & 0 \\
0 & 0 & 0 & 0 & 0 & 0 & 0 & 0 & 0 \\
\frac{1}{2} & \frac{1}{2} & 0 & \frac{1}{2} & 1 & 0 & 0 & 1 & 1
\end{pmatrix}, 
\qquad 
\boldsymbol{B}= 
\begin{pmatrix}
1 & 0 & 0 & 0 & 0 & 0 & 0 & 0 & 0 \\
0 & 1 & 0 & 0 & 0 & 0 & 0 & 0 & 0 \\
-1 & 0 & 0 & -2 & -1 & 0 & -2 & -2 & 0 \\
0 & 0 & 0 & 1 & 0 & 0 & 0 & 0 & 0 \\
0 & 0 & 0 & 0 & 1 & 0 & 0 & 0 & 0 \\
\frac{1}{2} & -\frac{1}{2} & 0 & \frac{3}{2} & 1 & 0 & 1 & 2 & 0 \\
0 & 0 & 0 & 0 & 0 & 0 & 1 & 0 & 0 \\
0 & 0 & 0 & 0 & 0 & 0 & 0 & 1 & 0 \\
-\frac{1}{2} & -\frac{1}{2} & 0 & -\frac{1}{2} & -1 & 0 & 0 & -1 & 0    
\end{pmatrix}.
\end{equation*}
%
Although these transformations are not unique, they guarantee that the surrogate operator satisfies the required conservation constraints without additional loss terms.

\subsubsection{Training Details}
\label{subsection:ablation_details}

Both \LBNN{} and the $\MLPSymCons$ surrogate are trained on the same first-$500$-step dataset, with the same optimizer, the same two-stage training pipeline described in \cref{sec:NN_LBNN,appx:training}, and rollout length \(N_r=25\).
The comparison is architectural rather than target-matched: \LBNN{} learns an equilibrium distribution, while the surrogate from \cite{corbetta2023toward} learns the post-collision distribution directly.
\Cref{tab:hyper-parameter} reports the corresponding architectures and hyperparameters.

\begin{table}[htb]
    \centering
    \begin{tabular}{lcccccc}\toprule
    &Input &Output &Activation &Loss &Layer Size &Model Size \\\hline
    \multirow{2}{*}{Ours} &$(\rho, \Velocity, \temperature)^\top$ &\multirow{2}{*}{$\Hlattice_i^{\mathrm{eq}}$} &\multirow{2}{*}{$
\relu$} &\multirow{2}{*}{MSE} &4x32, 32x32, 32x32, 32x32 &\multirow{2}{*}{6784} \\
    &$\{\latticevelocity_i\}_{i=1}^9$  & & & &9x32, 32x32, 32x32, 32x32 & \\
    \multirow{2}{*}{Model in \cite{corbetta2023toward}} &\multirow{2}{*}{$\Hlattice_i$} &\multirow{2}{*}{$\Hlattice_i^{\mathrm{coll}}$} &\multirow{2}{*}{$
\relu$} &\multirow{2}{*}{MSE} &\multirow{2}{*}{9x50, 50x50, 50x50, 50x9} &\multirow{2}{*}{6059} \\
    & & & & & & \\
    \bottomrule
    \end{tabular}
    \caption{Comparison of model hyperparameters and sizes for \LBNN{} and the surrogate from \cite{corbetta2023toward}.}
    \label{tab:hyper-parameter}
\end{table}

Our model takes as input the macroscopic observables $(\rho, \Velocity, \temperature)^\top$ and lattice velocities $\{\latticevelocity_i\}_{i=1}^9$, while the surrogate of \cite{corbetta2023toward} uses the pre-collision populations $\Hlattice_i$. 
Our model predicts the equilibrium distribution $\Hlattice_i^{\mathrm{eq}}$, while the surrogate of \cite{corbetta2023toward} predicts the post-collision distribution $\Hlattice_i^{\mathrm{coll}}$.


\

\subsubsection{Comparison of Results between \LBNN{} and the $\MLPSymCons$ Surrogate} \label{subsection:results_symm}

\begin{figure}[H]
    \centering
    \includegraphics[width=\linewidth]{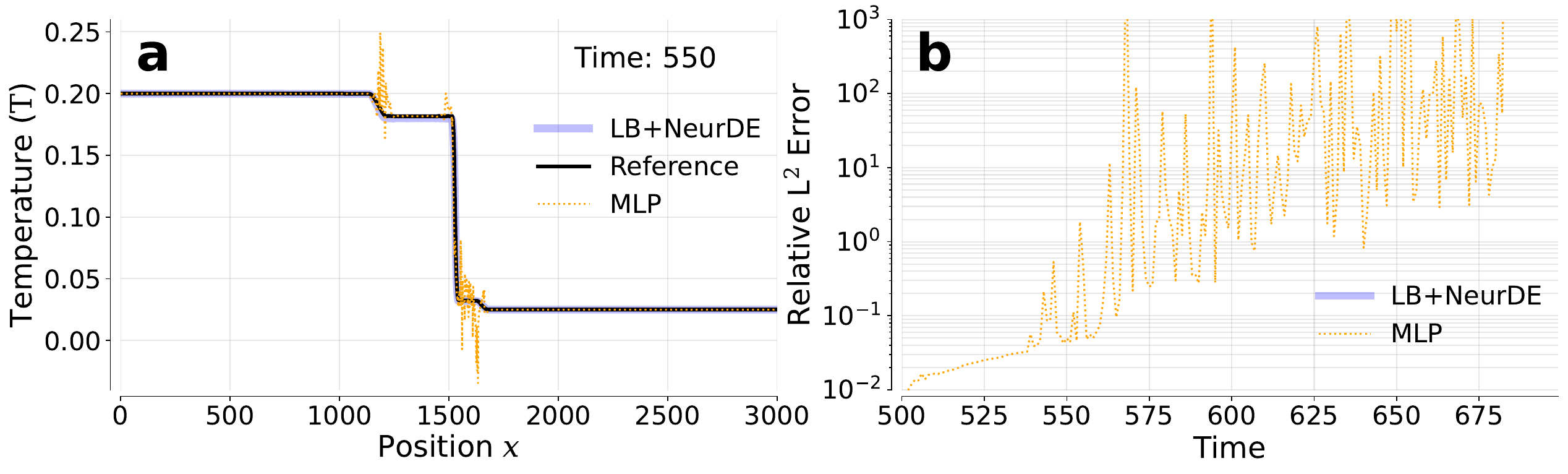}
    \caption{Comparison of the \emph{temperature} predictions for Sod case 1 between \LBNN, $\MLPSymCons$, and the numerical simulation for time-step 550. Here, both \LBNN{} and $\MLPSymCons$ are initialized at $t_0=500$. Panel a shows the predicted temperature; and panel b shows its relative $\Lp$-norm error with respect to the numerical simulation at different time-steps. The blue line represents \LBNN, the black line represents the numerical reference, and the dotted orange line represents $\MLPSymCons$.}
    \label{fig:temp_ablation_w_cons}
\end{figure}

The $\MLPSymCons$ model exhibits numerical instability arising from error accumulation, leading to unphysical temperature values ($\temperature < 0$) around time-step 550 (\cref{fig:temp_ablation_w_cons}a). 
The relative $\Lp$-norm error (\cref{fig:temp_ablation_w_cons}b) increases by several orders of magnitude within just 25 iterations.  
This instability can be traced to the algebraic correction defined in \cite{corbetta2023toward}(Eq.~27) (cf.~\cref{eq:appendix_postcollision}), which---while ensuring mass and momentum conservation---introduces negative post-collision populations $\G_i^{\mathrm{coll}}$, particularly in high-Mach regimes where temperature fluctuations are strong.\footnote{When the temperature exceeds the range supported by the lattice, negative populations can emerge, leading to numerical instability.}


\subsubsection{Without Algebraic Correction}
\label{appendix:without_algebraic}

To improve the $\MLPSymCons$ surrogate, we removed the algebraic correction layer, see \cref{eq:appendix_postcollision}, from the model in \cite{corbetta2023toward}. 
This modification compromised energy conservation (requiring a soft constraint), but it guaranteed a non-negative post-collision distribution, $ \G_i^{\mathrm{coll}}$. 
As shown in \cite{corbetta2023toward}(Fig.5), omitting this algebraic correction results in suboptimal simulation outcomes for the Taylor-Green vortex. 
However, in the specific regime studied here, we observed initial improvement in short-time predictions, compared with \cref{subsection:results_symm}, but the simulation still diverged into an unphysical result.  

\begin{figure}[htb!] 
    \centering
    \includegraphics[width=\linewidth]{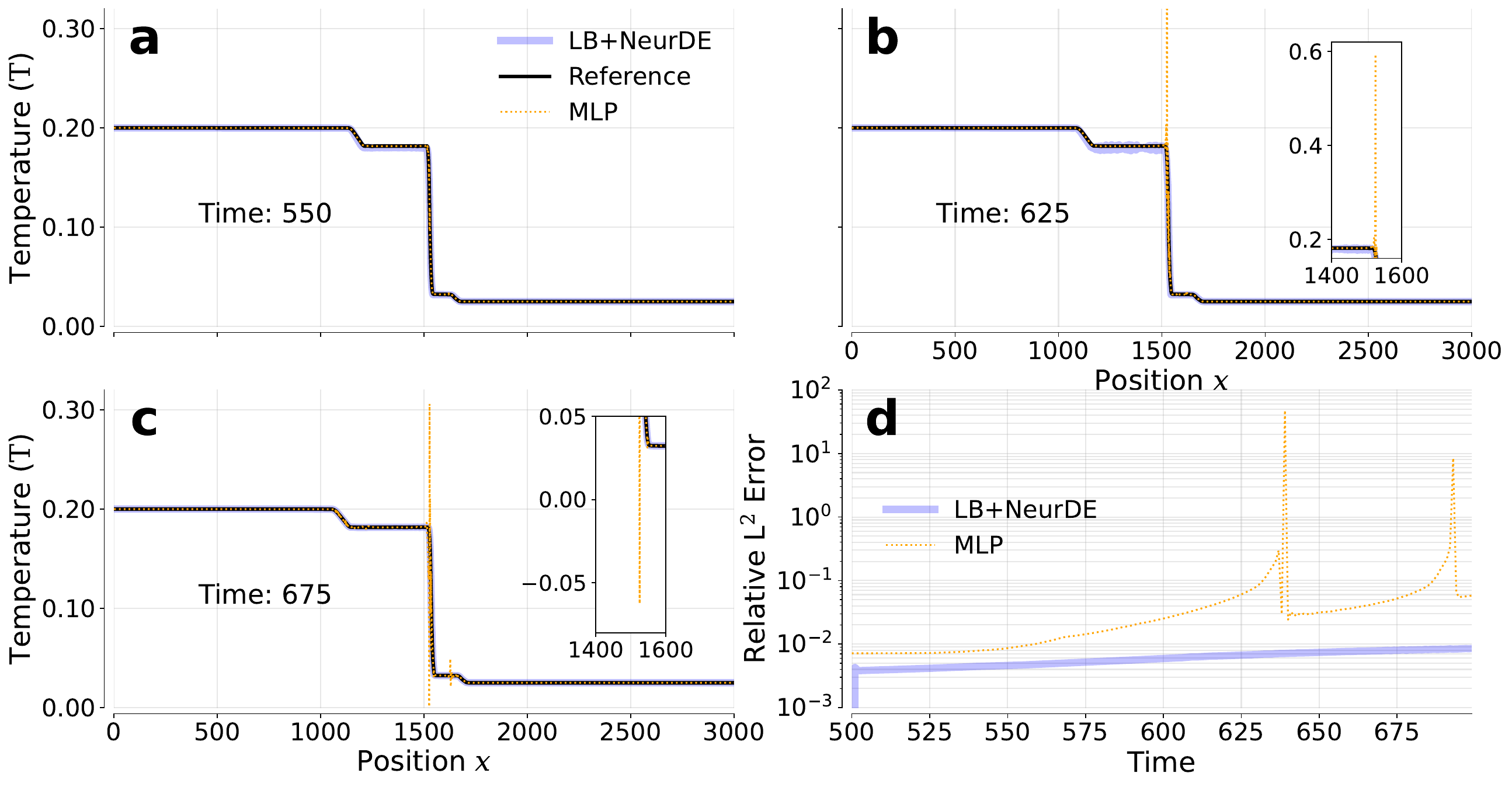}
    \caption{Comparison of the evolving \emph{temperature} predictions for Sod case 1 between \LBNN, $\MLPSymCons$, without the algebraic correction, and the numerical simulation for time-step 550, 625, and 675, in panels a, b, and c, respectively. Here, both \LBNN{} and $\MLPSymCons$ are initialized at $t_0=500$. Panel d shows the relative $\Lp$-norm error with respect to the numerical simulation as a function of prediction time. The blue line represents \LBNN, the black line represents the numerical reference, and the dotted orange line represents $\MLPSymCons$ without the algebraic correction.}
    \label{fig:temp_ablation_wo_cons}
\end{figure}

Compared to the result obtained with the algebraic correction (\cref{eq:appendix_postcollision} and shown in  \cref{fig:temp_ablation_w_cons}), this approach initially improved stability during $50$ time-steps, as illustrated in \cref{fig:temp_ablation_wo_cons}a.
However, at longer time-steps (up to the time-step $675$ time steps), a negative temperature value emerges, as shown in \cref{fig:temp_ablation_wo_cons}c. In \cref{fig:temp_ablation_wo_cons}b, we can see some numerical artifacts that start appearing in the simulation.
This led to a significant increase in relative error, as depicted in \cref{fig:temp_ablation_wo_cons}d. \

Upon inspection, the negative temperature result arises from an unphysical condition where the total energy density of the system, $\rho E = \bra \G_i/2\ket $, is smaller than the kinetic energy, $\rho \Velocity \cdot \Velocity /2$. This leads to instabilities, as:
$$0>\temperature = \dfrac{1}{\Cv} \left(E- \dfrac{\Velocity \cdot \Velocity }{2}\right), \qquad \text{ given that  }\qquad \rho E < \rho \dfrac{\Velocity \cdot \Velocity }{2}.$$

The authors of \cite{corbetta2023toward} acknowledged this problem and proposed an alternative approach based on soft constraints.
A more robust compressible-flow surrogate would need to enforce positivity of the post-collision population directly, for example through an exponential parameterization as in \cite{thyagarajan2023exponential}.
This would turn the replacement-collision baseline into a constrained-closure problem, making it closer in spirit to entropy-based moment closures. \

\subsection{Surrogate Model for $\phiS$} 
\label{appendix:streaming}

Here, we investigate the effectiveness of our translation based $\phiS$ operator \cref{eq:streaming_LBM} by replacing it with a surrogate neural network.
There are many potential models that propagate, or forecast, dynamical systems, including Transformers~\cite{nie2023a, wu2021autoformer, tian2022fedformer, liu2024itransformer} and ResNets.
The latter have been generally shown to have interesting connections with dynamical systems~\cite{chang2017multi, zhang2019towards} and ODE solvers~\cite{chen2018neuralODEs,krishnapriyan2023learning}.
Such models can become fairly large and difficult to interpret, which are misaligned with our physics-based approach that leverages much of the mathematical structure of the kinetic formulation.
Consequently, we use the so-called ``\emph{ContinuousNet}'' model~\cite{krishnapriyan2023learning}, which incorporates numerical integration techniques within the training and evaluation, as our streaming surrogate.

In \cref{phiS_continousnet}, 
we provide an overview of \emph{ContinuousNet} and how we incorporate this approach into the streaming. 
In \Cref{phiS_training details}, 
we describe our training procedure and setup for our experiment.
In \cref{phiS_numerical_results}, 
we present the numerical results.

\subsubsection{Temporal Streaming through \emph{ContinuousNet}}
\label{phiS_continousnet}

\textit{ContinuousNet}~\cite{krishnapriyan2023learning} is a model that predicts continuous-in-time solutions to ODEs by incorporating numerical integration schemes within the model.
Specifically, \textit{ContinuousNet} uses a neural network to calculate the time derivative of the solution to the ODE ($\partial_t \Hlattice_i$).
\textit{ContinuousNet} then evolves the function $\Hlattice_i(t, \mathbf{x})$ to $\Hlattice_i(t+\Delta t, \mathbf{x})$ using (explicit) numerical integration schemes, like Euler or Runga Kutta, which use the predicted $\partial_t \Hlattice$.
In this experiment, we use the forward Euler method as the ODE solver (although we note the point of \textit{ContinuousNet} is that higher-order schemes in the ODE solver typically perform better).
This \textit{ContinuousNet}-surrogate approach leverages the strengths of numerical integration schemes, making it methodologically well-aligned with our physics-based LB approach. \

Ultimately, \textit{ContinuousNet} learns the mapping from the distribution of current post-collision state, $\Hlattice_i^{\mathrm{coll}} $, to future pre-collision $N$ states. 
That is, 
$$
\left \{\Hlattice_i (t_{n+1},\bx), \Hlattice_i (t_{n+2},\bx), \ldots, \Hlattice_i (t_{n+N},\bx)\right \}  = \textit{ContinuousNet}(\Hlattice_i^{\mathrm{coll}} (t_n,\bx); \theta) .
$$ 
The equilibrium distribution for the collision $\phiC$ in \cref{eq:LBM} uses the $\Feqlattice_i$, and $\Geqlattice_i$, as described previously in \cref{eq:Feq_prod,eq:Geq_Levermore}. \

This \emph{ContinousNet} surrogate model is evaluated by first applying collision operation to the model's prediction at $t_{n+N}$, and then using this result as an input to the $\textit{ContinuousNet}$ model. This process is repeated autoregressively to extend the predictions over longer time steps; see \cref{alg:streamingNN}. The collision operator used during testing is identical to the one employed for data generation (\cref{eq:Feq_prod,eq:Geq_Levermore}).\

\begin{algorithm}[ht!]
    \SetAlgoLined
    \SetAlgoVlined
    \SetAlgoLongEnd
    \SetNlSty{textbf}{\{}{\}}
        $N \leftarrow 25$;
        $s \leftarrow t$\; 
    
     \While{$0\le r< N_{end}$}{
    $\Hlattice^{\mathrm{eq}}_i(s, \bx) \leftarrow \phi_i\left(M \Hlattice_i(s, \bx); \theta \right)$; \texttt{ // equilibrium \cref{eq:Feq_prod,eq:Geq_Levermore})}\; 
    $\Hlattice_i^{\mathrm{coll}} \leftarrow\phiC(\Hlattice_i, \Hlattice^{\mathrm{eq}}_i)$;\texttt{ //Collision}\; 
    $\left \{\Hlattice_i ^{\mathrm{pred}} (s+{\h},\bx), \Hlattice_i ^{\mathrm{pred}} (s+{2\h},\bx), \ldots, \Hlattice_i ^{\mathrm{pred}} (s+{N\h},\bx)\right \}  \leftarrow \textit{ContinuousNet}(\Hlattice_i^{\mathrm{coll}} (s,\bx); \theta)$; \;     
      $s \leftarrow s+N\h$ \;
       $r\leftarrow r+1$;\texttt{ //Increment $r$}\;
        $\Hlattice_i \leftarrow \Hlattice_i ^{\mathrm{pred}}; \texttt{ //Getting $\Hlattice_i$ prediction}$\;
        
     }
     
    \KwOut{ $\{\Hlattice_i (t,\bx), \ldots  \Hlattice_i (t_{n},\bx), \ldots, \Hlattice_i (t_{N_{end}+N},\bx) \}$}
    \caption{Replacing the entire streaming operator.}
    \label{alg:streamingNN}
\end{algorithm}

\subsubsection{Training Details} \label{phiS_training details}

The model is trained on the same dataset as previously discussed in \cref{subsection:ablation_details}. 
However, we opted not to employ a two-stage training strategy for this streaming operator due to the recursive nature of the \textit{ContinuousNet} model. This design allows for efficient training without compromising performance. For a fair comparison with our model, which is trained using a two-stage strategy, we set the predictive sequence length to $N = 25$, consistent with $N_r = 25$ (\cref{alg:LBM_NN_algorithm_full_training}). Once the model is trained, we evaluate the model performance on longer sequence.\

\subsubsection{Numerical Results}
\label{phiS_numerical_results}

\begin{figure}[ht!]
    \centering
    \includegraphics[width=\linewidth]{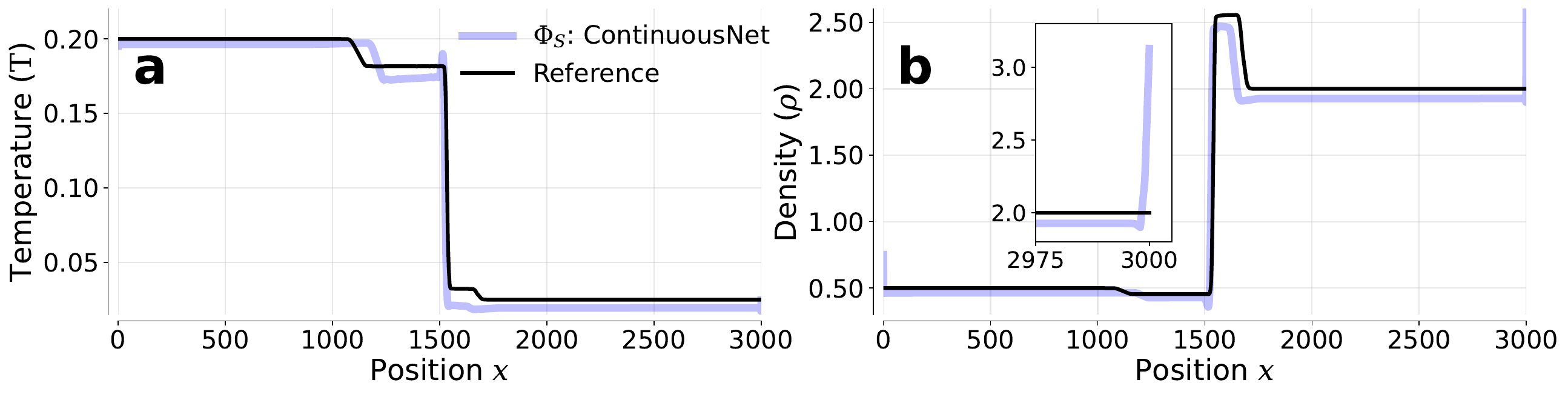}
    \caption{The \emph{temperature} comparison between the \textit{ContinuousNet} surrogate model and numerical simulation for the subsonic Sod shock tube (case 1 \cref{eq:Sod_case_1}). The results are presented at time-step 650 which is 150 time-steps beyond training data. The blue line represents the \textit{ContinuousNet} surrogate model and the black line represents the simulation.}
    \label{fig:Streaming_operator}
\end{figure}

As illustrated in \cref{fig:Streaming_operator}, the overall model (the Euler variant) exhibits significant inaccuracies in the temperature prediction, while the density prediction is relatively well-behaved, 
with the exception of the boundary values. 
Specifically, the model fails to accurately represent both the rarefaction wave and the shock wave in the temperature simulation. Notably, in the temperature field, the shock wave is entirely absent from the model's output. 
These findings are not unexpected when we consider the fundamental linearity of the streaming operation in the kinetic framework (cf., \cref{fig:Streaming_operator}).
The streaming operator is essentially a linear operation. 
It shifts discrete distributions $\Hlattice_i(t,\bx)$ along fixed, pre-defined lattice velocities without requiring iterative solvers or nonlinear processing. By replacing this exact mechanism with a learned neural surrogate, we introduce unnecessary complexity and parameterization. 

\section{2D Supersonic Flow} 
\label{appendix:cylinder}
This appendix provides supplementary information for the simulation of supersonic flow around a circular cylinder, as introduced in \cref{section:supersonic_flow}. 
The computational domain is a rectangular region of size $(15r,\, 25r)$, with the cylinder of radius $r = 20$ centered at $(166,\, 149)$.

Here, we present the remaining macroscopic observables---temperature, density, speed, and pressure---that were omitted from the main text for brevity. 
\Cref{appendix:cylinder_500training} shows these fields for the case where \LBNN{} is trained on the first $500$ time steps and visualized at \(t=700\). 
In the same configuration, \Cref{appx:cylinder_metrics} reports a separate $500$-step diagnostic probe initialized at \(t_0=500\) and evaluated through \(t=999\). 
\Cref{appendix:cylinder_150training} presents analogous results for the long-horizon experiment presented in \cref{section:evaluating_distant_time}, in which \LBNN{} is trained on only the first $150$ time steps (as described in \cref{subsub:cylinder_long}) and tested from the initial condition at $t_0 = 900$ to predict the following $100$ time steps.

\subsection{Cylinder Evaluation Protocol and Metrics}
\label{appx:cylinder_metrics}
The cylinder diagnostics in \Cref{fig:cylinder_main} are computed from a $500$-step autoregressive rollout initialized from the stored trajectory state at \(t=500\) and evaluated through final time index \(t=999\). The reference is the corresponding simulation using the exponential \(\Geqlattice_i\) closure described in \Cref{eq:Geq_Levermore}.

We use the following metric definitions.
\begin{itemize}[leftmargin=*]
    \item \textbf{Stable horizon, positivity violations, and extrema.} The stable horizon is the number of valid rollout steps completed before any non-finite value, \(\rho\le 0\), or \(T\le 0\). We also record the minimum density and temperature over the rollout.
    \item \textbf{Bow-shock position and standoff.} The bow-shock position is detected on the centerline pressure profile as the midpoint of the largest absolute pressure-gradient jump upstream of the cylinder nose, after applying a three-point smoothing stencil. The standoff distance is the distance, in grid cells, from the cylinder nose to this detected shock location. A zero-cell error means that the same discrete detector returns the same cell-level shock location for \LBNN{} and the reference.
    \item \textbf{Pressure jump and pressure ratio.} These are computed from centerline pressure averages in short upstream and downstream windows around the detected bow shock, separated from the shock by a one-cell gap. We report both the raw jump/ratio and their relative errors against the reference.
    \item \textbf{Bow-shock aligned centerline error.} This shifts the predicted final centerline profiles by the detected bow-shock offset and then averages relative profile errors over \(\rho\), \(u_x\), \(T\), \(P\), and local Mach number on the pre-cylinder centerline interval. This measures shock-shape fidelity after removing any detected phase offset.
    \item \textbf{Final extrema.} The maximum Mach number and pressure are computed over the fluid cells at the final reported time.
\end{itemize}

\subsection{Prediction Results for Training on the First 500 Time-Steps}
\label{appendix:cylinder_500training}

\begin{figure}[htb!]
    \centering
    \includegraphics[width=\linewidth]{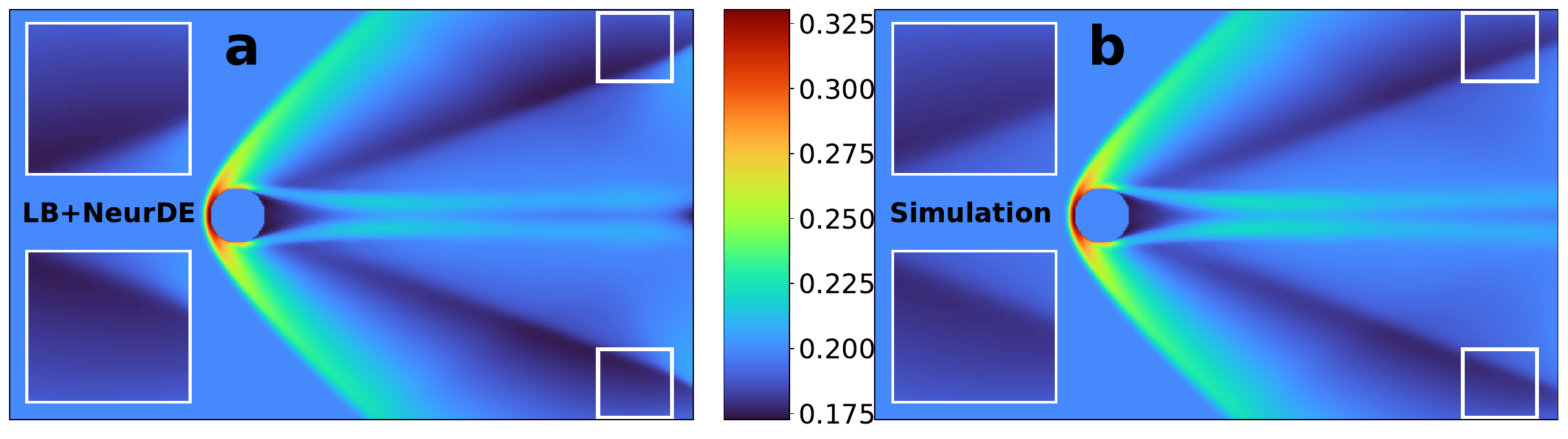}
    \caption{The \LBNN{} \emph{temperature} prediction at $t=700$ for the 2D supersonic flow experiment when trained on the first 500 time-steps and initialized at $t_0=500$.}
    \label{fig:Supersonic_Flow_cylinder_Temp}
\end{figure}

\begin{figure}[htb!]
    \centering
    \includegraphics[width=\linewidth]{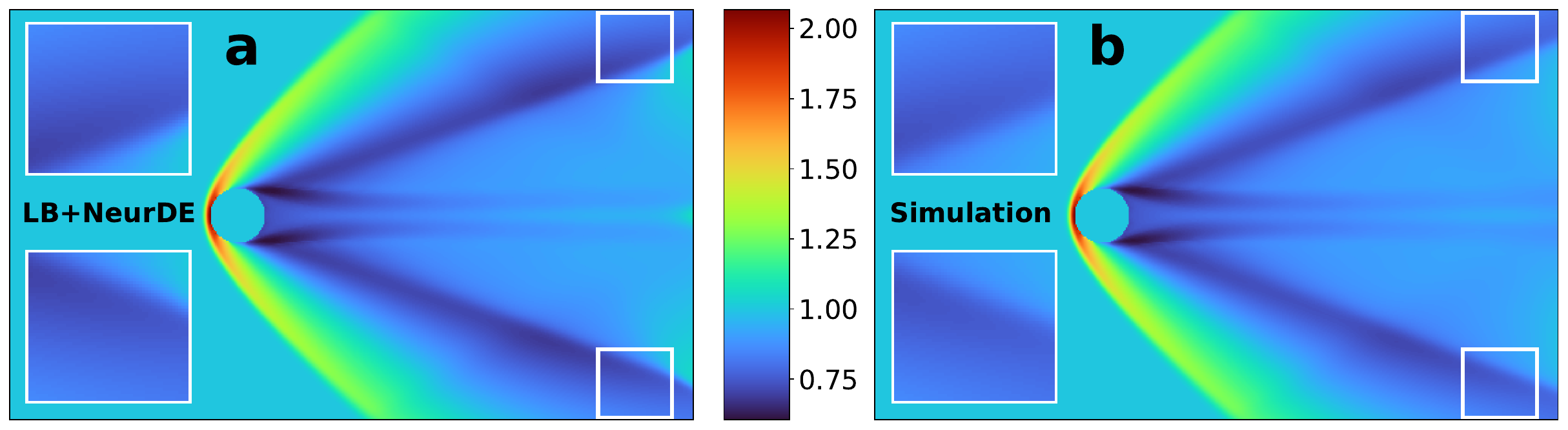}
    \caption{The \LBNN{} \emph{density} prediction at $t=700$ for the 2D supersonic flow experiment when trained on the first 500 time-steps and initialized at $t_0=500$.}
    \label{fig:Supersonic_Flow_cylinder_density}
\end{figure}

\begin{figure}[ht!]
    \centering
    \includegraphics[width=\linewidth]{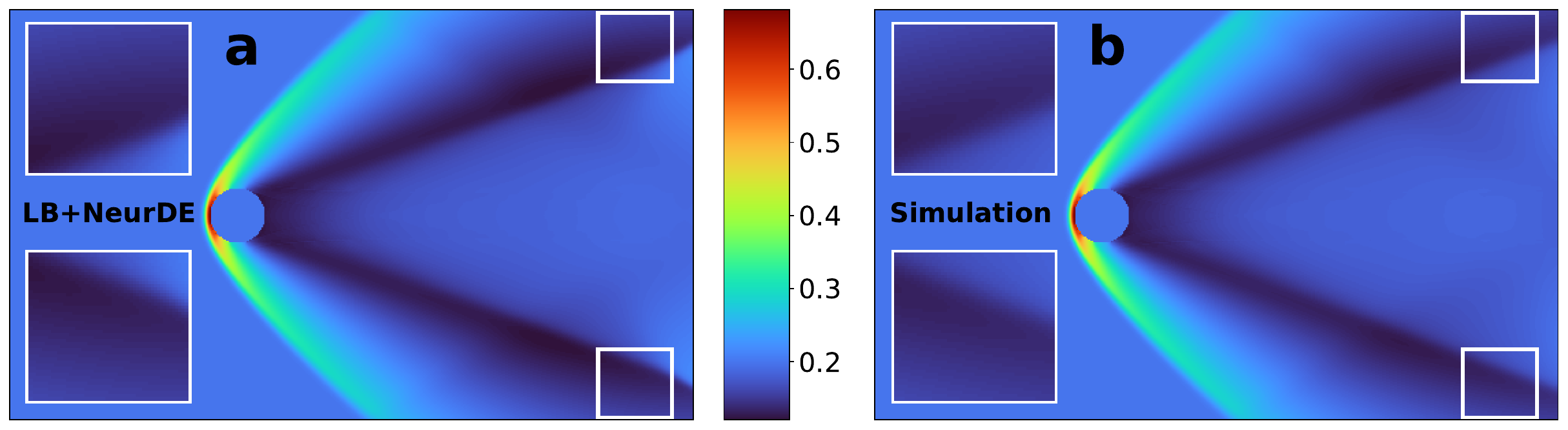}
    \caption{The \LBNN{} \emph{pressure} prediction at $t=700$ for the 2D supersonic flow experiment when trained on the first 500 time-steps and initialized at $t_0=500$.}
    \label{fig:Supersonic_Flow_cylinder_pressure}
\end{figure}

\begin{figure}[ht!]
    \centering
    \includegraphics[width=\linewidth]{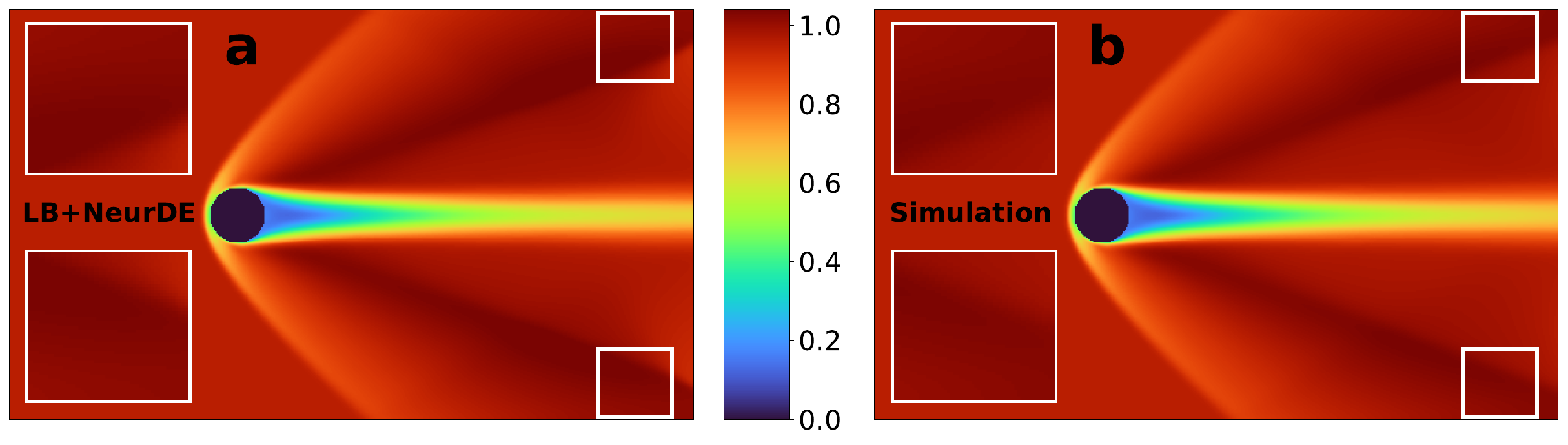}
    \caption{The \LBNN{} \emph{speed} ($\sqrt{\Velocity\cdot \Velocity}$) prediction at $t=700$ for the 2D supersonic flow experiment when trained on the first 500 time-steps and initialized at $t_0=500$.}
    \label{fig:Supersonic_Flow_cylinder_speed}
\end{figure}

\begin{figure}[ht!]
    \centering
    \includegraphics[width=\linewidth]{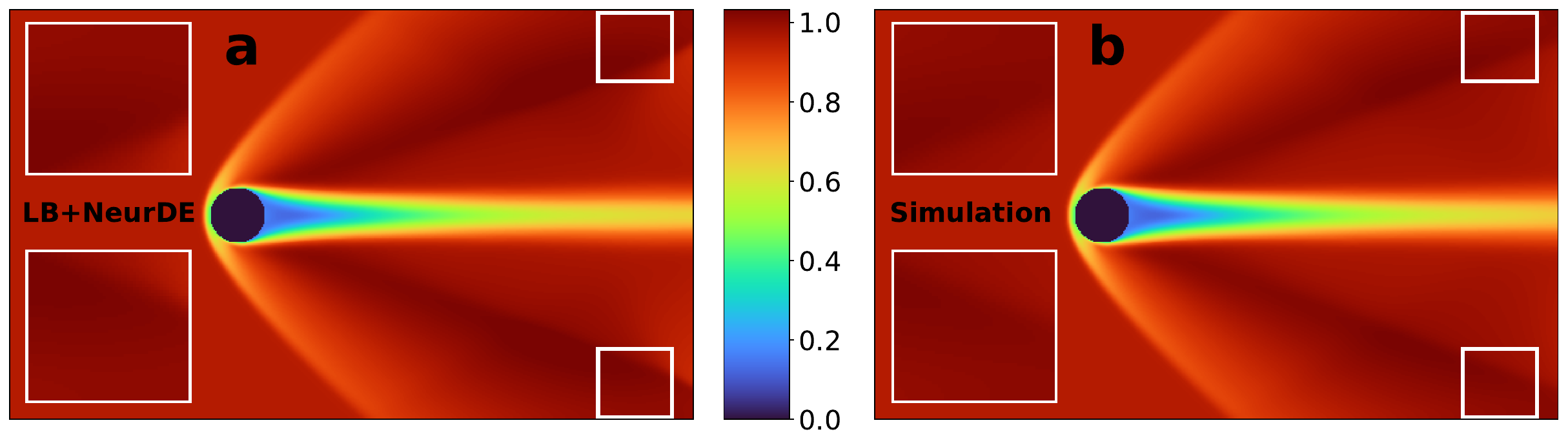}
    \caption{The \LBNN{} \emph{$x$ velocity} $(\Velocity_x)$ prediction at $t=700$ for the 2D supersonic flow experiment when trained on the first 500 time-steps and initialized at $t_0=500$.}
    \label{fig:Supersonic_Flow_cylinder_ux}
\end{figure}

\begin{figure}[ht!]
    \centering
    \includegraphics[width=\linewidth]{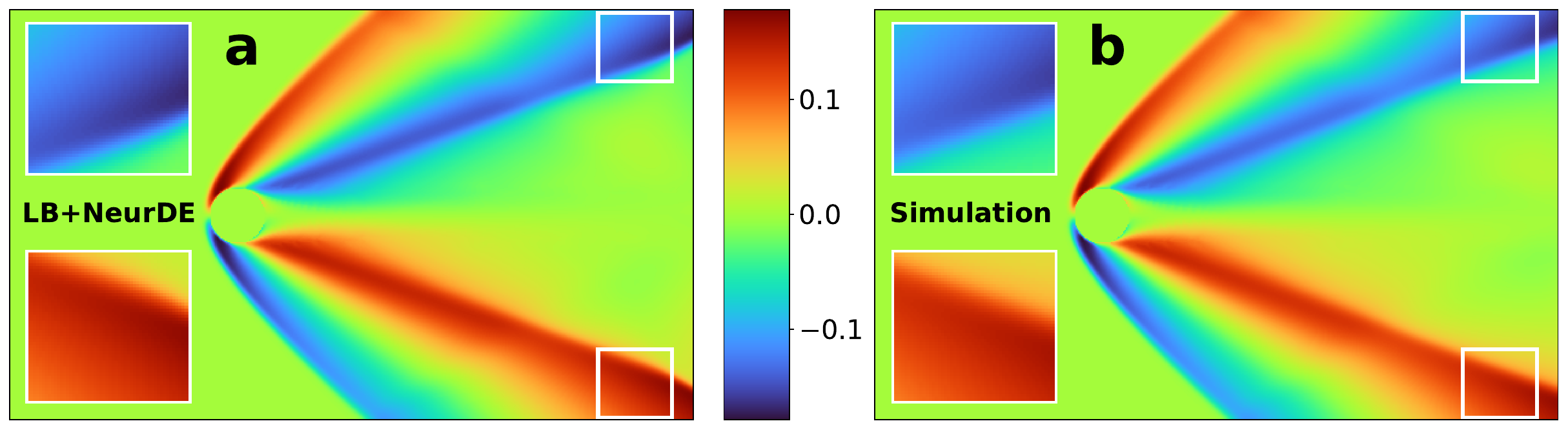}
    \caption{The \LBNN{} \emph{$y$ velocity} $(\Velocity_y)$ prediction at $t=700$ for the 2D supersonic flow experiment when trained on the first 500 time-steps and initialized at $t_0=500$.}
    \label{fig:Supersonic_Flow_cylinder_uy}
\end{figure}

We present further macroscopic results for the 2D cylinder under the same experimental constraints as \cref{section:supersonic_flow}.
\LBNN{} is trained on the first 500 time-steps and then predicts 200 time-steps into the future after being initialized at $t_0=500$.
When comparing the predictions at $t=700$, depicted in \cref{fig:Supersonic_Flow_cylinder_Temp,fig:Supersonic_Flow_cylinder_density,fig:Supersonic_Flow_cylinder_speed,fig:Supersonic_Flow_cylinder_pressure,fig:Supersonic_Flow_cylinder_ux,fig:Supersonic_Flow_cylinder_uy}, we see that \LBNN{} and the reference model exhibit good qualitative agreement.
We observe slight deviations near the right boundary of the outlet as highlighted in the insets.

\FloatBarrier

\FloatBarrier
\subsection{Results for Long-Term Predictions}
\label{appendix:cylinder_150training}

\begin{figure}[ht!]
    \centering
    \includegraphics[width=\linewidth]{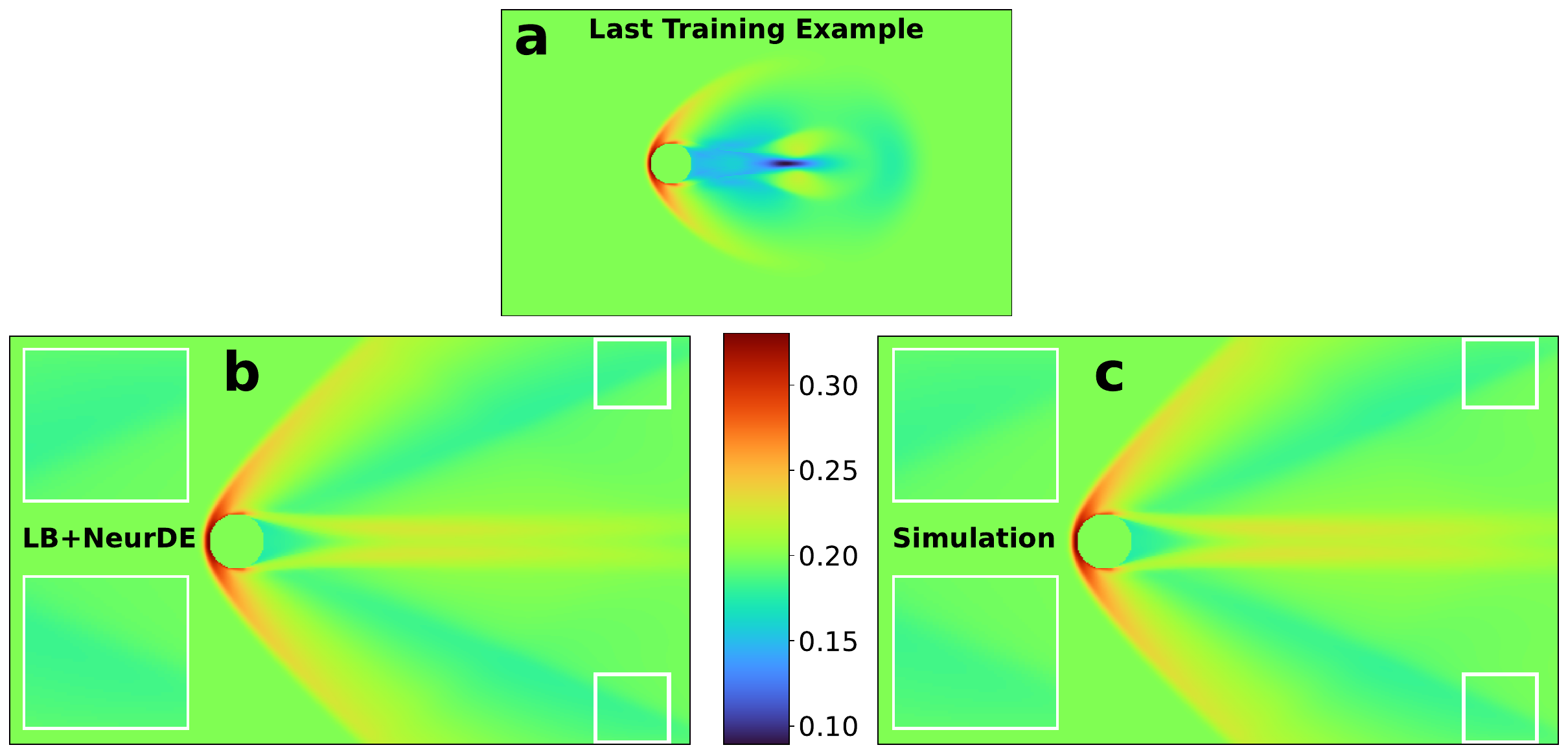}
    \caption{The \LBNN{} \emph{temperature} prediction at $t=999$ for the 2D supersonic flow experiment when trained on the first 150 time-steps and initialized at $t_0=900$.}
    \label{fig:Supersonic_Flow_cylinder_long_Temp}
\end{figure}

\begin{figure}[ht!]
    \centering
    \includegraphics[width=\linewidth]{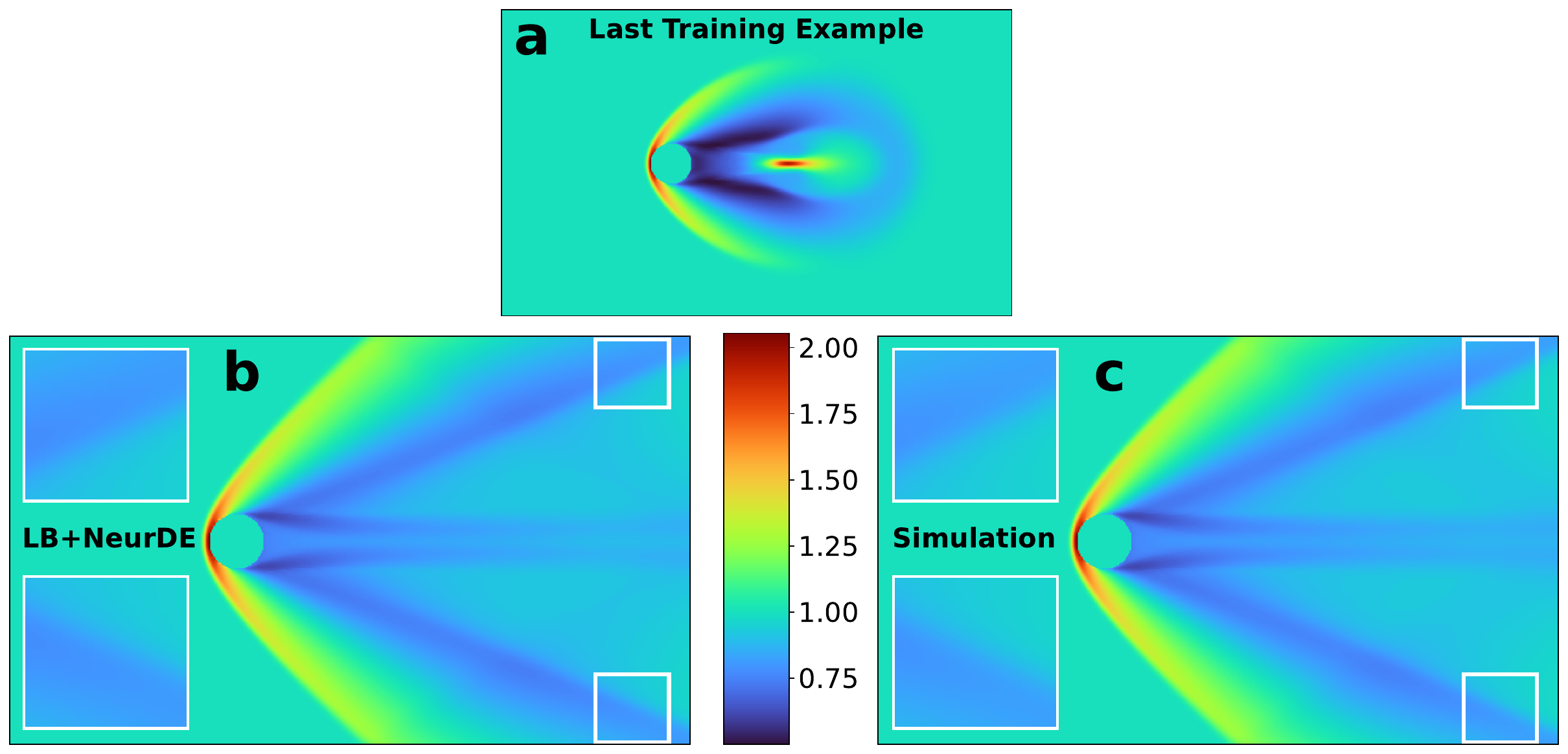}
    \caption{The \LBNN{} \emph{density} prediction at $t=999$ for the 2D supersonic flow experiment when trained on the first 150 time-steps and initialized at $t_0=900$.}
    \label{fig:Supersonic_Flow_cylinder_long_density}
\end{figure}

\begin{figure}[ht!]
    \centering
    \includegraphics[width=\linewidth]{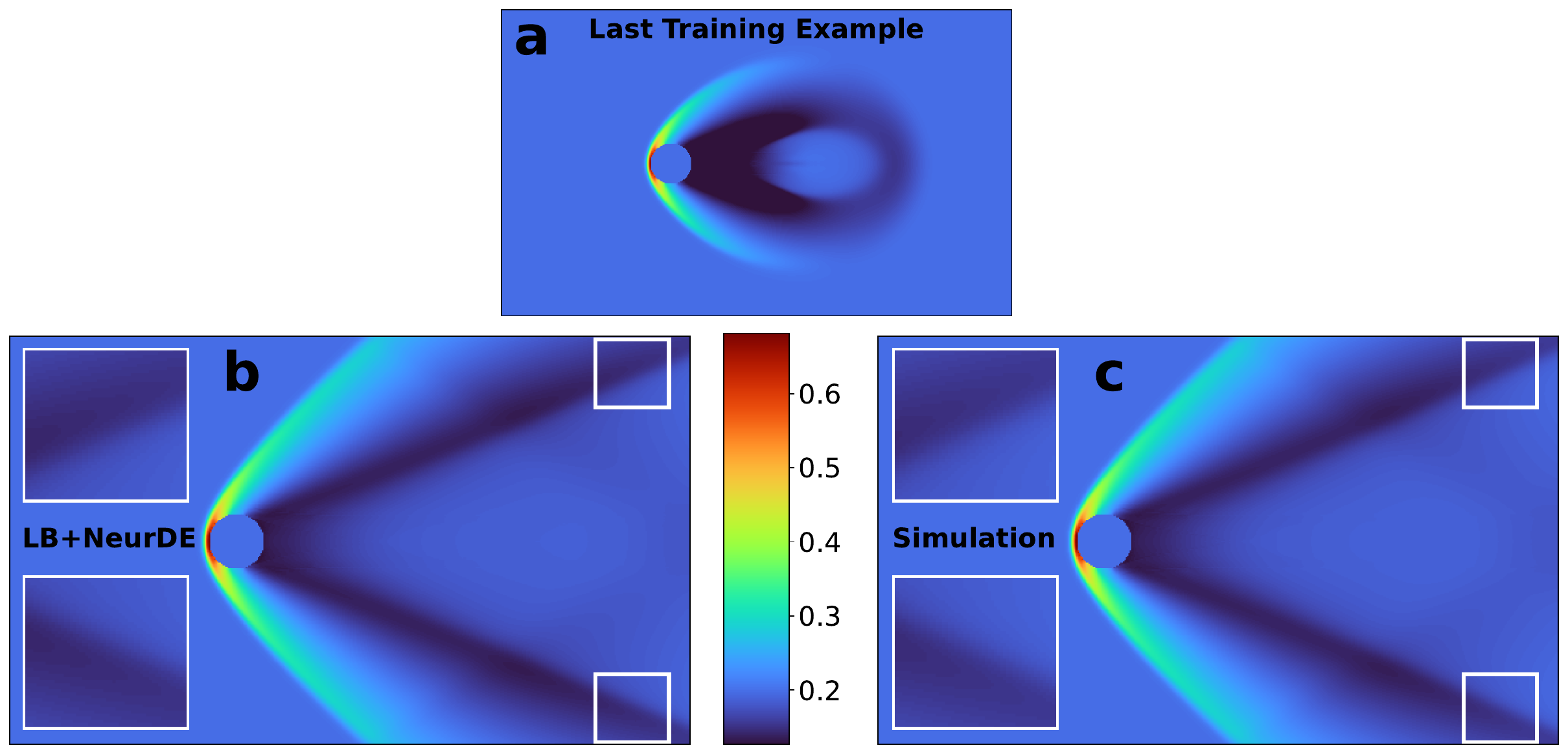}
    \caption{The \LBNN{} \emph{pressure} prediction at $t=999$ for the 2D supersonic flow experiment when trained on the first 150 time-steps and initialized at $t_0=900$.}
    \label{fig:Supersonic_Flow_cylinder_long_pressure}
\end{figure}

\begin{figure}[ht!]
    \centering
    \includegraphics[width=\linewidth]{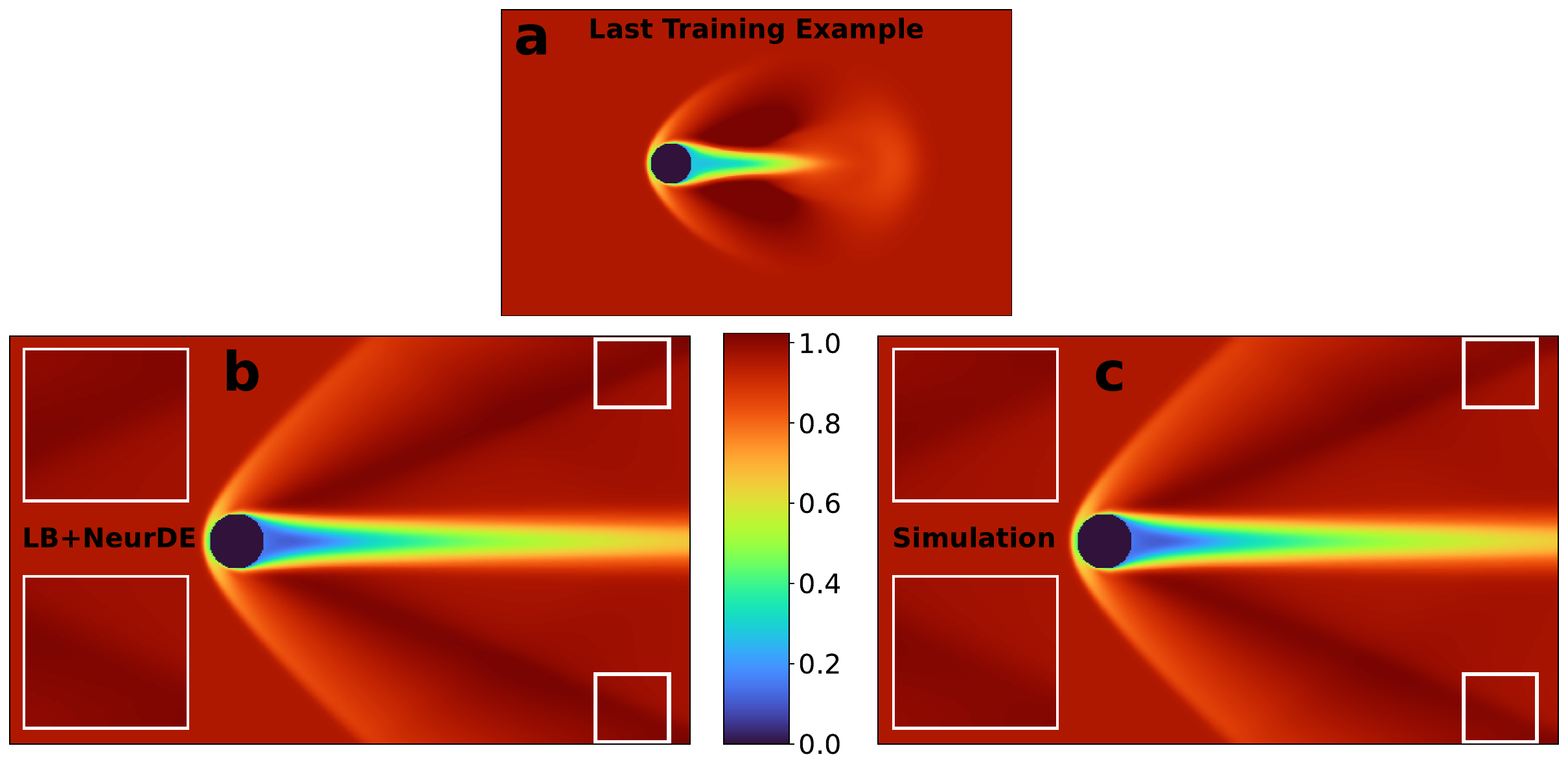}
    \caption{The \LBNN{} \emph{speed} prediction at $t=999$ for the 2D supersonic flow experiment when trained on the first 150 time-steps and initialized at $t_0=900$.}
    \label{fig:Supersonic_Flow_cylinder_long_speed}
\end{figure}

\begin{figure}[ht!]
    \centering
    \includegraphics[width=\linewidth]{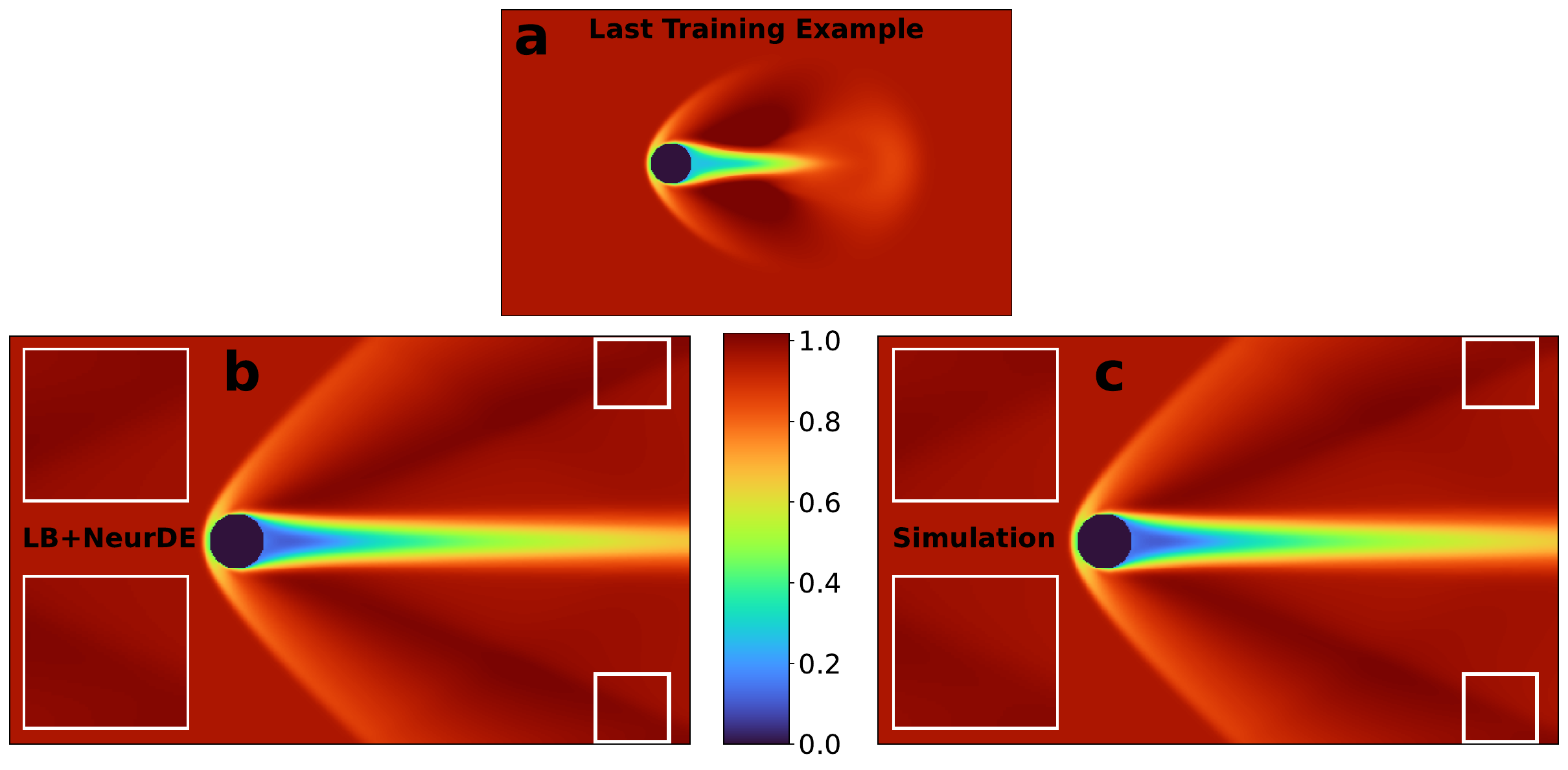}
    \caption{The \LBNN{} \emph{$x$ velocity} prediction at $t=999$ for the 2D supersonic flow experiment when trained on the first 150 time-steps and initialized at $t_0=900$.}
    \label{fig:Supersonic_Flow_cylinder_long_ux}
\end{figure}

\begin{figure}[ht!]
    \centering
    \includegraphics[width=\linewidth]{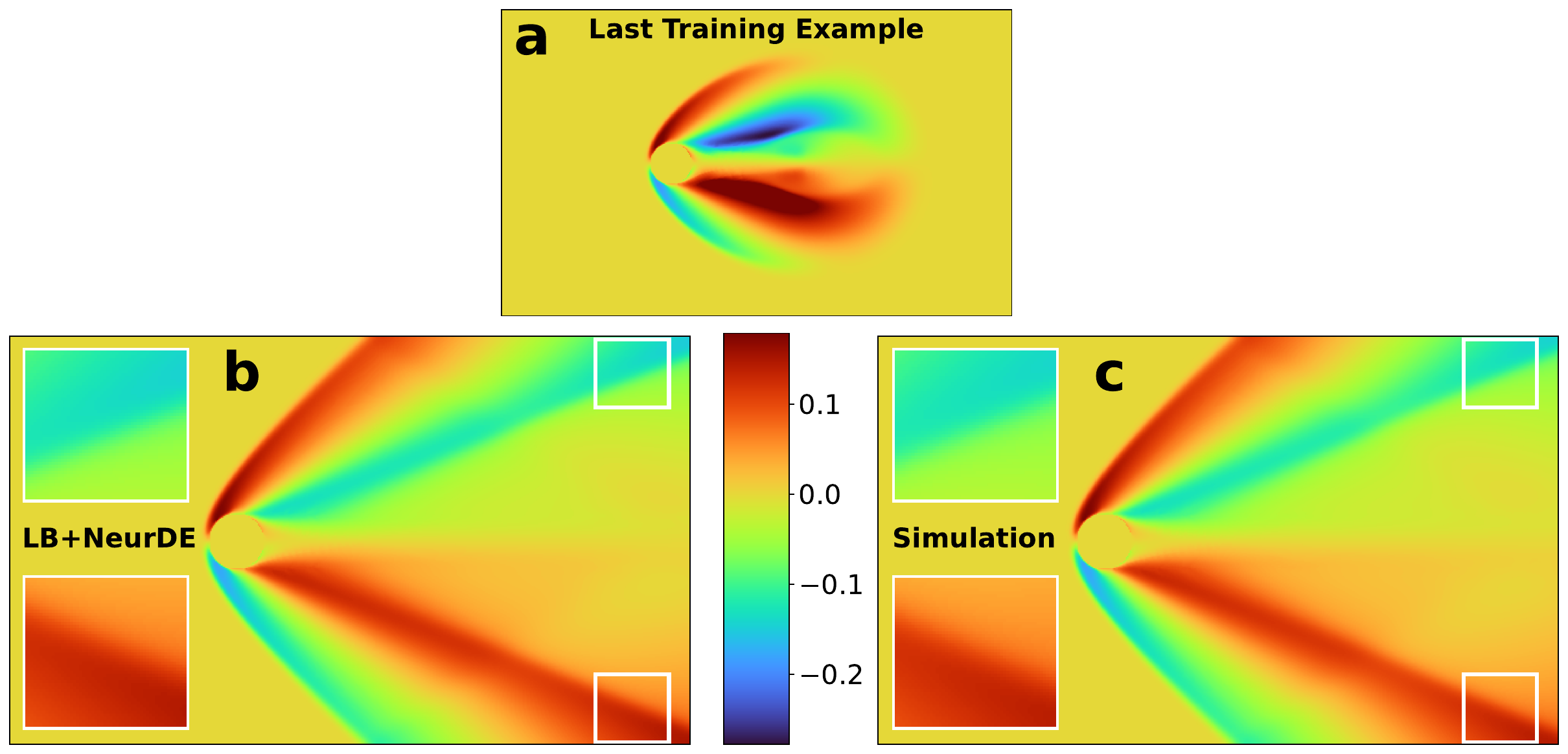}
    \caption{The \LBNN{} \emph{$y$ velocity} prediction at $t=999$ for the 2D supersonic flow experiment when trained on the first 150 time-steps and initialized at $t_0=900$.}
    \label{fig:Supersonic_Flow_cylinder_long_uy}
\end{figure}

We present further macroscopic results for the 2D cylinder under the same experimental constraints as \cref{subsub:cylinder_long}.
\LBNN{} is trained on the first 150 time-steps and then predicts 100 time-steps into the future after being initialized at $t_0=900$.
When comparing the predictions at $t=999$, depicted in \cref{fig:Supersonic_Flow_cylinder_long_Temp,fig:Supersonic_Flow_cylinder_long_density,fig:Supersonic_Flow_cylinder_long_speed,fig:Supersonic_Flow_cylinder_long_pressure,fig:Supersonic_Flow_cylinder_long_ux,fig:Supersonic_Flow_cylinder_long_uy}, \LBNN{} and the reference model again exhibit good qualitative agreement.
We observe slight deviations near the right boundary of the outlet as highlighted in the insets.

\FloatBarrier

\clearpage
\FloatBarrier

\section{Finite-Volume DUGKS Realization}
\label{appx:fvfd_dugks}

NeurDE is intended as a learned local equilibrium interface rather than as a
lattice-specific transport rule. In on-lattice LB, transport is an exact memory
shift; with shifted or off-lattice velocities it is a fixed interpolation or
reconstruction step. Either way, the transport is prescribed and comparatively
simple, while the learned object is the local closure. To make this point
concrete, we also ran a finite-volume, face-based thermal-compressible
realization based on the reduced DUGKS formulation of Guo, Wang, and
Xu~\cite{guo2015dugks_thermal}. In this setting the learned object is a local
Shakhov-type face closure~\cite{shakhov1968generalization} rather than a
lattice equilibrium evaluated inside a pure collide--stream scheme. Thus the
experiment tests whether the same equilibrium-learning interface can be queried
by a fixed FV transport update.

The implementation is a one-dimensional finite-volume thermal DUGKS with reduced distributions \(g(x,\xi,t)\) and \(h(x,\xi,t)\), Shakhov--BGK collision, a uniform Newton--Cotes velocity quadrature, and Van-Leer reconstruction at faces. The reported Sod test uses \(n_x=100\) cells, \(201\) velocities on \([-10,10]\), \(\gamma=1.4\), \(\mathrm{Pr}=2/3\), \(\mu_{\mathrm{ref}}=10^{-5}\), viscosity exponent \(1/2\), and \(\mathrm{CFL}=0.95\). The comparison in \Cref{fig:fvfd_dugks_shakhov_profiles} uses a learned-face mode: the cell-source Shakhov pair remains analytic, while the face-source pair is replaced by a learned correction evaluated from local features built from \(\log \rho\), \(u\), \(\log T\), \(\log \tau\), and a scaled heat-flux observable through a small residual network.

The profile comparison covers density, velocity, temperature, and pressure at \(t\approx 0.05, 0.10, 0.15,\) and \(0.20\), corresponding to snapshot steps \(58,116,174,\) and \(232\) in the plotting script. We compare the exact Sod reference solution, the analytic Shakhov face closure, and the learned-face Shakhov closure. Across all four observables, the learned-face profiles remain close to the analytic baseline and to the reference solution. The most discriminating feature is the temperature overshoot near the shock/contact structure, where the learned-face closure preserves the non-monotone thermal profile comparably to the analytic model and, in several snapshots, tracks the reference transition slightly more sharply. Density, velocity, and pressure overlap at the plotted resolution between the analytic and learned closures over the displayed horizon.

\begin{figure}[H]
    \centering
    \includegraphics[width=\linewidth]{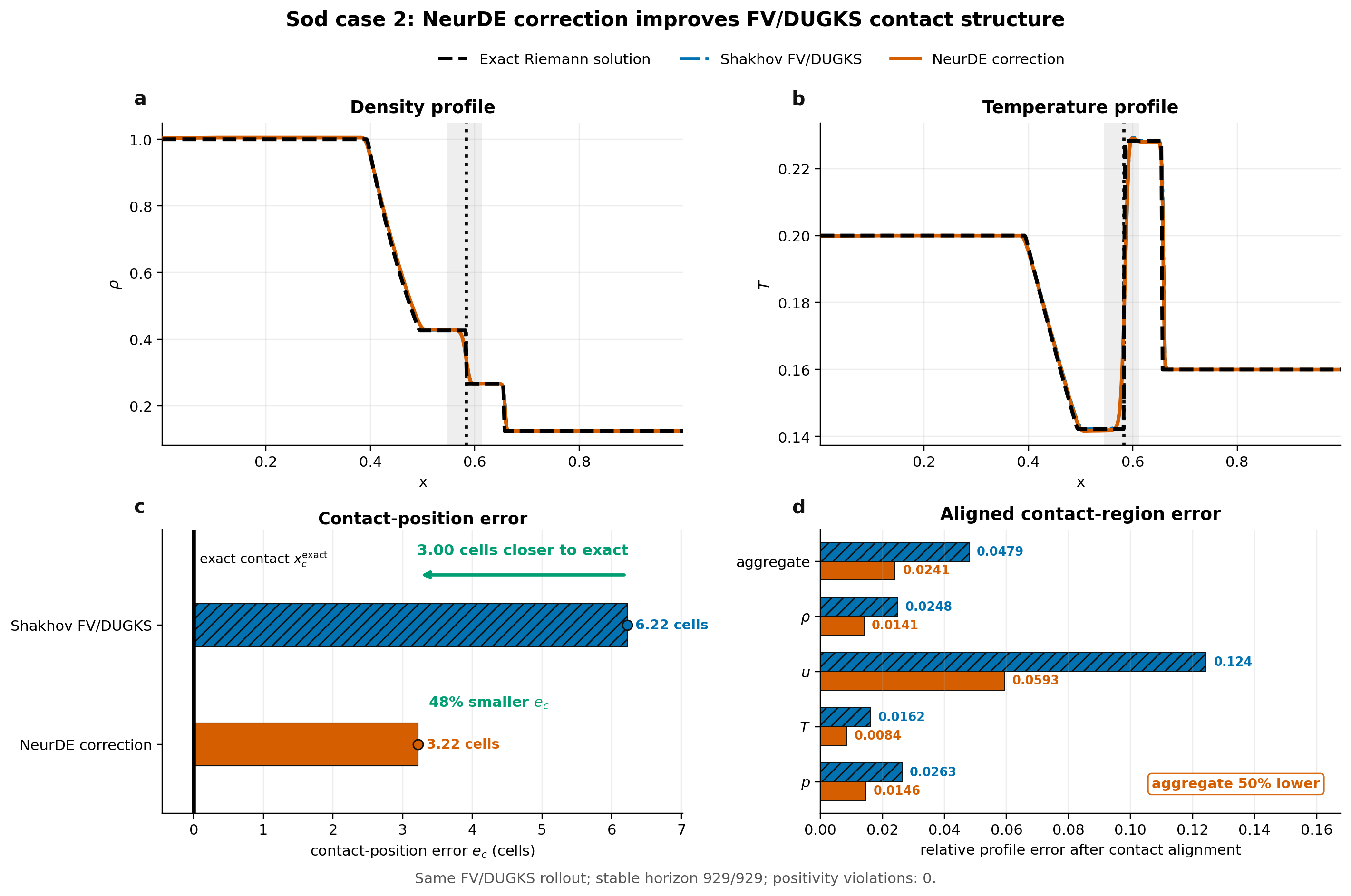}
    \caption{FV/DUGKS Sod case-2 contact-profile zoom for the exact-trained NeurDE correction. The top panels show the full density and temperature profiles, while the lower panels center coordinates at the exact contact and mark the contact locations inferred from analytic Shakhov and the NeurDE correction. The correction moves the contact marker \(3.00\) cells closer to the exact Riemann contact and lowers the contact-aligned profile error by about \(50\%\), while preserving a \(929/929\) stable horizon with zero positivity violations. Panel~(c) reports the contact-position error \(e_c = |\hat{x}_c - x_c^{\mathrm{exact}}|/\Delta x\). The detected contact location \(\hat{x}_c\) is the maximizer, over the contact window \(W_c\), of \(|\partial_x\tilde{\rho}|/(|\partial_x\tilde{p}|/P_c + 10^{-6})\), with \(P_c=\max_{W_c}|\partial_x\tilde{p}|\).}
    \label{fig:fvfd_dugks_case2_contact_correction}
\end{figure}

\begin{figure}[H]
    \centering
    \includegraphics[width=\linewidth]{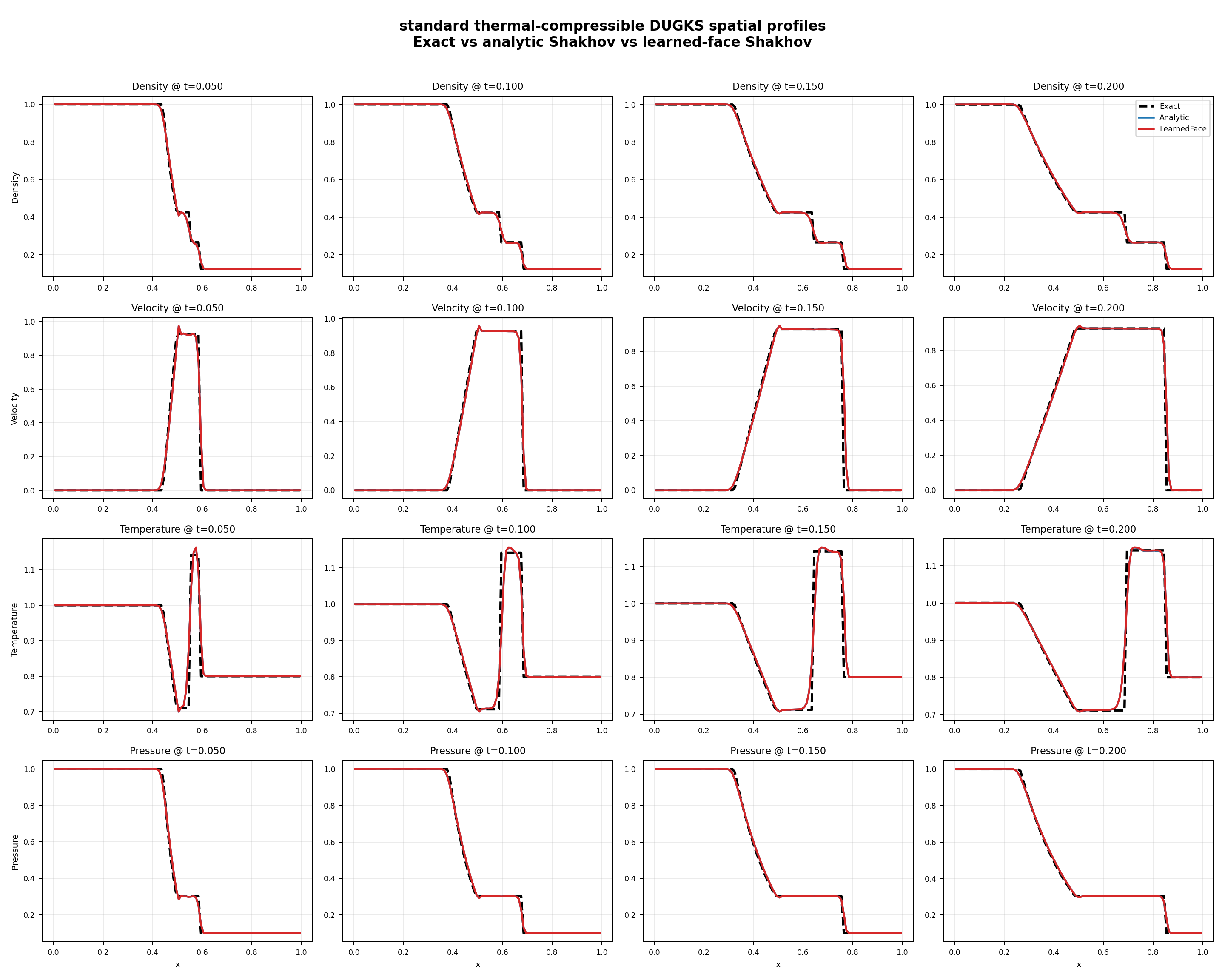}
    \caption{Finite-volume, face-based thermal-compressible DUGKS realization with a learned Shakhov face closure. Columns show spatial profiles at \(t\approx 0.05, 0.10, 0.15,\) and \(0.20\); rows report density, velocity, temperature, and pressure. The black dashed curve is the exact reference solution, the blue curve is the analytic Shakhov face closure, and the red curve is the learned-face Shakhov closure. The learned-face profiles remain close to the analytic baseline and to the reference solution across the displayed horizon, with the clearest differences appearing in the temperature overshoot and discontinuity locations.}
    \label{fig:fvfd_dugks_shakhov_profiles}
\end{figure}

\Cref{tab:fvfd_dugks_metrics} adds wave-structure diagnostics using the same
style of front-location and aligned-profile metrics as the LB Sod tests. The
case-2 learned-face rollout remains stable for the full \(929\)-step probe,
with zero positivity violations, minimum density \(0.1250\), and minimum
temperature \(0.1419\). Its shock and rarefaction-tail errors are both below
one cell; the larger residual is the contact location, while the
shock-aligned profile error remains \(9.3\times 10^{-3}\). To test whether the
learned FV closure can improve rather than only match the analytic face
closure, we also train a separate exact-supervised NeurDE face-source
correction on the identical case-2 FV/DUGKS discretization. This correction
reduces the contact error from \(6.22\) to \(3.22\) cells and the
contact-aligned profile error from \(4.79\times 10^{-2}\) to
\(2.41\times 10^{-2}\), while preserving zero positivity violations and the
full \(929/929\) stable horizon. The corresponding contact-profile zoom in
\Cref{fig:fvfd_dugks_case2_contact_correction} visualizes the three-cell
reduction in contact-location error without introducing a new positivity
failure. The standard Sod probe shows the same qualitative behavior over its
\(232\)-step evaluation horizon. We also record the domain-integrated drift
diagnostic produced by the evaluator, but for this non-periodic FV Riemann
setup it is used as a rollout diagnostic rather than as a periodic-domain
conservation theorem.

\begin{table}[H]
    \centering
    \scriptsize
    \renewcommand{\arraystretch}{0.95}
    \begin{tabular}{@{}llcccccccc@{}}
        \toprule
        Probe & Closure & Shock & Contact & Raref. & Plateau & Aligned S/C & Stable & Pos. viol. & Runtime \\
        & & [cells] & [cells] & [cells] & error & error & horizon & & [s] \\
        \midrule
        Case 2 & Exact-trained corr. & 0.6313 & 3.2223 & 0.4786 & 0.0014 & 0.0091/0.0241 & 929/929 & 0 & 1.68 \\
        Case 2 & Learned face & 0.6313 & 5.7777 & 0.4786 & 0.0015 & 0.0093/0.0570 & 929/929 & 0 & 1.62 \\
        Case 2 & Analytic face & 0.6313 & 6.2223 & 0.4786 & 0.0011 & 0.0091/0.0479 & 929/929 & 0 & 1.51 \\
        Standard & Learned face & 0.1068 & 4.5828 & 1.0920 & 0.0017 & 0.0142/0.1042 & 232/232 & 0 & 0.45 \\
        Standard & Analytic face & 0.1068 & 5.5828 & 1.0920 & 0.0017 & 0.0141/0.1257 & 232/232 & 0 & 0.38 \\
        \bottomrule
    \end{tabular}
    \caption{Finite-volume DUGKS Sod wave-structure diagnostics for learned-face, exact-trained correction, and analytic Shakhov closures. Front errors are reported in grid cells. The learned-face closure matches the analytic face closure closely while preserving positivity over the evaluated horizons; the exact-supervised correction improves the case-2 contact structure without reducing the stable horizon.}
    \label{tab:fvfd_dugks_metrics}
\end{table}

This appendix result establishes the FV/DUGKS form of the solver-interface
test: the NeurDE closure is evaluated locally at faces inside a finite-volume
kinetic update rather than streamed exactly along lattice links. Once a
transport discretization supplies local states, the same learned equilibrium
interface can therefore be queried outside an LB collide--stream update, and
the case-2 correction improves the contact structure over analytic Shakhov on
the matched FV/DUGKS rollout.

\FloatBarrier

\section{Additional Conservation Laws}
\label{appx:additional_conservation_laws}

To further illustrate that the proposed framework is not restricted to the compressible flow studied in the main text, we consider additional one-dimensional scalar conservation laws, namely the inviscid Burgers equation, a first LWR traffic-flow example, and a first Buckley--Leverett rollout. These examples show that the NeurDE construction extends naturally beyond the Euler setting, while providing a particularly transparent setting in which to compare the raw equilibrium ansatz with its conservative counterpart (see \Cref{appx:conservative_neurde}).

\subsection{Inviscid Burgers Equation}
\label{appx:burgers_setup_conservative}

We consider the one-dimensional inviscid Burgers equation \cite{leveque2002finite}
\[
\partial_t u+\partial_x\!\left(\frac{u^2}{2}\right)=0,
\qquad x\in[0,L],
\]
with periodic boundary conditions. In the sinusoidal tests used here, the initial condition is
\[
u(0,x)=u_0(x)
=
\bar u + A \sin\!\left(\frac{2\pi k x}{L}+\varphi\right),
\]
where \(\bar u\) is the mean state, \(A\) is the amplitude, \(k\) is the wave number, and \(\varphi\) is the phase. For smooth data of this form, the characteristic steepening time is
\[
t_{\mathrm s}=\frac{L}{2\pi |A|k},
\]
which marks the onset of shock formation.

At the kinetic level, the equilibrium populations
\[
\Feqlattice(u)=\bigl(\Feqlattice_1(u),\dots,\Feqlattice_Q(u)\bigr)^\top
\]
must reproduce both the Burgers conserved variable and its flux. Writing \(c_i\) for the discrete velocity associated with population \(i\), the equilibrium constraints are
\[
\sum_{i=1}^Q \Feqlattice_i(u)=u,
\qquad
\sum_{i=1}^Q c_i\,\Feqlattice_i(u)=\frac{u^2}{2}.
\]
Equivalently, defining
\[
\bm{C}_{\mathrm B}
=
\begin{bmatrix}
1 & \cdots & 1\\
c_1 & \cdots & c_Q
\end{bmatrix},
\qquad
\bm{U}_{\mathrm B}(u)
=
\begin{bmatrix}
u\\[2mm]
\frac12 u^2
\end{bmatrix},
\]
the Burgers equilibrium condition is
\[
\bm{C}_{\mathrm B}\Feqlattice(u)=\bm{U}_{\mathrm B}(u).
\]
The initial kinetic state is then taken as
\[
\bm{F}(0,x)=\Feqlattice\bigl(u_0(x)\bigr),
\]
or equivalently from the dataset populations associated with the same macroscopic initialization.

In the non-conservative NeurDE variant, the model outputs a raw positive equilibrium ansatz \cref{eq:levermore_closure_NN}
\[
\tilde{\Feqlattice}(u)
=
\bigl(\tilde{\Feqlattice}_1(u),\dots,\tilde{\Feqlattice}_Q(u)\bigr)^\top,
\qquad
\tilde{\Feqlattice}_i(u)
=
\exp\!\bigl(\underbalpha(u)\cdot \underbvarphi(\latticevelocity_i)\bigr),
\qquad i=1,\dots,Q.
\]
which guarantees positivity but does not, in general, satisfy the Burgers moment constraints exactly:
\[
\bm{C}_{\mathrm B}\tilde{\Feqlattice}(u)\neq \bm{U}_{\mathrm B}(u).
\]
The conservative variant is obtained by enforcing the Burgers moments exactly using the general conservative correction described in \cref{appx:conservative_neurde}. In the present scalar setting, this means specializing the general construction there to the Burgers moment operator \(\bm{C}_{\mathrm B}\) and moment vector \(\bm{U}_{\mathrm B}(u)\). Thus the conservative equilibrium satisfies
\[
\bm{C}_{\mathrm B}\Feqlattice(u)=\bm{U}_{\mathrm B}(u),
\]
while remaining close to the raw positive ansatz in the entropic sense described in \cref{appx:conservative_neurde}. We refer to \cref{appx:conservative_neurde} for the corresponding projection formula, its KL interpretation, the local quadratic approximation, and the discussion of positivity.

\begin{figure}[t]
    \centering
    \includegraphics[width=0.82\linewidth]{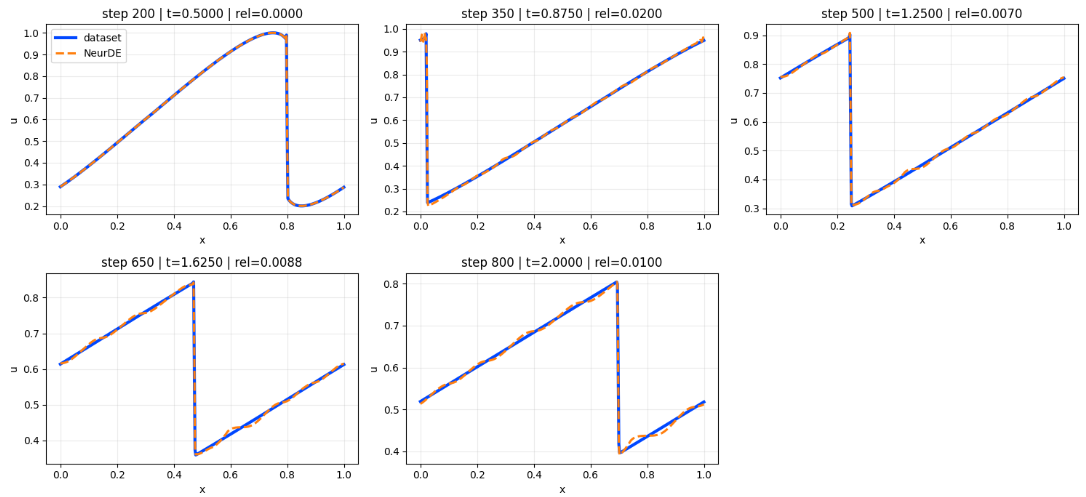}
    
    \vspace{0.8em}
    
    \includegraphics[width=0.82\linewidth]{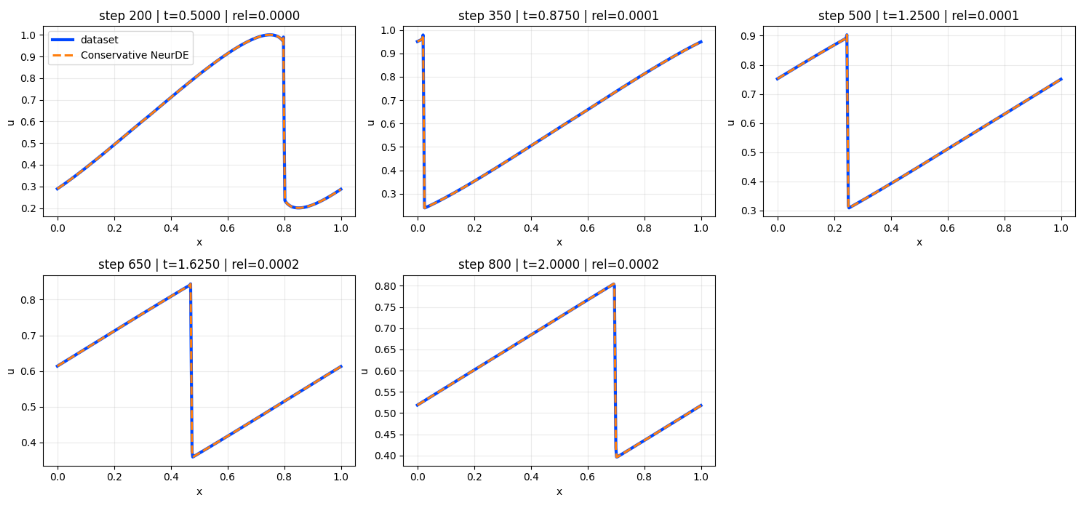}
    \caption{Burgers rollouts for sinusoidal initial data at multiple times after shock formation. \textbf{Top:} standard NeurDE using the raw positive equilibrium ansatz \(\tilde{\Feqlattice}\), which only approximately satisfies the Burgers moment constraints. Even without TVD regularization, the rollout remains accurate and the reported relative errors stay small, although slight discrepancies become visible near the shock and along the smooth branch at later times. \textbf{Bottom:} conservative NeurDE obtained by enforcing the Burgers moment constraints exactly through the conservative correction specialized from \cref{appx:conservative_neurde}. In this case the predicted profiles remain even closer to the dataset throughout the rollout, with uniformly smaller errors and easier optimization during training.}
    \label{fig:burgers_neurde_vs_conservative}
\end{figure}

\paragraph{Burgers rollout comparison.}
\Cref{fig:burgers_neurde_vs_conservative} compares the standard and conservative NeurDE rollouts for sinusoidal Burgers initial data at several times after the onset of shock formation. In both panels, the blue solid line denotes the dataset solution and the orange dashed line denotes the model prediction. The standard NeurDE rollout remains quite accurate, with only small discrepancies visible near the shock and along the smooth branch at later times. Unlike the transonic Sod case in \cref{appendix:Sod_case_2}, we do not use a TVD regularizer \cref{eq:TVD_condition} in these Burgers experiments. The conservative NeurDE variant is consistently closer to the dataset and also converges more readily in training, indicating that exact enforcement of the Burgers moments improves both optimization and long-horizon accuracy without requiring additional TVD-type stabilization.

\subsection{LWR Traffic Flow}
\label{appx:lwr_setup_conservative}

We also tested the conservative construction on a first one-dimensional LWR (Lighthill--Whitham--Richards) traffic-flow rollout \cite{leveque2002finite}, written in conservative form as
\[
\partial_t \rho+\partial_x Q(\rho)=0,
\qquad
x\in[0,1],
\]
with scalar density \(\rho\) and the Greenshields flux
\[
Q(\rho)
=
v_{\mathrm{free}}\rho\left(1-\frac{\rho}{\rho_{\max}}\right),
\qquad
\rho_{\max}=1,\quad v_{\mathrm{free}}=1.
\]
The data in this first test use the Riemann initialization \(\rho_L=0.7\), \(\rho_R=0.2\), and \(x_0=0.2\). At the kinetic level, the equilibrium populations must reproduce both the conserved density and its flux, so the conservative correction from \Cref{appx:conservative_neurde} is specialized through
\[
\bm{C}_{\mathrm{LWR}}
=
\begin{bmatrix}
1 & \cdots & 1\\
c_1 & \cdots & c_Q
\end{bmatrix},
\qquad
\bm{U}_{\mathrm{LWR}}(\rho)
=
\begin{bmatrix}
\rho\\[2mm]
Q(\rho)
\end{bmatrix},
\qquad
\bm{C}_{\mathrm{LWR}}\Feqlattice(\rho)=\bm{U}_{\mathrm{LWR}}(\rho).
\]

The resulting conservative rollout remains close to the dataset across the five reported times in \Cref{fig:lwr_conservative_rollout}. The relative error grows from \(0\) at \(t=0.4000\) to \(1.95\times 10^{-2}\) at \(t=0.4975\), while the predicted density profile continues to track the piecewise-linear structure of the reference solution closely. This first LWR test therefore supports the same conclusion as Burgers: exact enforcement of the scalar conserved moments yields a robust discrete-velocity realization beyond the compressible-flow regime.

\IfFileExists{figs/LWR_conservative.png}{
\begin{figure}[t]
    \centering
    \includegraphics[width=\linewidth]{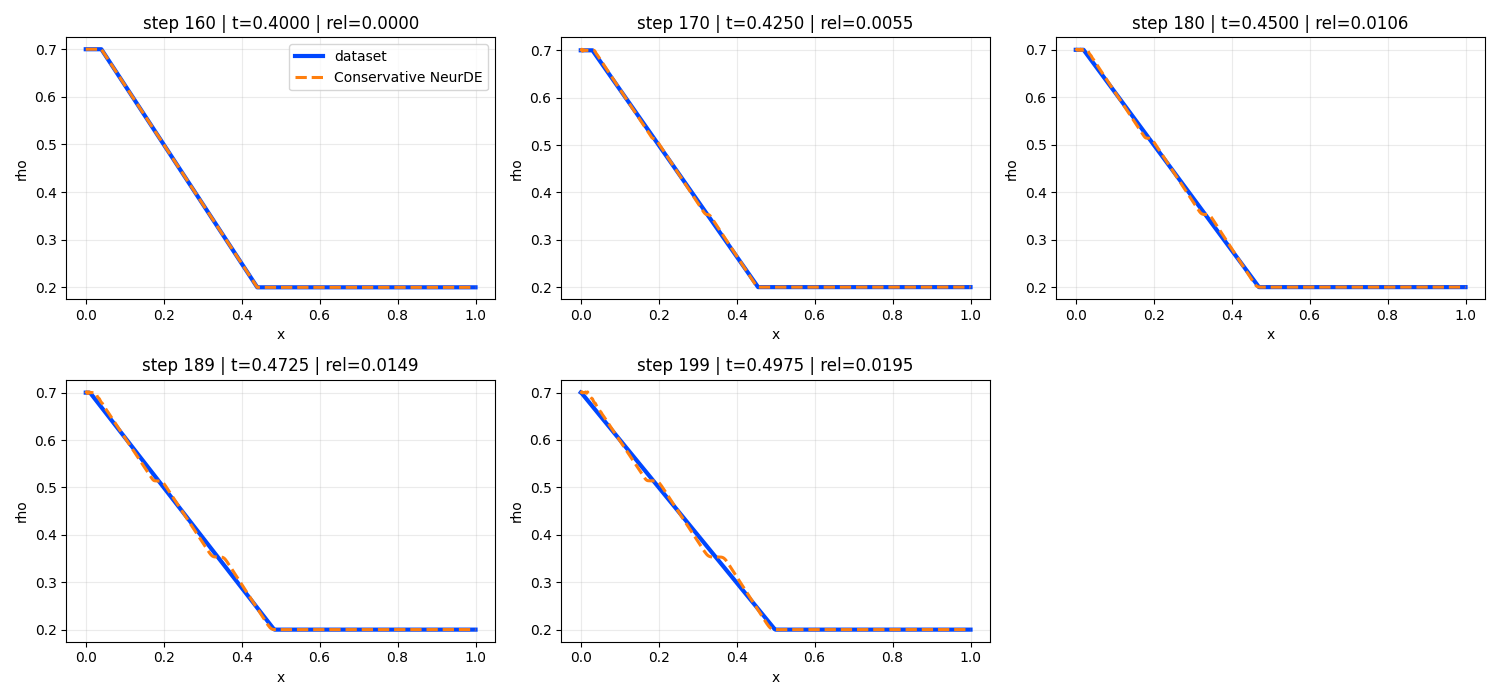}
    \caption{Conservative NeurDE rollout for a first LWR traffic-flow test at five times near \(t=0.5\). The blue solid line denotes the dataset solution and the orange dashed line denotes the conservative NeurDE prediction. Across the reported times \(t=0.4000, 0.4250, 0.4500, 0.4725,\) and \(0.4975\), the relative error increases from \(0\) to \(1.95\times 10^{-2}\), while the predicted density remains close to the reference profile throughout the rollout.}
    \label{fig:lwr_conservative_rollout}
\end{figure}
}{}

\subsection{Buckley--Leverett Flow}
\label{appx:buckley_setup_conservative}

We also tested the conservative construction on a first one-dimensional Buckley--Leverett rollout \cite{leveque2002finite}, governed by
\[
\partial_t u+\partial_x f_{\mathrm{BL}}(u)=0,
\qquad
x\in[0,1],
\]
with scalar saturation \(u\) and fractional-flow flux
\[
f_{\mathrm{BL}}(u)
=
\frac{u^2}{u^2 + M(1-u)^2},
\qquad M=0.5.
\]
The data in this first test use the Riemann initialization \(u_L=1\), \(u_R=0\), and \(x_0=0.2\). As in the Burgers and LWR examples, the equilibrium populations are required to match both the conserved scalar and its flux, so the conservative correction from \Cref{appx:conservative_neurde} is specialized through
\[
\bm{C}_{\mathrm{BL}}
=
\begin{bmatrix}
1 & \cdots & 1\\
c_1 & \cdots & c_Q
\end{bmatrix},
\qquad
\bm{U}_{\mathrm{BL}}(u)
=
\begin{bmatrix}
u\\[2mm]
f_{\mathrm{BL}}(u)
\end{bmatrix},
\qquad
\bm{C}_{\mathrm{BL}}\Feqlattice(u)=\bm{U}_{\mathrm{BL}}(u).
\]

\begin{figure}[t]
    \centering
    \includegraphics[width=\linewidth]{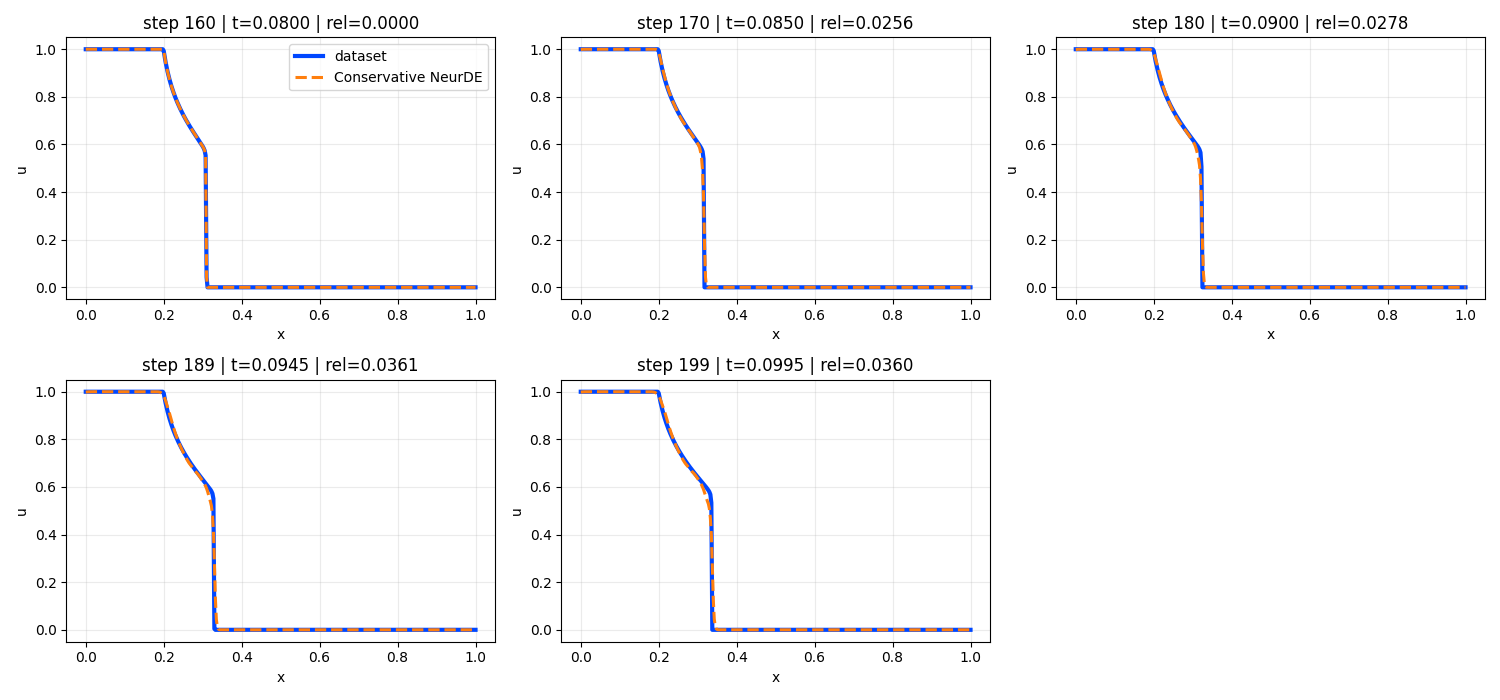}
    \caption{Conservative NeurDE rollout for a first Buckley--Leverett test at five times near \(t=0.1\). The blue solid line denotes the dataset solution and the orange dashed line denotes the conservative NeurDE prediction. Across the reported times \(t=0.0800, 0.0850, 0.0900, 0.0945,\) and \(0.0995\), the relative error remains between \(0\) and \(3.61\times 10^{-2}\), while the predicted profile remains close to the reference solution throughout the rollout.}
    \label{fig:buckley_conservative_rollout}
\end{figure}

The resulting conservative rollout remains close to the dataset across the five reported times in \Cref{fig:buckley_conservative_rollout}. The relative error is \(0\) at \(t=0.0800\), then stays in the range \(2.56\times 10^{-2}\) to \(3.61\times 10^{-2}\) through \(t=0.0995\). Throughout the rollout, the predicted profile tracks both the smooth branch and the sharp front of the reference solution closely, indicating stable behavior for this nonlinear fractional-flow benchmark.

\end{document}